\newif\ifconfver
\confverfalse      

\newif\ifcutshort      
\cutshorttrue

\newif\ifcutshortlvltwo  
\cutshortlvltwofalse

\ifconfver
\documentclass[10pt,twocolumn,twoside]{IEEEtran}
\else
\documentclass[11pt, draftclsnofoot, onecolumn]{IEEEtran}
\fi

\usepackage{amsmath,amsfonts,bm,amssymb,epsfig,psfrag,ifthen,color}
\usepackage{booktabs}
\usepackage{multirow}
\usepackage[table]{xcolor}  
\usepackage{array}
\usepackage{calc}
\usepackage{caption}
\usepackage{graphicx}
\usepackage{psfrag}
\usepackage[tight,footnotesize]{subfigure} 
\usepackage{stfloats}
\usepackage{url}
\usepackage{longtable}
\usepackage{amsmath,amsfonts,amssymb,amsbsy,bm,paralist,theorem,ifthen,color}
\usepackage{algorithmic,algorithm,theorem}
\usepackage{mathtools}

\newcommand\Hc{\ensuremath{\mathcal{H}}}

\newcommand\Tc{\ensuremath{{\mathcal{T}}}}

\newcommand\Nc{\ensuremath{{\mathcal{N}}}}
\newcommand\Kc{\ensuremath{{\mathcal{K}}}}

\newcommand\xb{\ensuremath{{\bf x}}}

\newcommand\wb{\ensuremath{{\bm w}}}
\newcommand\yb{\ensuremath{{\bm y}}}

\newcommand\ssb{\ensuremath{{\bm s}}}

\newcommand\Hb{\ensuremath{{\bf H}}}
\newcommand\hb{\ensuremath{{\bf h}}}
\newcommand\Ab{\ensuremath{{\bf A}}}
\newcommand\ab{\ensuremath{{\bf a}}}
\newcommand\Bb{\ensuremath{{\bm B}}}

\newcommand\db{\ensuremath{{\bm d}}}
\newcommand\Db{\ensuremath{{\bm D}}}
\newcommand\eb{\ensuremath{{\bm e}}}

\newcommand\gb{\ensuremath{{\bm g}}}
\newcommand\gbm{\ensuremath{{\bm g}}}
\newcommand\Ib{\ensuremath{{\bm I}}}
\newcommand\Ic{\ensuremath{{\mathcal{I}}}}

\newcommand\Tb{\ensuremath{{\bm T}}}
\newcommand\Xb{\ensuremath{{\bf X}}}

\newcommand\Qb{\ensuremath{{\bf Q}}}

\newcommand\Vb{\ensuremath{{\bf V}}}

\newcommand\Wb{\ensuremath{{\bf W}}}
\newcommand\Zb{\ensuremath{{\bf Z}}}

\newcommand\zerob{\ensuremath{{\bm 0}}}

\newcommand\oneb{\ensuremath{{\bf 1}}}

\newcommand\Xc{\ensuremath{{\mathcal{X}}}}
\newcommand{\wt}{\widetilde}

\newcommand{\ol}{\overline}

\newtheorem{Prop}{Proposition}
\newtheorem{Theorem}{Theorem}
\newtheorem{Def}{Definition}

\newtheorem{Rmk}{Remark}

\ifconfver

\else

\fi

\begin{document}

    \bibliographystyle{IEEEtran}

    \title{ {Clustering by Orthogonal NMF Model and Non-Convex Penalty Optimization}}

    \ifconfver \else {\linespread{1.1} \rm \fi

        \author{\vspace{0.8cm}Shuai~Wang,
        	~Tsung-Hui~Chang, ~Ying~Cui,
        	~and Jong-Shi~Pang\\
            \thanks{
   Shuai Wang and Tsung-Hui Chang are with the Shenzhen Research Institute of Big Data and School of Science and Engineering, The Chinese University of Hong Kong, Shenzhen 518172, China (e-mail: shuaiwang@link.cuhk.edu.cn,~tsunghui.chang@ieee.org).}
 \thanks{Ying Cui is with the Department of Industrial and Systems Engineering, University of Minnesota, Minneapolis, MN 55455, USA (e-mail: yingcui@umn.edu).}
 \thanks{Jong-Shi Pang is with the Department of Industrial and Systems Engineering, University of Southern California, Los Angeles, CA 90089, USA (e-mail: jongship@usc.edu).}
            }

        \maketitle
        
\vspace{-1.5cm}
\begin{center}
\today
\end{center}\vspace{0.5cm}

        \begin{abstract}
	       The non-negative matrix factorization (NMF) model with an additional orthogonality constraint on one of the factor matrices, called the orthogonal NMF (ONMF), has been found a promising clustering model and can outperform the classical K-means. 
	       However, solving the ONMF model is a challenging optimization problem because the coupling of the orthogonality and non-negativity constraints introduces a mixed combinatorial aspect into the problem due to the determination of the correct status of the variables (positive or zero). 
	       Most of the existing methods directly deal with the orthogonality constraint in its original form via various optimization techniques, but are not scalable for large-scale problems. In this paper, we propose a new ONMF based clustering formulation that equivalently transforms the orthogonality constraint into a set of norm-based non-convex equality constraints.  We then apply a non-convex penalty (NCP) approach to add them to the objective as penalty terms, leading to a problem that is efficiently solvable.
	       One smooth penalty formulation and one non-smooth penalty formulation are respectively studied. We build theoretical conditions for the penalized problems to provide feasible stationary solutions to the ONMF based clustering problem, as well as proposing efficient algorithms for solving the penalized problems of the two NCP methods.
	       Experimental results based on both synthetic and real datasets are presented to show that the proposed NCP methods are computationally time efficient, and either match or outperform the existing K-means and ONMF based methods in terms of the clustering performance.
            \\\\
        \end{abstract}


        \ifconfver \else \IEEEpeerreviewmaketitle} \fi

    \vspace{-0.5cm}

\section{Introduction}

Clustering is one of the most fundamental data mining tasks and has an enormous number of applications \cite{Cluster:2013}. Typically, clustering is a key intermediate step to explore underlying structure of massive data for subsequent analysis. For example, in internet applications, clustering is used to identify users of different interests who then can be provided with specific service recommendation \cite{CFCR_2013, CRSV_2014}. In biology, clustering can be used for pattern discovery of genes which can help identify subtypes of a certain disease or cancer \cite{ICANMF_2009, Brunet_2004, Shuai_VS-ADMM_2018}.

Among the existing clustering methods, the K-means \cite{KM_1982} is the most widely used one, thanks to its simplicity \cite{Dis_kmeans_2016}. However, the K-means may not always yield satisfactory clustering results. 
On one hand, from an optimization perspective, the iterative steps of finding the cluster centroids and cluster assignment in K-means are equivalent to solving a binary integer constrained matrix factorization problem by alternating optimization \cite{Kmeans_NMF_2015,JNKM_2017}. Due to the non-convex matrix factorization model and binary integer constraint, the iterates of K-means are likely to be stuck at an unsatisfactory local point, and are sensitive to the choice of initial points \cite{K++_2007}. On the other hand, the K-means overlooks the inherent low-rank structure and prior information which are usually owned by high-dimensional real data. Therefore, various dimension-reduction techniques such as  principal component analysis (PCA), spectral clustering \cite{Spectral_2007}, non-negative matrix factorization (NMF) \cite{Lee_NMF_2001,Trkmen_2015} and deep neural networks \cite{UDE_Xie_2016,DCN_Yang_2017} are proposed.
However, these methods are merely used as a preprocessing stage to find a clustering-friendly representation for the data, and the K-means is still often used for clustering the dimension-reduced data.
Thus, the intrinsic drawback of the K-means caused by the non-convex nature and discrete constraints is not addressed.

Recently, as an variant of NMF, the orthogonal NMF (ONMF) model has been considered for data clustering  \cite{Ding_Orth_2006,Yoo_Choi_2010,Choi_ONMF_2008,HALS_2015,Pomplili_2014,ONMFS_2015,Shuai_VS-ADMM_2018}. 
The ONMF model imposes an additional orthogonality constraint on one of the factor matrices in NMF. It turns out that, like the K-means, the orthogonally constrained factor matrix functions the same as an indicator matrix that shows how the data samples are assigned to different clusters \cite{Ding_Orth_2006,Pomplili_2014}. 
Therefore, the ONMF model can be regarded as a continuous relaxation (which has no discrete constraint) of the K-means.
Studies on various data mining tasks have found that the ONMF model can outperform the K-means and NMF based clustering methods \cite{VideoQuery_2014,ONMF_cancer_2014, Yoo_Choi_2010,HALS_2015,Pomplili_2014,ONMFS_2015,Shuai_VS-ADMM_2018}.

\subsection{Related Works}
{Despite} its widespread use, solving the ONMF problem is challenging due to the existence of both the orthogonality and non-negativity constraints. Many of the existing ONMF algorithms extend upon the classical multiplicative update (MU) rule by Lee and Seung \cite{Lee_NMF_2001} for the vanilla NMF to accommodate the additional orthogonality constraint. For example, reference \cite{Ding_Orth_2006} penalized the orthogonality constraint followed by applying the MU rule. 
The authors of \cite{Choi_ONMF_2008} derived the MU rule directly using the gradient vector over the Stiefel manifold. 
Reference \cite{Pomplili_2014} employed the augmented Lagrangian (AL) method that penalizes the non-negativity constraint and applies the gradient projection method for the orthogonally constrained subproblem. 
Since projection onto the set of orthonormal matrices involves singular value decomposition (SVD), the orthogonal nonnegatively penalized matrix factorization (ONP-MF) method in \cite{Pomplili_2014} can be computationally inefficient.
The hierarchical alternating least squares (HALS) method proposed in  \cite{HALS_2015} is claimed to achieve a better balance between the orthogonality and non-negativity constraints. The HALS method updates one column and one row of the two respective factor matrices at the same time in each iteration, subject to the orthogonality and non-negativity constraints. Since the subproblem in HALS is handled by the same MU rule in  \cite{Ding_Orth_2006}, it is computationally efficient in general.
Reference \cite{ONMFS_2015} proposed to approximate the ONMF solution by solving a low-rank non-negative PCA problem, which however involves generating a large number of candidate solutions. While the method in \cite{ONMFS_2015} is the first that can provide provable approximation guarantee for the ONMF problem, it can be computationally inefficient especially when a high-quality solution is sought.

On the other hand, it is noticed that some recent  Riemannian optimization methods on the Stiefel manifold \cite{PG_Stiefel_2020,TConjugate_Stiefel_2020} can handle problems with orthogonality constraints together with non-smooth, Lipschitz-continuous regularizations. However,  since the indicator function induced by the non-negativity constraint is not Lipschitz continuous,  these methods are not applicable to the ONMF problem.
\subsection{Contributions}

In this paper, we propose a new approach to handle the ONMF based data clustering problem, aiming at achieving high-quality clustering performance with reasonable computational cost. Specifically, 
by recognizing the fact that the coupling of the orthogonality constraint and non-negativity constraint introduces disjunctions into the problem
and projection onto the orthogonal set requires expensive SVD, we avoid directly dealing with the orthogonality constraint as done in the existing methods \cite{Ding_Orth_2006, HALS_2015,Choi_ONMF_2008,Pomplili_2014}.
Instead, we propose in \cite{Shuai_SNCP_2019} a novel problem reformulation for the ONMF problem that replaces the orthogonality constraint by a set of norm-based (non-convex) equality constraints. The second ingredient of the proposed approach is the use of the penalty method in optimization \cite{Wright_NO} to add these non-convex norm-equality constraints as penalty terms in the objective function while keeping the non-negativity in the constraints.  
The advantages of the proposed method  are twofold. First, the penalty method allows to find a feasible solution to satisfy the norm-equality constraints gradually in a gentle fashion, and is less likely to be stuck at bad local solutions.
Second, since only simple non-negativity constraints are left, the penalized problem can be efficiently handled by the existing proximal alternating linearized minimization (PALM) method \cite{PALM_2014} without involving any SVD. As will be shown shortly, the proposed algorithms are inherently parallel and are more computationally efficient than most of the existing methods in practice.

In particular, we consider two novel types of non-convex penalty (NCP) formulations - one is smooth that has squared $\ell_1$-norm minus squared $\ell_2$-norm as the penalty term,
and the other one is non-smooth that has $\ell_1$-norm minus $\ell_\infty$-norm as the penalty term. The two penalties lead to different theoretical properties and algorithm developments.
We obtain theoretical conditions under which the two penalty methods can yield a feasible and meaningful solution to the considered ONMF based clustering problem.
We also develop computationally efficient PALM algorithms to solve the penalized problems for the smooth and non-smooth NCP methods, respectively. It is worthy to mention that the penalized problem of the non-smooth NCP method is a non-convex and non-smooth problem with a negative infinity norm regularization. We propose a novel non-convex proximal operator of the negative infinity-norm function, and show that it has a simple closed-form solution. 

%
%

Extensive experiments are conducted based on a synthetic data set \cite{JNKM_2017}, the gene dataset TCGA \cite{website_TCGA,TCGA_CGCD} 
and the document dataset TDT2 \cite{LCCF_2011}. The experimental results demonstrate that the proposed smooth and non-smooth NCP methods can provide either comparable or greatly better performance than the existing K-means based methods and ONMF based methods in \cite{Ding_Orth_2006,Pomplili_2014,Choi_ONMF_2008, HALS_2015}, and at the same time are more time efficient than most of the existing ONMF based methods. 


{\bf Synopsis:} Section \ref{sec: onmf formulation} reviews the existing K-means and ONMF model, and presents the proposed clustering problem formulation.
The proposed NCP optimization framework is presented in Section \ref{sec: proposed ncp}, where theoretical conditions for the smooth and non-smooth NCP methods to yield feasible stationary solutions are analyzed. The PALM algorithms used for solving the smooth and non-smooth penalized problems are given in Section \ref{sec:update}. Experimental results are presented in Section \ref{sec: sim results}, and lastly the conclusions are drawn in Section \ref{sec: conclusion}.

{\bf Notation:} We use boldface lowercase
letters and boldface uppercase letters to represent column vectors and
matrices, respectively. $\mathbb{R}^{m \times n}$ denotes the set of $m$ by $n$ real-valued matrices. The $(i,j)$th entry of matrix $\Ab$ is denoted by 
$[\Ab]_{ij}$; the $i$th element of vector $\ab$ is denoted by $(\ab)_i$ or $a_i$. Superscript $\top$ stands for matrix transpose.
For a matrix $\Ab \in \mathbb{R}^{m \times n}$, its column vectors are denoted by $\ab_j \in \mathbb{R}^{m}$, $j=1,\ldots,n$, and its row vectors are denoted by
$\wt\ab_i \in \mathbb{R}^{n}$, $i=1,\ldots,m$; that is,
$\Ab 
=
\begin{bmatrix}
	\ab_1, \ldots,\ab_n
\end{bmatrix}
=
\begin{bmatrix}
	\wt\ab_1,
	\ldots,
	\wt\ab_m
\end{bmatrix}^{\top}.$
We denote $\Ab \geq \zerob$ as a non-negative matrix, i.e., $[\Ab]_{ij} \geq 0$ for all $i=1,\ldots,m,j=1,\ldots,n$.
$\oneb$ denotes the all-one vector, $\zerob$ is the all-zero vector, 
$ \Ib_{m}$ is the $m$ by $m$ identity matrix, and
$\eb_n$ denotes the elementary vector with one in the $n$th entry and zero otherwise.
$\|\cdot\|_F$ and $\|\cdot\|_p$ are the matrix Frobenius norm and vector $p$-norm, respectively.
$\langle \Ab,\Bb\rangle$ denotes the inner product between matrices $\Ab$ and $\Bb$.
$\lambda_{\max}(\Ab)$ stands for the maximum eigenvalue of matrix $\Ab$. For a convex function $f\colon\mathbb{R}^n \rightarrow \mathbb{R}$, $\partial f(\xb)$ denotes the subdifferential of $f$ at $\xb$ as in standard convex analysis \cite{BT_ConvexAnalysis}. Lastly, $[\Ab]^+$ denotes $\max\{\Ab,\zerob\}$ which is a matrix that reserves the non-negative elements of $\Ab$. 

\section{Data Clustering and ONMF Model}\label{sec: onmf formulation}

\subsection{K-Means and ONMF Model}

Let $\Xb \geq \zerob$ be a non-negative data matrix that contains $N$ data samples  
and each of the samples has $M$ features, i.e., $\Xb=[\xb_1,\ldots, \xb_N] \in \mathbb{R}^{M \times N}$.
The task of data clustering is to assign the $N$ data samples into a predefined number of $K$ clusters in the sense that the samples belonging to one cluster are close to each other based on certain distance metric. 
The most popular setting is to consider the Euclidean distance and the use of the K-means due to its simplicity.

From an optimization point of view \cite{Kmeans_NMF_2015,JNKM_2017}, the iterative procedure of K-means can be interpreted as an alternating optimization algorithm applied to the following matrix factorization problem
\begin{subequations}\label{eqn: kmeans prob}
	\begin{align}
		\min_{ \substack{\Wb, \Hb}}~ & \|\Xb - \Wb\Hb\|_{F}^2,\label{eqn: kmeans prob obj} \\
		{\rm s.t.}~ & \oneb^{\top}\hb_j = 1, ~[\Hb]_{ij}\in \{0, 1\},  ~\forall i \in \Kc, j\in \Nc, \label{eqn: kmeans prob C1} \\
		&\Wb\geq \zerob,
	\end{align}
\end{subequations}
where $\Kc \triangleq \{1,\ldots,K\}$ and $\Nc \triangleq \{1,\ldots,N\}$.  
Here, columns of $\Wb \in \mathbb{R}^{M \times K}$ represent centroids of the $K$ clusters, while
the matrix $\Hb \in \mathbb{R}^{K \times N}$
indicates the cluster assignment of samples. 
Specifically, 
$[\Hb]_{ij}=1$ indicates that the $j$th sample $\xb_j$ is uniquely assigned to cluster $i$, and $[\Hb]_{ij}=0$ otherwise. 

One can see from \eqref{eqn: kmeans prob} that, when $\Wb=[\wb_1,\ldots,\wb_K]$ is given, the optimal $\Hb$ is obtained by assigning each sample to the cluster that has the nearest centroid, while when $\Hb$ is given, the optimal $\wb_i$ is given by the centroid (average of samples) assigned to the $i$th cluster, for all $i\in \Kc$. The two steps are exactly the well-known K-means algorithm.
However, due to the non-convex binary constraint \eqref{eqn: kmeans prob C1} and the hard clustering assignment during the iterative steps of K-means, the K-means is sensitive to the initial conditions and
may not always yield satisfactory clustering performance \cite{K++_2007}.

Regarded as a relaxation of \eqref{eqn: kmeans prob}, the following vanilla NMF model is also considered for data clustering \cite{Trkmen_2015,Pomplili_2014,JNKM_2017}
\begin{subequations}\label{eqn: nmf prob}
	\begin{align}
		\min_{\Wb,\Hb}~&\|\Xb - \Wb\Hb\|_{F}^{2} \\
		{\rm s.t.}~&\Wb \geq  \zerob, \Hb \geq \zerob. \label{eqn:nmf  C1} 
	\end{align}
\end{subequations}
However, since \eqref{eqn: kmeans prob C1}  is completely removed from \eqref{eqn: kmeans prob},
the obtained $\Hb$ from the above vanilla NMF model cannot reveal clear clustering assignment. 

It was shown in \cite{Ding_Orth_2006,Pomplili_2014} that the ONMF model, which has an additional orthogonality constraint on $\Hb$, is closely related to the K-means model \eqref{eqn: kmeans prob}.
In particular, the ONMF problem is given by
\begin{subequations}\label{eqn: onmf prob}
	\begin{align}
		\min_{\Wb,\Hb}~&\|\Xb - \Wb\Hb\|_{F}^{2} \\
		{\rm s.t.}~&\Wb \geq  \zerob , \Hb \geq  \zerob, \label{eqn:onmf  C1} \\
		&\Hb\Hb^{\top}= \Ib_{K}.
		\label{eqn:onmf  C2}
	\end{align}
\end{subequations}
A key observation is given as below.

\vspace{0.4cm}
\noindent \fbox{\parbox{\linewidth}{ {\bf Observation:} \ Any $\Hb$ satisfying $\Hb \geq  \zerob$ and $\Hb\Hb^{\top}= \Ib_{K}$ has at most one non-zero entry in each column.
		
}} ~
\vspace{0.2cm}

\noindent Thus, matrix $\Hb$ in \eqref{eqn: onmf prob} functions similarly as that in \eqref{eqn: kmeans prob} whose nonzero elements indicate the cluster assignment of data samples. Nevertheless, different from \eqref{eqn: kmeans prob}, the non-zero entries of $\Hb$ in \eqref{eqn: onmf prob} are not restricted to be either zero or one but can be scaled. Owing to the two facts, the ONMF model is 
less sensitive to data scaling and is preferred than the K-means especially for data where clustering results are independent of data scaling; see \cite{Pomplili_2014} for more discussions.

However, the ONMF problem \eqref{eqn: onmf prob} is intrinsically challenging to solve since the above Observation indicates that the intersection of $\Hb \geq  \zerob$ and $\Hb\Hb^{\top}= \Ib_{K}$ is mixed combinatorial due to the determination of the correct status of the variables (positive or zero).  In view of this,  the philosophy of the existing methods is to handle the two constraints separately either by the penalty method  \cite{Ding_Orth_2006,HALS_2015,Choi_ONMF_2008} or by the AL method  \cite{Pomplili_2014} which requires repetitive projections onto the orthogonal set via expensive SVD.
Unlike the existing methods, we present a novel problem reformulation of \eqref{eqn: onmf prob} which avoids dealing with the orthogonality constraint directly. Based on the reformulated problem, we propose a non-convex penalty (NCP) optimization framework that is not only amenable to efficient computation but also able to provide favorable clustering performance.
\subsection{Proposed Clustering Formulation}

We note that any vector $\xb \in \mathbb{R}^{n}$ has at most one non-zero entry if and only if
\begin{align} \label{eqn: x one nonzero}
	\|\xb\|_{p} = \|\xb\|_{q}, ~1 \leq p <q.
\end{align}
According to the Observation, any $\Hb$ satisfying \eqref{eqn:onmf  C2} and \eqref{eqn:onmf  C1} also lies in the following set
\begin{align}\label{eqn: set equiv}
	&\left\{  \Hb\geq \zerob~\bigg| \begin{array}{ll}
		& \|\wt \hb_i\|_2 =1,~i\in \Kc, \\
		& \|\hb_j\|_p=\|\hb_j\|_q,~j \in \Nc
	\end{array} \right \}.
\end{align}
Thus, the ONMF model \eqref{eqn: onmf prob} can be equivalently written as
\begin{subequations}\label{eqn: onmf prob eq}
	\begin{align}
		\min_{\Wb,\Hb}~&\|\Xb - \Wb\Hb\|_{F}^{2} \label{eqn: onmf prob obj} \\
		{\rm s.t.}~&\Wb\geq \zerob, \Hb\geq \zerob, \label{eqn:onmf  C1 eq} \\
		& \|\wt \hb_i\|_2 =1,~i\in \Kc, \\
		& \|\hb_j\|_p=\|\hb_j\|_q,~j \in \Nc.
		\label{eqn:onmf  C2 eq}
	\end{align}
\end{subequations}

Firstly, note that the condition $\|\wt \hb_i\|_2 =1,~i\in \Kc$, is not intrinsic to the data clustering task. 
In essence, both $\Hb$ and $\Qb\Hb$, where $\Qb\geq 0$ is a diagonal matrix, indicate the same cluster assignment, and both  $(\Wb,\Hb)$ and $(\Wb\Qb^{-1},\Qb\Hb)$ have the same objective values in \eqref{eqn: onmf prob obj}. 
Therefore, without loss of the clustering performance, we remove $\|\wt \hb_i\|_2 =1,~i\in \Kc$, from \eqref{eqn: onmf prob eq}.
Secondly, for bounded solution, we add the regularization term  $\frac{  \mu_w}{2}\|\Wb\|_F^2 + \frac{ \mu_h}{2}\|\Hb\|_F^2$ to \eqref{eqn: onmf prob eq}, where $\mu_w,\mu_h\geq 0$ are two parameters\footnote{Adding the regularization term makes the objective coercive, which subsequently can guarantee bounded solutions to the optimization problem. 
}.
Thirdly, without loss of generality, we replace $ \|\hb_j\|_p=\|\hb_j\|_q$ with $\|\hb_j\|_p^v=\|\hb_j\|_q^v$ by adding a power exponent $v> 0$.
As a result, we have the following problem formulation for data clustering:

\vspace{0.4cm}

\noindent \fbox{\parbox{\linewidth}{ {\bf Proposed clustering formulation:} 
		\begin{subequations}\label{eqn: onmf prob P2}
			\begin{align} 
				&\min_{\Wb,\Hb}~   \|\Xb-\Wb\Hb\|_F^2 + \frac{\mu_w}{2}\|\Wb\|_F^2 + \frac{\mu_h}{2}\|\Hb\|_F^2\\
				&~~~{\rm s.t.}~
				\Wb \geq \zerob, \label{eqn: onmf prob P2 C3} \\
				&~~~~\begin{array}{ll}
					&\Hb\geq \zerob, \\
					& \|\hb_j\|_p^v=\|\hb_j\|_q^v,~j \in \Nc, 
				\end{array}\bigg\} \triangleq \Hc_{p,q}^v. \label{eqn: onmf prob P2 C1} 
			\end{align}
		\end{subequations}
	}
} 
\vspace{0.2cm}

One clear advantage of formulation \eqref{eqn: onmf prob P2} is its scalability. Concisely, when $\Wb$ is fixed, problem \eqref{eqn: onmf prob P2} can be fully decoupled across the columns of $\Hb$ into $N$ subproblems. 
This makes it easy to apply some decomposition methods for dealing with large-scale clustering tasks. 
Nevertheless, it is still challenging to solve \eqref{eqn: onmf prob P2}. The main challenge lies in two interwound issues. The first is 
how to deal with the non-convex objective function and the (possibly non-smooth) norm-equality constraint \eqref{eqn: onmf prob P2 C1} together with the non-negativity constraint \eqref{eqn: onmf prob P2 C3}. The second is how to choose proper values of $p$,  $q$ and $v$ 
since different choices lead to different theoretical and computational properties. 

Before proceeding with the algorithmic development, we first present basic definitions for characterizing a proper solution of the non-convex and possibly non-smooth  problem \eqref{eqn: onmf prob P2}. 
\begin{Def} {(Tangent cone)} \label{def: tangent cone}
	Let $\Xc \subseteq \mathbb{R}^n$ and $\xb \in \Xc$. 
	A vector $\db$ is a tangent of $\Xc$ at $\xb$ if either $\db = \zerob$ or there exists a sequence $\{\xb^k\} \subset \Xc$ 
	and positive scalars $\{\tau^k\}$ such that $
	\lim_{k\to \infty}\frac{\xb^k - \xb}{\tau^k} = \db
	$ when $\xb^k \to  \xb,~\tau^k \searrow 0$.	
	The tangent cone of $\Xc$ at $\xb$, denoted by $\Tc_{\Xc}(\xb)$, contains all the tangents of $\Xc$ at $\xb$. 
\end{Def}
\begin{Def}{(Directional derivative)}  \label{def: directional derivative}
	Let $f:\mathbb{R}^n\to \mathbb{R}$ be a possibly non-smooth function.
	Then $f$ is directionally differentiable if the directional derivative of $f$ along any direction $\db \in \mathbb{R}^n$ 
	\begin{align}
		f'(\xb;\db)=\lim_{\tau \searrow 0} \frac{f(\xb+\tau \db) - f(\xb)}{\tau}
	\end{align}
	exists at any $\xb  \in \mathbb{R}^n$. 
\end{Def}
If $f$ is differentiable, then the directional derivative $f'(\xb; \db)$ in the above definition reduce to $\nabla f(\xb)^{\top}\db$, where $\nabla f(\xb)$ is the gradient vector of $f$ at $\xb$.

\begin{Def} (B-stationary solution)\cite{BK:VIC1_JSPang}  \label{def: unsmooth stationary point}
	For an optimization problem $\min_{\xb \in \Xc} f(x)$, where $f:\mathbb{R}^n\to \mathbb{R}$ is  directionally differentiable, and $\Xc$ is a closed set. Then, $\ol\xb \in \Xc$ is a B-stationary point if 
	$f'(\ol\xb; \db) \geq 0, \forall \db \in \Tc_{\Xc}(\ol\xb).$
\end{Def}
If $\Xc$ is a convex set, then the condition is equivalent to $f'(\ol\xb; \xb-\ol\xb) \geq 0, \forall \xb \in \Xc,$ and $\ol \xb$ is known as the \emph{d-stationary point}.  If $f$ is differentiable and $\Xc$ is convex, then the condition reduces to
$\nabla f(\ol\xb)^{\top}( \xb-\ol\xb) \geq 0, \forall \xb \in \Xc,$ and $\ol \xb$ is simply called a \emph{stationary point} \cite[(1.3.3)]{BK:VIC1_JSPang}.

As proved in Appendix \ref{appdix: proof of prop nonzero cols}, the following proposition shows that any B-stationary point of \eqref{eqn: onmf prob P2} has non-zero columns for $\Hb$.
\begin{Prop} \label{thm: non-zero columns}
	Let $(\ol\Wb, \ol\Hb)$ be a B-stationary point of \eqref{eqn: onmf prob P2} satisfying $(\ol\Wb)^{T}\xb_{j} \ne \zerob, \forall j \in \Nc$. Then, ${\ol\hb_j} \ne \zerob, \forall j \in \Nc$.
\end{Prop}
The condition of $(\ol\Wb)^{T}\xb_{j} \ne \zerob, \forall j \in \Nc$
is actually mild as it is known that the centroid $\Wb$ should usually lie in the space spanned by the cluster samples \cite{Convex_SNMF_2010}.
Proposition \ref{thm: non-zero columns} implies that a B-stationary point of \eqref{eqn: onmf prob P2} is clustering meaningful since, at a B-stationary point,  each data sample must be properly assigned to one of the clusters. 

\vspace{-0.0cm}
\section{Proposed Non-Convex Penalty Method}\label{sec: proposed ncp}
\vspace{-0.0cm}


In this section, we present the proposed NCP framework to handle problem \eqref{eqn: onmf prob P2}. Let $\phi(\hb_j)\triangleq \|\hb_j\|_p^{v} - \|\hb_j\|_q^{v}\geq 0$ for $1\leq p<q$. Since $\phi(\hb_j)=0$ is equivalent to making mixed binary decisions for each entry of $\hb_j$ (being zero or non-zero), our intuition is to have an algorithm that can gently arrive at the mixed binary decision so that it can be less sensitive to bad initial points. Also, it is desirable to handle a problem with simple constraint sets.
Motivated by this,
we consider the following penalized formulation
\begin{subequations}\label{eqn: onmf prob PP}
	\begin{align} 
		\min_{\Wb,\Hb}~ &F(\Wb, \Hb) + \frac{\rho}{v} \sum_{j=1}^N \bigg(\|\hb_j\|_p^{v} - \|\hb_j\|_q^{v}\bigg) \label{eqn: onmf prob PP obj} \\
		{\rm s.t.}~
		& \Wb \geq \zerob, \Hb \geq \zerob, \label{eqn: onmf prob PP C2}
	\end{align}
\end{subequations}
where  $F(\Wb,\Hb)\triangleq \|\Xb-\Wb\Hb\|_F^2 + \frac{ \mu_w}{2}\|\Wb\|_F^2 + \frac{ \mu_h}{2}\|\Hb\|_F^2$ , and $\rho > 0$ is a penalty parameter.
As seen from \eqref{eqn: onmf prob PP}, since the non-convex norm-equality constraints in \eqref{eqn: onmf prob P2 C1}  are penalized in the objective function, the penalized problem \eqref{eqn: onmf prob PP} involves simple convex constraint set only. 

Like the classical penalty method \cite[Chapter 17]{Wright_NO}, we attempt to reach a good clustering solution of \eqref{eqn: onmf prob P2} through solving a sequence of penalized subproblems in \eqref{eqn: onmf prob PP}, by gradually increasing $\rho$.
The proposed NCP framework is shown in Algorithm \ref{table: SNCP}. It is expected that the norm-equality constraint \eqref{eqn: onmf prob P2 C1} can be gradually satisfied as $\rho$ increases, implying that the clustering assignment is achieved step by step in a smooth manner. This is in contrast to the classical K-means which has hard decision on the clustering assignment in the iterative process. This property also makes the proposed NCP method less sensitive to the choice of the initial point, which is one of the key advantages.  An illustrative example demonstrating this point is given in \cite[Section 5]{SPD} of the supplementary material.

We study two types of penalty functions. In particular, we respectively consider a smooth penalty and a non-smooth penalty for \eqref{eqn: onmf prob PP}. The smooth penalty is inspired by the classical quadratic penalty method \cite[Chapter 17.1]{Wright_NO} where we choose $p=1, q=2, v=2$ for \eqref{eqn: onmf prob PP}. 
For the non-smooth penalty, we choose $p=1, q=\infty$ and $v=1$. As will be shown in Section \ref{subsec: nonsmooth ncp},  such choice can provide so called \emph{exact penalty} property  \cite{Wright_NO}\footnote{We say the penalized problem like \eqref{eqn: onmf prob PP} has exact penalty if there exists a finite value of $\rho^*$ such that for any $\rho>\rho^*$ the solution of \eqref{eqn: onmf prob PP} is also a (feasible) solution of \eqref{eqn: onmf prob P2}. }
while only requiring relaxed conditions on the solution of \eqref{eqn: onmf prob PP}. 
To justify the effectiveness of the two penalties, in the ensuing two subsections, we present theoretical conditions for which the penalty method and Algorithm \ref{table: SNCP} can yield a feasible stationary solution to problem \eqref{eqn: onmf prob P2}. In Section \ref{sec:update}, efficient algorithms for solving \eqref{eqn: onmf prob PP} are presented.

\subsection{Smooth NCP (SNCP) Method}\label{subsec: smooth ncp}

The proposed SNCP is obtained by choosing $p = 1, q = 2$ and $v = 2$ in \eqref{eqn: onmf prob P2} and \eqref{eqn: onmf prob PP}.
Since $\Hb\geq \zerob$, $\|\hb_j\|_1=\oneb^{\top}\hb_j, j \in \Nc$, the corresponding penalized problem \eqref{eqn: onmf prob PP}
is given by
\vspace{-0.2cm}
\begin{subequations}\label{eqn: onmf prob PP smooth}
	\begin{align} 
		\min_{\Wb,\Hb}~ 
		&F(\Wb, \Hb) +  \frac{\rho}{2} \sum_{j=1}^N \bigg((\oneb^{\top}\hb_j)^2 - \|\hb_j\|_2^{2}\bigg) 
		\label{eqn: onmf prob PP smooth obj} \\
		{\rm s.t.}~
		& \Wb\geq \zerob, \Hb\geq \zerob, \label{eqn: onmf prob PP smooth C2}
	\end{align}
\end{subequations}
which has $G_\rho(\Wb, \Hb)\triangleq F(\Wb, \Hb) +  \frac{\rho}{2} \sum_{j=1}^N \big((\oneb^{\top}\hb_j)^2 - \|\hb_j\|_2^{2}\big)$ as
the smooth objective function.

\begin{algorithm}[t!]
	\caption{Proposed NCP framework for problem \eqref{eqn: onmf prob P2}.}
	\begin{algorithmic}[1]\label{table: SNCP}
		\STATE {\bf Set} $r = 1$, and given a parameter $\gamma>1$ and initial values of $\rho > 0$ and $(\Wb^{(0)},\Hb^{(0)})$.
		\REPEAT
		\STATE Obtain a d-stationary point  $(\Wb^{(r)},\Hb^{(r)})$ of problem \eqref{eqn: onmf prob PP}, using 
		$(\Wb^{(r-1)},\Hb^{(r-1)})$ as the initial point (see Algorithm \ref{table: smooth ALM} and Algorithm \ref{table: non-smooth ALM} in Section \ref{sec:update}).
		\STATE Increase the penalty parameter $\rho =\gamma \rho$ if the row vectors of $\Hb^{(r)}$ are not sufficiently orthogonal (see Section \ref{sec: sim results}).
		\STATE Set $r = r + 1$.	
		\UNTIL a predefined stopping criteria is satisfied.
	\end{algorithmic}
	\vspace{-0.1cm}
\end{algorithm}


%
The following proposition provides the theoretical justification for the SNCP method.

\begin{Theorem}\label{prop: local min} 
	There exists a finite $\rho^{\star} > 0$ such that for any $\rho > \rho^{\star}$, any local minimum solution of \eqref{eqn: onmf prob PP smooth} is a feasible and a local minimum solution to \eqref{eqn: onmf prob P2} (with $p=1$, $q=2$ and $v=2$).
\end{Theorem}

The proof of Theorem \ref{prop: local min}  is given in Appendix \ref{appdix: proof of prop local min}. 
%
%
%
Theorem \ref{prop: local min} shows that the SNCP has the exact penalty as long as a local minimum of \eqref{eqn: onmf prob PP smooth} can be reached.
Unfortunately, this cannot be guaranteed in general.
The following theorem asserts a relaxed condition that if only a stationary solution (not local minimum) is obtained, \eqref{eqn: onmf prob PP smooth} can still yield a feasible B-stationary solution of   \eqref{eqn: onmf prob P2} as long as $\rho$ goes to infinity. 

\begin{Theorem}\label{thm: stationary pont}
	Let $(\Wb^{\rho},\Hb^{\rho})$ denote a stationary point of \eqref{eqn: onmf prob PP smooth}, and  assume that $(\Wb^{\rho},\Hb^{\rho})$ is bounded and it has a limit point $(\Wb^\infty,\Hb^\infty)\neq \zerob$ when $\rho \to \infty$.
	Then, $(\Wb^\infty,\Hb^\infty)$ is a feasible B-stationary point to \eqref{eqn: onmf prob P2}. 
\end{Theorem}

In practice, a finite value of $\rho$ would be sufficient to obtain a reasonably good solution as one will see in Section \ref{sec: sim results}. It is worthwhile to note that for the general quadratic penalty method \cite[Chapter 17.1]{Wright_NO} to claim the same result as in Theorem \ref{thm: stationary pont}, one
requires conditions such as $(\Wb^\infty,\Hb^\infty)$ satisfies certain constraint qualification.  Our Theorem \ref{thm: stationary pont} does not have such requirement thanks to the special constraint structure of \eqref{eqn: onmf prob P2}; see the proof of Theorem \ref{thm: stationary pont} in Appendix \ref{appdix: proof of thm statioanry point}.



\subsection{Non-smooth NCP (NSNCP) method} \label{subsec: nonsmooth ncp}

In this subsection, we consider a non-smooth NCP method by setting $p = 1, q = \infty$ and $v = 1$ in \eqref{eqn: onmf prob P2} and \eqref{eqn: onmf prob PP}.
The corresponding penalized problem \eqref{eqn: onmf prob PP}
is given by
\begin{subequations}\label{eqn: onmf prob PP nsmooth}
	\begin{align} 
		\min_{\Wb,\Hb}~ 
		&F(\Wb, \Hb) +  {\rho} \sum_{j=1}^N \big(\oneb^{\top}\hb_j - \|\hb_j\|_\infty\big) 
		\label{eqn: onmf prob PP nsmooth obj} \\
		{\rm s.t.}~
		& \Wb \geq \zerob, \Hb \geq \zerob, \label{eqn: onmf prob PP nsmooth C2}
	\end{align}
\end{subequations}
where the objective function $F_\rho(\Wb, \Hb) \triangleq F(\Wb, \Hb) +  {\rho} \sum_{j=1}^N \big(\oneb^{\top}\hb_j - \|\hb_j\|_\infty\big)$ is non-smooth.
Interestingly, analogous to the exact penalty in Theorem \ref{prop: local min} but without the need of a local minimum solution,
the NSNCP method allows one to obtain a feasible B-stationary solution of \eqref{eqn: onmf prob P2} if a d-stationary solution of \eqref{eqn: onmf prob PP nsmooth} can be obtained.
\vspace{-0.0cm}
\begin{Theorem} \label{thm: exact penalty}
	Assume that $(\Wb^{\rho},\Hb^{\rho})$ is bounded. Then, there exists a finite $\rho^\star > 0$ such that for all $\rho > \rho^{\star}$, if $(\Wb^{\rho}, \Hb^{\rho})$ is a d-stationary point of \eqref{eqn: onmf prob PP nsmooth}, then 
	$(\Wb^{\rho}, \Hb^{\rho})$ is feasible and a B-stationary point to \eqref{eqn: onmf prob P2}.
\end{Theorem}

The proof of Theorem \ref{thm: exact penalty} is presented in Appendix \ref{appdix: proof of thm nonsmooth}.
By comparing Theorem \ref{thm: exact penalty} with Theorem \ref{prop: local min}, one can see that the NSNCP method in \eqref{eqn: onmf prob PP nsmooth} is theoretically preferred since it does not require the penalized problem \eqref{eqn: onmf prob PP nsmooth} to provide a locally optimal solution (which computationally cannot be ascertained) but only a d-stationary point; by comparing Theorem 3 with Theorem \ref{thm: stationary pont}, we see that the NSNCP requires a finite value of $\rho$ in the non-smooth formulation \eqref{eqn: onmf prob PP smooth obj} versus an infinite value of $\rho$ in the smooth formulation \eqref{eqn: onmf prob PP nsmooth obj}.
However, since  \eqref{eqn: onmf prob PP nsmooth} is non-convex and non-smooth, it is more challenging to handle than its smooth counterpart \eqref{eqn: onmf prob PP smooth}.  In the next section, we present efficient algorithms that can be used to obtain a proper stationary solution of the penalized problems \eqref{eqn: onmf prob PP smooth} and \eqref{eqn: onmf prob PP nsmooth}, respectively. 

\begin{Rmk}{\rm (On the choice of $p,q,$ and $v$)
		As mentioned, the smooth penalty in \eqref{eqn: onmf prob PP smooth} with $p = 1, q = 2, v = 2$ 
		is inspired by the classical quadratic penalty method \cite[Chapter 17.1]{Wright_NO}.  A natural question is -- can we have other choices? In fact, one can verify that as long as 
		$p, q, v$ satisfies
		\begin{align} \label{eqn: p_q_v}
			p \geq 1, q > p, v \geq q,
		\end{align}
		then \eqref{eqn: onmf prob PP} will be a smooth problem.  Moreover, one can have the same claim as Theorem \ref{thm: stationary pont} based on a straightforward extension of the proof of Theorem \ref{thm: stationary pont}.  However,  we prefer the choice in \eqref{eqn: onmf prob PP smooth}  since  we are not aware of any theoretical result in the literature that suggests one can benefit from higher orders of penalty functions.  In addition, our experiences in numerical experiments in fact indicate that higher order choices may yield poor performance.  As shown in the supplementary material \cite[Section 3.1]{SPD}, a large value of $v$ would make the landscape of $\phi(\hb_j)=\|\hb_j\|_p^v - \|\hb_j\|_1^v$ has a flat valley around the origin. Besides,  as shown in \cite[Section 3.2]{SPD},  large values of $v$ and $q$ make Algorithm \ref{table: SNCP} less stable and more difficult to reach a feasible solution satisfying $\phi(\hb_j)=0$.
		
		Our choice of  $p=1,q=\infty,v=1$ for the non-smooth penalty \eqref{eqn: onmf prob PP nsmooth} is constructed on purpose in order to achieve the exact penalty property in Theorem \ref{thm: exact penalty}.  As one can see from \eqref{eqn: staionary equality 1} that a key property to prove Theorem \ref{thm: exact penalty} is that 
		the directional derivative of $\phi(\xb^\rho) = \oneb^{\top} \xb^\rho - \|\xb^\rho\|_\infty$ has an upper bound linear in $\db$ and independent of the exact values of $\xb^\rho$.  Such property would not hold for $v>1$ or other choices of $q>1$.
		In Section \ref{subsec: nonsmooth update}, we will further show that \eqref{eqn: onmf prob PP nsmooth} with the negative infinity norm can have a simple proximal operator which enable us to solve \eqref{eqn: onmf prob PP nsmooth}  efficiently.
	}
	
\end{Rmk}

\section{Obtaining a Stationary Point of \eqref{eqn: onmf prob PP}} 
\label{sec:update}

In accordance with Theorem \ref{thm: stationary pont} and Theorem \ref{thm: exact penalty}, we need to obtain a stationary solution for  problem \eqref{eqn: onmf prob PP smooth}  and a d-stationary solution for  problem \eqref{eqn: onmf prob PP nsmooth}, respectively.  In view of the separable constraint structure for $\Wb$ and $\Hb$, the PALM algorithm in \cite{PALM_2014} is particularly efficient in handling the two problems. 

\subsection{Algorithm for Solving \eqref{eqn: onmf prob PP smooth}}\vspace{-0.0cm}

By applying the PALM algorithm to the smooth penalized problems \eqref{eqn: onmf prob PP smooth}, one simply performs block-wise gradient projection with respect to $\Wb$ and $\Hb$ iteratively, as shown in Algorithm \ref{table: smooth ALM}. Here, $t^{k}$ and $c^{k}$ are two step size parameters. Denote $L_G(\Wb^k)$ and $L_G(\Hb^k)$ as the Lipschitz constants of $\nabla_{\Hb}G_{\rho}(\Wb^{k}, \Hb)$ and $\nabla_{\Wb}G_{\rho}(\Wb, \Hb^{k})$, respectively. Convergence of Algorithm \ref{table: smooth ALM} can be established based on \cite{PALM_2014}.

\vspace{-0.0cm}
\begin{Theorem} \label{thm: convergence of PALM}
	Let $\{\Wb^k, \Hb^k\}$ be the sequence generated by Algorithm \ref{table: smooth ALM}
	with $t^k > \frac{L_G(\Wb^k)}{2}$ and $c^k > \frac{L_G(\Hb^k)}{2}$. Then $\{G_{\rho}(\Wb^k, \Hb^k)\}$ is non-increasing, $\{\Wb^k, \Hb^k\}$ is bounded, 
	and $\{\Wb^k, \Hb^k\}$ converges to a stationary point of \eqref{eqn: onmf prob PP smooth}.
\end{Theorem}

\vspace{-0.0cm}
{\bf Proof:} Since $G_\rho(\Wb,\Hb)$ is a coercive function, 
and the PALM updates in Algorithm \ref{table: smooth ALM} guarantees descent of the objective function $G_\rho(\Wb^k,\Hb^k)$ \cite[Remark 4(iii)]{PALM_2014} under $t^k > \frac{L_G(\Wb^k)}{2}$ and $c^k > \frac{L_G(\Hb^k)}{2}$,
$\{\Wb^k, \Hb^k\}$ are bounded. Besides, because $G_\rho(\Wb,\Hb)$ satisfies the Kurdyka-Lojasiewicw (KL) property, and $\Wb\geq \zerob$, $\Hb\geq \zerob$ are convex sets,
we can obtain the desired results by \cite[Theorem 1]{PALM_2014}.
\hfill $\blacksquare$

\begin{algorithm}[t!]
	\caption{PALM for solving penalized problem \eqref{eqn: onmf prob PP smooth}. }
	\begin{algorithmic}[1]\label{table: smooth ALM}
		\STATE {\bf Set} $k = 0, \Wb^{0} = \Wb^{(r-1)},~\Hb^{0} = \Hb^{(r-1)}$.	
		\REPEAT
		\STATE 
		$
		\Hb^{k+1} =(\Hb^{k} - \frac{1}{t^{k}}\nabla_{\Hb}G_{\rho}(\Wb^{k}, \Hb^{k}))^+,
		$
		
		$	\Wb^{k+1} =(\Wb^{k} - \frac{1}{c^{k+1}}\nabla_{\Wb}G_{\rho}(\Wb^{k}, \Hb^{k+1}))^+, 
		$
		\STATE Set $k =k + 1$,
		\UNTIL a predefined stopping criteria is satisfied.
	\end{algorithmic}
\end{algorithm}

\subsection{Algorithm for Solving \eqref{eqn: onmf prob PP nsmooth}}
\label{subsec: nonsmooth update}

The non-smooth penalized problem \eqref{eqn: onmf prob PP nsmooth} is more challenging to handle due to the non-smooth and non-convex term $-\|\hb_j\|_\infty$. In particular, when applying the same PALM strategy to problem \eqref{eqn: onmf prob PP nsmooth}, the corresponding subproblem for updating $\Hb$ is given by
\begin{align} 
	\Hb^{k+1} 
	&=\arg\min_{\Hb  \geq \zerob} \frac{t^k}{2}\|\Hb - \Bb^{k+1}\|_F^2 - {\rho} \sum_{j=1}^N \|\hb_j\|_\infty, \label{eqn: H update nonsmooth}
\end{align}
where $\Bb^{k+1} = \Hb^k - \frac{1}{t^k}\nabla_{\Hb} \wt F_\rho(\Wb^k, \Hb^k)$ and 
$\wt F_\rho(\Wb, \Hb) \triangleq F(\Wb, \Hb) + \rho\sum_{j = 1}^{N}\oneb^{\top}\hb_j$ is the smooth component in \eqref{eqn: onmf prob PP nsmooth obj}.
Problem \eqref{eqn: H update nonsmooth} is a proximal operator associated with the non-smooth and concave function $-\rho\sum_{j = 1}^N\|\hb_j\|_\infty$. Intriguingly, while being non-convex, the proximal operator \eqref{eqn: H update nonsmooth} has a simple closed-form solution as stated below.

\begin{Prop} \label{prop: prox map negative infinity norm}
	Consider the following problem
	\begin{subequations}\label{eqn: prox prob}
		\begin{align}
			\min_{\xb \in \mathbb{R}^{n}} &~\frac{1}{2}\|\xb - \yb\|_{2}^{2} - c\|\xb\|_{\infty}~~\\
			{\rm s.t.}~~& \xb \geq \zerob,
		\end{align}
	\end{subequations}
	where $\yb=[y_1,\ldots,y_n]^{\top} \in \mathbb{R}^{n}$ is given and $c > 0$ is a scalar. Denote $\xb^{\star} = [x_1^{\star}, \dots, x_{n}^{\star}]^{\top}$ as an optimal solution of \eqref{eqn: prox prob}, and let $i^{\star}$ be the unique index such that $\|\xb^{\star}\|_{\infty} = (\xb^{\star})_{i^{\star}}$. Then,  $\displaystyle i^{\star} \in  \arg\max_{i=1,\ldots,n}y_i$ and
	\begin{align} \label{prop: the optimal solution}
		(\xb^{\star})_i = \begin{cases} (y_i + c)^+, ~{\rm if} ~i = i^{\star}, \\
			(y_i)^+, ~{\rm otherwise},
		\end{cases}
	\end{align}	for $i=1,\ldots,n$.
\end{Prop}

Proposition \ref{prop: prox map negative infinity norm} is proved in Appendix \ref{appdix: proof of Prop nonsmooth}. 
By applying Proposition \ref{prop: prox map negative infinity norm} to \eqref{eqn: H update nonsmooth}, we obtain the PALM method for solving problem \eqref{eqn: onmf prob PP nsmooth} in Algorithm \ref{table: non-smooth ALM}.

We show in the following theorem that Algorithm \ref{table: non-smooth ALM} can yield a d-stationary solution of the penalized problem \eqref{eqn: onmf prob PP nsmooth}, as desired by Theorem \ref{thm: exact penalty}. Denote  $ L_{\wt F}(\Wb^k)$ and $L_F(\Hb^k)$ as the Lipschitz constant of $\nabla_{\Hb}\wt F(\Wb^{k}, \Hb)$ and $\nabla_{\Wb}F_{\rho}(\Wb, \Hb^{k})$, respectively. 

\begin{Theorem} \label{thm: convergence of non-smooth ALM}
	Let $\{\Wb^k, \Hb^k\}$ be the sequence generated by Algorithm \ref{table: non-smooth ALM} with $t^k > L_{\wt F}(\Wb^k)$ and $c^k > \frac{L_F(\Hb^k)}{2}$. Then $\{F_{\rho}(\Wb^k, \Hb^k)\}$ is non-increasing, $\{\Wb^k, \Hb^k\}$ is bounded, and 
	$\{\Wb^k, \Hb^k\}$ converges to a d-stationary point of \eqref{eqn: onmf prob PP nsmooth}.
\end{Theorem}

The proof is presented in Appendix \ref{appdix: proof of conv of non-smooth ALM}.  By combining Theorem \ref{thm: exact penalty} and Theorem \ref{thm: convergence of non-smooth ALM}, we see that the NSNCP method with Algorithm \ref{table: SNCP} and Algorithm \ref{table: non-smooth ALM} can yield a feasible B-stationary solution of 
problem \eqref{eqn: onmf prob P2}.

We should emphasize again that Proposition \ref{prop: prox map negative infinity norm} is critical to satisfy the condition of Theorem \ref{thm: exact penalty} since it enables Algorithm \ref{table: non-smooth ALM} to output a d-stationary solution of problem \eqref{eqn: onmf prob PP nsmooth} as stated in Theorem \ref{thm: convergence of non-smooth ALM}.
On the contrary,  if one uses the subgradient method for problem \eqref{eqn: onmf prob PP nsmooth}, it can only achieve a critical point defined based on the subdifferential of the non-smooth terms \cite[Theorem 1]{PALM_2014}, which is weaker than the d-stationary point due to the difference-of-convex (DC) objective function of problem \eqref{eqn: onmf prob PP nsmooth}; see \cite{Pang_2017} for detailed discussions.


\begin{algorithm}[t!]
	\caption{PALM for solving penalized problem \eqref{eqn: onmf prob PP nsmooth}.}
	\begin{algorithmic}[1]\label{table: non-smooth ALM}
		\STATE {\bf Set} $k = 0, \Wb^{0} = \Wb^{(r-1)},~\Hb^{0} = \Hb^{(r-1)}$.	
		\REPEAT
		\STATE 
		$\displaystyle \Hb^{k+1} =(\Bb^{k+1} + \frac{\rho}{t^k}\Zb^{k+1})^+ ,$\\
		where 
		$$[\Zb^{k+1}]_{ij} = \begin{cases}
			1 & {\rm if~}i=i^\star_j, \\
			0 & \text{otherwise}.
		\end{cases},~\forall i=1,\ldots,M,j\in \Nc,$$
		and $i^\star_j$ is any one of the indices in $\displaystyle \arg\max_{\ell=1,\ldots,K}[\Bb^{k+1}]_{\ell j}$.		
		\STATE  $	\Wb^{k+1} =(\Wb^{k} - \frac{1}{c^{k+1}}\nabla_{\Wb}F_{\rho}(\Wb^{k}, \Hb^{k+1}))^+, $\\
		
		\STATE Set $k =k + 1$,
		\UNTIL a predefined stopping criteria is satisfied.
		
	\end{algorithmic}
\end{algorithm}


Before ending the section, we have the following remarks.

\begin{Rmk}\label{rmk pqv}{\rm (Comparison between SNCP and NSNCP)
		As presented in Section \ref{sec: proposed ncp}, the NSNCP method has a stronger theoretical result since it can guarantee exact penalty without necessarily achieving a local minimum solution of the penalized problem  \eqref{eqn: onmf prob PP}. However, numerically we found the two methods in fact perform comparably to each other in terms of data clustering; see Section \ref{sec: sim results} for comparison details of these two versions of the penalty approach.
		Moreover, both methods can often yield more favorable results than the classical K-means, and outperform the existing ONMF  methods in terms of both clustering performance and computation time. This will be discussed  in Section \ref{sec: sim results}.
	}
\end{Rmk}

\begin{Rmk}\label{efficient}{\rm (Complexity comparison with existing ONMF methods) 
		As discussed in \cite{HALS_2015}, the DTPP \cite{Ding_Orth_2006},  ONMF-S \cite{HALS_2015} have the same per-iteration complexity of the order $\mathcal{O}(MNK + K^2M)$, and the HALS \cite{HALS_2015} has the per-iteration complexity of the order  $\mathcal{O}(MNK + K^2(M + N))$ whereas ONP-MF \cite{Pomplili_2014} has a much higher per-iteration complexity which is no less than $\mathcal{O}(N^3 + KN^2 + MNK)$ due to the SVD of a $K \times N$ matrix.
		%
		One can easily verify that the per-iteration complexity of both the proposed Algorithms 2 and 3 is $\mathcal{O}(MNK + K^2(M + N))$,  which is comparable to the complexity order of DTPP, ONMF-S and HALS since $K\ll \min\{M,N\}$. 
		In Section \ref{sec: sim results}, we will further demonstrate that the proposed SNCP/NSNCP methods have faster convergence speed than most of the existing ONMF methods,  and therefore are computation-time more efficient.

	}
\end{Rmk}

\section{Experiment Results} \label{sec: sim results}

In this section, we examine the clustering performance of the proposed SNCP and NSNCP methods against 6 existing clustering methods, namely, K-means (KM), K-means++ \cite{K++_2007}, DTPP \cite{Ding_Orth_2006}, ONP-MF \cite{Pomplili_2014}, ONMF-S \cite{Choi_ONMF_2008} and HALS \cite{HALS_2015}. Note that the later four methods are all based on the ONMF model \eqref{eqn: onmf prob}.
The adjusted Rand index (ARI) \cite{ARI_2011} and clustering accuracy (ACC) \cite{LCCF_2011} are adopted for performance evaluation, both of which are widely-used metrics for clustering validation.  ACC is the ratio of the number of correctly clustered data samples to the total number of samples.  A data sample is said to be correctly clustered if it is assigned to the same cluster as that in the ground truth.
To define ARI,  let $S = \{s_1, s_2, \ldots, s_n\}$ be  a set of data samples,  $C = \{c_1, \ldots, c_K\}$ be a clustering result obtained by an algorithm, and $L = \{l_1, \ldots, l_K\}$ is the clustering result by the ground truth. The ARI of the clustering $C$ is defined as

\vspace{-0.3cm}
{\small 
	\begin{align} 
		\frac{\sum\limits_{i,j}\dbinom{n_{ij}}{2} -\bigg[\sum\limits_{i}\dbinom{n_{i\cdot}}{2}\sum\limits_{j}\dbinom{n_{\cdot j}}{2}\bigg ]\bigg/\dbinom{n}{2}}{\frac{1}{2}\bigg[\sum\limits_{i}\dbinom{n_{i\cdot}}{2}+\sum\limits_{j}\dbinom{n_{\cdot j}}{2}\bigg ] -\bigg[\sum\limits_{i}\dbinom{n_{i\cdot}}{2}\sum\limits_{j}\dbinom{n_{\cdot j}}{2}\bigg ]\bigg/\dbinom{n}{2}},\notag 
\end{align}}

\noindent where $n_{i,j}$ is the number of samples that are in both cluster $c_i$ of $C$ and $l_j$ of $L$, $n_{i \cdot} = \sum_{j}n_{i,j}$ is the number of samples in cluster $c_i$ of $C$,  $n_{\cdot j} = \sum_{i}n_{i,j}$ is the number of samples in cluster $l_j$ of $L$.  A larger value of ARI indicates a better clustering result. 
However, due to limited space, we only present the results of ACC here and relegate the results of ARI in \cite[Section 4]{SPD}.
Both of these methods are implemented with python and have been uploaded to \url{https://github.com/wshuai317/NCP\_ONMF}.

\subsection{Performance with Synthetic Data} 
\label{sec: synthetic data}

We follow  the   linear model $\Xb = \Wb\Hb + \Vb$ in \cite{JNKM_2017}  to generate the synthetic data where $\Vb\in \mathbb{R}^{M\times N}$ denotes the measurement noise.
The signal to noise ratio (SNR) is defined as $10\log_{10}(\|\Wb\Hb\|_F^{2}/\|\Vb\|_F^{2})$ dB.
We follow the same procedure as in \cite{JNKM_2017} to generate $\Wb$, $\Vb$ and the cluster assignment matrix $\Hb$, with $M=2000$, $N=1000$ and $K=10$ (10 clusters). The number of data samples in the 10 clusters are $117, 62, 36, 124, 15, 24, 119, 43, 122$ and $338$, respectively. Like \cite{JNKM_2017}, $5\%$ of the data samples are replaced by randomly generated outliers. 

{\bf Parameter setting:} 
In Algorithm \ref{table: SNCP}, the initial penalty parameter $\rho$ is set to $10^{-8}$. For Step 4 of Algorithm \ref{table: SNCP}, the orthogonality of $\Hb^{(r)}$ is measured by
\begin{align}
	&{\rm (Orthogonality)}~\epsilon_{\rm orth} =\dfrac{\|\Qb^{(r)}\Hb^{(r)}(\Qb^{(r)}\Hb^{(r)})^{\top}- \Ib_K\|_F}{K^2},
\end{align}
where $\Qb^{(r)}$ is a diagonal matrix such that rows of $\Qb^{(r)}\Hb^{(r)}$ have unit 2-norm.
One updates $\rho=\gamma \rho$ in Step 4 of Algorithm \ref{table: SNCP} whenever $\epsilon_{\rm orth} \geq 10^{-10}$.
Besides, we define
\begin{align}\label{normal residual}
	&\text{(Normalized Residual)~}  \epsilon_{\rm NR}= \dfrac{\|\Wb^{(r)} - \Wb^{(r-1)}\|_F}{\|\Wb^{(r-1)}\|_F} + \dfrac{\|\Hb^{(r)} - \Hb^{(r-1)}\|_F}{\|\Hb^{(r-1)}\|_F}. 
\end{align}	
The stopping condition of Algorithm \ref{table: SNCP} is when both $\epsilon_{\rm orth}$ and $\epsilon_{\rm NR}$ are sufficiently small.	

For Algorithm \ref{table: smooth ALM}, $t^k$ and $c^{k+1}$ are chosen as $\frac{1}{2}\lambda_{\max}(\nabla_{\Hb}^2 G_{\rho}(\Wb^k, \Hb^k))$  and $\frac{1}{2}\lambda_{\max}(\nabla_{\Wb}^2 G_{\rho}(\Wb^k, \Hb^{k+1}))$, respectively; while for Algorithm \ref{table: non-smooth ALM}, $t^k$ and $c^{k+1}$ are chosen as $\lambda_{\max}(\nabla_{\Hb}^2 \wt F(\Wb^k, \Hb^k))$ and $\frac{1}{2}\lambda_{\max}(\nabla_{\Wb}^2 F_{\rho}\\(\Wb^k, \Hb^{k+1}))$, respectively. The stopping condition for Algorithm \ref{table: smooth ALM} and Algorithm \ref{table: non-smooth ALM} is the normalized residual of $(\Wb^k,\Hb^k)$ which is defined in the same way as \eqref{normal residual} and is denoted as $\epsilon_{\rm PALM}$.

If not mentioned specifically, we choose $\mu_w = 0$ and $\mu_h = 10^{-10}$ in \eqref{eqn: onmf prob P2}. The parameter $\gamma$ for increasing $\rho$ is set to 1.1. When it comes to the stopping condition, if not mentioned specifically, for the SNCP method, we set $\max\{\epsilon_{\rm orth},\epsilon_{\rm NR}\} \leq 10^{-5}$ for Algorithm \ref{table: SNCP} and $\epsilon_{\rm PALM} < 3 \times 10^{-3}$ for Algorithm \ref{table: smooth ALM}; for the NSNCP method, we set $\max\{\epsilon_{\rm orth},\epsilon_{\rm NR}\}\leq 10^{-3}$ for Algorithm \ref{table: SNCP} and $\epsilon_{\rm PALM} < 3 \times 10^{-3}$ for Algorithm \ref{table: non-smooth ALM}. For the four ONMF methods under comparison,  the stopping condition is  $\epsilon_{\rm NR} < 10^{-5}$ or the maximum iteration number of 2000 is achieved. All algorithms under test are initialized with 10 common, randomly generated initial points, and the presented results are averaged over the 10 experimental trials.

{\bf Effect of $\mu_w$ and $\mu_h$ as well as $\epsilon_{\rm PALM}$ and $\gamma$:}  The parameters $\mu_w$ and $\mu_h$ as well as $\epsilon_{\rm PALM}$ and $\gamma$ do have impact on the algorithm convergence. To justify our choices above, we conducted extensive experiments with different combinations of these parameters. The numerical results are presented in \cite[Section 1]{SPD} of the supplementary material. The messages that one can infer from these experiments are that 1) smaller values of $\mu_w$ and $\mu_h$ are preferred, and 2) a larger $\gamma$ can speed up to satisfy the orthogonality of $\Hb$, but requires a smaller value of $\epsilon_{\rm PALM}$ in order to achieve a lower objective value of \eqref{eqn: onmf prob P2}. Interested readers may refer to the supplementary material \cite{SPD} for more discussions.

\begin{figure} [t!]
	\centering
	\subfigure[]{\label{fig:nr_comp}
		\includegraphics[width=7cm]{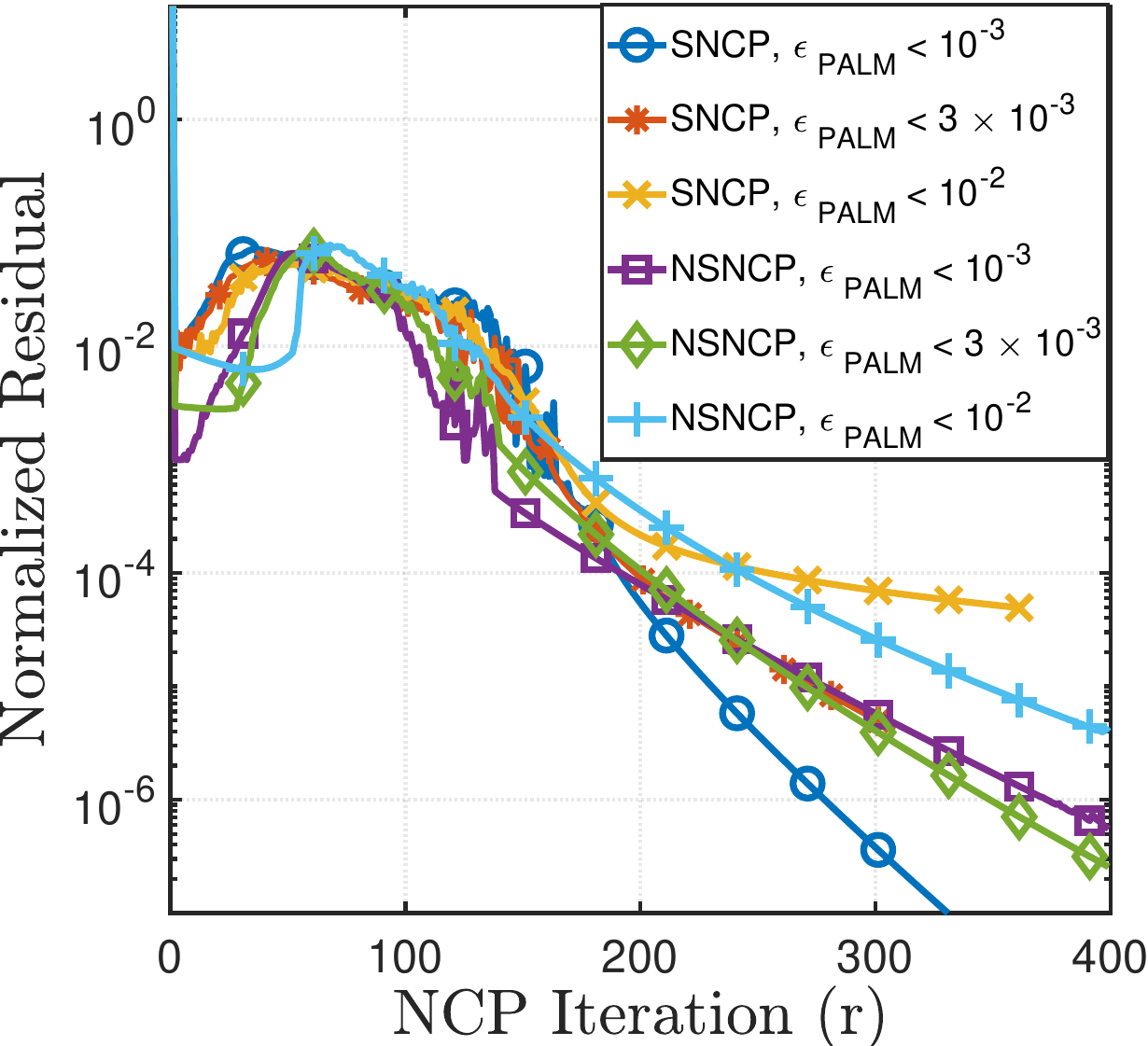}
	}
	\subfigure[]{ \label{fig:ortho_comp}
		\includegraphics[width=7cm]{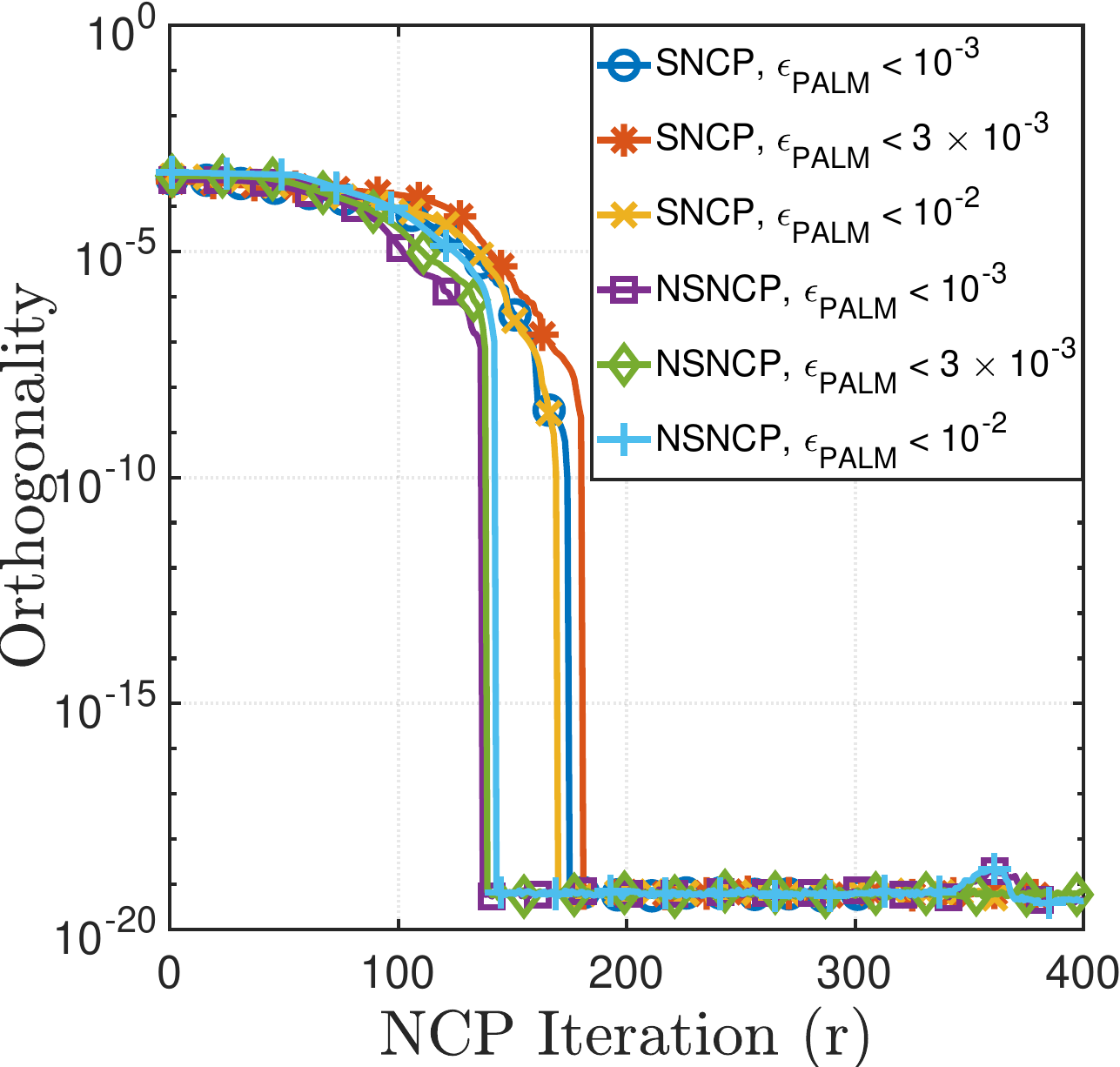}
	}
	\vspace{-0.4cm}
	\caption{Convergence curves of normalized residual and orthogonality achieved by the proposed SNCP and NSNCP methods for the synthetic data with SNR = -3 dB.}
	\label{fig:converge_comp}\vspace{-0.3cm}
\end{figure}

{\bf SNCP vs NSNCP:}
Let us first examine the convergence behaviors of the proposed SNCP and NSNCP methods. Fig. \ref{fig:nr_comp} and Fig. \ref{fig:ortho_comp} respectively display the normalized residual and orthogonality achieved versus the iteration number of Algorithm \ref{table: SNCP} when the SNCP and NSNCP methods are used. One can see from Fig. \ref{fig:nr_comp} that both SNCP and NSNCP converge and they can converge faster for smaller values of $\epsilon_{\rm PALM}$. As seen in Fig. \ref{fig:ortho_comp}, both methods indeed can achieve an orthogonal $\Hb$. Moreoever, it can be observed that the NSNCP usually converge faster, and it takes about 130 iterations to reach $\epsilon_{\rm orth} < 10^{-15}$, which is faster than the SNCP method.

\begin{figure}[t!]
	\centering
	\subfigure[]{ \label{fig:acc_comp(b)}
		\includegraphics[width=7cm]{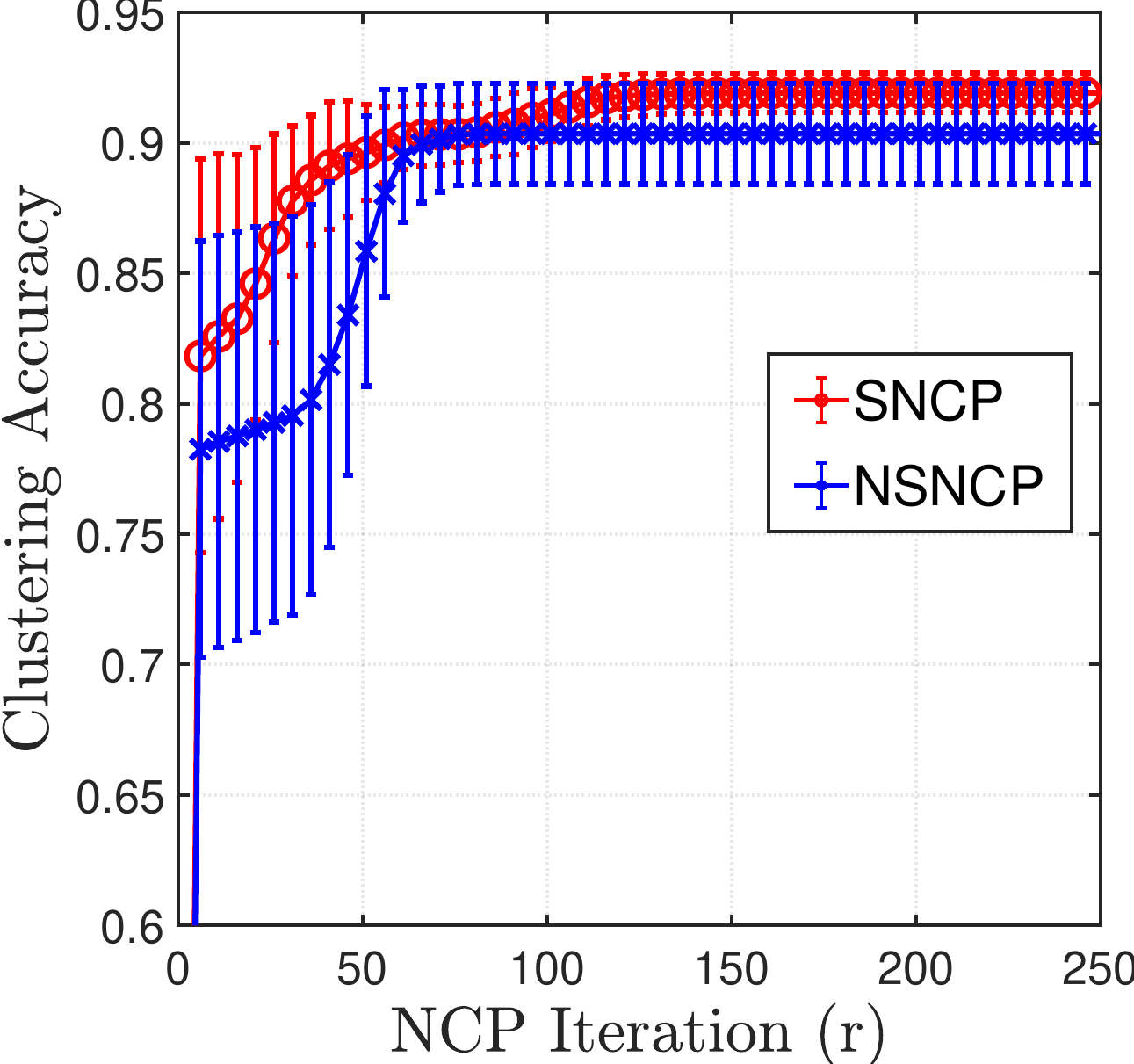}
	}
	\subfigure[]{ \label{fig:acc_comp(a)}
		\includegraphics[width=7cm]{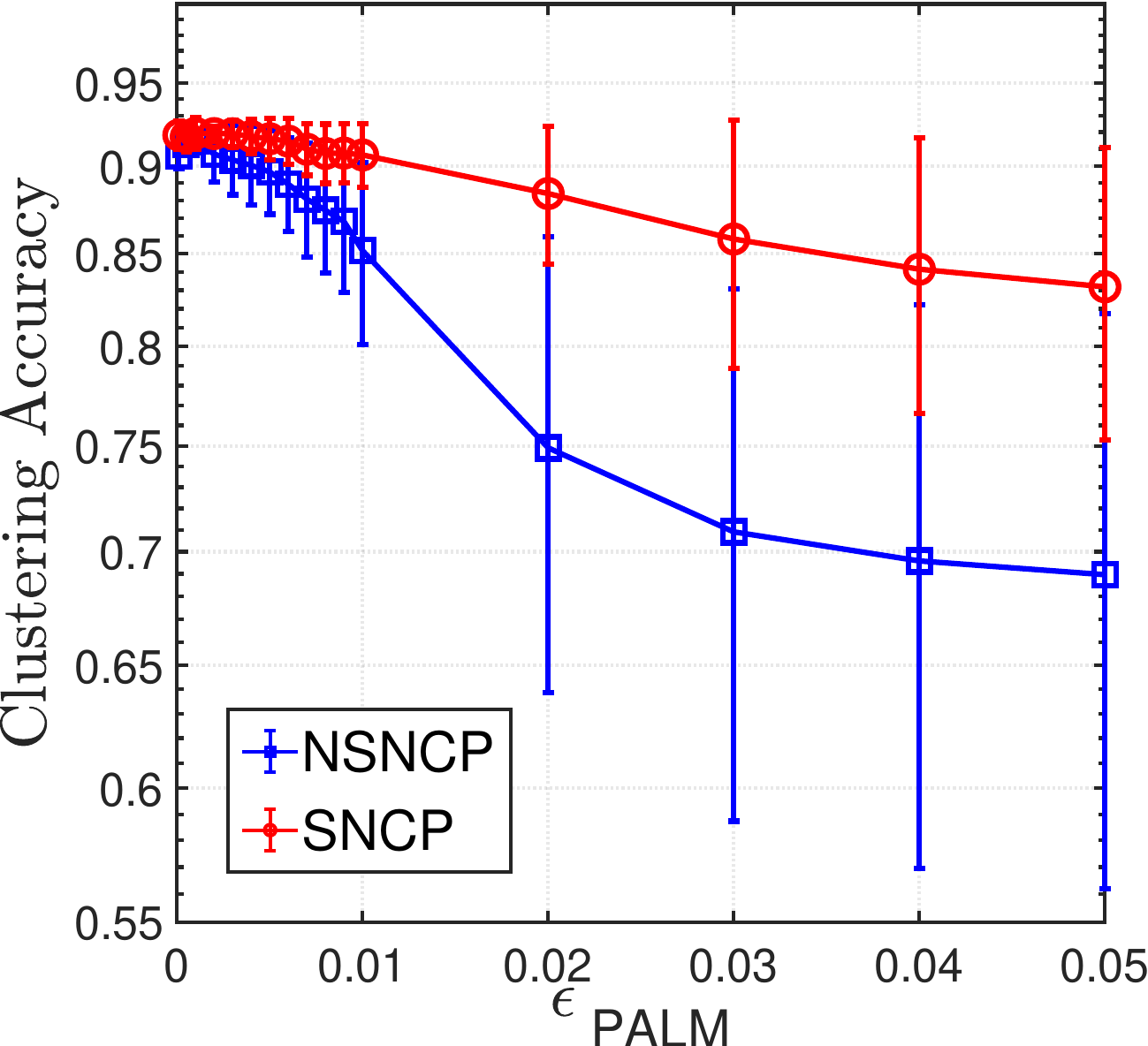}
	}
	\vspace{-0.2cm}
	\caption{Comparison of the proposed SNCP and NSNCP methods in terms of clustering accuracy. Each point is averaged over 10 runs with different initials. The error bar shows the standard deviation of the 10 results. }
	\label{fig:acc_comp}
	\vspace{-0.4cm}
\end{figure}

To further examine the difference between the SNCP and NSNCP methods, 
we plot in Fig. \ref{fig:acc_comp(b)} the convergence curves of clustering accuracy (ACC) versus the iteration number of Algorithm \ref{table: SNCP} for both the SNCP and NSNCP methods, on the synthetic data with SNR = $-3$ dB. The results are averaged over 10 experiments, each of which uses a different initial point for Algorithm \ref{table: SNCP}. 
The error bars (dashed line) are the standard deviation of the 10 experimental results.  Intriguingly, one can observe that the NSNCP method actually converges faster than the SNCP method in terms of clustering accuracy, which echos Theorem \ref{thm: exact penalty} that the NSNCP method requires a finite $\rho$ only to achieve a feasible and meaningful solution of \eqref{eqn: onmf prob P2}. However, as seen from the figure, the SNCP method eventually can reach a slightly higher clustering accuracy than the NSNCP method on the synthetic data.

In Fig. \ref{fig:acc_comp}(b), we further present the clustering accuracy (ACC) versus the stopping condition $\epsilon_{\rm PALM}$ of these two methods.  
One can observe that the clustering accuracy improves with a smaller $\epsilon_{\rm PALM}$ for both methods. 
We also see that the performance gap between the two methods reduces with smaller $\epsilon_{\rm PALM}$, and when $\epsilon_{\rm PALM} < 10^{-3}$, they yield comparable clustering performance. This implies that the NSNCP method may require a more stringent $\epsilon_{\rm PALM}$ in order to reach a desired clustering performance.

%

\begin{figure}[t!]
	\centering
	\subfigure[Syn, SNR = -3 dB]{ \label{fig: iter_comp_snr-3}
		\includegraphics[width=7cm]{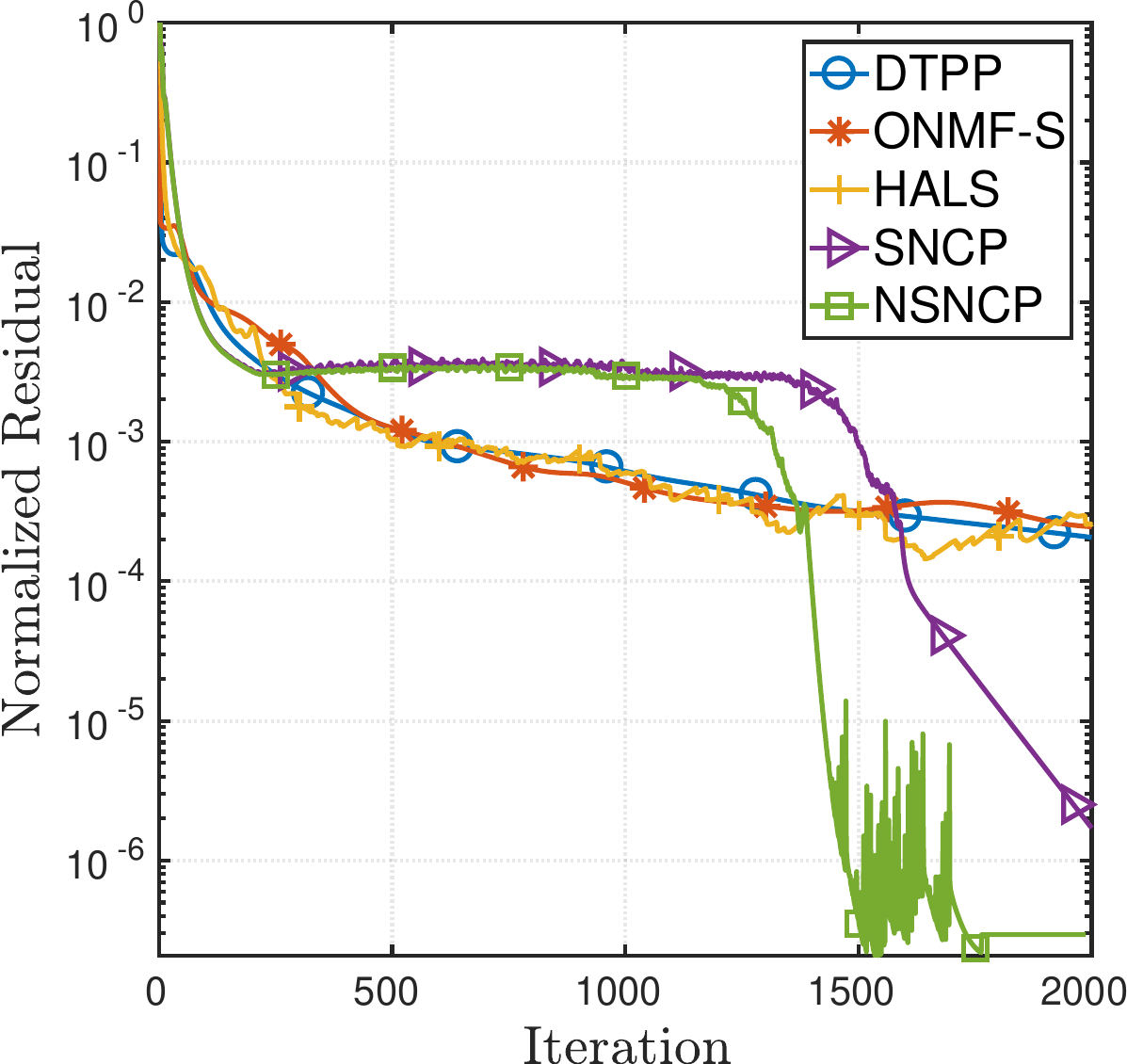}
	}
	\subfigure[Syn, SNR = -1 dB]{ \label{fig: iter_comp_snr-1}
		\includegraphics[width=7cm]{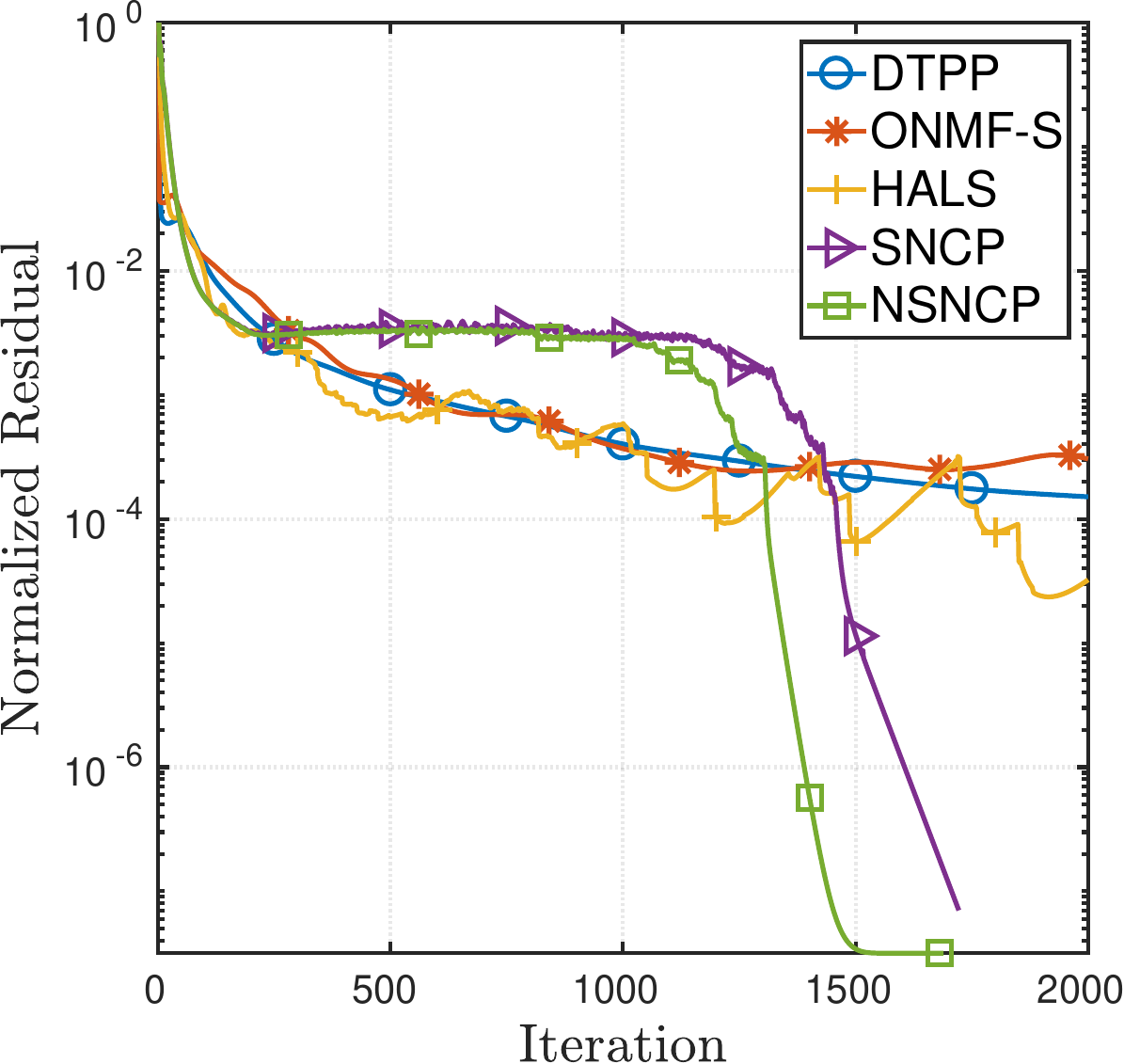}
	}
	\vspace{-0.4cm}
	\caption{Convergence curves of normalized residual achieved by the proposed methods and the ONMF methods for the synthetic data with SNR = -3 dB and SNR = -1 dB. Each point is averaged over 10 runs with different initials.}
	\label{fig:iter_conv}
	\vspace{-0.2cm}
\end{figure}

{\bf Comparison with the existing ONMF methods:} 
We compare the convergence speed of the proposed SNCP/NSNCP methods with the ONMF methods. Fig. \ref{fig: iter_comp_snr-3} and Fig. \ref{fig: iter_comp_snr-1} respectively show the average normalized residual achieved versus the (total) iteration number.  For the SNCP/NSNCP,  the iteration number refers to the accumulated iteration number of Algorithm \ref{table: smooth ALM} (resp. Algorithm \ref{table: non-smooth ALM}) when the outer loop Algorithm \ref{table: SNCP} is applied.  It can be observed from Fig. \ref{fig:iter_conv} that these traditional ONMF methods suffer from slow convergence, whereas the SNCP/NSNCP
can converge faster although at the beginning their movements are not so fast.
Table \ref{table: iter_syn} further shows the average iteration numbers of all methods under test to reach the normalized residual $\epsilon_{\rm NR} < 10^{-5}$.  One can see that DTPP and ONMF-S both require more than 2000 iterations while the HALS requires less iteration numbers on average.
The SNCP/NSNCP both require less iteration numbers than the other three ONMF methods, and the NSNCP has the least.


%
%

{\bf Clustering performance:} 
Table \ref{table: clustering quality} lists 
the average clustering performance (ACC) of the eight methods under test on the synthetic data with different SNR values. All results are obtained by averaging over  20 experiments. 
In each experiment, all methods use the same randomly generated initial point. 


First of all, one can observe that K-means++ does not perform better than the K-means for the synthetic dataset.
The ONMF based methods (i.e., DTPP, ONP-MF, ONMF-S, HALS, and proposed SNCP/NSNCP) significantly outperform the K-means and K-means++.
Nevertheless, one can see from Table \ref{table: clustering quality}  that the proposed SNCP and NSNCP consistently yield the best clustering performance.
\vspace{-0.0cm}

{ \begin{table}[t!] 
		\centering 
		\caption{Average clustering performance (\%) and CPU time (s) on the synthetic data for different values of SNR. 
		}\vspace{-0.2cm}
		\setlength{\tabcolsep}{3.0mm}
		\label{table: clustering quality}
		\begin{tabular}{|c|c|c|c|c|c|c|}
			\hline \rowcolor{gray!50}  
			\multicolumn{2}{|c|}{SNR (dB)}              
			&       -5       &       -3       &       -1       &       1        &       3      	       \\ \hline\hline

			\multirow{9}{0.7cm}{ACC}   
			&   KM    &     63.4      &     69.7     &     74.7      &     74.3      &     75.6            \\ \cline{2-7}
			&     KM++      &     64.1      &     70.1      &     70.3      &     70.8      &     69.2           \\ \cline{2-7}
			&     DTPP     &     81.6      &     85.1      &     85.9      &     87.0      &     86.6           \\ \cline{2-7}
			&     ONP-MF     &     66.8      &     88.3      &   83.1  &     89.9      &     90.0            \\ \cline{2-7}
			&     ONMF-S     &     77.4      &     79.0      &     80.2      &     81.3      &     81.7           \\ \cline{2-7}
			&     HALS     &     76.3      &     86.0      &     88.1      &     89.6      &     89.4          \\ \cline{2-7}
			&     SNCP     & \textbf{91.5} & \textbf{91.9} &  \textbf{92.0} & \textbf{92.5}& \textbf{92.8}   \\
			\cline{2-7}
			& NSNCP     &  90.1 &   90.7 &91.8 &  92.0 &  92.2  \\ \hline\hline
			\multirow{9}{0.7cm}{Time}   
			&   KM    &     3.10      &     1.96     &     1.52      &     1.16      &     1.11         \\ \cline{2-7}
			&     KM++      &  3.09    &   2.44        &  2.05         &  1.70         &  1.59            \\ \cline{2-7}
			&     DTPP     &     30.4      &     34.0      &     38.6      &     31.1      &     35.1         \\ \cline{2-7}
			&     ONP-MF     &     1097      &     1123      &     1124    &     1153      &     1148          \\ \cline{2-7}
			&     ONMF-S     &     68.4      &     97.1      &     116      &     116      &     90.5           \\ \cline{2-7}
			&     HALS     &     22.6      &     25.6      &     23.3      &     15.7      &     19.3          \\ \cline{2-7}
			&     SNCP     &  20.6 &  15.6 &  13.4 & 13.3&  12.4 \\ \cline{2-7}
			&     NSNCP    &  14.8 &  15.5 &  11.5 & 10.5 &  10.6  \\ \hline		
		\end{tabular}
		\vspace{-0.4cm}
\end{table}}

\begin{table}[t!] 
	\centering 
	\caption{Average iteration number on the synthetic datasets for different values of SNR.} 
	\vspace{-0.2cm}
	\setlength{\tabcolsep}{3mm}
	\label{table: iter_syn}
	\begin{tabular}{|c|c|c|c|c|c|c|}
		\hline \rowcolor{gray!50} 
		\multicolumn{2}{|c|}{SNR (dB)}              
		&       -5       &       -3       &       -1       &       1        &       3      	       \\ \hline\hline
		\multirow{5}{1cm}{\#Iteration}   
		&     DTPP     &   2000      &  2000        &   2000      &   2000        &      2000       \\ \cline{2-7}
		&     ONMF-S     &    2000       &  2000       &      2000    &      2000    &           2000     \\ \cline{2-7}
		&     HALS     &    1931      &     1788     &    1521     &     1442      &     1325          \\ \cline{2-7}
		&     SNCP     &  1921 &  1601 &  1401 & 1310 &  1260 \\ \cline{2-7}
		&     NSNCP    &  1478 &  1322 &  1130 & 1063 &  1047  \\ \hline		
	\end{tabular}
	\vspace{-0.45cm}
\end{table}

{\bf Computational time:} 
Table \ref{table: clustering quality} also shows the CPU time taken by each method. 
The computer used in the experiments has a Ubuntu 16.04 OS, and equipped with 3.40 GHz Intel Core i7-6700 CPU and 52 GB RAM.
As seen, other than the K-means and K-means++ which are well known computationally cheap,
the proposed SNCP and NSNCP methods are more computationally time efficient than the other five ONMF based methods.
In particular, while the ONP-MF can provide competitive clustering performance when SNR $\geq 1$ dB, its computation time is long.
By contrast, the HALS can provide reasonably good clustering performance when SNR $\geq 1$ dB and its computation time is moderate. 
Lastly, we note that the NSNCP method is slightly faster than the SNCP method, though the latter can provide the best clustering performance.
All these results are consistent with the discussion in Remark \ref{efficient} and the results in Table \ref{table: iter_syn}.


{\bf Clustering stability:} 
We evaluate the stability of the clustering methods against different initial points. In particular, we adopt the consensus map and cophenetic correlation (CC) coefficient \cite{Brunet_2004} to measure the stability. 
The consensus map is based on the consensus matrix ${C} \in \mathbb{R}^{n \times n}$, which has each entry $[C]_{i,j} = 1$ if sample $i$ and sample $j$ is assigned to the same cluster, and entry $[C]_{i,j} = 0$ otherwise.  
Then the consensus map, denoted by $\bar C$, is the heatmap of the average of consensus matrices obtained by 10 runs of experiments with different initial points.
Thus, roughly speaking, in the consensus map $\bar C$, the $(i,j)$th entry will be close to 1 if sample $i$ and sample $j$ are consistently assigned to the same cluster even under different initial conditions.  
The CC coefficient (between 0 and 1) is a qualitative measure of the consensus matrix and it is defined as the Pearson correlation of two distance matrices of data samples: one is given by using the off-diagonal entries of $\bar{C}$ and 
the other one utilizes the cophenetic distances of samples after performing average linkage hierarchical clustering; details can be referred to \cite{Dendrogram_1962}.  The CC coefficient would approach 1 if the consensus map $\bar{C}$ have entries closer to $0$ or $1$.

\begin{figure}[t!]
	\centering
	\includegraphics[width=8cm]{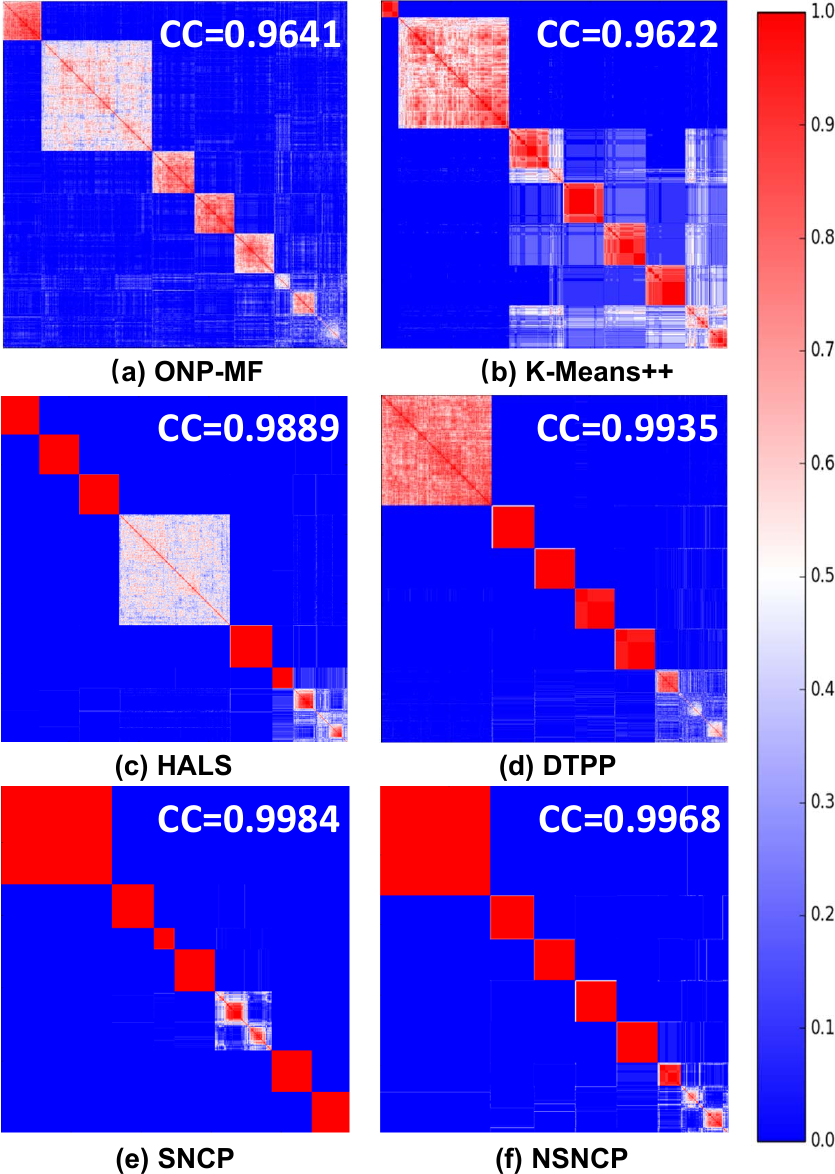}
	\caption{The consensus map of clustering results for the synthetic data with SNR = -5 dB. 
	}
	\label{fig:cons_map}
	\vspace{-0.6cm}
\end{figure}

Fig. \ref{fig:cons_map} shows the results for SNR = $-3$ dB, and one can see that the proposed SNCP method gives the stablest clustering results, which is followed by the NSNCP method. In particular,  clustering inconsistency only happens in small-sized clusters for these two methods.


\vspace{-0.1cm}

\subsection{Application to Biological Data Analysis} 
In the experiment, we apply the various clustering methods to the The Cancer Genome Atlas (TCGA) database \cite{website_TCGA,TCGA_CGCD}, which contains the expression data of 20531 genes on 11135 cancer samples belonging to 33 cancer types. We consider 5 subsets of the TCGA dataset with different numbers of cancer types as shown in the 3nd row of Table \ref{table: clustering TCGA data}. 
The Pearson’s Chi-Squared Test \cite{TELSL2009} is applied to select 5000 genes ($M=5000$) for each data sample. 

For the proposed SNCP method and NSNCP method, an additional constraint is added to upper bound each entry of $\Wb$ by the maximum value of the data.  
The other parameters are set the same as that in Section \ref{sec: synthetic data}.

{\bf Performance comparison:}
One can see from Table \ref{table: clustering TCGA data} that the proposed NSNCP method performs best on the TCGA data, especially for the first three datasets. The SNCP method  provides very close performances as the NSNCP method on datasets 2, 3 and 4. Compared with the ONMF based methods, the K-means and K-means++ yield relatively poor performance on this TCGA data. 
In terms of computation time, the HALS is still the most time efficient one, especially when the sample size is small. However, there exists considerable performance gap between HALS and the proposed NCP methods and the computation advantage of HALS can diminish when the data size increases. Lastly, one can observe from the table that the other three ONMF methods are quite computationally expensive, particularly the ONP-MF and ONMF-S.

\begin{table}[t!] 
	\centering 
	\caption{Average clustering performance ($\%$) and CPU time (s) on the TCGA data. } 
	\setlength{\tabcolsep}{3mm}
	\label{table: clustering TCGA data}\vspace{-0.2cm}
	\begin{tabular}{|c|c|c|c|c|c|c|}
		\hline \rowcolor{gray!50}  
		\multicolumn{2}{|c|}{Dataset}              
		&       1       &       2       &       3       &       4        &       5            \\ \hline
		\multicolumn{2}{|c|}{\#samples $N$}              
		&       1667       &       3086       &    3660    &  5314       &       11135          \\ \hline
		\multicolumn{2}{|c|}{\#cancers $K$}              
		&       5       &       10       &    15    &  20       &       33           \\ \hline	\hline
		\multirow{8}{0.7cm}{ACC}
		&   KM    &     75.0      &   67.0      &   57.0      &  52.2      &    34.4    \\ \cline{2-7}
		&   KM++    &    75.5      &    55.3      &    53.7      &     48.4     &    34.5      \\\cline{2-7}   
		&   DTPP    &     79.2      &     58.5      &    58.0      &     57.0      &  \textbf{43.1}  \\ \cline{2-7}
		&     ONMF-S      &    85.8  &   71.8       &   61.9       & 58.2       &     38.4       \\ \cline{2-7}
		&     ONP-MF      &  84.0      &  68.5    &    48.1    &  32.9   & 14.8  \\ \cline{2-7}
		&     HALS     &  86.2  & 68.8   &  56.1   & 56.3   & 39.1 \\ \cline{2-7}
		&     SNCP     & 85.6 & 79.3 & 64.0 & 61.1 &41.1 \\ \cline{2-7}
		&     NSNCP     & \textbf{89.2} & \textbf{81.2} & \textbf{68.2} & \textbf{64.3} & 42.7 \\ \hline	\hline
		\multirow{8}{0.7cm}{Time}
		&   KM    &    3.02      &    7.54     &   12.3      &   34.7      &     115     \\ \cline{2-7}
		&   KM++    &     3.08      &     7.83      &     26.9      &  40.3  &     331         \\\cline{2-7}   
		&   DTPP    &   249      &  434    &    1357    &   2135      &     2454       \\ \cline{2-7}
		&     ONMF-S      & 89.9    &   772     &   1440     &  2492     &   19794  \\ \cline{2-7}
		&     ONP-MF      & 4886  &  10414       &   14232   &    16092  &       77222    \\ \cline{2-7}
		&     HALS     &     2.26     &   11.4      &    98.1     &  260     &   1605      \\ \cline{2-7}
		&     SNCP     & 41.6 &  249 & 307 & 486 & 1118 \\ \cline{2-7}
		&     NSNCP     & 22.1 & 153 & 240 & 526 &  1756 \\ \hline	
	\end{tabular}
	\vspace{-0.35cm}
\end{table}

\vspace{-0.2cm}
\subsection{Application to Document Clustering}

We here examine the performance of the proposed methods on the document dataset TDT2 corpus \cite{LCCF_2011} which consists of 10212 on-topic documents in total with 56 semantic categories. 
We extract 6 subsets, each of which contains 10 randomly picked categories ($K=10$). Each document sample is normalized and represented as a term-frequency-inverse-document-frequency (tf-idf) vector \cite{LCCF_2011}. 
The dimension of each test data is shown in the first three rows of Table \ref{table: clustering doc data}. 
In the experiment, we set $\mu_w =0$ and $\mu_h=10^{-8}$ for the proposed SNCP and NSNCP methods. 

{\bf Performance comparison:}
As seen from Table \ref{table: clustering doc data}, except for Datasets 5 and 6, the ONMF based methods yield better performance than the K-means/K-means++. In particular, K-means++ yields the highest clustering accuracy on Dataset 5, and K-means gives comparable performance on Dataset 6. A close inspection shows that for Dataset 5, the initial points picked by K-means++ happen to be close to the cluster centroids of the ground truth. 

In addition, we can see that, except for Datasets 3 and 5, the SNCP method provides higher accuracy than 
the ONMF-S/ONP-MF. Besides, as seem from Table \ref{table: clustering doc data}, the computation time of the ONMF-S/ONP-MF are large and can be 50 times slower than the SNCP method on Dataset 6. It is also seen that the HALS is most time efficient on this experiment although it only provides moderate clustering performance. 
While the NSNCP method does not perform as well as its smooth counterpart, it gives comparable clustering performance for the first four dataset, and is computationally time efficient compared to ONMF-S and ONP-MF.

\begin{table}[t!] 
	\centering 
	\caption{Average clustering performance ($\%$) and CPU time (s) on TDT2 data. } 
	\setlength{\tabcolsep}{2mm}
	\label{table: clustering doc data}\vspace{-0.1cm}
	\begin{tabular}{|c|c|c|c|c|c|c|c|}
		\hline \rowcolor{gray!50}  
		\multicolumn{2}{|c|}{Dataset}              
		&     1      &       2      &    3     &   4  &   5      &   6  \\ \hline	
		\multicolumn{2}{|c|}{\#terms $M$}              
		&       13133      &       24968       &      11079       &   20431    &   16067      &   29725      \\ \hline
		\multicolumn{2}{|c|}{\#docs $N$}              
		&       842       &      3292      &    631     &    1745       &       1079       &     4779       \\ \hline	\hline
		\multirow{7}{0.5cm}{ACC}
		&   KM    &     77.9      &     51.7      &     84.4      &     49.3      &     61.3      & 70.2  \\ \cline{2-8}
		&   KM++    &     73.5      &     48.9      &     84.7      &     47.1      &     \textbf{61.7}      & 65.7  \\
		\cline{2-8} 
		&   DTPP    &     70.0      &     50.4      &     77.4      &     52.1      &     45.1     & 66.6  \\ \cline{2-8}
		&     ONMF-S      &    81.8  &   52.0       &    83.6       & 59.6          &     50.9      &     68.8        \\ \cline{2-8}
		&     ONP-MF      &   85.3        &   46.1        &  \textbf{89.5}         &    60.9       &  59.4    &  66.8          \\ \cline{2-8}
		&     HALS     &  77.9   & 47.8   &  85.1   & 54.3   &  48.7  &  64.4  \\ \cline{2-8}
		&     SNCP     & \textbf{86.1} & \textbf{56.8} & 88.1 & \textbf{62.0} & 58.2 & \textbf{70.7}  \\ \cline{2-8}
		&     NSNCP     &  79.7 &  54.1 &  88.6 &  60.8 &  56.4 &  64.8   \\ \hline\hline 
		\multirow{7}{0.7cm}{Time}
		&   KM    &     13.6      &     195      &     6.82      &     65.0      &     27.4      &     318        \\ \cline{2-8}
		&   KM++    &     10.6      &     199      &     8.15      &    53.6  &     28.9      &     401       \\
		\cline{2-8}  
		&   DTPP    &     126      &     726     &     81      &    413      &     204      &   2282         \\ \cline{2-8}
		&     ONMF-S      & 405    &   8124     &   157     &   1915     &  633    &   17603   \\ \cline{2-8}
		&     ONP-MF      &    1200     &     7854      &     757      &     3660      &     1951      &  14302        \\ \cline{2-8}
		&     HALS     &     12.9      &    35.6      &     6.07      &   27.4      &    20.0      &     162        \\ \cline{2-8}
		&     SNCP     & 37.3 &  407 & 20.4 & 64.0 & 41.1  &  342  \\ \cline{2-8}
		&     NSNCP     &  24.7 &  335 &  16.3 & 50.0 &   37.6 &  424 \\ \hline	
	\end{tabular}
	\vspace{-0.4cm}
\end{table}

{\bf Clustering on dimension-reduced data:}
\begin{table}[t!] 
	\centering 
	\caption{Average clustering performance ($\%$) and CPU time (s) on dimension-reduced TDT2 data } 
	\label{table: doc data after DR}
	\setlength{\tabcolsep}{2.5mm}
	\label{table: clustering real data_DR}\vspace{-0.2cm}
	\begin{tabular}{|c|c|c|c|c|c|c|c|}
		\hline \rowcolor{gray!50}  
		\multicolumn{2}{|c|}{Dataset}              
		&       1       &       2       &       3       &       4        &       5        &       6        \\ \hline\hline 
		\multicolumn{2}{|c|}{Time of SC }              
		&       13.8       &    402       &       7.11       &      90.7       &     30.7        &       1114       \\ \hline	\hline 		
		\multirow{7}{0.7cm}{ACC}
		&   KM    &     81.6      &    69.0   &   83.1      &   86.3   &     89.1   & 80.2   \\ \cline{2-8}
		&   KM++    &     98.3      &    83.3      &     98.9     &     \textbf{99.6}      &     98.9      & 90.3  \\
		\cline{2-8}
		&  HALS    & 95.8   & 82.8  &  97.8  & 98.8  & 99.4   & 90.8   \\ \cline{2-8}	
		&   ONP-MF    &   98.6  &  82.3 &   \textbf{99.0}  &  99.3   &  \textbf{99.5}  & 89.7  \\ \cline{2-8}
		&   SNCP    & \textbf{99.0} & \textbf{84.3} & 98.0   &  99.2 & 99.3  & \textbf{92.7}  \\ \hline\hline 
		\multirow{7}{0.7cm}{Time}
		&   KM    &   0.24  & 1.34 &   0.17 &   0.45    &  0.33 & 2.33  \\ \cline{2-8}
		&   KM++    &  0.25     & 1.96 & 0.18  &     0.46  &   0.34   &  2.18    \\ \cline{2-8}
		&   HALS    &  0.78  &  2.45  & 0.81  & 1.65   & 0.75  & 1.52  \\ \cline{2-8}	
		&   ONP-MF    &     13.6  &  385  &   7.29     &   90.0   &     26.0      & 790  \\ \cline{2-8}
		&   SNCP    & 0.59 & 2.00 &0.55 & 1.31   & 1.21  & 3.17  \\ \hline
	\end{tabular}
	\vspace{-0.3cm}
\end{table}
Since the feature size of the TDT2 data is large and 
in view of the fact that dimension reduction techniques can extract low-rank structure of data and improve the clustering performance, we apply the spectral clustering (SC) \cite{Spectral_2007} to the TDT2 data followed by applying the various clustering methods to the dimension-reduced data. In the experiment, we set $\mu_w = \mu_h = 0$ and each entry of $\Wb$ is lower (resp. upper) bounded by the minimum (resp. maximum) value of the dimension-reduced data. The experimental results are displayed in Table \ref{table: doc data after DR}.
As observed, all methods have a great leap in the clustering performance when compared to Table \ref{table: clustering doc data}, and there is no significant performance gap between various methods. Nevertheless, the proposed SNCP method can provide competitive clustering results and performs best on Dataset 1, 2, and 6. Although the ONP-MF performs best on Dataset 3 and 5, it remains to be expensive in computation time. While the K-means++ gives the best performance on Dataset 4,  one can see that the proposed SNCP method is only slighter worse. 

\begin{figure} [t!]
	\centering
	\subfigure[\scriptsize TDT2\_2]{
		\includegraphics[width=5cm]{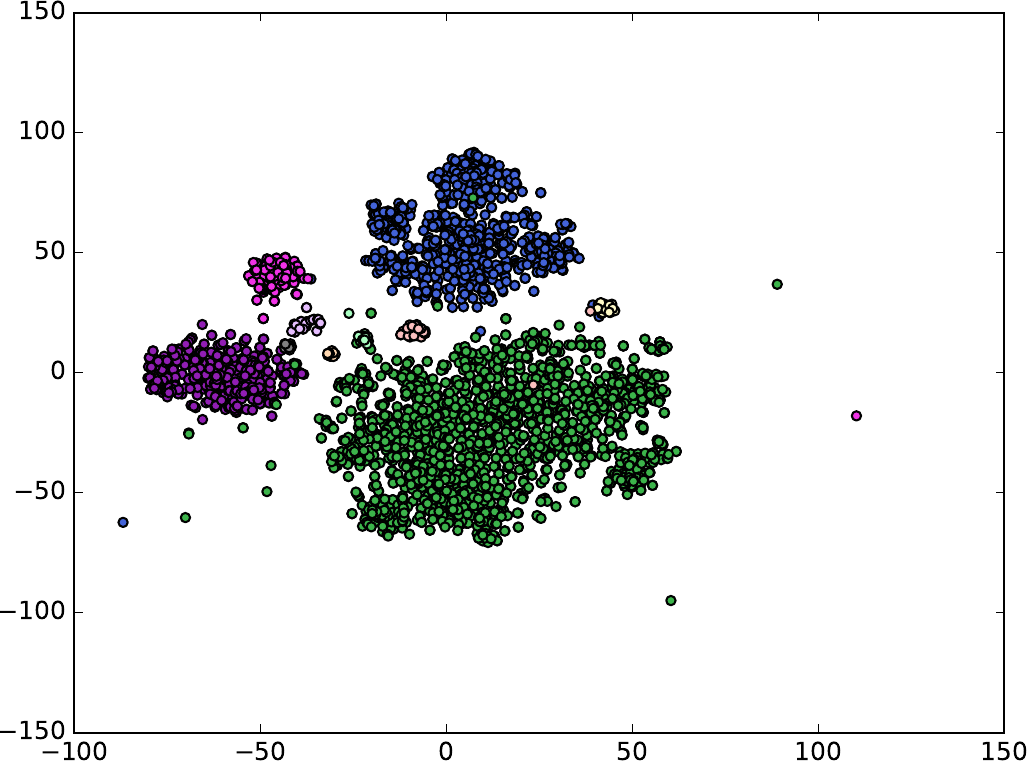}
		\label{fig:visualize_TDT2 a}
	}
	\subfigure[\scriptsize TDT2\_4]{
		\includegraphics[width=5cm]{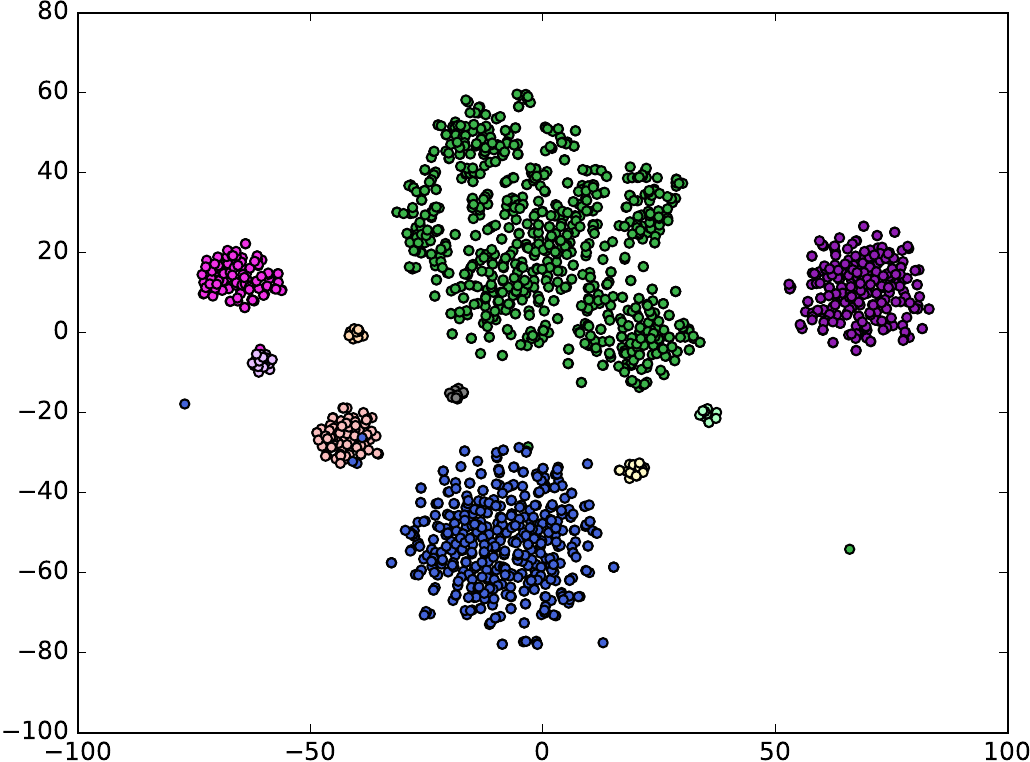}
		\label{fig:visualize_TDT2 b}
	}
	\subfigure[\scriptsize TDT2\_6]{
		\includegraphics[width=5cm]{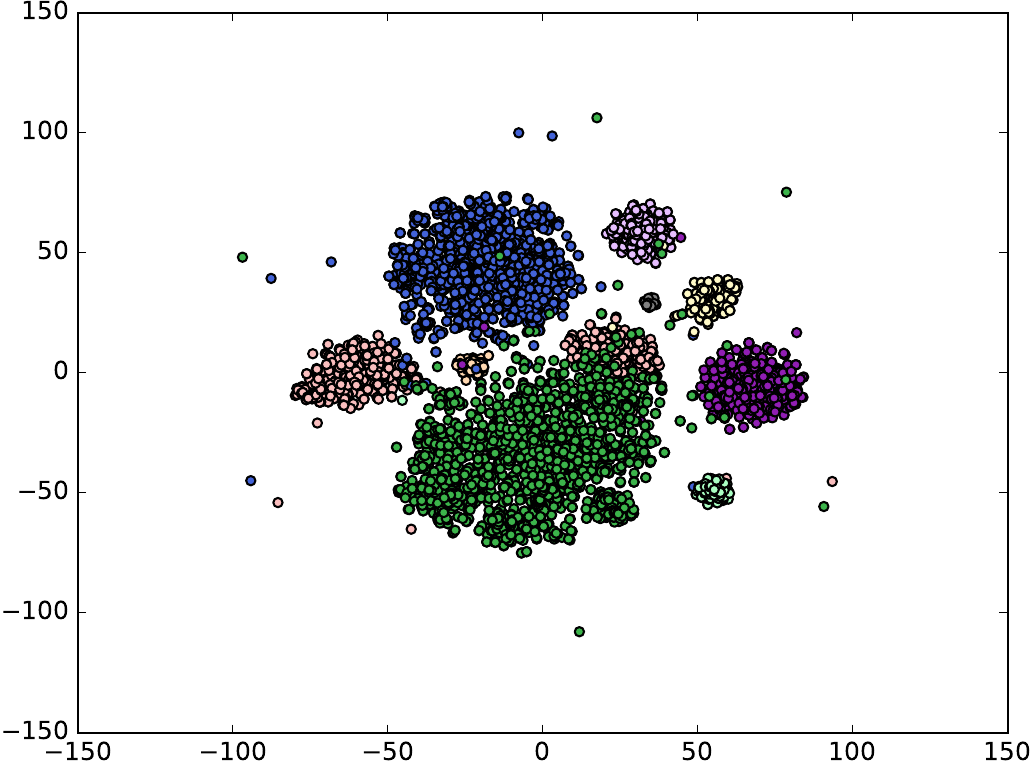}
		\label{fig:visualize_TDT2 c}
	}
	\subfigure[\scriptsize TDT2\_DR2]{
		\includegraphics[width=5cm]{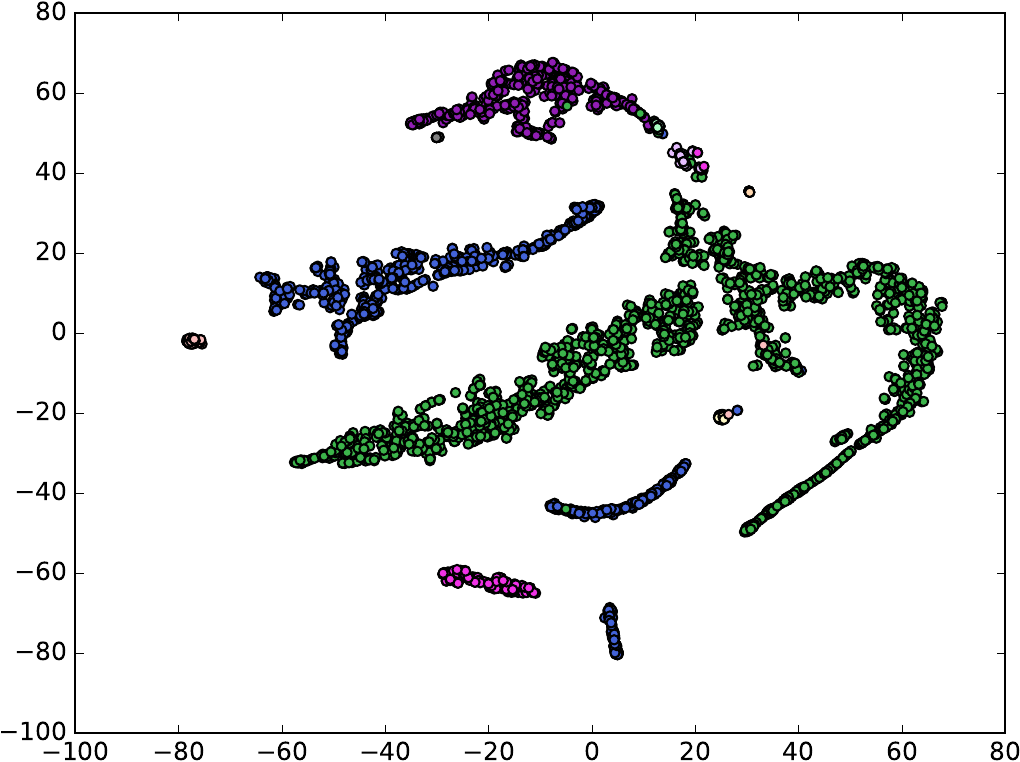} \label{fig:visualize_TDT2 d}
	}
	\subfigure[\scriptsize TDT2\_DR4]{
		\includegraphics[width=5cm]{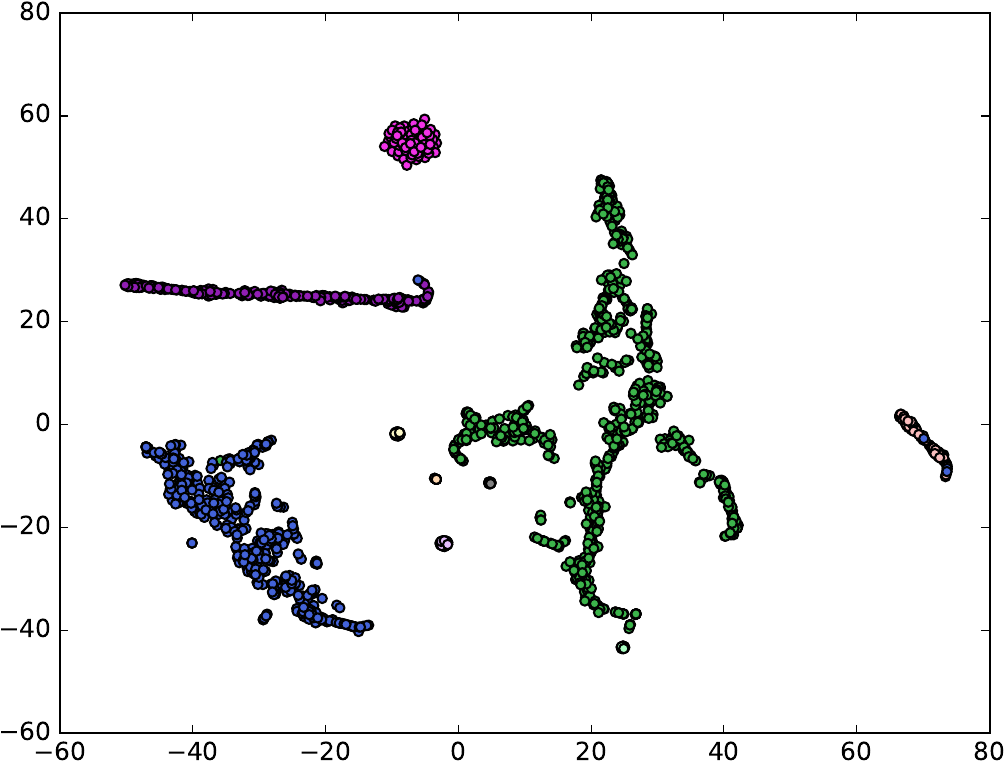} \label{fig:visualize_TDT2 e}
	}
	\subfigure[\scriptsize TDT2\_DR6]{
		\includegraphics[width=5cm]{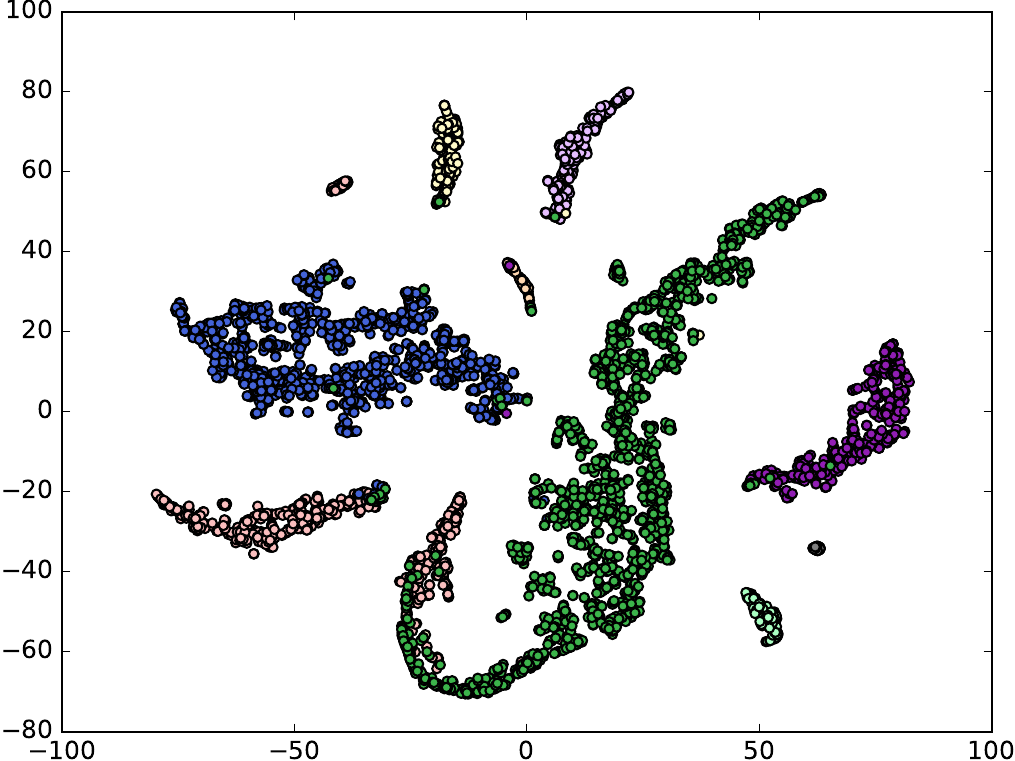} \label{fig:visualize_TDT2 f}
	}
	\caption{Visualization of dataset 2, dataset 4, and dataset 6 using t-SNE  \cite{TSNE_2008} before and after dimension reduction. The true cluster labels are indicated using different colors.}
	\vspace{-0.2cm}
	\label{fig:visualize_TDT2}
\end{figure}

To look into the reason why SC can greatly improve the clustering performance, we employ t-SNE \cite{TSNE_2008} \footnote{Although the performance of t-SNE may vary with different initial points and hyperparameters, we have tested different initial points in the experiments and found most of the results are consistent to each other.}
and visualize some of the TDT2 datasets in Fig. \ref{fig:visualize_TDT2}. Here, TDT2\_2 denotes Dataset 2 in Table \ref{table: clustering doc data} while TDT2\_DR2 denotes dimension-reduced Dataset 2 by SC. 
We can see from Fig. \ref{fig:visualize_TDT2} that, although  all of Dataset 2, 4 and 6 exhibit much clearer cluster structures after dimension reduction by SC,  data samples in both TDT2\_DR2 and TDT2\_DR6 still overlap with each other which is challenging for K-means and K-means++. 
This may explain to some extent why all methods in Table \ref{table: doc data after DR} have significantly improved clustering performance comparing to Table \ref{table: clustering doc data}, and why the K-means++ does not perform as well as the SNCP on TDT2\_DR2 and TDT2\_DR6.

In particular, we see that data samples in TDT2\_DR4 in Fig. \ref{fig:visualize_TDT2 e} becomes almost separable, and therefore the K-means++ can achieve a nearly perfect performance and outperforms the SNCP method.  By contrast, as seen in Fig. \ref{fig:visualize_TDT2 d} and \ref{fig:visualize_TDT2 f}, data samples in both TDT2\_DR2 and TDT2\_DR6 still overlap with each other after dimension reduction, and their data samples are unevenly spread with complex shapes. Such case is known to be challenging for K-means and K-means++ \cite{DCN_Yang_2017}.
As seen from Table \ref{table: doc data after DR}, the SNCP method, which outperforms the K-means++ on the two datasets, is more capable of capturing complicated cluster structures. 

\section{Conclusion}\label{sec: conclusion}
In this paper, we have proposed the ONMF based data clustering formulation \eqref{eqn: onmf prob P2} and the NCP approach \eqref{eqn: onmf prob PP} and Algorithm \ref{table: SNCP}. We have considered one smooth NCP formulation \eqref{eqn: onmf prob PP smooth} and one non-smooth NCP formulation \eqref{eqn: onmf prob PP nsmooth}, and analyzed the theoretical conditions of the two methods for which a feasible and meaningful solution of \eqref{eqn: onmf prob P2}  can be obtained. Efficient implementations of the proposed methods based on the PALM algorithm (Algorithm \ref{table: smooth ALM} and Algorithm \ref{table: non-smooth ALM}) have also been devised.

Extensive experiments have been conducted based on the synthetic data and real data TCGA and TDT2.
In particular, when comparing to the existing K-means and ONMF based methods, the proposed methods can perform either significantly better or comparably in terms of clustering accuracy, while being much more time efficient than most of the other ONMF based methods. 

It is worthwhile to mention some interesting directions for future research. Firstly, while the current  PALM algorithms seem to work well, it is possible to employ some more advanced optimization algorithms such as Nesterov's accelerated method \cite{APCG_2014, ADCA_2018, APPCD_2015} to improve algorithm convergence speed. Secondly, while parallel and distributed implementations of the proposed methods are feasible, it is important to reduce the communication overhead between cluster nodes for large-scale scenarios \cite{Distributed_KM_GT_2013, DC_LSH_2018}. It is also interesting to study joint dimension reduction and clustering methods \cite{JNKM_2017} using the proposed NCP methods, and extend the current Euclidean distance measure to more general cost functions such as the $\beta$-divergence \cite{NMF_beta_2011, Beta_NMF_2012} or the cohesion measure in \cite{Cluster_Metric_2016}. 

\appendices

\section{Proof of Proposition 1}\label{appdix: proof of prop nonzero cols}

{\bf Proof:} Since $(\ol\Wb, \ol\Hb)$ is a B-stationary point of problem (7) in the manuscript, we have
\begin{align} \label{thm: nonzeor column 0}
	\!\!\!\langle 2(\ol\Wb)^T(\ol\Wb~\ol\Hb - \Xb) + \nu\ol \Hb, ~\Db\rangle \geq 0,
	\forall \Db \in \Tc_{\Hc^v_{p,q}}(\ol\Hb).
\end{align}
Suppose that there exist $j'\in \Nc$ such that $\ol\hb_{j'} = \zerob$. Then, \eqref{thm: nonzeor column 0} becomes
\begin{align} \label{thm: nonzeor column 3}
	&\sum_{j \ne j^{\prime}}\langle 2(\ol\Wb)^{T}(\ol\Wb~\ol\hb_j - \xb_j) + \nu\ol\hb_j, ~\db_{j}\rangle  \notag \\
	&~~~~~~- \langle 2(\ol\Wb)^{T} \xb_{j^{\prime}}, ~\db_{j^{\prime}}\rangle \geq 0,~\forall \Db \in \Tc_{\Hc^v_{p,q}}(\ol\Hb).
\end{align}
Suppose that $((\ol\Wb)^{T}\xb_{j^{\prime}})_{\ell} > 0$ for some $\ell \in \Kc$.
Choose a direction $\ol \Db$ such that $\ol \db_{j} = \zerob, \forall j \ne j^{\prime}$, $\ol \db_{j^{\prime}} = \alpha\eb_{\ell}$ for some $\alpha  > 0$. It is obvious to see that $\ol \Db \in \Tb_{\Hc^v_{p,q}}(\ol\Hb)$ based on Definition 1. 
By substituting $\ol \Db$ into \eqref{thm: nonzeor column 3}, we obtain
\begin{align} \label{thm: nonzeor column 4}
	0 > -  2 \langle (\ol\Wb)^{T} \xb_{j^{\prime}}, \alpha\eb_{\ell} \rangle = -2 \alpha((\ol\Wb)^{T} \xb_{j^{\prime}})_{\ell} \geq 0,
\end{align}
which, however, is a contradiction.
\hfill $\blacksquare$

\vspace{-0.2cm}
\section{Proof of Theorem \ref{prop: local min}}\label{appdix: proof of prop local min}

As $(\Wb^{\star}, \Hb^{\star})$ is a local minimizer of problem \eqref{eqn: onmf prob PP smooth}, there exists a neighborhood of $\Hb^{\star}$, denoted by $\Nc_\epsilon(\Hb^\star)=\{\Hb\geq \zerob~|~\|\Hb-\Hb^\star\|_F^2\leq \epsilon\}$, such that 
\begin{align} \label{eqn: prop4 local opt}
	G_\rho(\Wb^{\star}, \Hb^{\star}) \leq G_\rho(\Wb^{\star}, \Hb), \forall \Hb \in \Nc_\epsilon(\Hb^{\star}).
\end{align} 
Suppose there exists a $j^{\prime}$ such that $\oneb^{\top}\hb_{j^{\prime}}^{\star} \ne \|\hb_{j^{\prime}}^{\star}\|_2$. 
Let $c \triangleq \oneb^{\top}\hb_{j^{\prime}}^{\star}$,
$$\Phi \triangleq \{  {\ell} \in \Kc~ |~ [\Hb^{\star}]_{{\ell}, j^{\prime}} > 0 \}=\{i_1,i_2,\ldots,i_{|\Phi|}\},$$ and define a hyperplane $\Hc_c \triangleq \{\Hb\geq \zerob~|~\oneb^{\top}\hb_{j'}=c\}$.
We will show a contradiction that there exists a neighbor of $\Hb^\star$ that lies in the hyperplane $\Hc_c$ and achieves a smaller objective than $\Hb^\star$.

To the end, for $\ell=1,\ldots,|\Phi|$, let 
\begin{align} \label{eqn: construct Hl}
	\Hb^{({\ell})} = \sum_{j \ne j^{\prime}} \hb_j^{\star}\eb_j^{\top} + \hb_{j^{\prime}}^{({\ell})}\eb_{j^{\prime}}^{\top},
\end{align} 
which is obtained by replacing the $j'$th column of $\Hb^\star$ by
\begin{align} \label{eqn: construct hj'}
	\hb_{j^{\prime}}^{({\ell})} \triangleq \alpha_{\ell} \ssb^{({\ell})}  + (1 - \alpha_{\ell})\hb_{j^{\prime}}^{\star}
\end{align} 
where $ 0 < \alpha_{\ell} < 1$ and $\ssb^{({\ell})} \triangleq c\eb_{i_\ell}.$
One can see that $\Hb^{(\ell)} \in \Hc_c$ since
\begin{align}
	\oneb^{\top}\hb_{j^{\prime}}^{({\ell})} = \alpha_{\ell}c + (1 - \alpha_{\ell})c = c.
\end{align}

Firstly, we show that $\Hb^{\star}$ is a convex combination of $\Hb^{({\ell})}, \forall \ell=1,\ldots,|\Phi|$. 
By \eqref{eqn: construct hj'}, we have
$\ssb^{({\ell})} = \frac{\hb_{j^{\prime}}^{({\ell})} - (1 - \alpha_{\ell})\hb_{j^{\prime}}^{\star}}{\alpha_{\ell}}$. 
Thus, we can obtain
\begin{align} \label{eqn: prop4 c1}
	\hb_{j^{\prime}}^{\star} = \sum_{{\ell} = 1}^{|\Phi|} [\Hb^{\star}]_{{i_\ell}, j^{\prime}} \eb_{i_\ell} = \sum_{{\ell} = 1}^{|\Phi|} \beta^{{\ell}}\bigg(\frac{\hb_{j^{\prime}}^{({\ell})} - (1 - \alpha_{\ell})\hb_{j^{\prime}}^{\star}}{\alpha_{\ell}} \bigg),
\end{align}
where $\beta_{\ell} \triangleq  \frac{[\Hb^{\star}]_{{i_\ell}, j^{\prime}}}{c}$.
Rearranging terms in \eqref{eqn: prop4 c1} gives rise to
\begin{align}\label{eqn: prop4 c11}
	\bigg(1 + \sum_{{\ell} = 1}^{|\phi|} \frac{\beta_{\ell}}{\alpha_{\ell}}(1- \alpha_{\ell}) \bigg) \hb_{j^{\prime}}^{\star} = \sum_{{\ell} = 1}^{|\phi|} \frac{\beta_{\ell}}{\alpha_{\ell}} \hb_{j^{\prime}}^{({\ell})}.
\end{align}
Notice that $\sum_{{\ell} = 1}^{|\phi|}\beta_{\ell} = \sum_{{\ell} = 1}^{|\phi|}\frac{[\Hb^{\star}]_{{i_\ell}, j^{\prime}}}{c}= \frac{\oneb^{\top}\hb_{j^{\prime}}^{\star}}{c} = 1$.
So \eqref{eqn: prop4 c11} reduces to
\begin{align}
	\hb_{j^{\prime}}^{\star} = \sum_{{\ell} = 1}^{|\phi|} \frac{\beta_{\ell}}{\alpha_{\ell}} \hb_{j^{\prime}}^{({\ell})} \bigg/ \sum_{{\ell} = 1}^{|\phi|} \frac{\beta_{\ell}}{\alpha_{\ell}}
	\triangleq 
	\sum_{{\ell} = 1}^{|\phi|} \gamma_\ell \hb_{j^{\prime}}^{({\ell})},
\end{align} 
which implies that $\Hb^{\star}$ is a convex combination of $\Hb^{(\ell)}, {\ell} =1,\ldots,|\Phi|$. 

Secondly, we show that $G_\rho(\Wb^\star, \Hb)$ is strongly concave with respect to $\Hb$ on the hyperplane $\Hc_c$ as long as $\rho$ is sufficiently large.
It is sufficient to show that $G_\rho(\Wb^\star, \Hb)-\frac{\rho}{2}\sum_{j=1}^N \oneb^{\top}\hb_j =F(\Wb^\star, \Hb) - \frac{\rho}{2}\sum_{j = 1}^{N} \|\hb_j\|_2^2$ is 
strongly concave in $\Hb$. 
This is true since $\nabla_{\Hb} F(\Wb^\star, \Hb)$ is Lipschitz continuous with a bounded Lipschitz constant (denoted by $\rho^\star$), and thus $F(\Wb^\star, \Hb) - \frac{\rho}{2}\sum_{j = 1}^{N} \|\hb_j\|_2^2$
will be a strongly concave function as long as $\rho>\rho^\star$.

By the above two facts, we obtain that
\begin{align} \label{eqn: convex combination1}
	\!\!\!\!\!G_{\rho}(\Wb^{\star}, \Hb^{\star}) > \sum_{{\ell} = 1}^{|\phi|} \gamma_\ell G_{\rho}(\Wb^{\star}, \Hb^{({\ell})})
	\geq G_{\rho}(\Wb^{\star}, \Hb^{({\ell'})}),
\end{align}
where 
$
{\ell}^{\prime} = \arg \min_{{\ell}=1,\ldots, |\Phi|} G_\rho(\Wb^{\star}, \Hb^{({\ell})}).
$
Since by \eqref{eqn: construct hj'}, $ \Hb^{({\ell'})}$ will lie in $\Nc_\epsilon(\Hb^\star)$ for $\alpha_{\ell'} < \frac{\epsilon}{c-[\Hb^\star]_{i_{\ell'},j'}}$, \eqref{eqn: convex combination1} shows a contradiction to the local optimality of $\Hb^\star$. 
Thus, for $\rho>\rho^\star$, $\Hb^\star$ must be feasible and a local minimizer to \eqref{eqn: onmf prob P2}.
\hfill $\blacksquare$

\section{Proof of Theorem \ref{thm: stationary pont}}\label{appdix: proof of thm statioanry point}
{
	
	To show that $(\Wb^\infty, \Hb^\infty)$ is a B-stationary point to \eqref{eqn: onmf prob P2}, it suffices to show
	\begin{subequations}
		\begin{align}  \label{thm2: W_bstationary}
			&\langle\nabla_{\Wb}F(\Wb^{\infty}, \Hb^{\infty}), \Wb-\Wb^{\infty}  \rangle \geq 0,  \forall \Wb\geq \zerob, \\
			\label{thm2: H_bstationary}
			&\langle\nabla_{\Hb}F(\Wb^{\infty}, \Hb^{\infty}), \Db \rangle  \geq 0,  \forall \Db \in \Tc_{\Hc^2_{1,2}}(\Hb^{\infty}),
		\end{align}
	\end{subequations}
	where $\Hc^2_{1,2}=\{\Hb\geq \zerob~|~(\oneb^\top \hb_j)^2 - \|\hb_j\|_{2}^2 =0,~\forall j\in \Nc \}$.
	Since $(\Wb^{\rho},\Hb^{\rho})$ is a stationary point of \eqref{eqn: onmf prob PP smooth}, we have
	\begin{align}  \label{thm2: W_bstationary rho}
		&\langle\nabla_{\Wb}F(\Wb^{\rho}, \Hb^{\rho}), \Wb-\Wb^{\rho}  \rangle \geq 0,  \forall \Wb\geq \zerob,
	\end{align}
	which directly implies \eqref{thm2: W_bstationary} by taking $\rho \to \infty$. To prove \eqref{thm2: H_bstationary}, we employ the following proposition and applies it to \eqref{eqn: onmf prob P2} for each column of $\Hb$.
	\begin{Prop}\label{prop: smooth penalty}
		Consider the following problem
		\begin{subequations}\label{eqn: problem}
			\begin{align}
				\min_{\xb\in \mathbb{R}^n} ~&f(\xb) \\
				{\rm s.t.~}&\hspace{-0.3cm}
				\left.
				\begin{array}{ll}
					& \xb \geq 0, \\
					& \oneb^{\top}\xb = \|\xb\|_2,
				\end{array}\right\} \triangleq \Xc,
			\end{align}
		\end{subequations}
		where $f$ is smooth, and the corresponding penalized problem
		\begin{align}\label{eqn: partial penalty problem}
			\min_{\xb \geq \zerob} ~f(\xb) + \frac{\rho}{2} \bigg((\oneb^{\top}\xb)^2 - \|\xb\|_2^2\bigg)
		\end{align}
		where $\rho>0$ is a penalty parameter.
		Let $\xb^\rho$ be a stationary point of \eqref{eqn: partial penalty problem}. Assume that $\xb^\rho$ is bounded and $\xb^\rho \to \xb^\infty \ne \zerob$ as $\rho \to \infty$. Then, $\xb^\infty$ is a feasible B-stationary point of \eqref{eqn: problem}.	
	\end{Prop}
	
	%
	%


	{\bf Proof:} Since $\xb^\rho$ is a stationary point of  \eqref{eqn: partial penalty problem},  which is also a Karush-Kuhn-Tucker (KKT) solution to \eqref{eqn: partial penalty problem}. By the complementary slackness, $\xb^\rho$ satisfies
	\begin{align}
		\nabla f(\xb^{\rho})^{\top}\xb^{\rho} + \rho \left[( \oneb^{\top}\xb^{\rho})^2 - \|\xb^{\rho}\|_2^2 \right] =  0.
	\end{align}
	Let $\xb^{\infty} \neq 0$ be the limit of a subsequence $\{ \xb^{\rho_k} \}$ with $\{ \rho_k \} \to \infty$.  We then have
	\begin{align}\label{proof E1}
		\nabla f(\xb^{\rho_k})^{\top}\xb^{\rho_k}+ 
		\rho_k \left[ (\oneb^{\top}\xb^{\rho_k} )^2 - \|\xb^{ \rho_k} \|_2^2 \right] = 0, \forall \, k.
	\end{align}
	Since $\nabla f(\xb^{\rho_k})^{\top}\xb^{\rho_k}$ is bounded due to bounded $\xb^\rho$, it follows from \eqref{proof E1} that 
	$(\oneb^{\top}\xb^{\rho_k} )^2 - \|\xb^{ \rho_k} \|_2^2 \to 0$ as $\rho_k \to \infty$.
	Hence, 
	$\xb^\infty$ is feasible to \eqref{eqn: problem}, and all components of $\xb^{\, \infty}$ are zero except one component, say $\xb^{\, \infty}_{\bar{i}} > 0$. 
	
	We next show that 
	\begin{align}\label{proof E2}\nabla f(\xb^\infty) ^{\top}\db \geq 0, \forall \db \in \Tc_{\Xc}(x^{\, \infty}),
	\end{align} i.e., $\xb^\infty$ is a B-stationary point of \eqref{eqn: problem}.
	Before that, we claim that the tangent cone $ \Tc_{\Xc}(\xb^{\, \infty}) = \left\{ \db \mid d_i = 0, \forall  i \neq \bar{i} \right\}$.  
	Indeed, any $\db \in \Tc_{\Xc}(\xb^{\, \infty})$ must satisfy $d_i \geq 0$ for all $i \neq \bar{i}$ and 
	$\oneb^{\top}\db = 
	\frac{\db^{\top}\xb^{\infty}}{\| \xb^{\infty} \|_2}
	= \db_{\bar{i}}$.  This yields $d_i = 0$ for all $i \neq \bar{i}$; thus
	the tangent cone is contained in the right-hand cone.  The reverse inclusion is obvious because
	$\xb^{\infty} + \tau \db \in \Xc$ for all sufficiently small $\tau > 0$. 
	Thus, \eqref{proof E2} boils down to
	\begin{align}\label{proof E3}
		\frac{\partial f(\xb^\infty)}{\partial x^\infty_{\bar i}}=0.
	\end{align}
	
	We next prove that there cannot exist an index $\bar{j} \neq \bar{i}$ and a subsequence $\{ \xb^{k} \triangleq \xb^{\rho_k} \}_{k \in \kappa}$
	such that $x^{k}_{\bar{j}} > 0$ for all $k \in \kappa$. If not, we 
	may assume without loss of generality that this subsequence is the entire sequence $\{ \xb^k \}$.  Thus
	$x^{k}_{\bar{j}} > 0$ for all $k$.  This is in addition to $x^{k}_{\bar{i}} > 0$ for all sufficiently large $k$. By the complementary slackness, we have
	\begin{align}
		0 =& 
		\frac{\partial f(\xb^{k})}{\partial x_{\bar{j}}}
		+ \rho_k \left[  \oneb^{\top}\xb^{k} - x^{ k}_{\bar{j}}  \right] \,  \\
		\geq& \displaystyle{
			\frac{\partial f(\xb^{k})}{\partial x_{\bar{j}}}
		} + \rho_k x^k_{\bar{i}},~\forall k. \label{thm2: ineq}
	\end{align}
	Since $\frac{\partial f(\xb^{\, k})}{\partial x_{\bar{j}}}$ is bounded, and $\rho_k \to \infty$ and $ x_{\bar{i}}^{k} \to x_{\bar{i}}^\infty > 0$ as $k \to \infty$, we arrive at a contraction after taking the limit on both sides of inequality \eqref{thm2: ineq}. Hence, for any converging subsequence $\{\xb^{\rho_k}\}$, there are only finite terms that have at least one more nonzero component in addition to $\bar{i}$. 
	
	By restricting to a further subsequence of $\{\xb^{\rho_k}\}$ satisfying $x^{\rho_k}_{\bar{i}} > 0$ and $x^{\rho_k}_{i} = 0$ for all $i\neq \bar{i}$, we have
	\begin{align}
		\frac{\partial f(\xb^{k})}{\partial x_{\bar{i}}}
		+ \rho_k \left[ \oneb^{\top}\xb^{k} - x^{k}_{\bar{i}} \right] = 
		\frac{\partial f(\xb^{k})}{\partial x_{\bar{i}}}
		=0,
	\end{align}
	which implies \eqref{proof E3}, and the proof is complete.
	\hfill $\blacksquare$

	\section{Proof of Theorem \ref{thm: exact penalty}} \label{appdix: proof of thm nonsmooth}
	
	Analogous to the proof of Theorem \ref{thm: stationary pont}, it is sufficient to consider the following proposition.
	\begin{Prop}\label{prop: nonsmooth penalty}
		Consider the following problem
		\begin{subequations}\label{eqn: nonsmooth problem}
			\begin{align}
				\min_{\xb\in \mathbb{R}^n} ~&f(\xb) \\
				{\rm s.t.~}&\hspace{-0.3cm}
				\left.
				\begin{array}{ll}
					& \xb \geq 0, \\
					& \oneb^{\top}\xb = \|\xb\|_\infty,
				\end{array}\right\} \triangleq \Xc,
			\end{align}
		\end{subequations}
		where $f$ is smooth, and the corresponding penalized problem
		\begin{align}\label{eqn: nonsmooth partial penalty problem}
			\min_{\xb \geq \zerob} ~f(\xb) + \rho (\oneb^{\top}\xb - \|\xb\|_\infty)
		\end{align}
		where $\rho>0$ is a penalty parameter.
		Let $\xb^\rho$ be a d-stationary point of \eqref{eqn: nonsmooth partial penalty problem}. Assume that $\xb^\rho$ is bounded. Then there exists a finite $\rho^\star$ such that for all $\rho > \rho^\star$, 
		$\xb^\rho$ is a feasible B-stationary point to problem \eqref{eqn: nonsmooth problem}.	
	\end{Prop}
	%
	
	{\bf Proof:} Denote $\phi(\xb) \triangleq \oneb^{\top} \xb - \|\xb\|_\infty$. Since $\xb^\rho$ is a d-stationary point of \eqref{eqn: nonsmooth partial penalty problem}, we have
	\begin{align} \label{proof of thm nonsmooth 1}
		\nabla f(\xb^\rho)^\top \db + \rho  \phi^{\prime}(\xb^\rho; \db) \geq 0, \forall \db \in \Tc_{+}(\xb^\rho),
	\end{align}
	where $\Tc_{+}(\xb^{\rho}) = \{\db ~|~ d_i \geq 0 ~\text{if}~ x_i  = 0\}$. Because 
	\begin{align} 
		\phi^{\prime}(\xb^{\rho}; \db)= \oneb^{\top}\db -\max_{\gbm \in \partial \|\xb^\rho\|_\infty} \gbm^{\top}\db,
	\end{align}
	and $ \eb_{i^{\star}} \in \partial  \|\xb^\rho\|_\infty$, where
	$\displaystyle i^{\star} \in \arg\max_{i} x_i^\rho$, condition \eqref{proof of thm nonsmooth 1} implies
	\begin{align} \label{eqn: staionary equality 1}
		\nabla f(\xb^{\rho})^{\top} \db + \rho( \oneb^{\top}\db- d_{i^{\star}})\geq 0, \forall \db \in \Tc_{+}(\xb^{\rho}).
	\end{align}
	On the other hand, for bounded $\xb^\rho$, we can upper bound
	\begin{align} \label{eqn: bound of grad}
		\nabla f(\xb^{\rho})^{\top} \db 
		\leq &\|\nabla f(\xb^{\rho})\|_2\|\db\|_2 \\
		\leq &\|\nabla f(\xb^{\rho})\|_2\|\db\|_1 
		\triangleq \rho^\star \|\db\|_1.
	\end{align}
	By substituting \eqref{eqn: bound of grad} into \eqref{eqn: staionary equality 1}, we obtain
	\begin{align} \label{eqn: stationary inequality 2}
		\!\!\!\rho^\star \|\db\|_1
		+ \rho ( \oneb^{\top}\db - d_{i^{\star}})\geq 0, \forall \db \in \Tc_{+}(\xb^{\rho}).
	\end{align}
	
	We argue that $\phi(\xb^{\rho}) = 0$ for $\rho > \rho^\star$. Suppose not. Let us choose a tangent $\bar \db \in \Tc_{+}(\xb^{\rho})$ as follows
	\begin{align} \label{eqn: stationary inequality 3}
		\bar d_i = 
		\begin{cases} 
			-1, ~\text{for~} x^\rho_{i}>0 {\rm ~and~} i\neq i^\star,\\
			~~~0,~{\rm otherwise}.
		\end{cases}
	\end{align}
	Substituting \eqref{eqn: stationary inequality 3} into \eqref{eqn: stationary inequality 2} gives rise to
	\begin{align} \label{eqn: contradiction}
		(\rho^{\star} - \rho) \|\bar \db\|_1 \geq 0,
	\end{align}
	which is a contradiction since $\rho > \rho^{\star}$. Thus $\phi(\xb^{\rho}) = 0$ must hold, i.e., $\xb^\rho$ is feasible to \eqref{eqn: nonsmooth problem} if $\rho > \rho^{\star}$.  
	
	Next, we show that $\xb^\rho$ is a B-stationary point to  \eqref{eqn: nonsmooth problem}.	
	%
	Since $\xb^\rho$ is feasible to $\Xc$ in 
	\eqref{eqn: nonsmooth problem} for $\rho > \rho^\star$. Thus, the tangent cone of $\Xc$ at $\xb^\rho$ is given by
	\begin{align}
		\Tc_{\Xc}(\xb^{\rho}) 
		&= \left\{  \db~\bigg |\!\!\!\!\!
		\begin{array}{ll}
			&\db = \eb_l, \forall l = 1, \ldots, n, \text{if}~ \|\xb^{\rho}\|_2 = 0; \\
			&d_{i} = \pm 1, \text{if}~ [x^{\rho}]_{i} > 0; \\
			&d_{i} = 0, \text{otherwise}.
		\end{array}\right\} \notag \\
		& \subset \Tc_{+}(\xb^{\rho}). \label{thm1: proof c3}
	\end{align}
	Besides, one can verify that
	for all $\db \in \Tc_{\Xc}(\xb^{\rho})$, 
	\begin{align}\label{thm1: proof c2}
		\phi^{\prime}(\xb^{\rho}; \db)  = \oneb^{\top}\db - \max_{\gb \in \partial (\|\xb^{\rho}\|_{\infty})} \gb^{\top}\db = 0.
	\end{align}
	Therefore, by applying \eqref{thm1: proof c3} and \eqref{thm1: proof c2} to \eqref{proof of thm nonsmooth 1}, we obtain
	\begin{align} 
		\nabla f(\xb^\rho) ^{\top}\db \geq 0, \forall \db \in \Tc_{\Xc}(\xb^\rho)
	\end{align}
	which is the desired result.
	\hfill $\blacksquare$ 
	
	{
		\section{Proof of Proposition \ref{prop: prox map negative infinity norm}}
		\label{appdix: proof of Prop nonsmooth}
		Denote $f(\xb) = \frac{1}{2}\|\xb - \yb\|_{2}^{2} - c\|\xb\|_{\infty}$. 
		Since $\xb^{\star}$ is an optimal solution of \eqref{eqn: prox prob}, we have 
		\begin{align} \label{prop: neg infty 1}
			f^{\prime}(\xb^{\star}; \db) \geq 0, \forall \db \in \Tc_{+}(\xb^{\star}),
		\end{align}where $f^{\prime}(\xb; \db)$ is the directional derivative of $f(\xb)$ at direction $\db$, and 
		\begin{align}\label{proof of prop 1 C0}
			\Tc_{+}(\xb^{\star}) = \{\db \in \mathbb{R}^n| d_i \geq 0, ~\text{if}~ (\xb^{\star})_i = 0\}.
		\end{align}	
		As the directional derivative of $ \|\xb\|_{\infty}$ is $\max_{\gb \in \partial(\|\xb\|_{\infty})}\gb^{T}\db$, we have from \eqref{prop: neg infty 1} that
		\begin{align}\label{proof of prop 1 C1}
			\langle\nabla f(\xb^{\star}), \db \rangle - c \max_{\gb \in \partial(\|\xb^\star\|_{\infty})}\gb^{T}\db \geq 0, \forall \db \in \Tc_{+}(\xb).
		\end{align}
		Let $i^\star$ be an index such that $(\xb^{\star})_{i^\star}=\|\xb^\star\|_\infty$. Then, $\eb_{i^{\star}} \in \partial(\|\xb^\star\|_{\infty})$, and \eqref{proof of prop 1 C1} implies
		\begin{align} \label{eqn: opt}
			\langle \xb^{\star} - (\yb + c\eb_{i^{\star}}), \db\rangle \geq 0, \forall \db \in \Tc_{+}(\xb^{\star}).
		\end{align}

		Now let us consider the case that $(\xb^{\star})_{i^{\star}} = 0$. Then, by  \eqref{proof of prop 1 C0}, it holds that
		\begin{align}\label{proof of prop 1 C2}
			\db =d_{i^\star} \eb_{i^\star} \in  \Tc_+(\xb^\star),~\forall d_{i^\star}\geq 0.
		\end{align}
		Substituting \eqref{proof of prop 1 C2} into \eqref{eqn: opt} gives rise to
		\begin{align}
			d_{i^{\star}} (y_{i^{\star}} + c) \leq 0, \forall d_{i^{\star}} \geq 0.
		\end{align}
		which implies
		$y_{i^{\star}} + c \leq 0.$
		On the other hand, suppose $(\xb^{\star})_{i^{\star}} \ne 0$. Then, 
		\begin{align}\label{proof of prop 1 C3}
			\db =d_{i^\star} \eb_{i^\star} \in  \Tc_+(\xb^\star),~\forall d_{i^\star}\in \mathbb{R}.
		\end{align}
		Substituting \eqref{proof of prop 1 C3} into \eqref{eqn: opt} gives rise to
		\begin{align}\label{proof of prop 1 C4}
			d_{i^{\star}} ((\xb^\star)_{i^\star} - y_{i^{\star}} - c) \geq 0, \forall d_{i^{\star}} \in \mathbb{R},
		\end{align}
		which implies
		$(\xb^{\star})_{i^{\star}} = y_{i^{\star}} + c.$
		By combining the above two implications, we obtain $(\xb^{\star})_{i^{\star}} = (y_{i^{\star}} + c)^{+}$.
		In addition, by following a similar argument as in \eqref{proof of prop 1 C2} to \eqref{proof of prop 1 C4}, one can obtain $(\xb^{\star})_{i} = [y_{i^{\star}}]_{+}$ for all $i \ne i^{\star}$.
		Therefore, \eqref{prop: the optimal solution} is true and is recapitulated here  
		\begin{align} \label{prop: the optimal solution 2}
			(\xb^{\star})_i = \begin{cases} (y_i + c)^+, ~{\rm if} ~i = i^{\star}, \\
				(y_i)^+, ~{\rm otherwise},
			\end{cases}
		\end{align}	for $i=1,\ldots,n$.
		If one substitutes the optimal $\xb^\star$ in \eqref{prop: the optimal solution 2} into the objective of \eqref{eqn: prox prob}, one can easily verify that the index $i^\star$ must be unique.
		
		Next, let us show that 
		\begin{align}\label{proof prop 1 C10}
			i^\star \in  \arg\max_{i=1,\ldots,n} y_i.
		\end{align} We respectively consider three cases.

		\underline{Case (a):  $\yb \geq \zerob$:} For $\yb \geq \zerob$, and by \eqref{prop: the optimal solution 2},
		the optimal objective value of \eqref{eqn: prox prob} is given by
		\begin{align}
			f(\xb^{\star}) = \frac{c^2}{2} - c(y_{i^{\star}} + c) = -c y_{i^{\star}} - \frac{c^2}{2}. 
		\end{align}
		Obviously, we must have $i^\star \in  \arg\max_{i=1,\ldots,n} y_i$.
		
		\underline{Case (b):  $\yb < \zerob$:} If   $\yb < \zerob$ and 
		$y_i + c \leq 0, \forall i = 1, \ldots, n$,
		we have the trivial solution $\xb^{\star} = \zerob$ according to \eqref{prop: the optimal solution 2} and the definition $(\xb^\star)_{i^\star} = \|\xb^\star\|_\infty$.
		
		Suppose that $\yb < \zerob$ and the set $\Ic = \{i \in \{1,\ldots,n\}|y_i + c > 0\}$ is non-empty. We claim that $i^{\star} \in \arg\max_{i \in \Ic} y_i$. Suppose $i^{\star} \notin \Ic$. Then we have $\xb^{\star} = \zerob$ and it leads to
		\begin{align}\label{proof prop 1 C5}
			f(\zerob) = \frac{1}{2} \sum_{i = 1}^{n} y_i^2 =  \frac{1}{2}\sum_{i \ne i^{\star}} y_i^2 +  \frac{1}{2}y_{i^{\star}}^2.
		\end{align}
		Suppose$i^{\star} \in \Ic$. Then by \eqref{prop: the optimal solution 2}, we have
		\begin{align}\label{proof prop 1 C6}
			f(\xb^\star) = & \frac{1}{2}\sum_{i \ne i^\star}y_i^2 + \frac{c^2}{2} - c(y_{i^\star} + c) \notag  \\
			= &\frac{1}{2}\sum_{i \ne i^\star}y_i^2 - cy_{i^\star} - \frac{c^2}{2}.
		\end{align}
		From \eqref{proof prop 1 C5} and \eqref{proof prop 1 C6}, It is apparent that $f(\zerob) > f(\xb^\star)$, and thereby $i^{\star} \in \Ic$. 
		
		Further consider two possible solutions, namely, $\xb_1^\star$ and $\xb_2^\star$ with $(\xb^\star_1)_{i_1} = \|\xb^\star_1\|_\infty$ and $(\xb^\star_2)_{i_2} = \|\xb^\star_2\|_\infty$, respectively, where $i_1,i_2\in \Ic$. Then, by \eqref{proof prop 1 C6}, we have
		\begin{align}\label{proof prop 1 C7}
			f(\xb_1^\star) - f(\xb_2^\star) = &~\frac{1}{2}(y_{i_2}^2 - y_{i_1}^2) + c(y_{i_2} - y_{i_1}) \notag  \\
			= & ~(y_{i_2} - y_{i_1})(\frac{1}{2}(y_{i_2} + y_{i_1}) + c) < 0,
		\end{align}
		whenever $y_{i_2} < y_{i_1}$ since $\frac{1}{2}(y_{i_2} + y_{i_1}) + c>0$ for $i_1, i_2 \in \Ic$. In other words, $i^{\star}$ with larger $y_{i^{\star}}$ leads to a smaller objective value, and thus we conclude that $i^{\star} \in \arg\max_{i \in \Ic} y_i$ .
		
		\underline{Case (c): $\yb \ngeq \zerob$ and $\yb \nless \zerob$}. 
		There are three possible situations: 1) $i^{\star} \in \Ic_0 = \{i|y_i < 0, y_i + c \leq 0\}$; 2) $i^{\star} \in \Ic_1 = \{i|y_i \geq 0\}$; and 3) $i^{\star} \in \Ic_2 = \{i|y_i < 0, y_i + c > 0\}$. 
		
		1) Suppose $i^{\star} \in \Ic_0$. Then $\xb^{\star} = \zerob$, and it leads to \eqref{proof prop 1 C5}.
		
		2) Suppose $i^{\star} \in \Ic_1$. Then, by \eqref{prop: the optimal solution 2}, we have
		\begin{align}\label{proof prop 1 C8}
			f(\xb^\star) 
			= &~\frac{1}{2}\sum_{i \notin \Ic_1}y_i^2 - cy_{i^\star} - \frac{c^2}{2}.
		\end{align}
		Clearly, $i^\star\in  \arg\max_{i \in \Ic_1}y_i$. 
		
		3) If $i^{\star} \in \Ic_2$. Then, by \eqref{prop: the optimal solution 2}, we have
		\begin{align}\label{proof prop 1 C9}
			f(\xb^\star) 
			= &~\frac{1}{2}\sum_{i \notin \Ic_1, i \ne i^\star}y_i^2 - cy_{i^\star} - \frac{c^2}{2}
		\end{align}
		Based on a similar argument as in \eqref{proof prop 1 C7}, one can conclude $i^\star \in \arg\max_{i \in \Ic_2} y_i$. 
		
		From \eqref{proof prop 1 C8} and \eqref{proof prop 1 C9}, it is obvious to that $f(\zerob) > f(\xb^\star)$ for $i^\star \in \Ic_1 \cup \Ic_2$. Besides, consider two possible solutions, namely, $\xb_1^\star$ with $(\xb^\star_1)_{i_1} = \|\xb^\star_1\|_\infty$ where $i_1 \in \Ic_1$, and  $\xb_2^\star$ 
		with $(\xb^\star_2)_{i_2} = \|\xb^\star_2\|_\infty$ where $i_2 \in \Ic_2$. Then, by  \eqref{proof prop 1 C8} and \eqref{proof prop 1 C9}, we have
		\begin{align}
			f(\xb_1^\star) - f(\xb_2^\star) &= \frac{1}{2}y_{i_2}^2 + c(y_{i_2} - y_{i_1}) \notag \\
			&= y_{i_2}(\frac{y_{i_2} }{2}+ c) - cy_{i_1} < 0
		\end{align}
		where the last inequality holds since $y_{i_2} < 0, y_{i_1} > 0$ and $\frac{1}{2}y_{i_2} + c > y_{i_2} + c > 0$. Thus, we obtain $i^{\star} = i_1 \in \arg\max_{i \in \Ic_1} y_i$.

		By summarizing the results from Case (a) to (c), we conclude that \eqref{proof prop 1 C10} holds true. Proposition \ref{prop: prox map negative infinity norm} is proved.
		\hfill $\blacksquare$
	}

	\vspace{-0.2cm}
	\section{Proof of Theorem 5}
	\label{appdix: proof of conv of non-smooth ALM}
	
	By a similar argument as the proof of Theorem 4, we obtain that the sequence $(\Wb^k, \Hb^k)$ generated by Algorithm 3 with $t^k > L_{\wt F}(\Wb^k)$ and $c^k > \frac{L_F(\Hb^k)}{2}$  is a bounded sequence.
	We assume that $(\Wb^\infty, \Hb^\infty)$ is a limit point of $(\Wb^k, \Hb^k)$, and $t^\infty > L_{\wt F}(\Wb^\infty)$  is a limit value of $t^k$.
	
	At iteration $k$, since $\wt F(\Wb^k, \Hb)$ is $L_{\wt F}(\Wb^k)$-smooth and $t^k > L_{\wt F}(\Wb^k)$, by the  descent lemma \cite[Lemma 3.2]{PALM_2014}, we have 
	\vspace{-0.2cm}
	\begin{align}
		&\wt F(\Wb^k, \Hb^{k+1}) - \rho \sum_{j = 1}^{N}\|\hb_j^{k+1}\|_\infty\notag \\
		\leq& \wt F(\Wb^k, \Hb^{k}) + \langle \nabla_{H}\wt F(\Wb^k, \Hb^{k}), \Hb^{k+1} -\Hb^{k} \rangle\notag \\
		&~ + \frac{t^k}{2}\|\Hb^{k+1} - \Hb^k\|_F^2 - \rho \sum_{j = 1}^{N}\|\hb_j^{k+1}\|_\infty \label{thm4: L-smooth}
	\end{align}
	Since $\Hb^{k+1}$ is the optimal solution of (13), 
	we have
	\begin{align}\label{thm4:opt}
		0  \leq& \langle \nabla_{H}\wt F(\Wb^k, \Hb^{k}), \Hb -\Hb^{k+1} \rangle \notag \\
		\leq &  \langle \nabla_{H}\wt F(\Wb^k, \Hb^{k}), \Hb -\Hb^{k} \rangle \notag \\
		&~+ \langle \nabla_{H}\wt F(\Wb^k, \Hb^{k}),  \Hb^{k} - \Hb^{k+1} \rangle.
	\end{align}
	
	By \eqref{thm4:opt},  \eqref{thm4: L-smooth} further infers 
	\begin{align}
		&\wt F(\Wb^k, \Hb^{k+1}) - \rho \sum_{j = 1}^{N}\|\hb_j^{k+1}\|_\infty\notag \\
		\leq& \wt F(\Wb^k, \Hb^{k}) + \langle \nabla_{H}\wt F(\Wb^k, \Hb^{k}), \Hb -\Hb^{k} \rangle  \notag \\
		&~+ \frac{t^k}{2}\|\Hb - \Hb^k\|_F^2 - \rho \sum_{j = 1}^{N}\|\hb_j\|_\infty, ~\forall \Hb \geq \zerob. \label{thm4: H_opt}
	\end{align}
	By taking the limits of both sides of \eqref{thm4: H_opt} as $k \rightarrow \infty$, we have
	\begin{align}
		&\wt F(\Wb^\infty, \Hb^{\infty}) - \rho \sum_{j = 1}^{N}\|\hb_j^{k+1}\|_\infty\notag \\
		\leq& \wt F(\Wb^\infty, \Hb^{\infty}) + \langle \nabla_{H}\wt F(\Wb^\infty, \Hb^{\infty}), \Hb -\Hb^{\infty} \rangle \notag \\
		&~+ \frac{t^\infty}{2}\|\Hb - \Hb^\infty\|_F^2 - \rho \sum_{j = 1}^{N}\|\hb_j\|_\infty, ~\forall \Hb \geq 0.
	\end{align}
	which implies that 
	\begin{align}
		\Hb^\infty = &\arg\min_{\Hb \geq \zerob}~	\langle \nabla_{H}\wt F(\Wb^\infty, \Hb^{\infty}), \Hb -\Hb^{\infty} \rangle \notag \\
		&~~~~~~~~~~~~+ \frac{t^\infty}{2}\|\Hb - \Hb^\infty\|_F^2 - \rho \sum_{j = 1}^{N}\|\hb_j\|_\infty, \label{thm4: H_opt1}
	\end{align}
	The optimality condition of \eqref{thm4: H_opt1} yields
	\begin{align}
		&\langle \nabla_{H} F(\Wb^\infty, \Hb^{\infty}), \Hb-
		\Hb^\infty \rangle \notag \\
		&~+ \rho \sum_{j = 1}^{N} \phi^\prime(\hb_j^\infty; \hb_j-
		\hb^\infty_j) \geq 0, ~\forall \Hb \geq \zerob. \label{thm4: H_dstat}
	\end{align}
	Similarly, for the update of $\Wb$, one can show that
	\begin{align}
		&\langle \nabla_{W} F(\Wb^\infty, \Hb^{\infty}), \Wb-\Wb^\infty \rangle \geq 0, \forall \Wb \geq \zerob. \label{thm4: W_dstat}
	\end{align}
	Combing \eqref{thm4: H_dstat} and \eqref{thm4: W_dstat}, we obtain that  $(\Wb^\infty, \Hb^\infty)$ is a d-stationary point to problem \eqref{eqn: onmf prob PP nsmooth}.
	\hfill $\blacksquare$

\vspace{-0.1cm}

\bibliography{refs20,refs10}

\end{document}


\bibliographystyle{IEEEtran}
%

\begin{center}
\LARGE
\bf Supplementary Materials of Manuscript "Clustering by Orthogonal NMF Model and Non-Convex Penalty Optimization"

\end{center}

\begin{center}
	\large
	Shuai Wang, Tsung-Hui Chang, Ying Cui, and Jong-Shi Pang	
	
\end{center}
%
%

{\setcounter{section}{0}}

{\setcounter{equation}{0}
	\renewcommand{\theequation}{S.\arabic{equation}}

\section{Algorithm convergence with the effect of $\mu_w$ and $\mu_h$}

We examine the performance of the proposed SNCP and NSNCP on the synthetic data with SNR = -3 dB using different values of $\mu_w$ and $\mu_h$. With fixed stopping condition $\epsilon_{\rm PALM} < 3 \times 10^{-3}$ and $\gamma = 1.1$, Fig. \ref{fig:mu&v} shows the clustering accuracy, the normalized residual, and the orthogonality versus the iteration number of Algorithm 1 when the SNCP is used under different combinations of $\mu_w$ and $\mu_h$ from $\{0, 10^{-10}, 10^{-8}, 10^{-6}, 10^{-4}\}$. 

One can observe from Fig. \ref{fig:mu&v}(a)-(f) that, with fixed $\mu_w$ and a smaller $\mu_h$, the proposed SNCP not only converges faster but also achieves a higher clustering accuracy. On the other hand, it is observed from Fig. \ref{fig:mu&v}(g)-(l) that a bigger $\mu_w$ may speed up the convergence of SNCP when the $\mu_h$ is fixed. However, Fig. \ref{fig:acc_nu0} and Fig. \ref{fig:acc_nu1e_6} tell us that the clustering accuracy achieved by SNCP does not increase when the $\mu_w$ increases and it becomes the worst when $\mu_w = 10^{-4}$ since some rows of $\Hb$ become $\zerob$ with a large $\mu_w$ and $\mu_h$, as seen from Fig. \ref{fig:mu&v}(g)-(l).

As show in Fig. \ref{fig:nsncp_mu&v}, similar behaviors of the proposed NSNCP corresponding to different choices of $\mu_w$ and $\mu_h$ can be observed. As a result, it is reasonable to choose small values of $\mu_w$ and $\mu_h$ for both SNCP and NSNCP if they are not set to zero.

\begin{figure}[t!]
	\centering
	\subfigure[\scriptsize ]{
		\includegraphics[width=4cm]{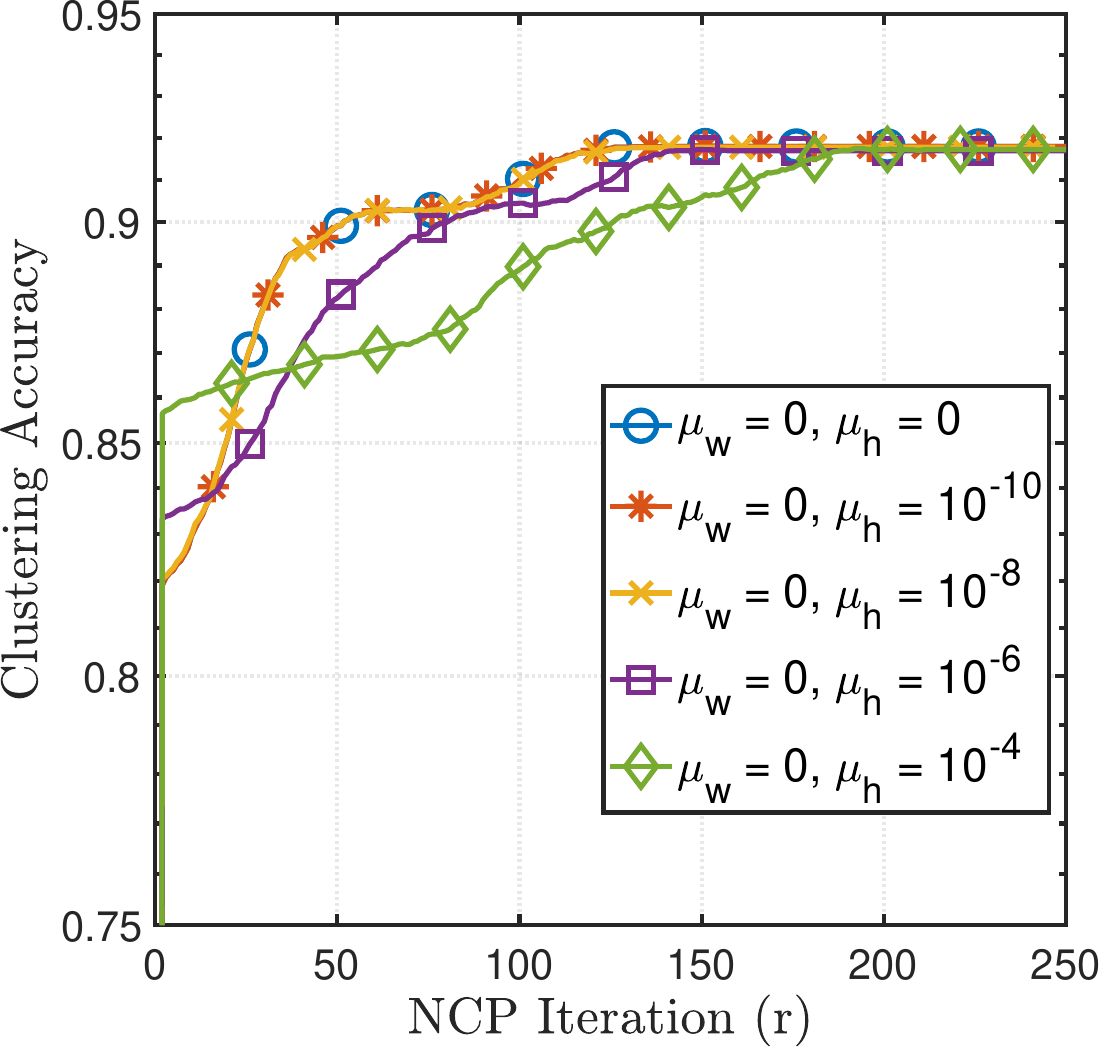}
		\label{fig:acc_mu0}
	}
	\subfigure[\scriptsize ]{
		\includegraphics[width=4cm]{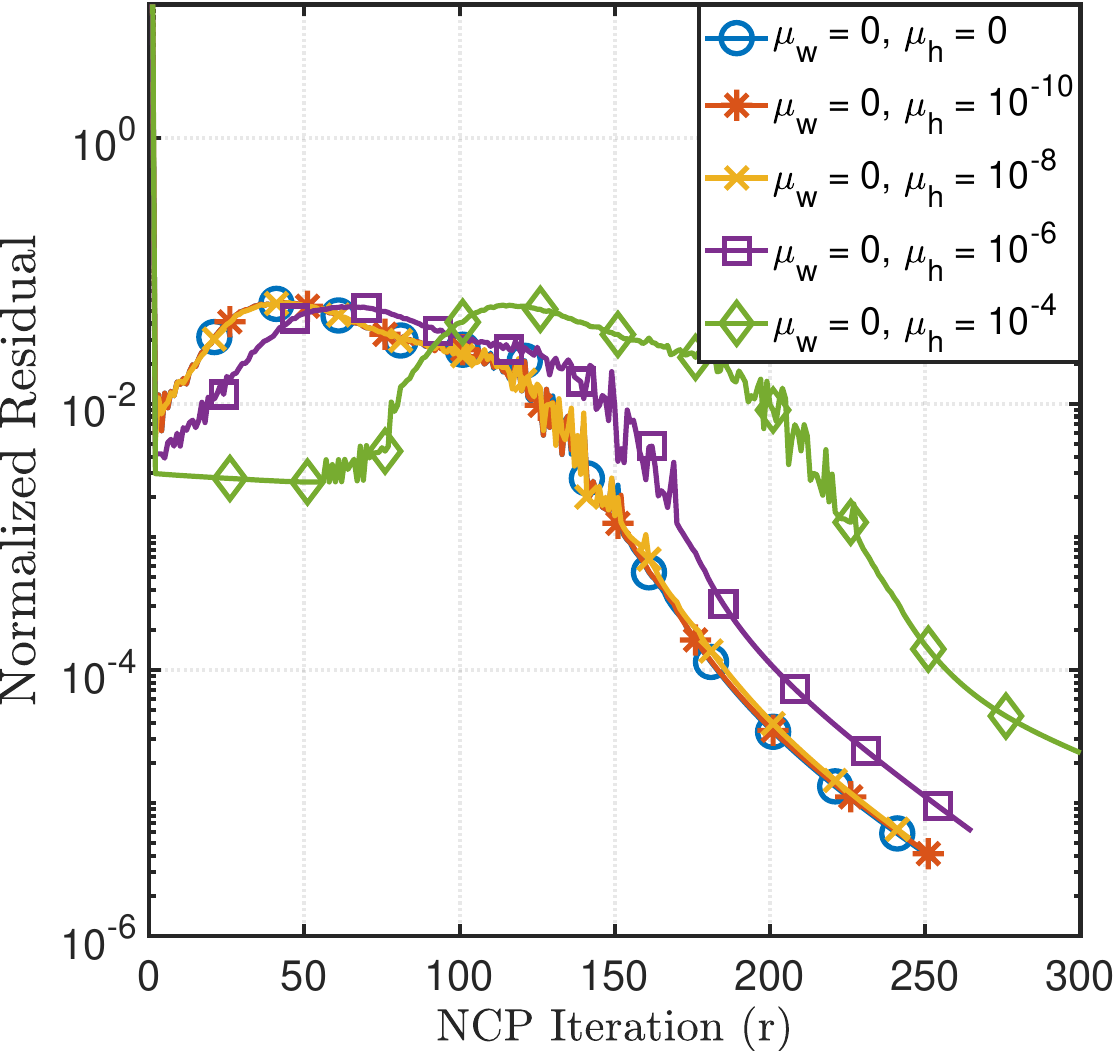}
		\label{fig:nr_mu0}
	}
	\subfigure[\scriptsize ]{
		\includegraphics[width=4cm]{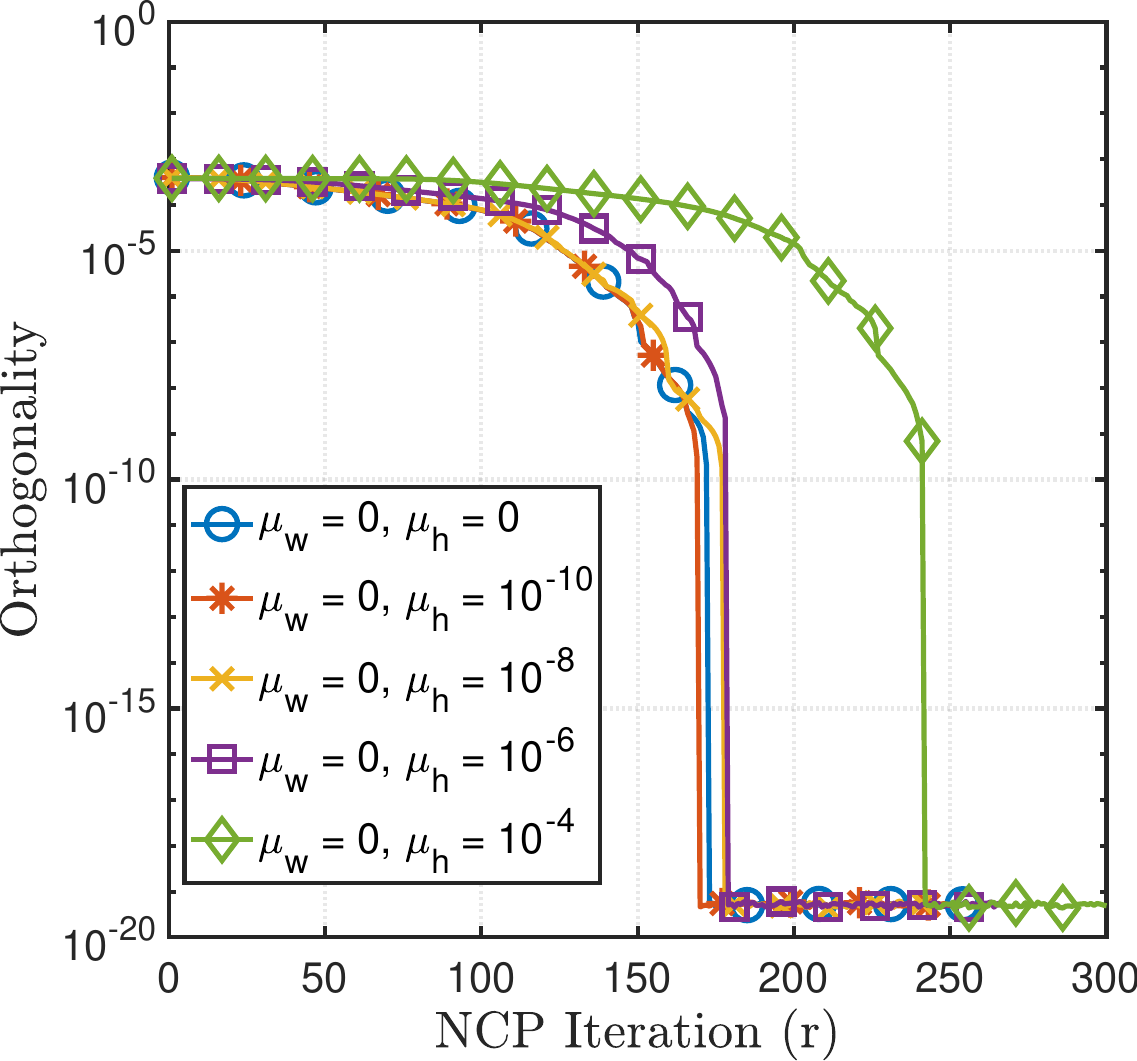}
		\label{fig:ortho_mu0}
	}
	\subfigure[\scriptsize ]{
		\includegraphics[width=4cm]{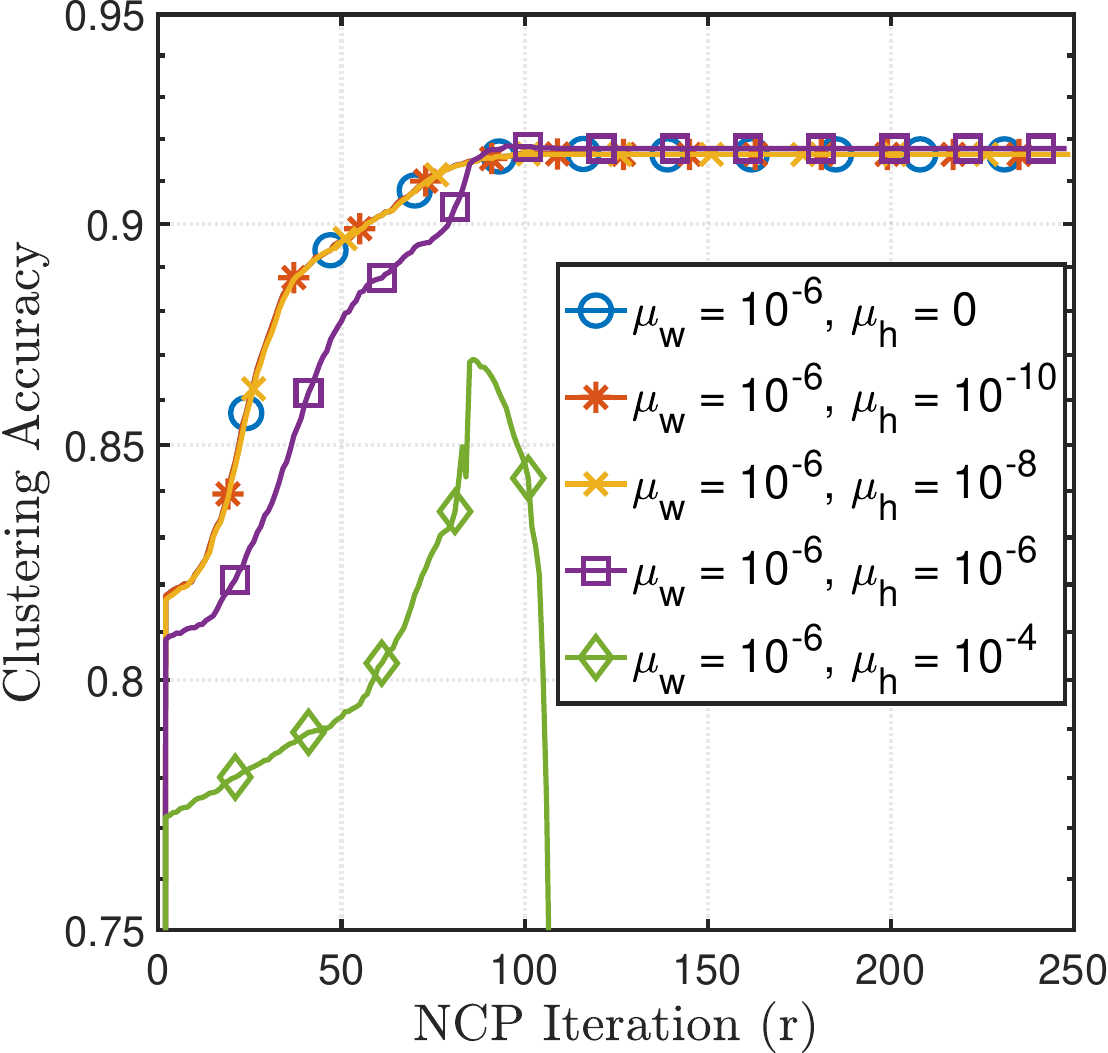} 
		\label{fig:acc_mu1e_6}
	}
	\subfigure[\scriptsize ]{
		\includegraphics[width=4cm]{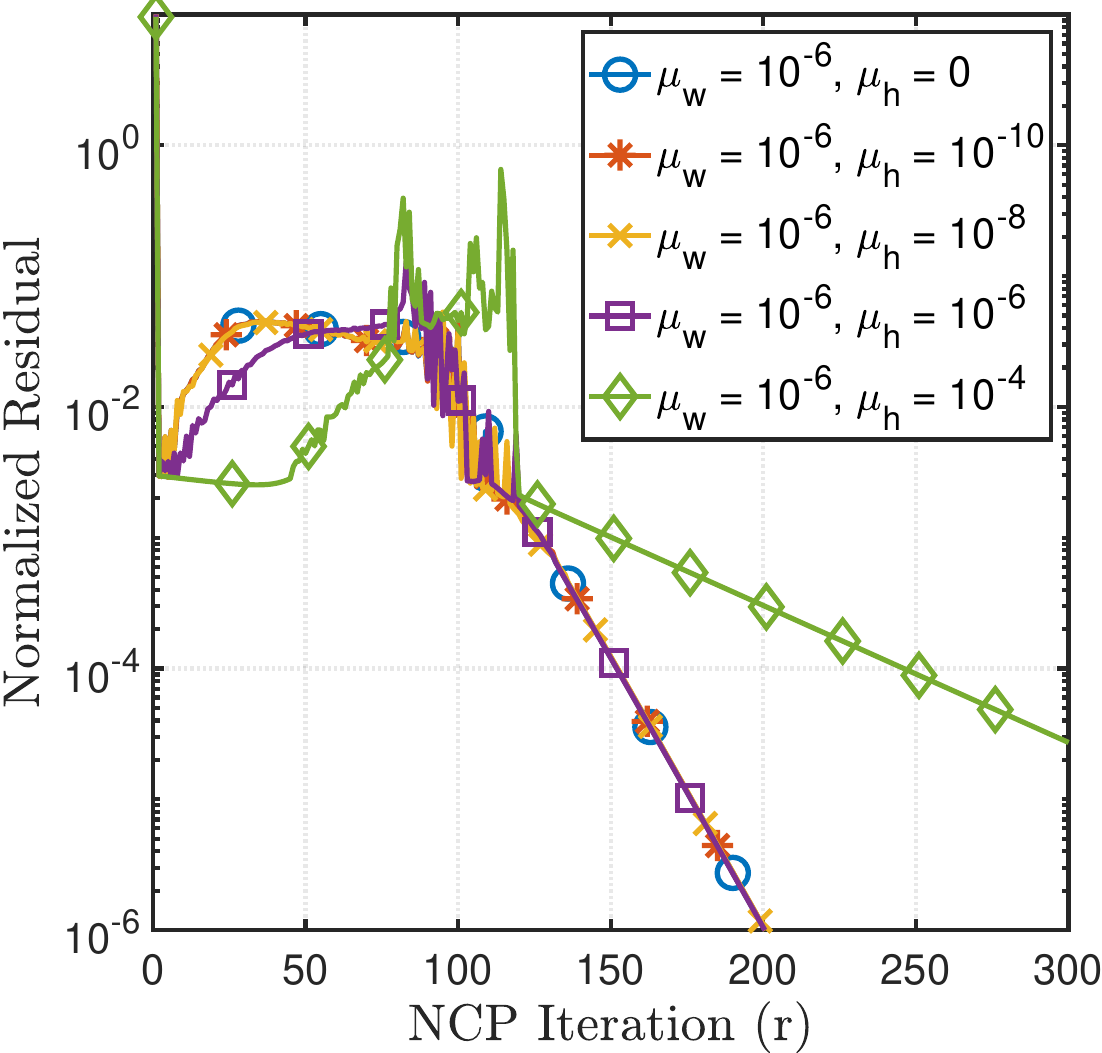} 
		\label{fig:nr_mu1e_6}
	}
	\subfigure[\scriptsize ]{
		\includegraphics[width=4cm]{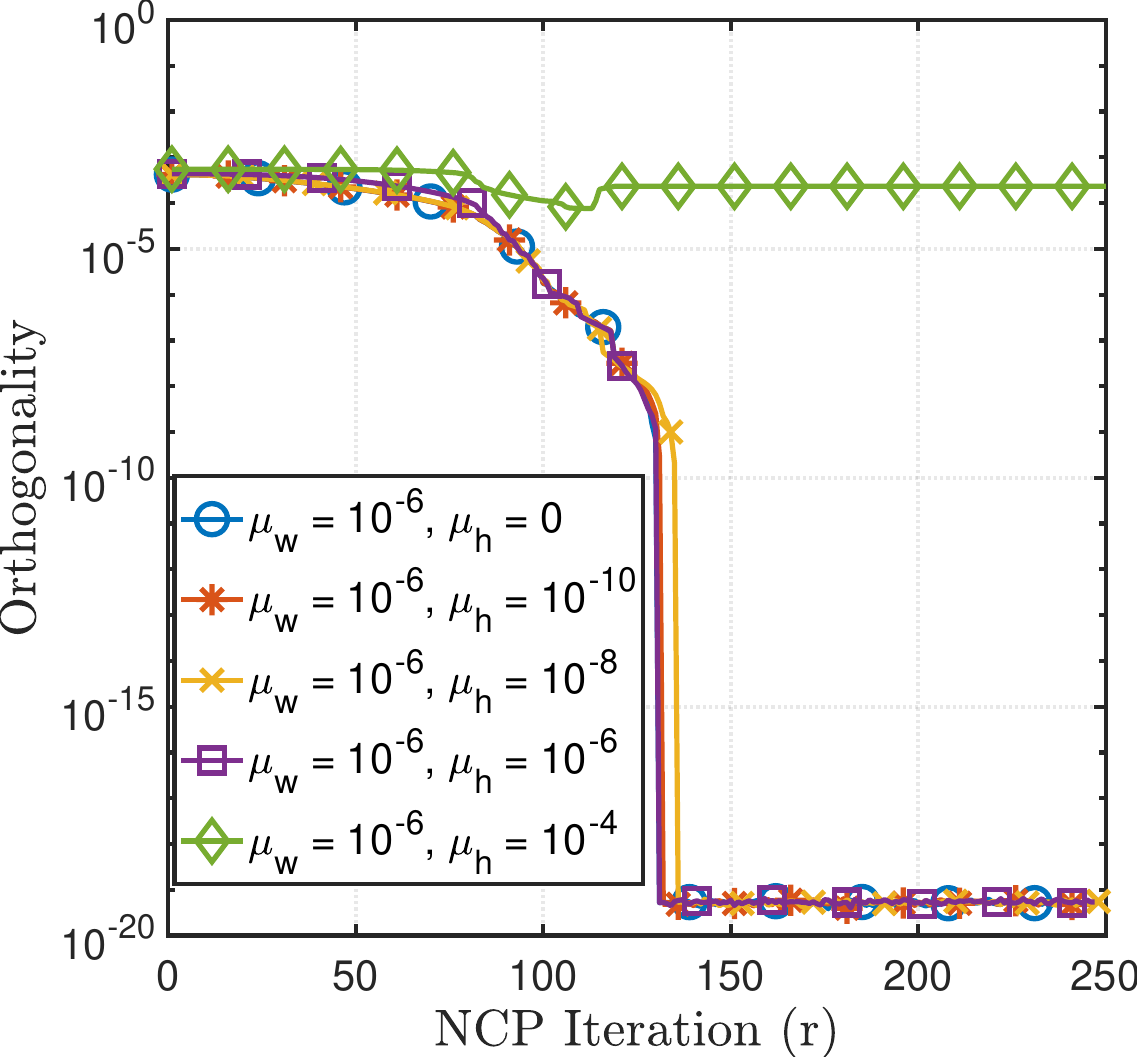} 
		\label{fig:ortho_mu1e_6}
	}
	\subfigure[\scriptsize ]{
		\includegraphics[width=4cm]{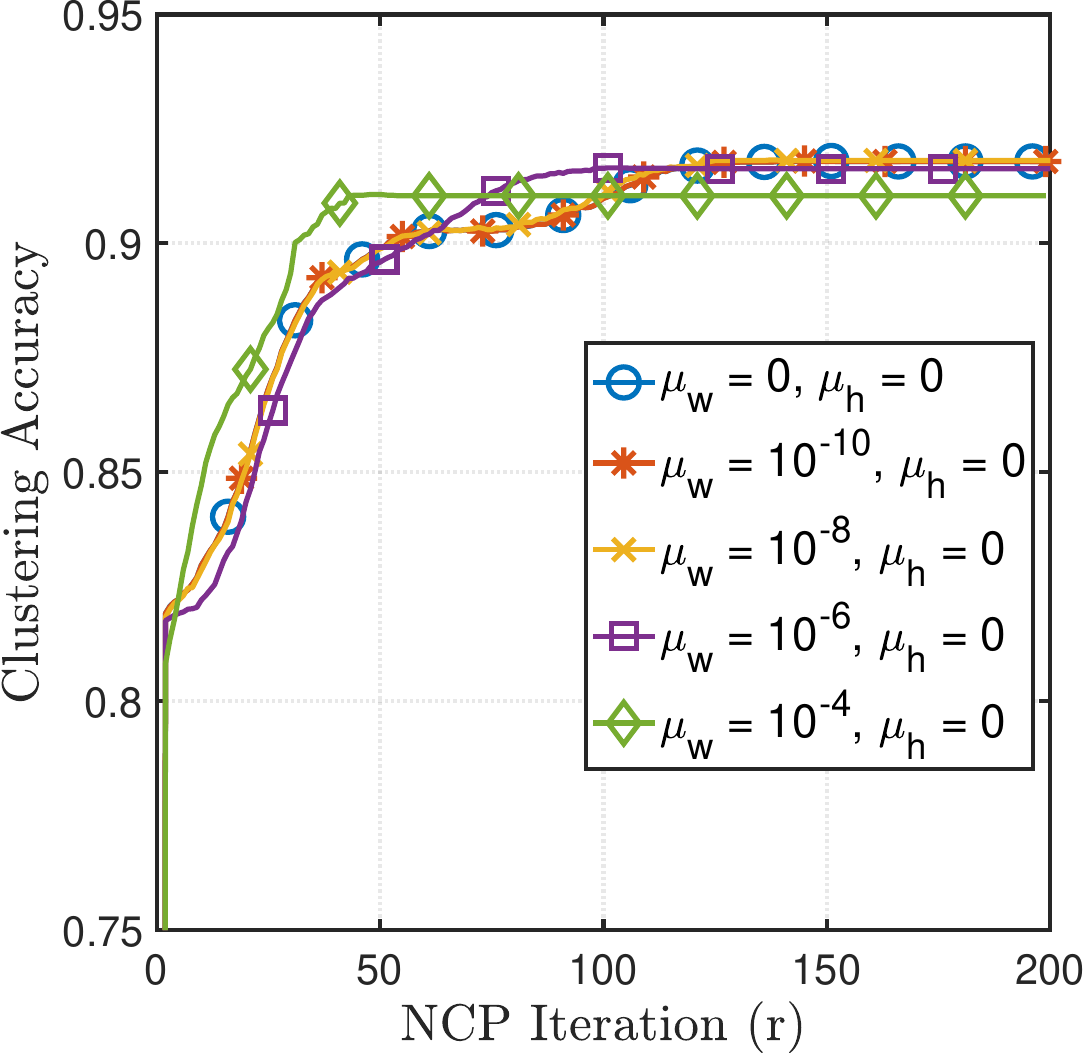} 
		\label{fig:acc_nu0}
	}
	\subfigure[\scriptsize ]{
		\includegraphics[width=4cm]{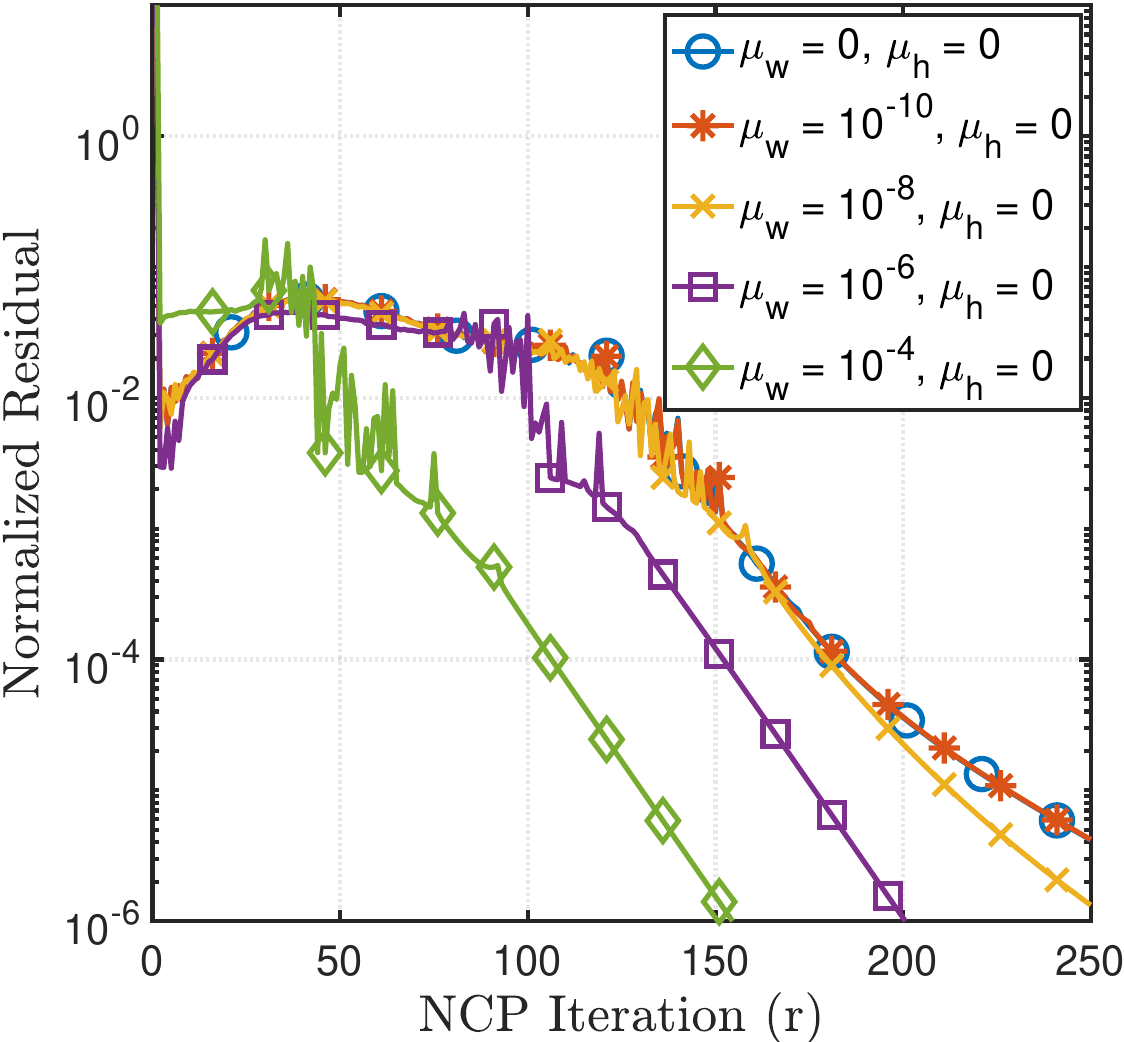} 
		\label{fig:nr_nu0}
	}
	\subfigure[\scriptsize ]{
		\includegraphics[width=4cm]{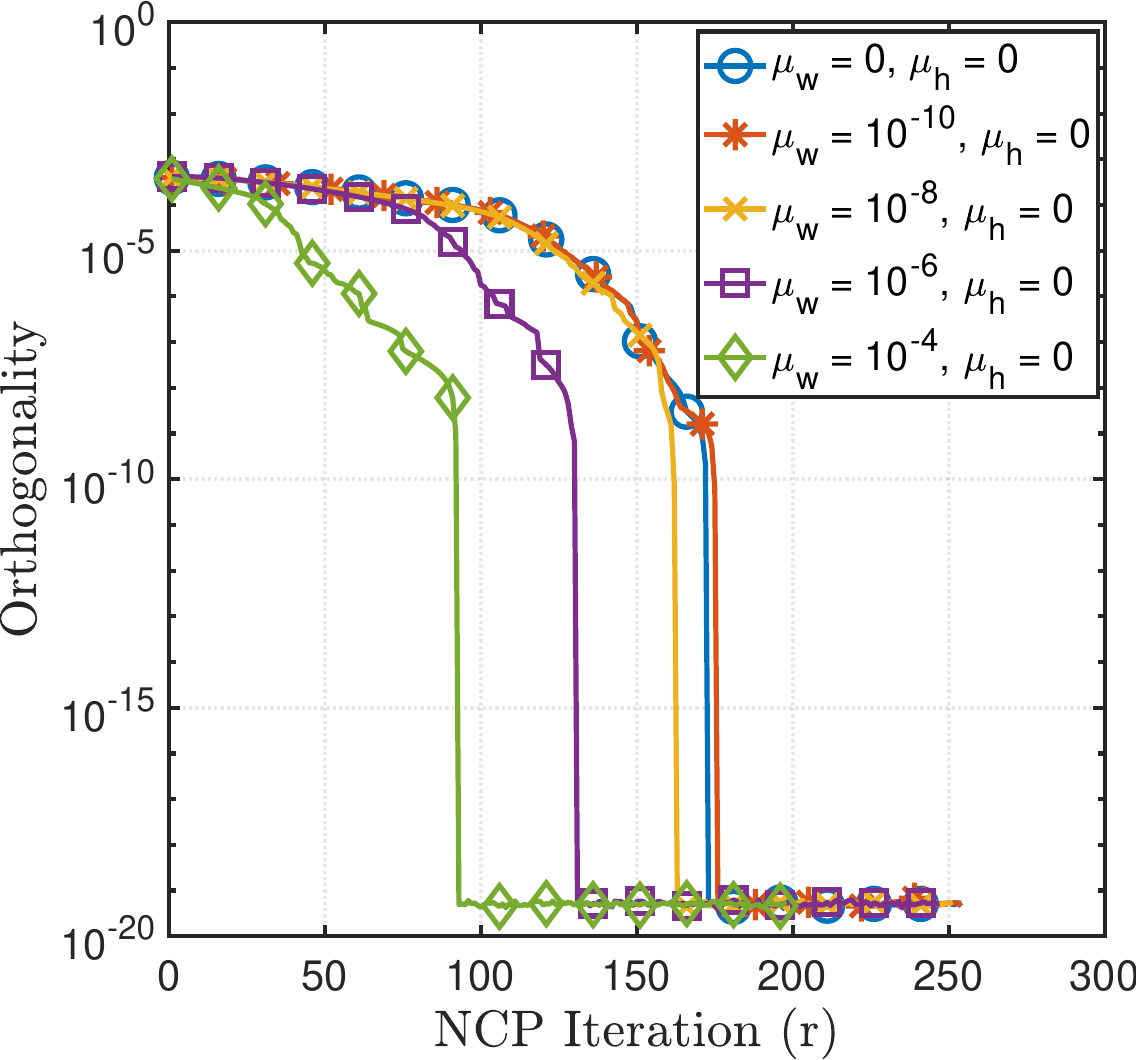} 
		\label{fig:ortho_nu0}
	}
	\subfigure[\scriptsize ]{
		\includegraphics[width=4cm]{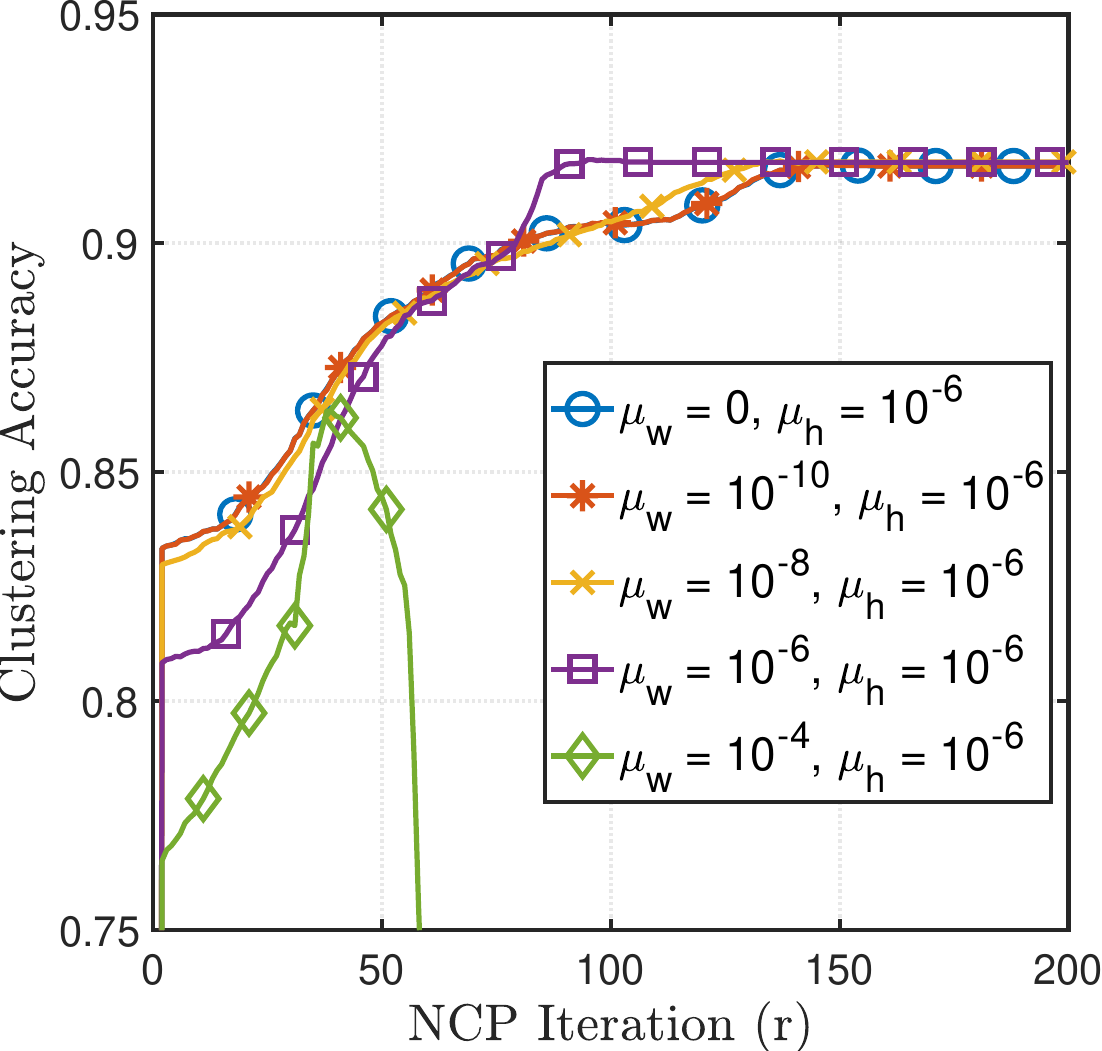} 
		\label{fig:acc_nu1e_6}
	}
	\subfigure[\scriptsize ]{
		\includegraphics[width=4cm]{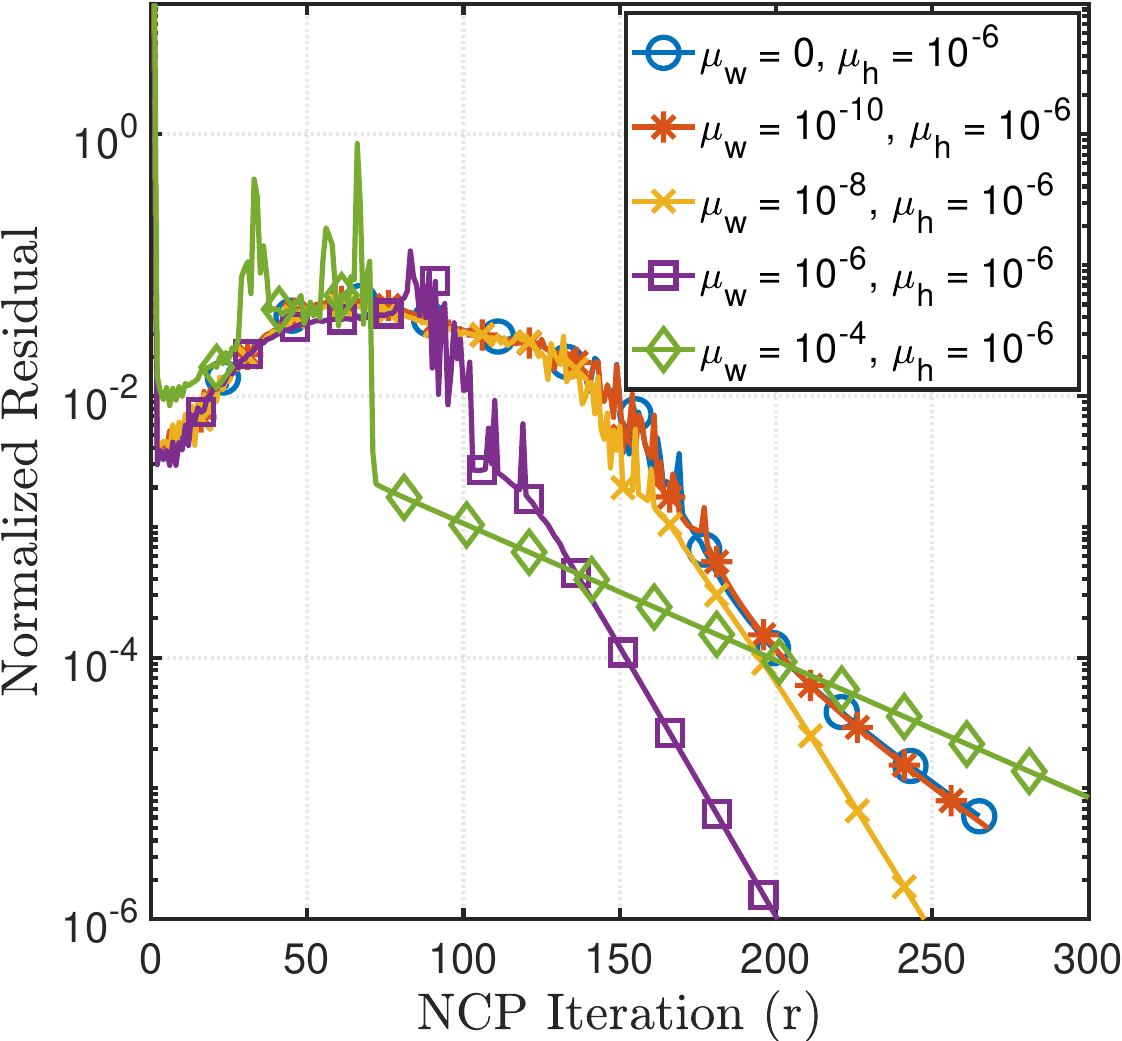} 
		\label{fig:nr_nu1e_6}
	}
	\subfigure[\scriptsize ]{
		\includegraphics[width=4cm]{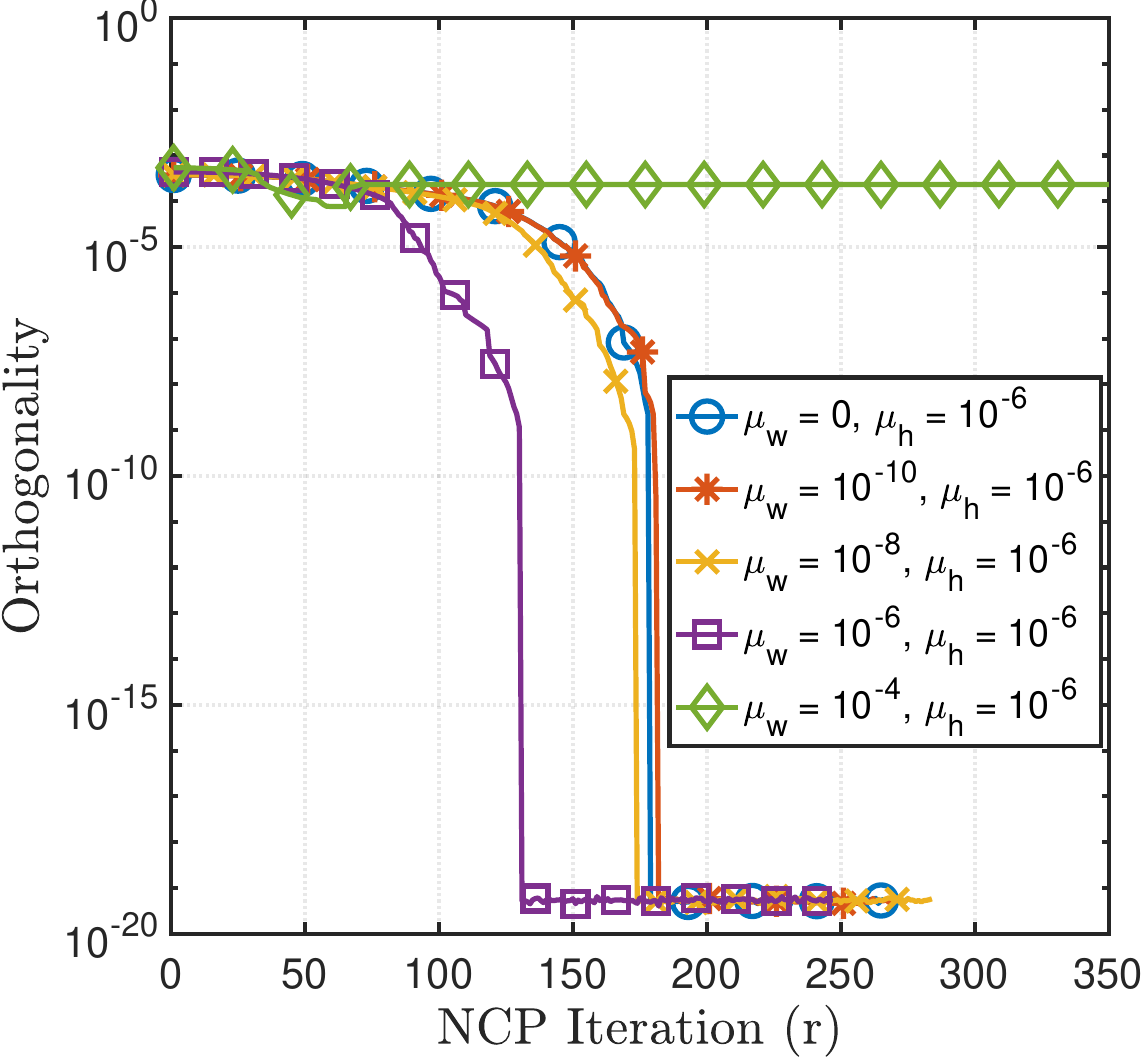} 
		\label{fig:ortho_nu1e_6}
	}
	\caption{Convergence curves of SNCP under different values of $\mu_w$ and $\mu_h$.}
	\vspace{-0.5cm}
	\label{fig:mu&v}
\end{figure}

\begin{figure} [t!]
	\centering
	\subfigure[\scriptsize ]{
		\includegraphics[width=4cm]{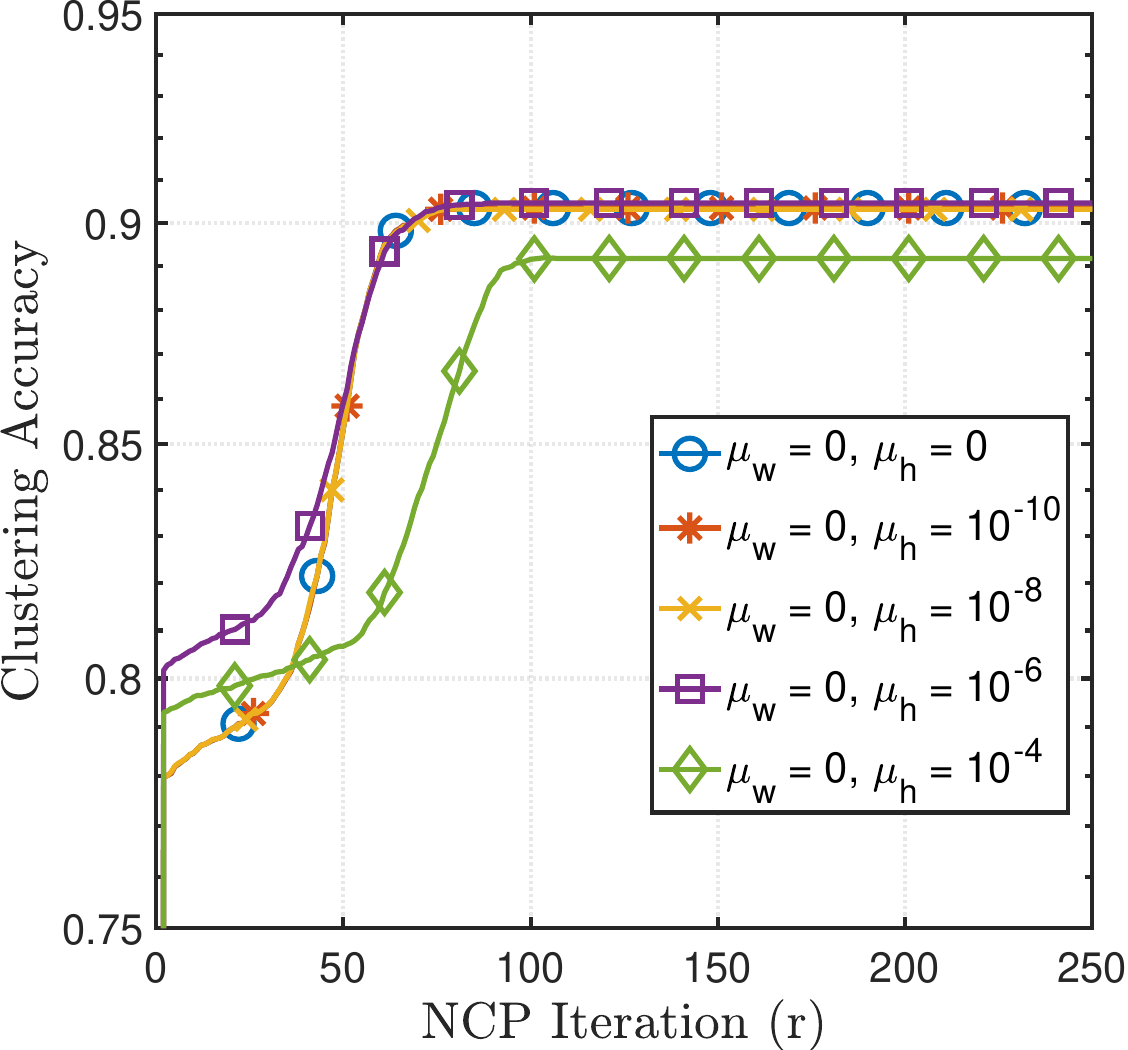}
		\label{fig:nsncp_acc_mu0}
	}
	\subfigure[\scriptsize ]{
		\includegraphics[width=4cm]{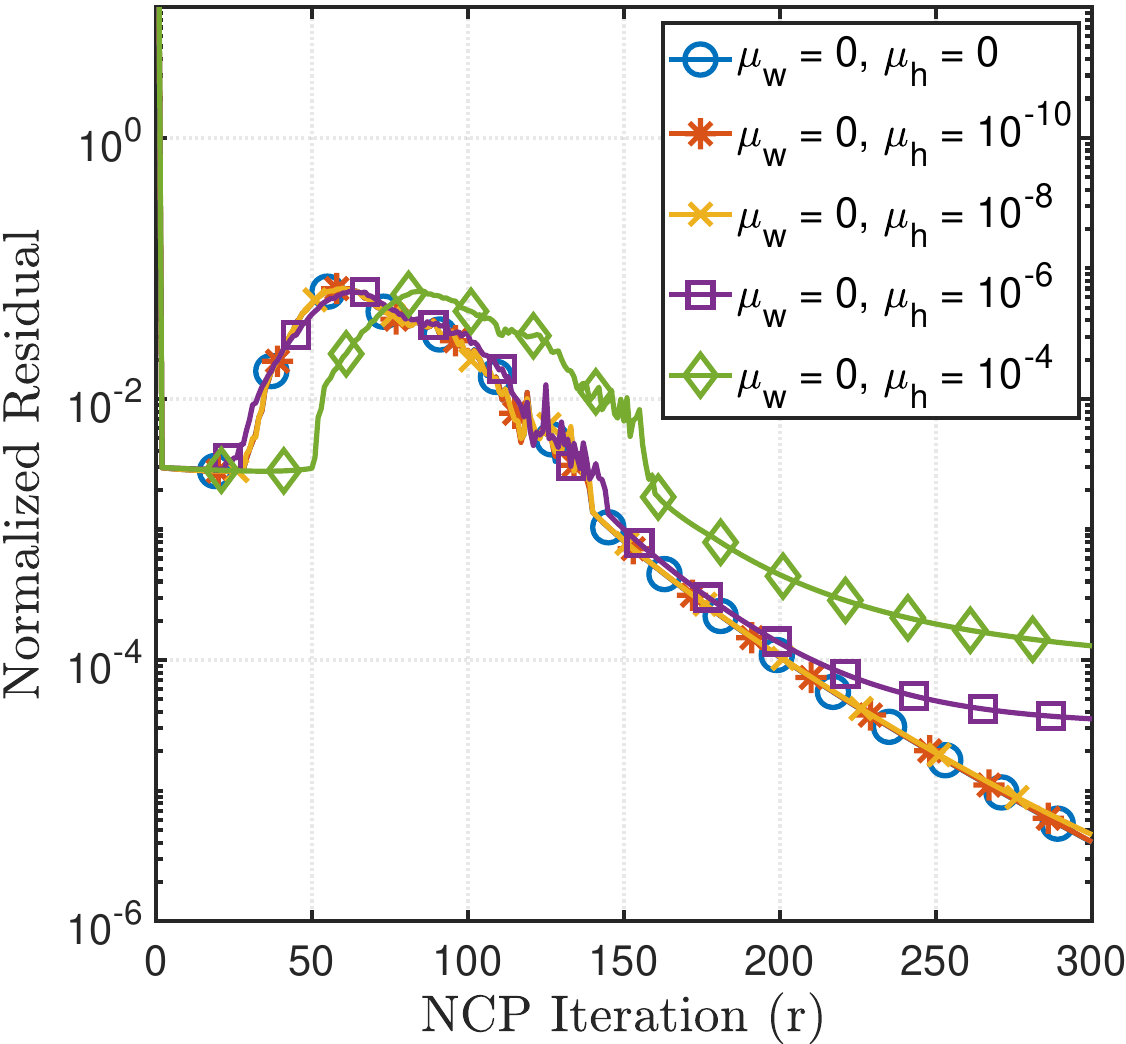}
		\label{fig:nsncp_nr_mu0}
	}
	\subfigure[\scriptsize ]{
		\includegraphics[width=4cm]{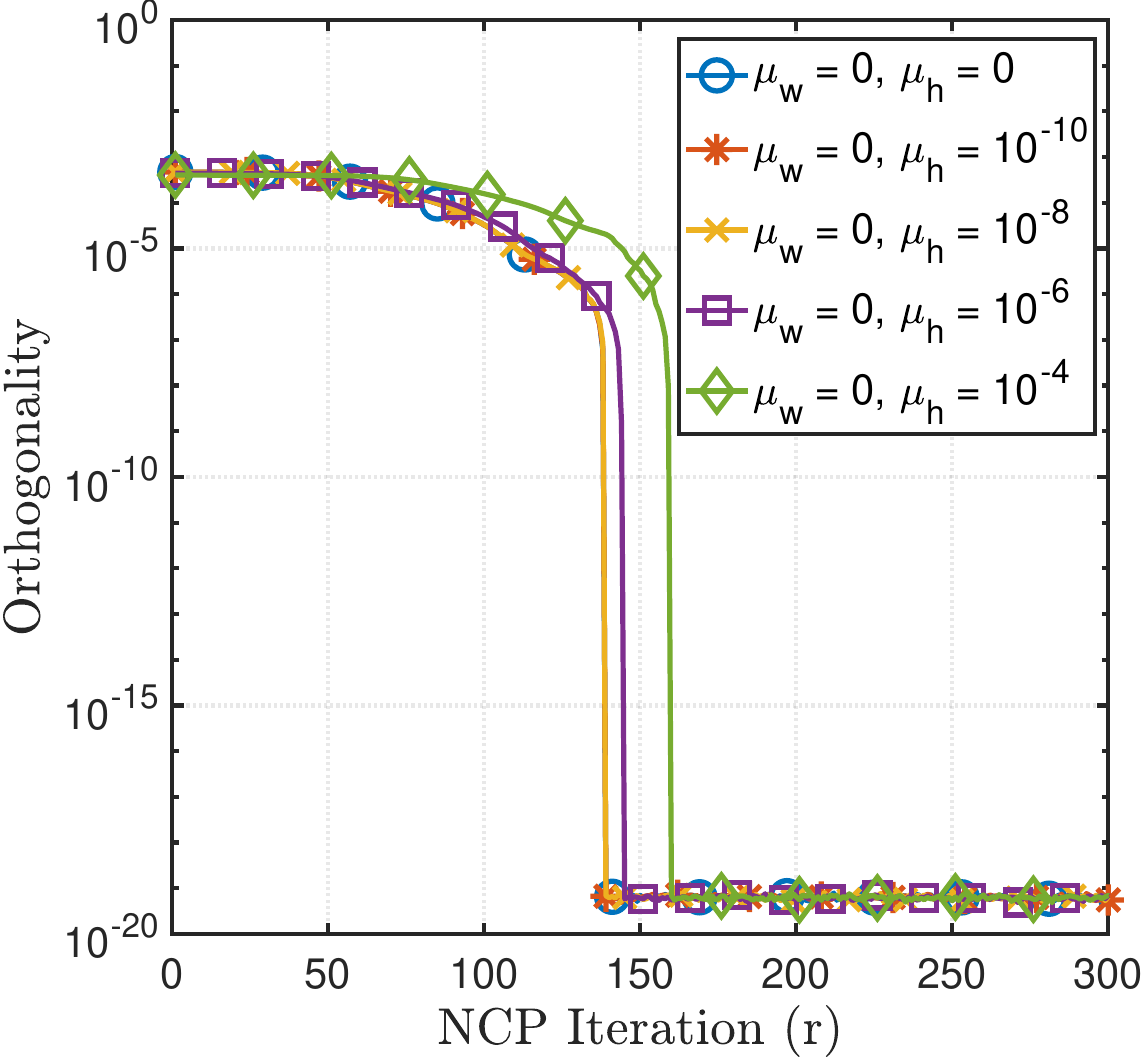}
		\label{fig:nsncp_ortho_mu0}
	}
	\subfigure[\scriptsize ]{
		\includegraphics[width=4cm]{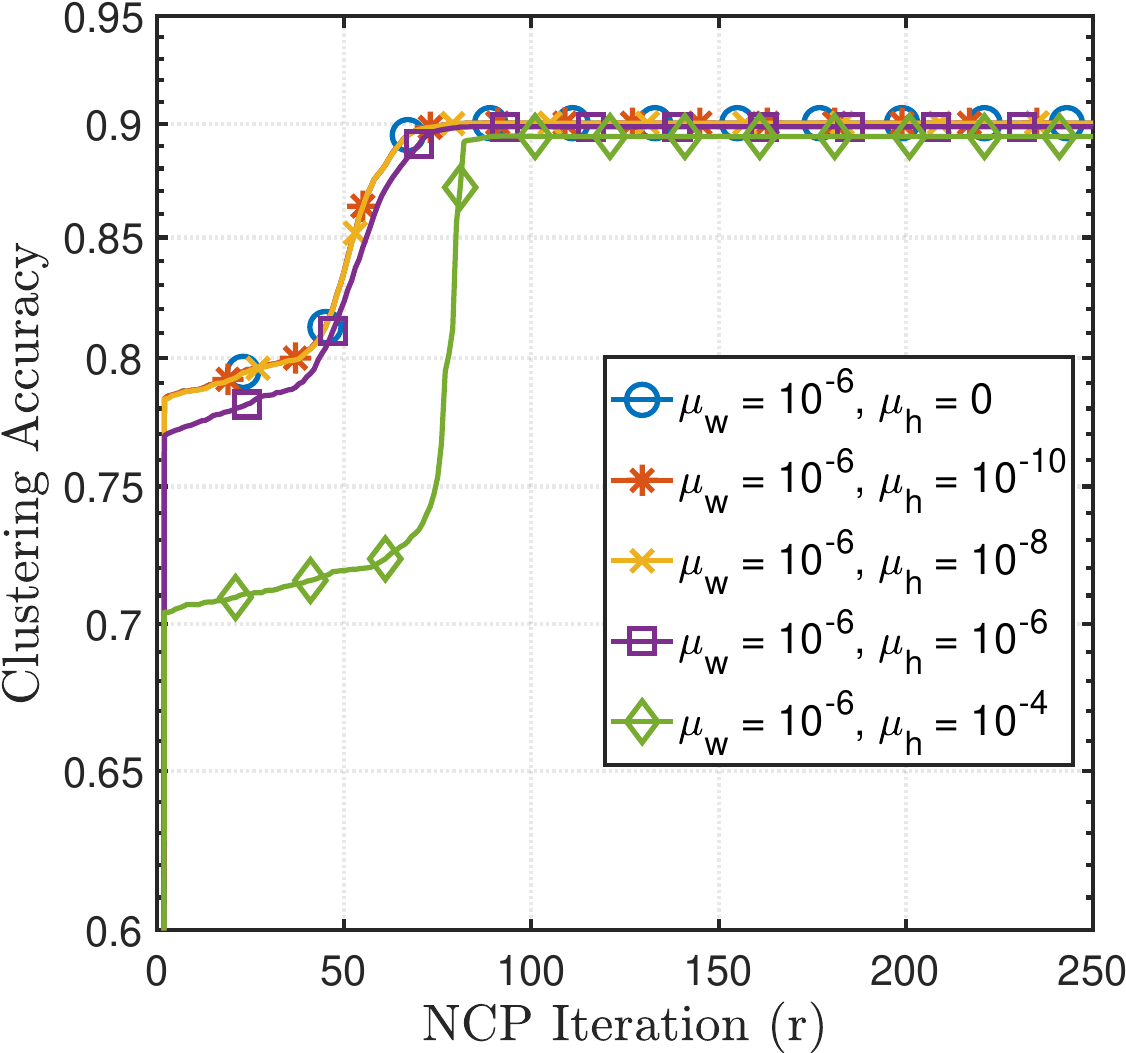} \label{fig:nsncp_acc_mu1e_6}
	}
	\subfigure[\scriptsize ]{
		\includegraphics[width=4cm]{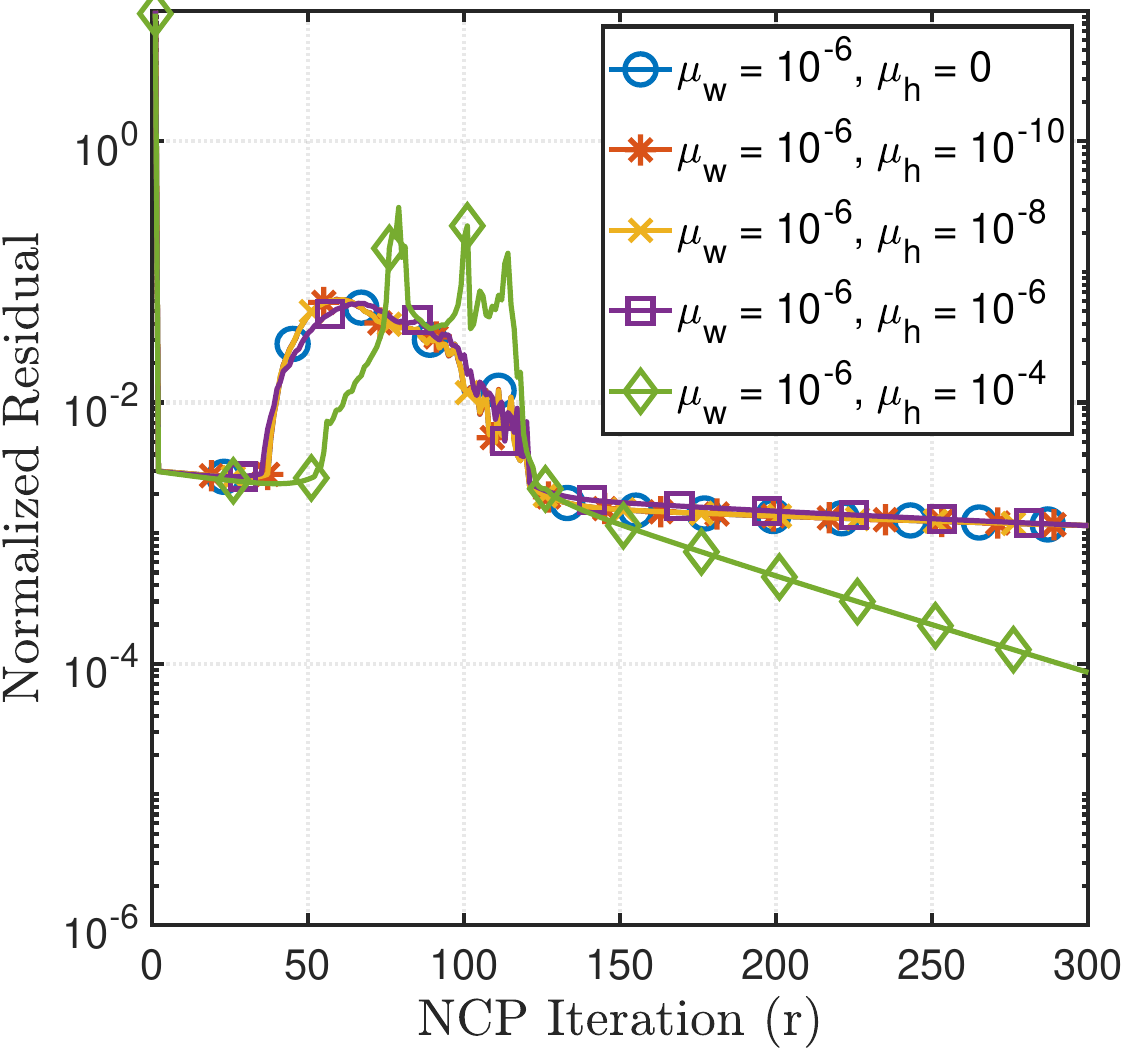} \label{fig:nsncp_nr_mu1e_6}
	}
	\subfigure[\scriptsize ]{
		\includegraphics[width=4cm]{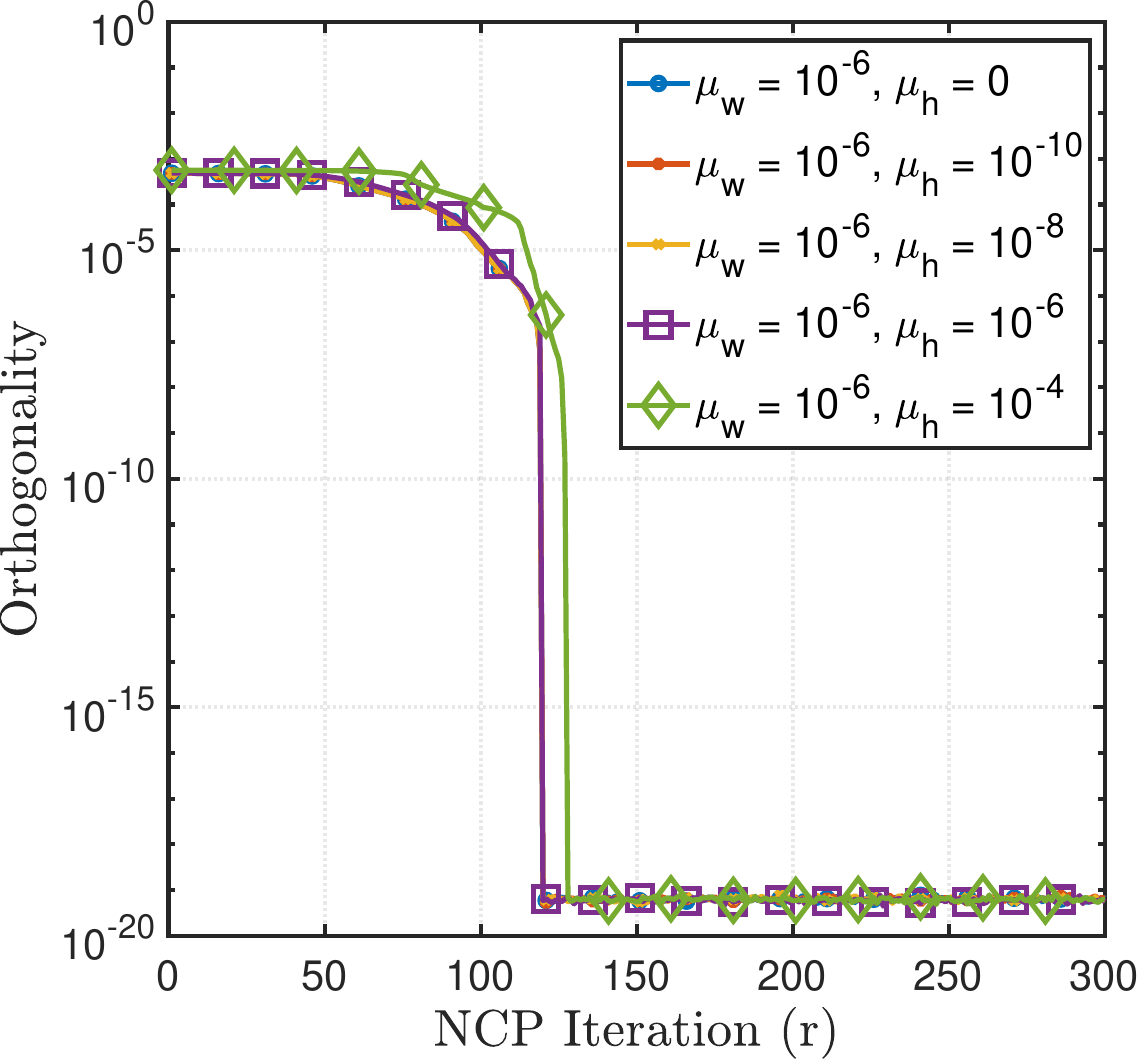} \label{fig:nsncp_ortho_mu1e_6}
	}
	\subfigure[\scriptsize ]{
		\includegraphics[width=4cm]{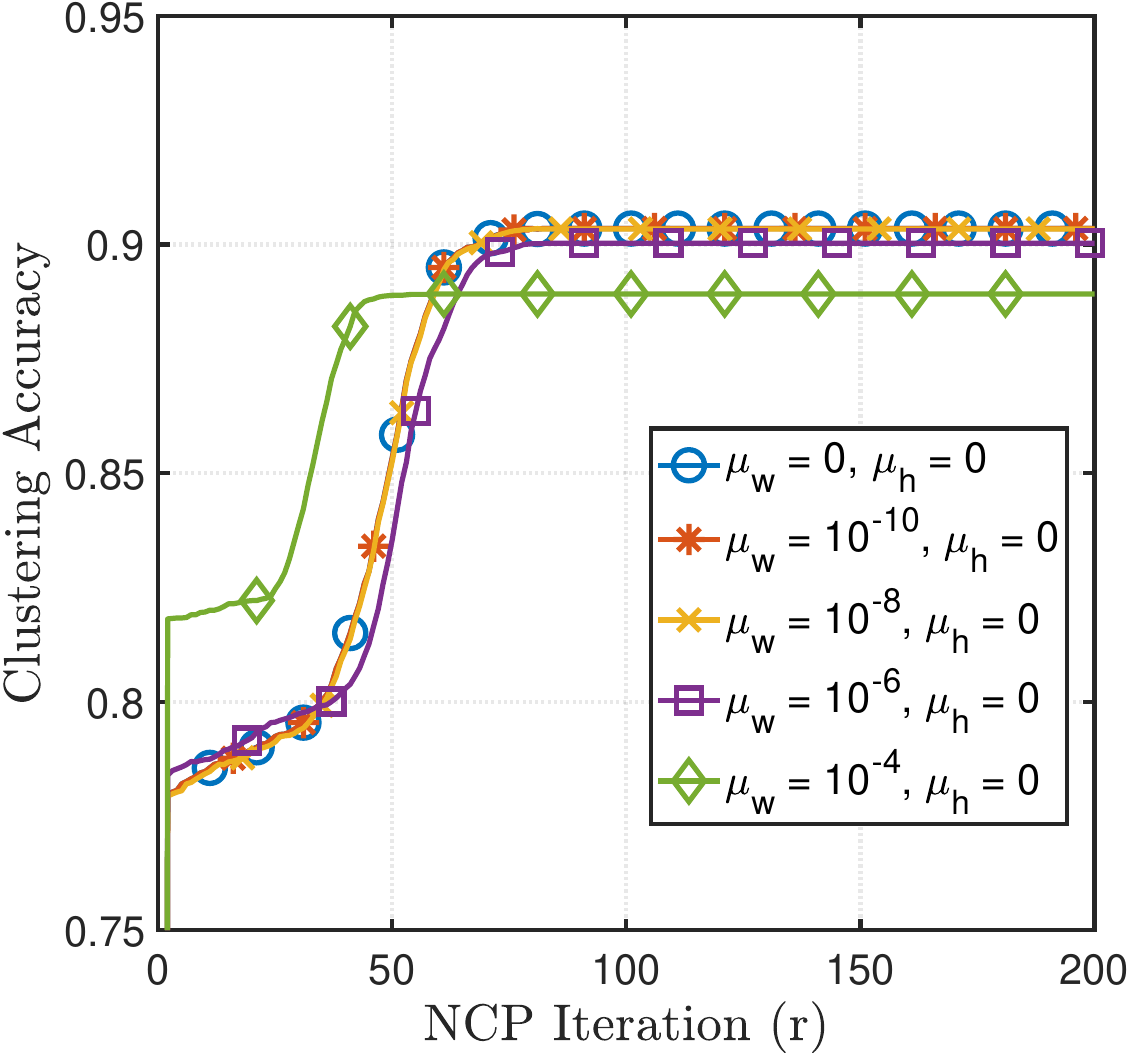} \label{fig:nsncp_acc_nu0}
	}
	\subfigure[\scriptsize ]{
		\includegraphics[width=4cm]{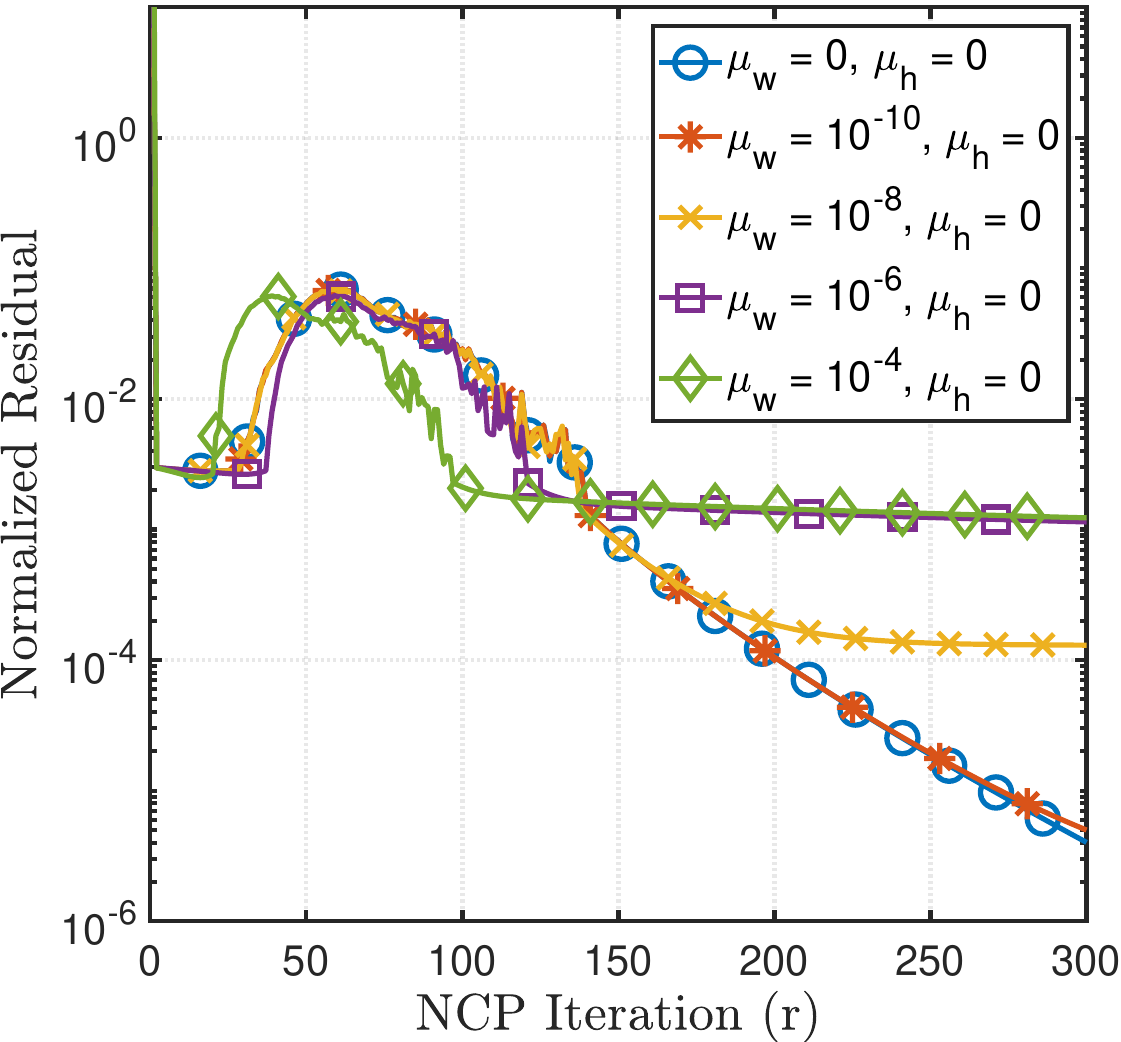} \label{fig:nsncp_nr_nu0}
	}
	\subfigure[\scriptsize ]{
		\includegraphics[width=4cm]{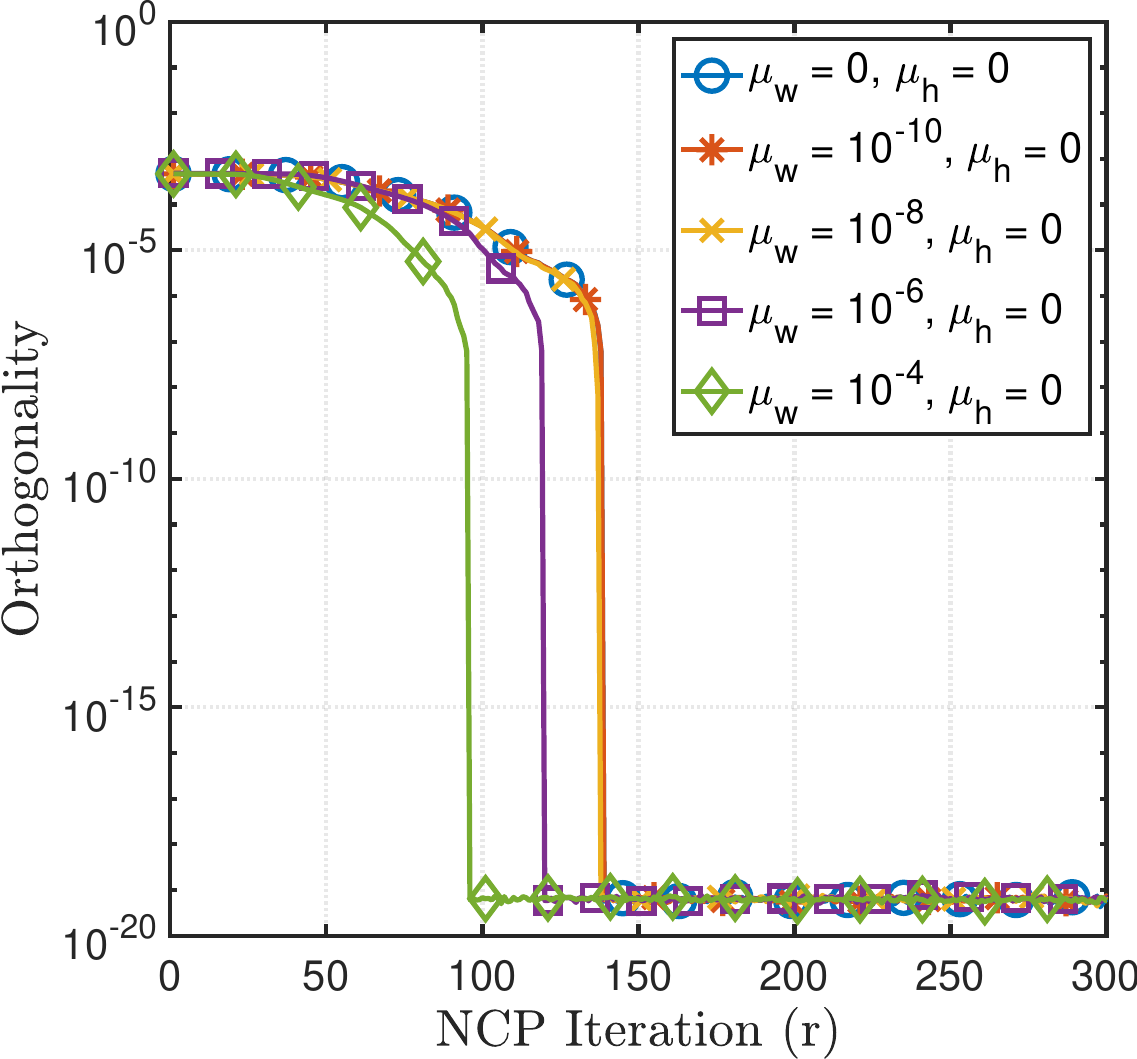} \label{fig:nsncp_ortho_nu0}
	}
	\subfigure[\scriptsize ]{
		\includegraphics[width=4cm]{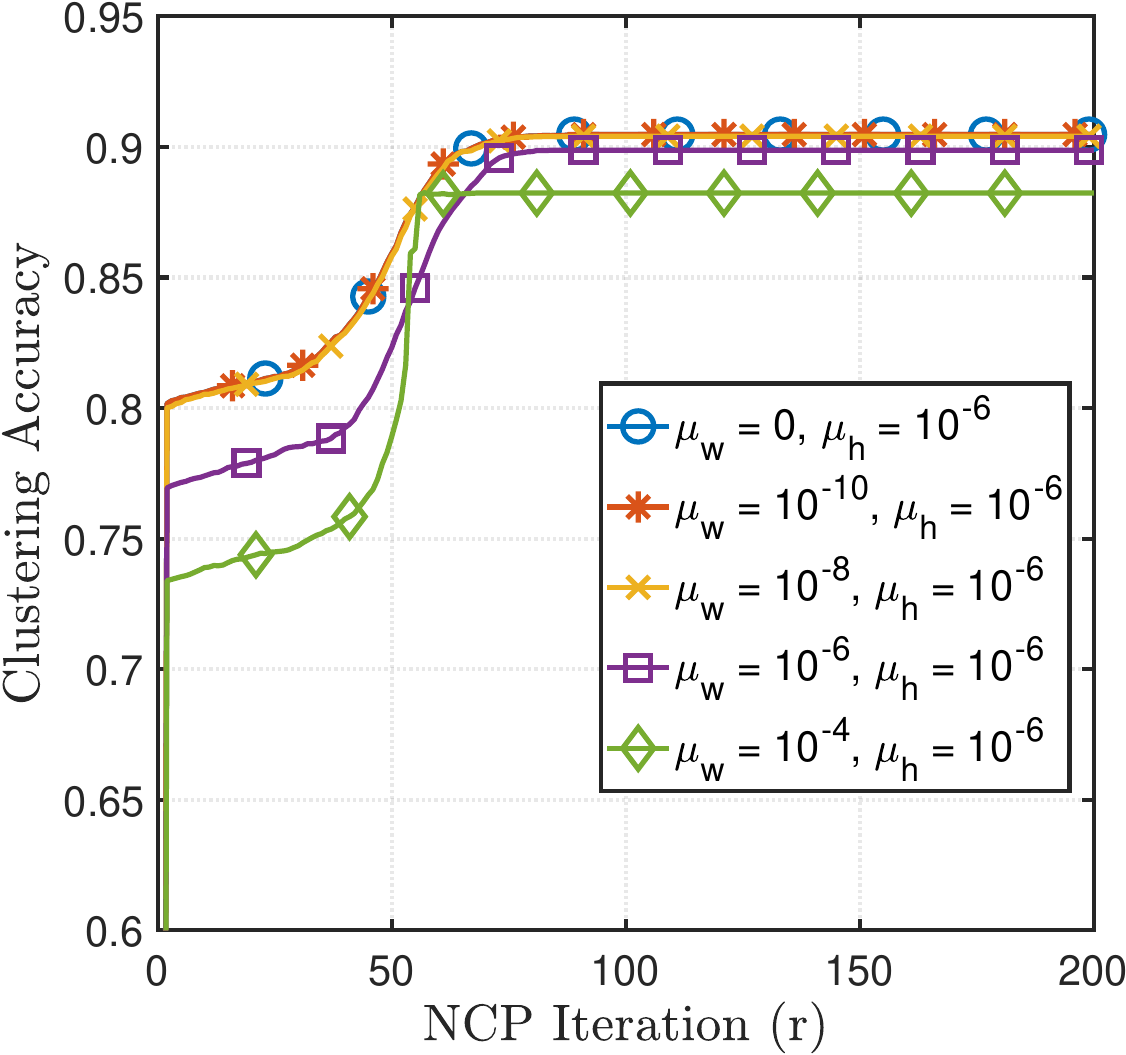} \label{fig:nsncp_acc_nu1e_6}
	}
	\subfigure[\scriptsize ]{
		\includegraphics[width=4cm]{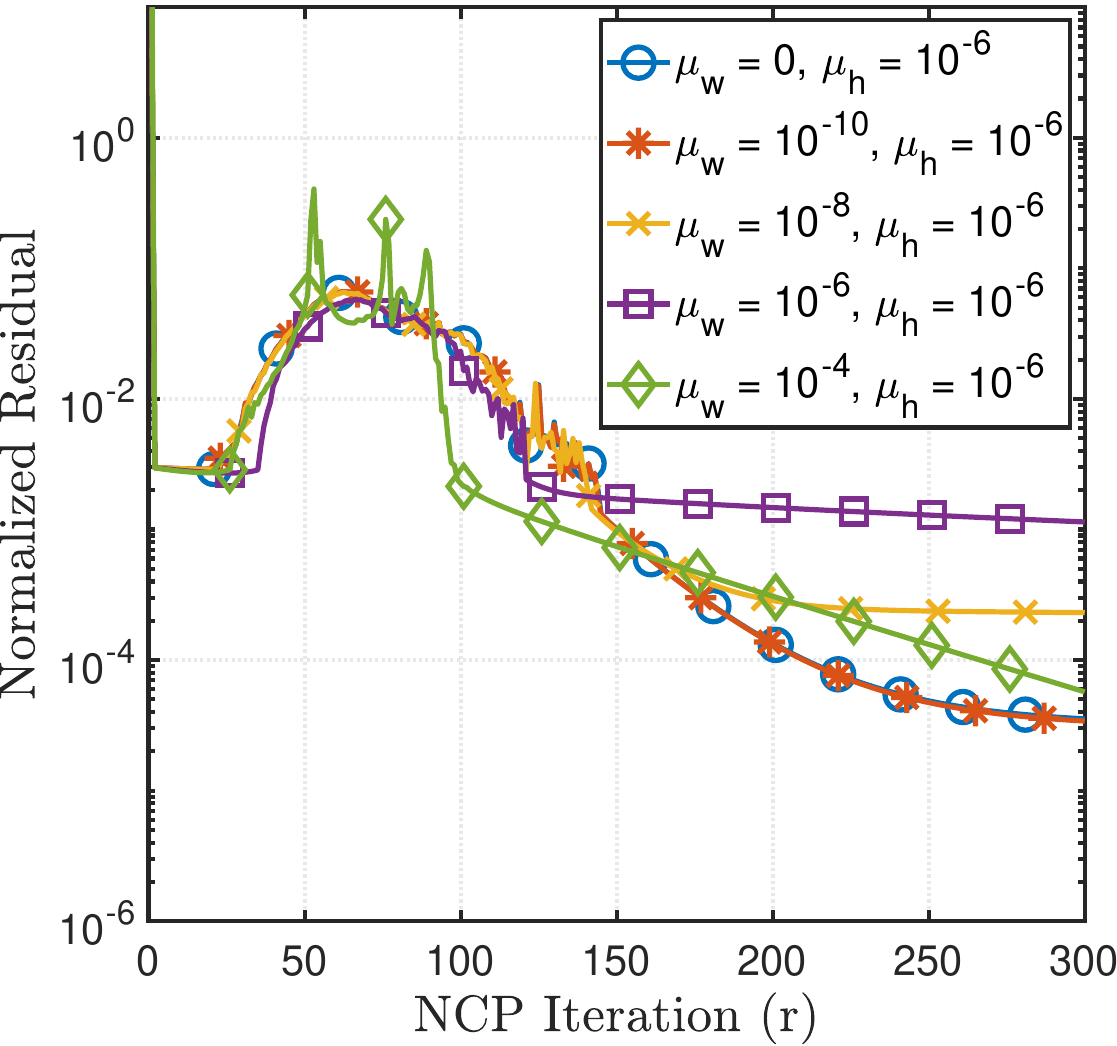} \label{fig:nsncp_nr_nu1e_6}
	}
	\subfigure[\scriptsize ]{
		\includegraphics[width=4cm]{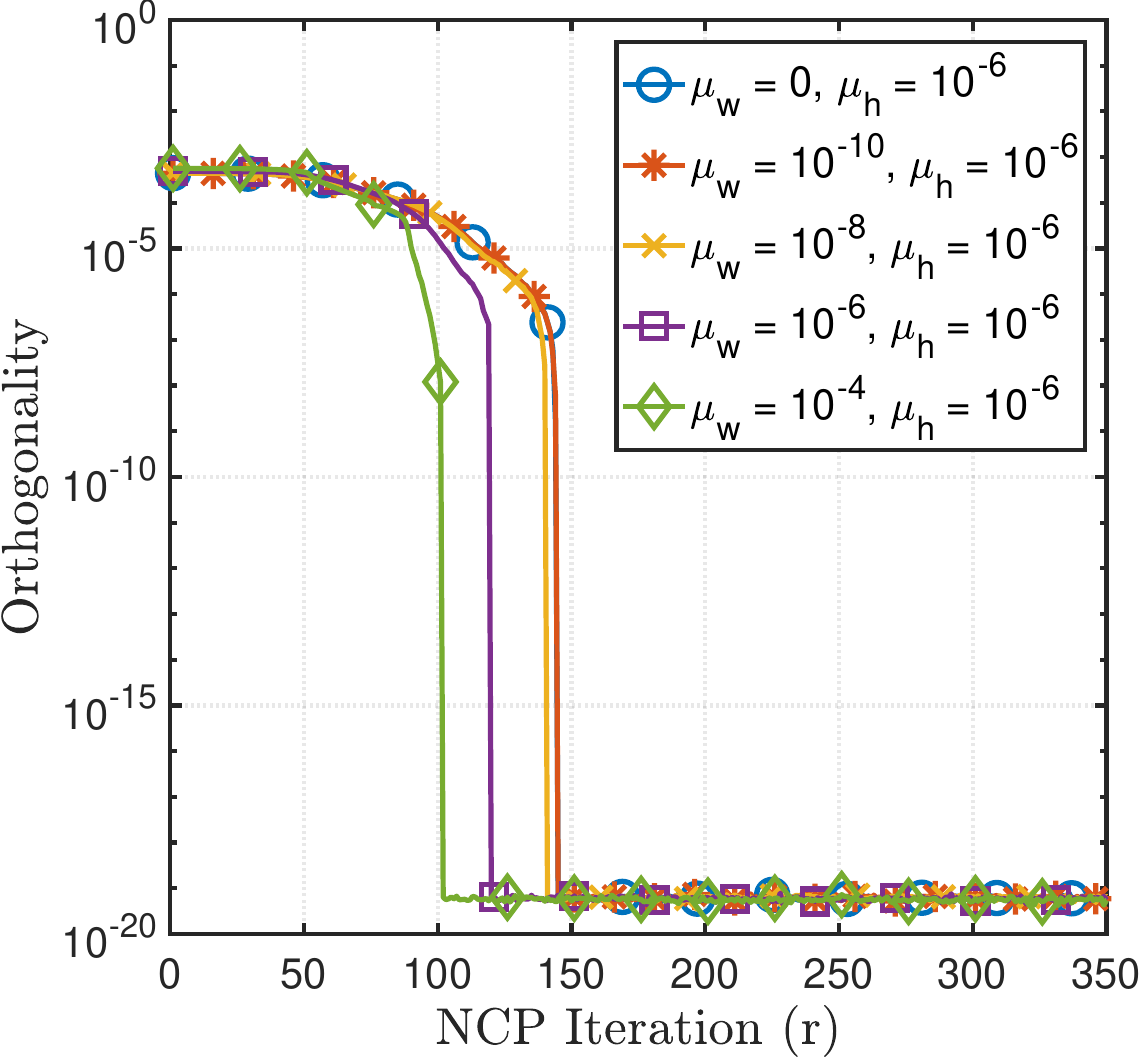} \label{fig:nsncp_ortho_nu1e_6}
	}\vspace{-0.3cm}
	\caption{Convergence curves of NSNCP under different values of $\mu_w$ and $\mu_h$.}
	\vspace{-0.5cm}
	\label{fig:nsncp_mu&v}
\end{figure}

\section{ Algorithm convergence with the effect of $\epsilon_{\rm PALM}$ and $\gamma$}
Let us examine the convergence behaviors of the proposed SNCP and NSNCP methods for different values of $\epsilon_{\rm PALM}$ and $\gamma$. The synthetic data with SNR = -3 is considered and $\mu_w = 0$ and $\mu_h = 0$. are set for Algorithm 1. Fig. \ref{fig:sncp_epsilon&gamma} and Fig. \ref{fig:nsncp_epsilon&gamma} respectively display convergence curves versus the iteration number of  the SNCP and NSNCP methods under different values of $\epsilon_{\rm PALM}$ and $\gamma$.

As shown in Fig. \ref{fig:sncp_epsilon&gamma}(a)-(f), with fixed $\epsilon_{\rm PALM}$, the larger $\gamma$ leads to a faster convergence of SNCP. However, it is observed in Fig. \ref{fig:sncp_epsilon&gamma}(a) and \ref{fig:sncp_epsilon&gamma}(d) that when $\epsilon_{\rm PALM} < 10^{-2}$, the performance is sensitive to the value of $\gamma$ whereas when $\epsilon_{\rm PALM} < 10^{-3}$ the performance becomes quite stable with respect to $\gamma$. The same behaviors are observed for the NSNCP method which is shown in Fig. \ref{fig:nsncp_epsilon&gamma}(a)-(f).

From Fig. \ref{fig:sncp_epsilon&gamma}(h) and \ref{fig:sncp_epsilon&gamma}(k), with fixed $\gamma$, the SNCP with smaller values of $\epsilon_{\rm PALM}$ converge faster but it may be slowed for $\epsilon_{\rm PALM} < 10^{-4}$. However, from Fig. \ref{fig:sncp_epsilon&gamma}(g) and \ref{fig:sncp_epsilon&gamma}(j),  one can see that smaller values of $\epsilon_{\rm PALM}$ always leads to smaller objective value. Similar observations apply to the NSNCP as shown in Fig. \ref{fig:nsncp_epsilon&gamma}(g)-(l).

%

\begin{figure} [t!]
	\centering
	\subfigure[\scriptsize ]{
		\includegraphics[width=4cm]{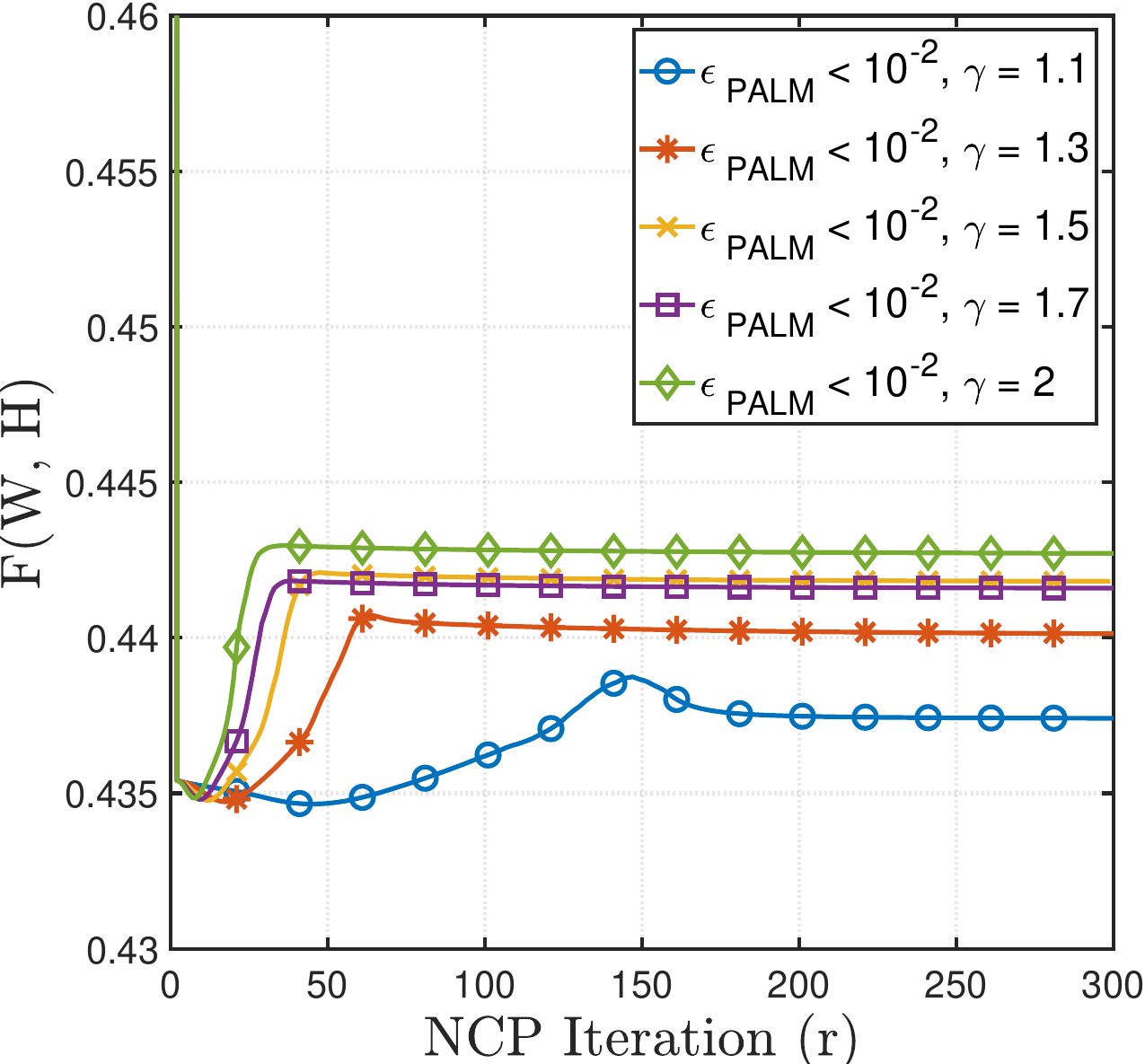}
		\label{fig:cost_epsilon1e_2}
	}
	\subfigure[\scriptsize ]{
		\includegraphics[width=4cm]{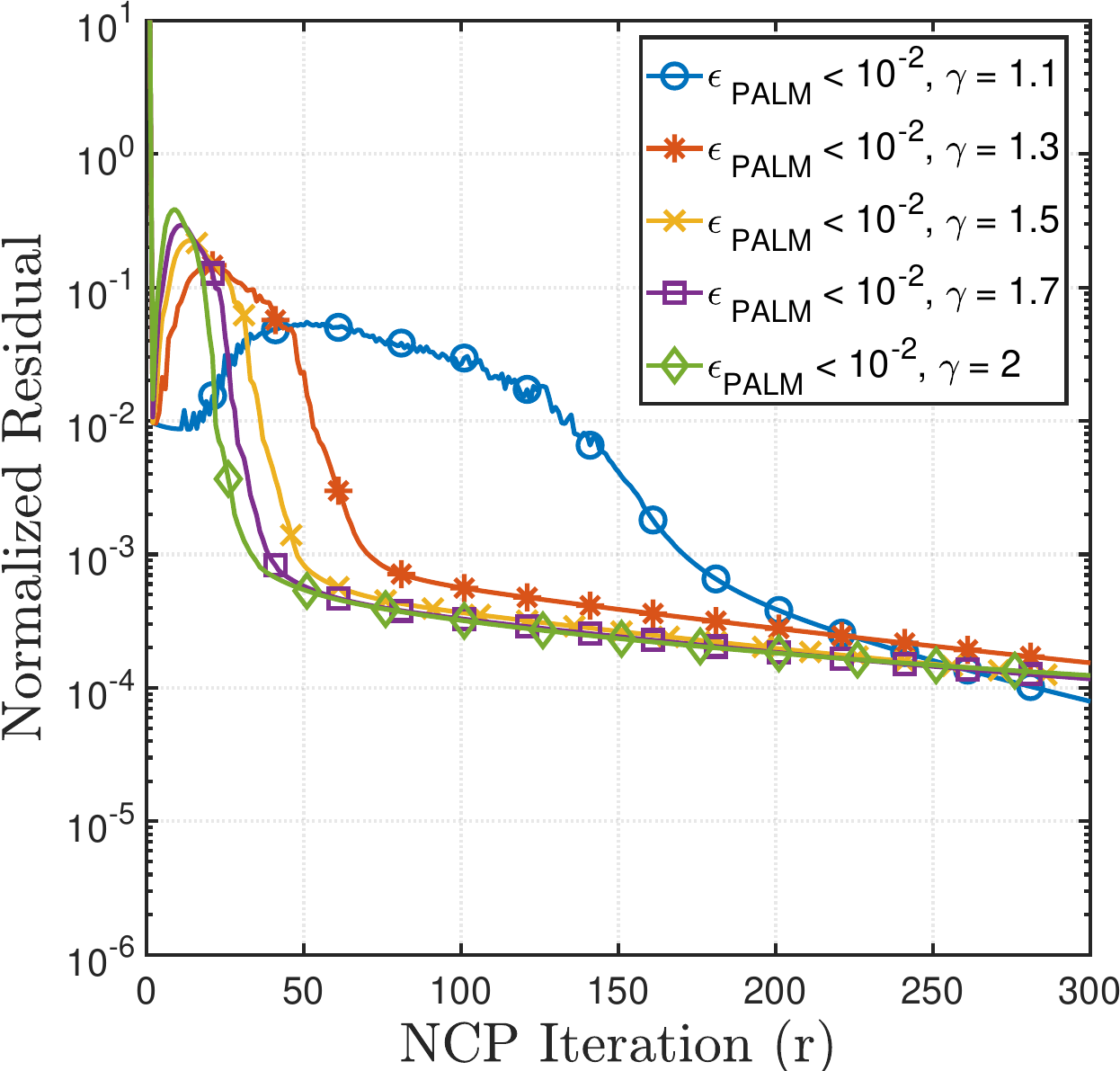}
		\label{fig:nr_epsilon1e_2}
	}
	\subfigure[\scriptsize ]{
		\includegraphics[width=4cm]{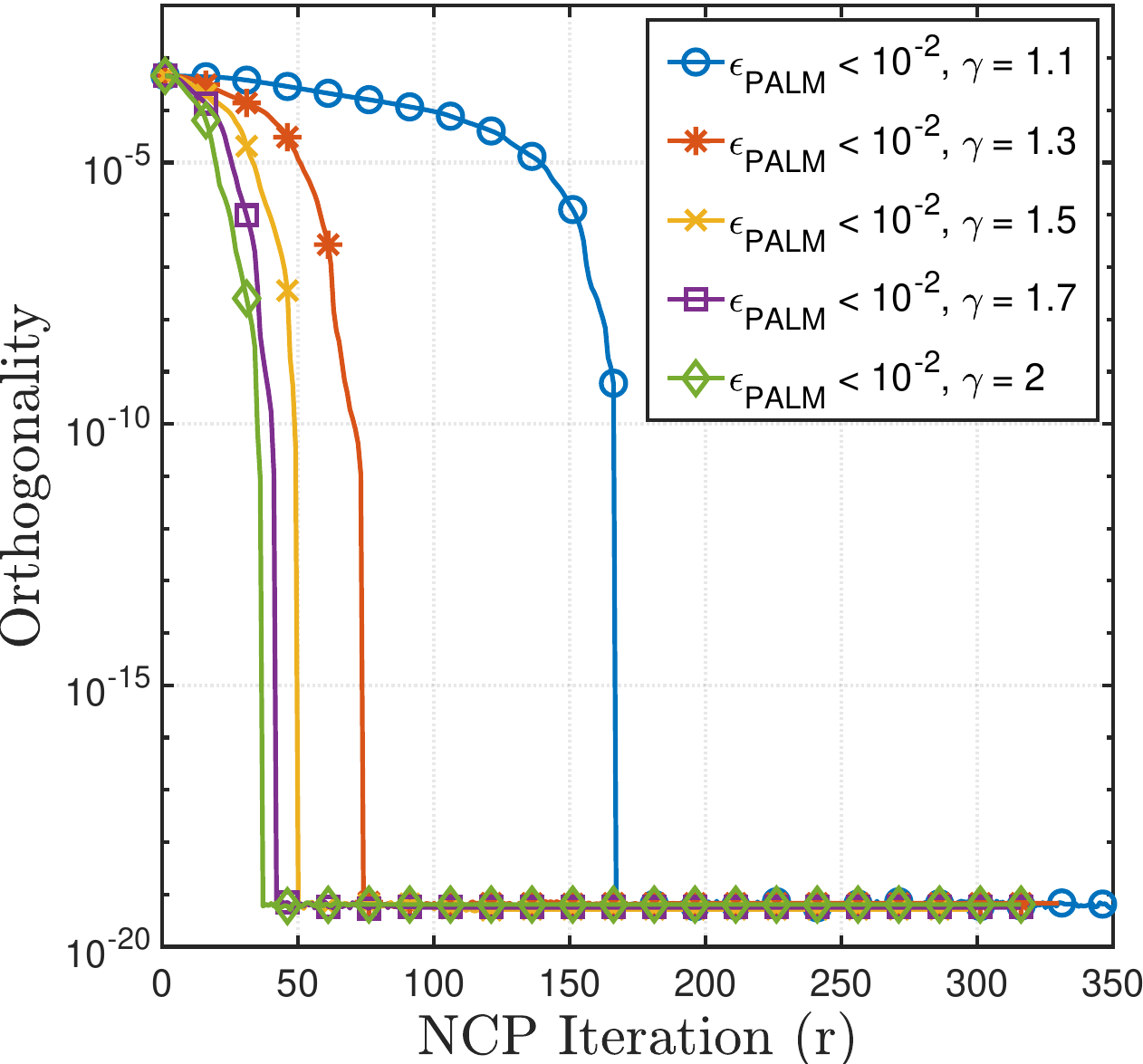}
		\label{fig:ortho_epsilon1e_2}
	}
	\subfigure[\scriptsize ]{
		\includegraphics[width=4cm]{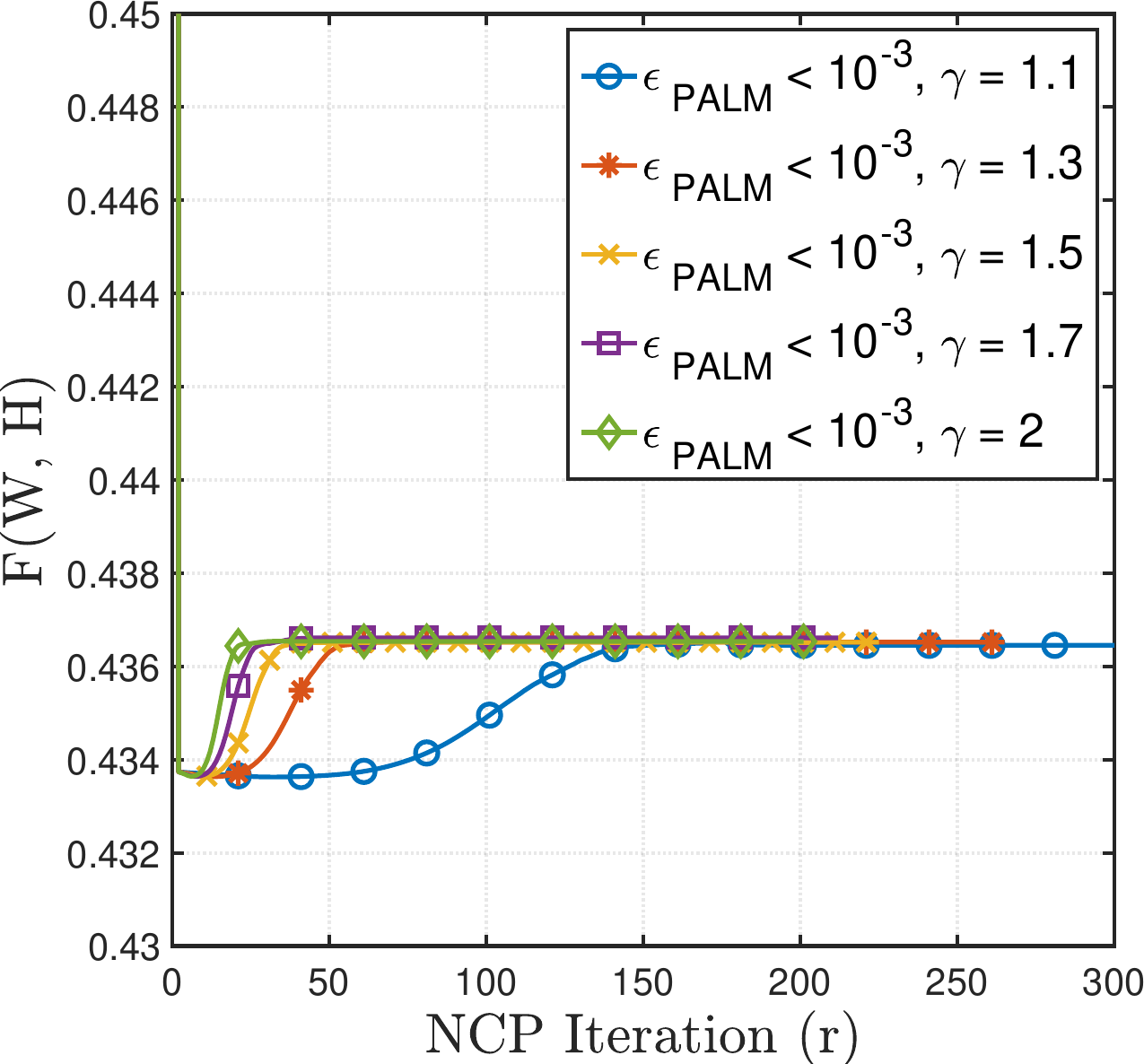}
		\label{fig:cost_epsilon1e_3}
	}
	\subfigure[\scriptsize ]{
		\includegraphics[width=4cm]{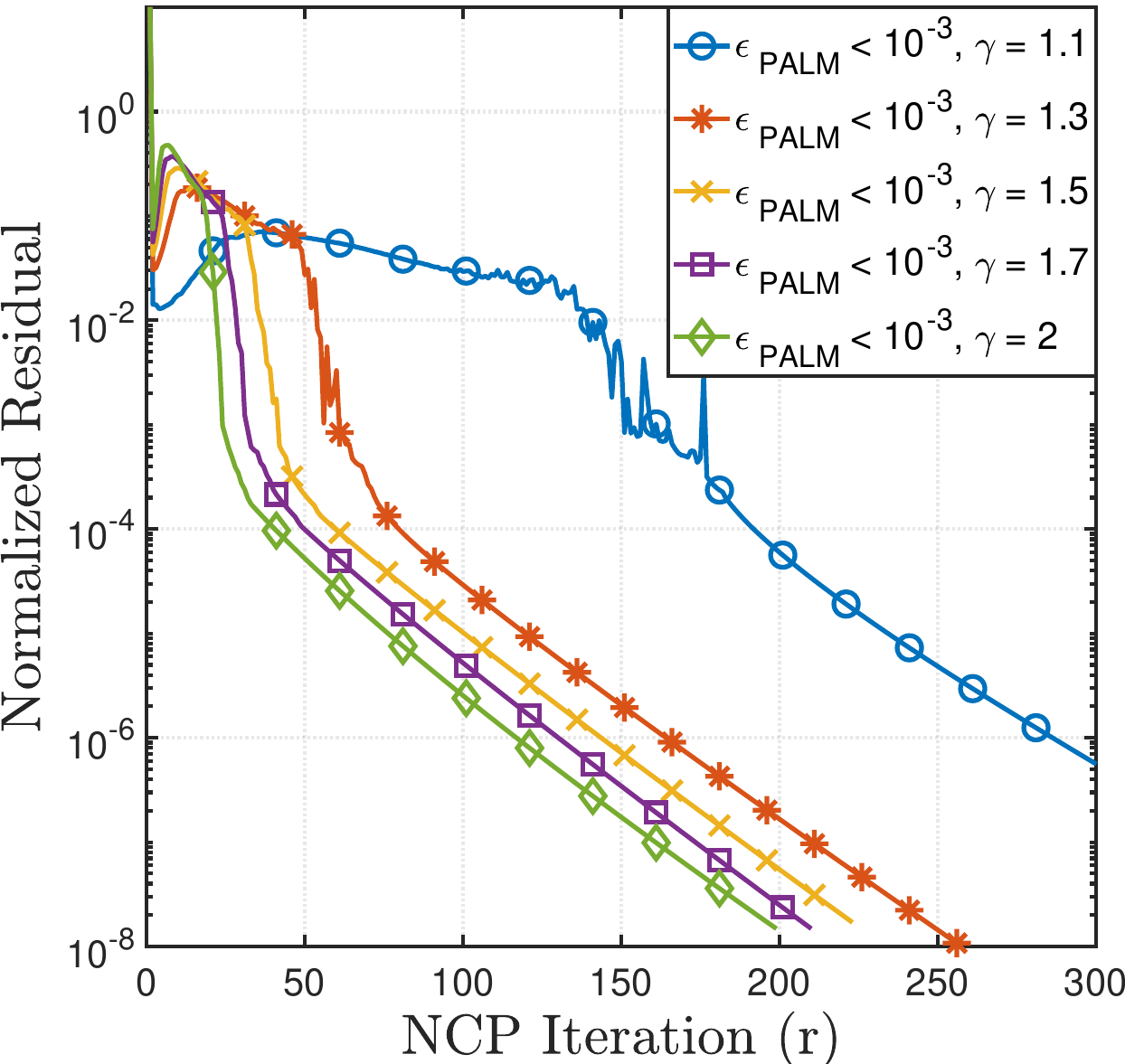}
		\label{fig:nr_epsilon1e_3}
	}
	\subfigure[\scriptsize ]{
		\includegraphics[width=4cm]{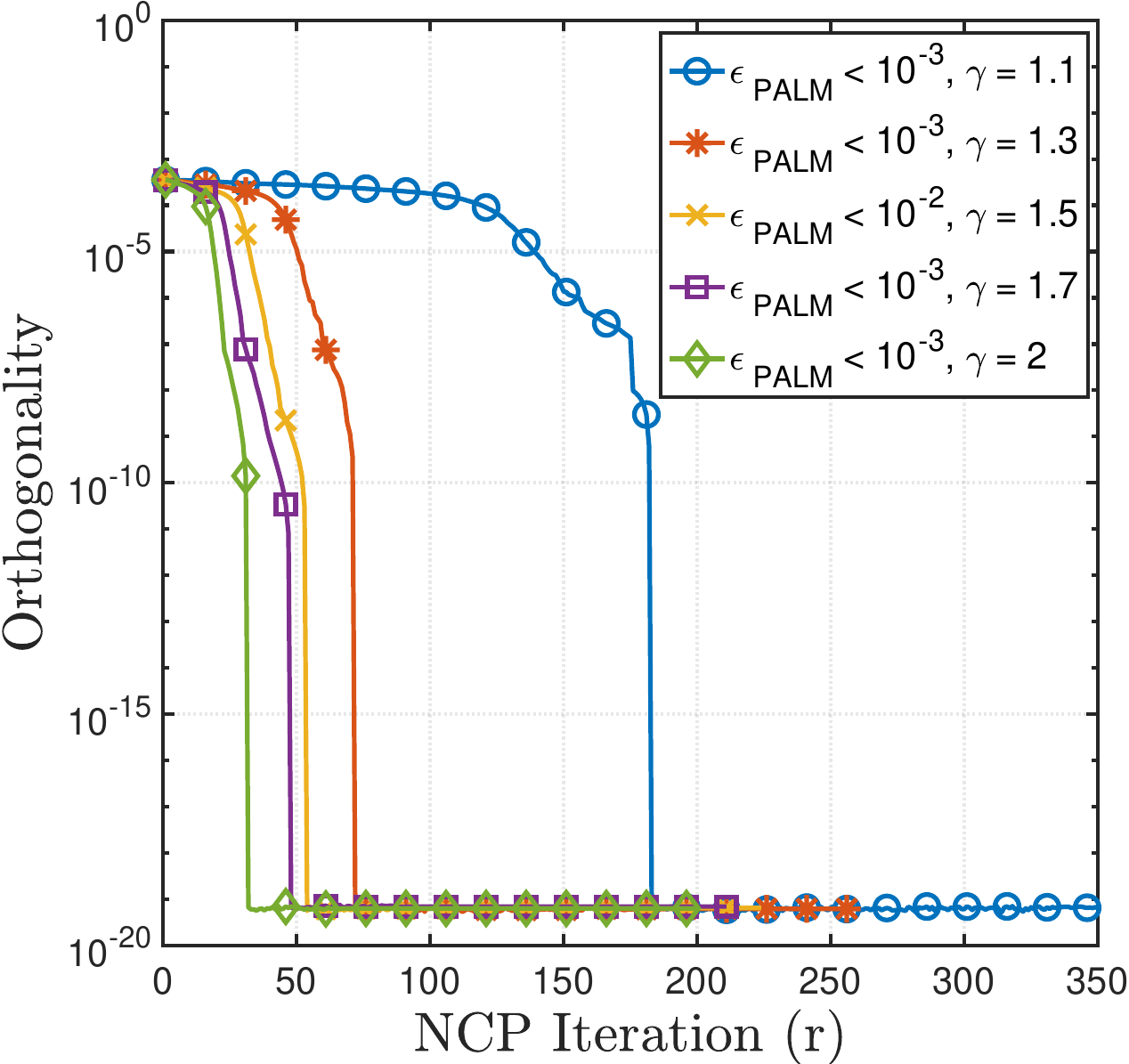}
		\label{fig:ortho_epsilon1e_3}
	}
	\subfigure[\scriptsize ]{
		\includegraphics[width=4cm]{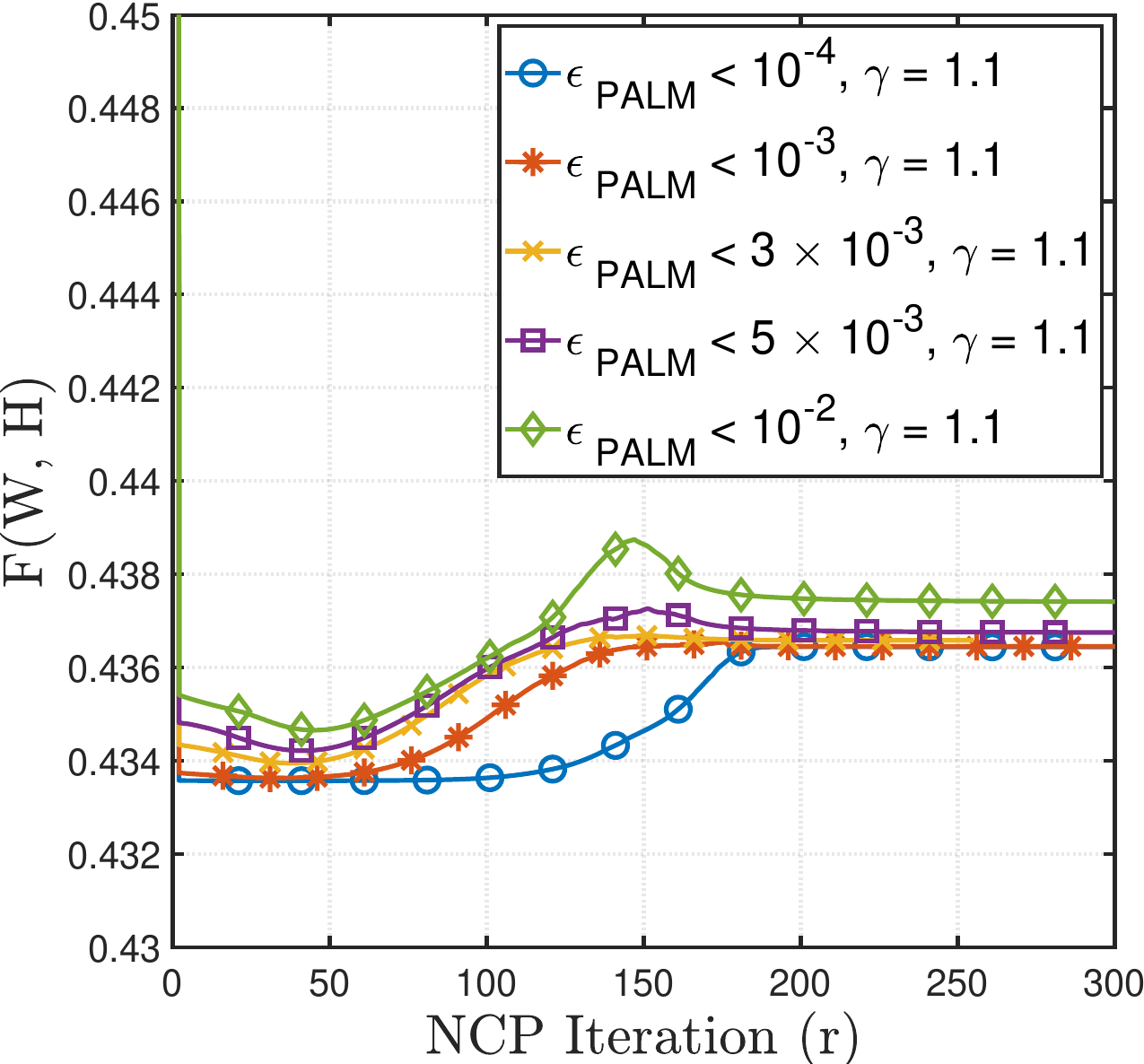}
		\label{fig:cost_gamma1_1}
	}
	\subfigure[\scriptsize ]{
		\includegraphics[width=4cm]{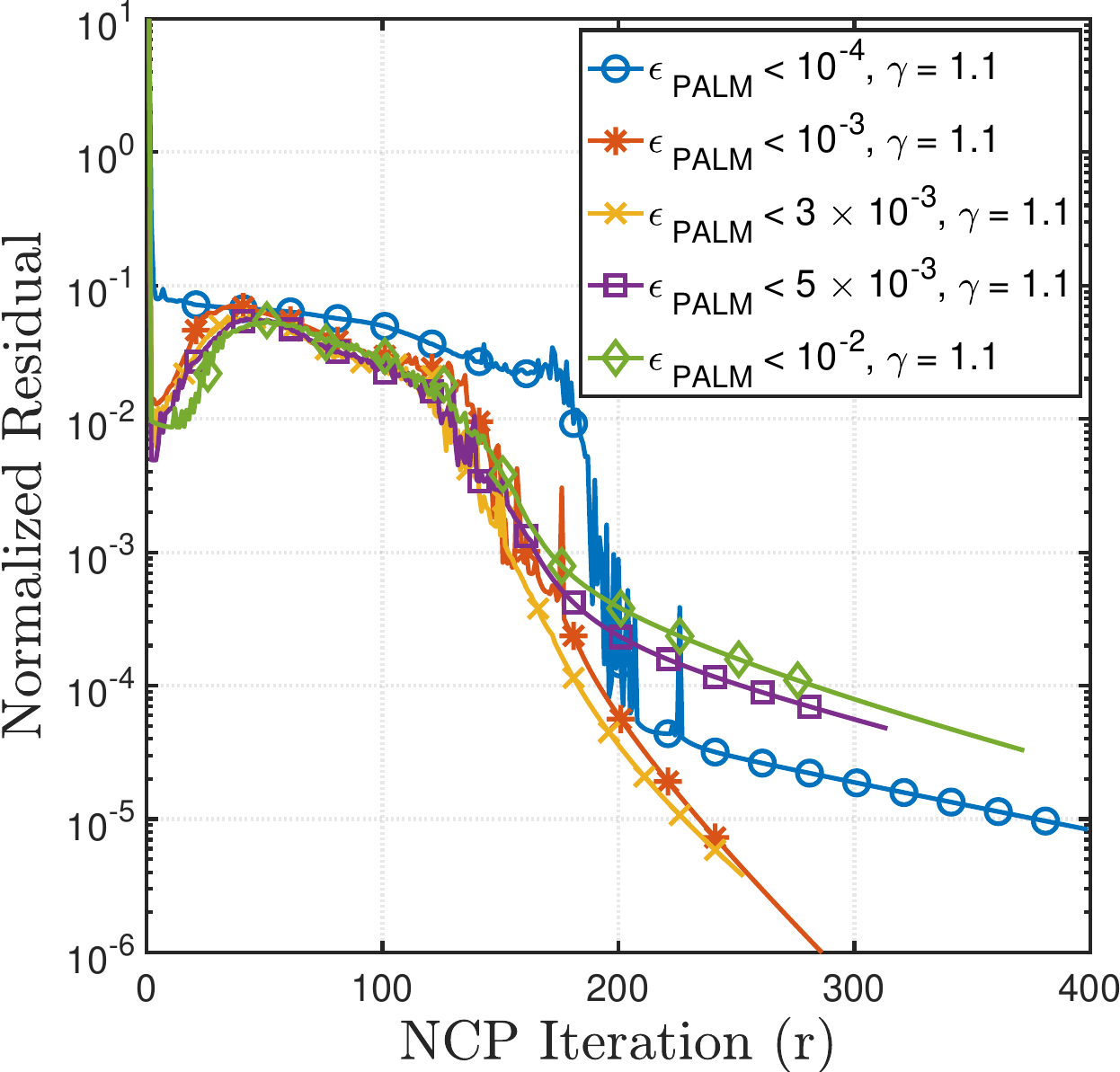}
		\label{fig:nr_gamma1_1}
	}
	\subfigure[\scriptsize ]{
		\includegraphics[width=4cm]{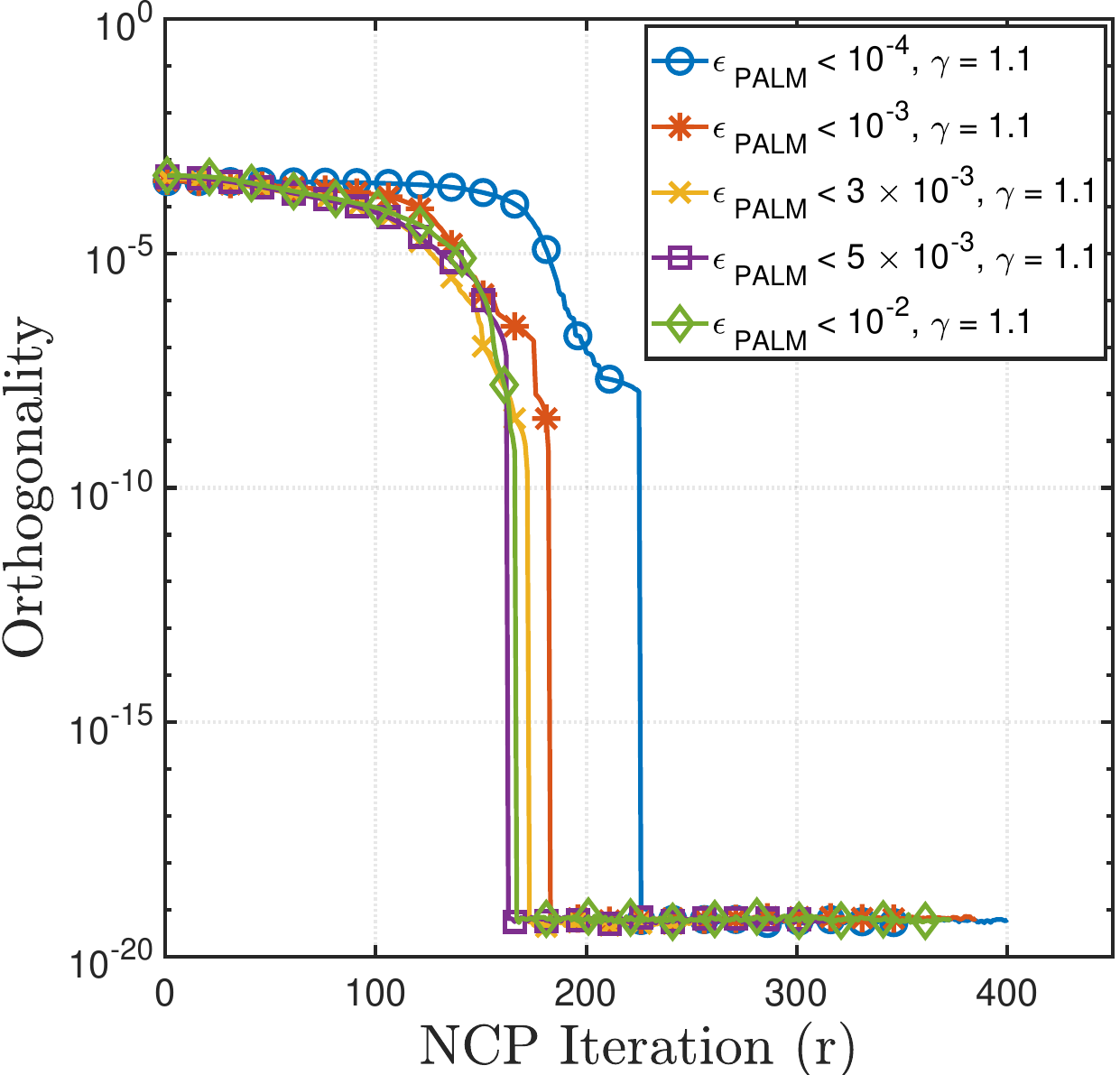}
		\label{fig:ortho_gamma1_1}
	}
	\subfigure[\scriptsize ]{
		\includegraphics[width=4cm]{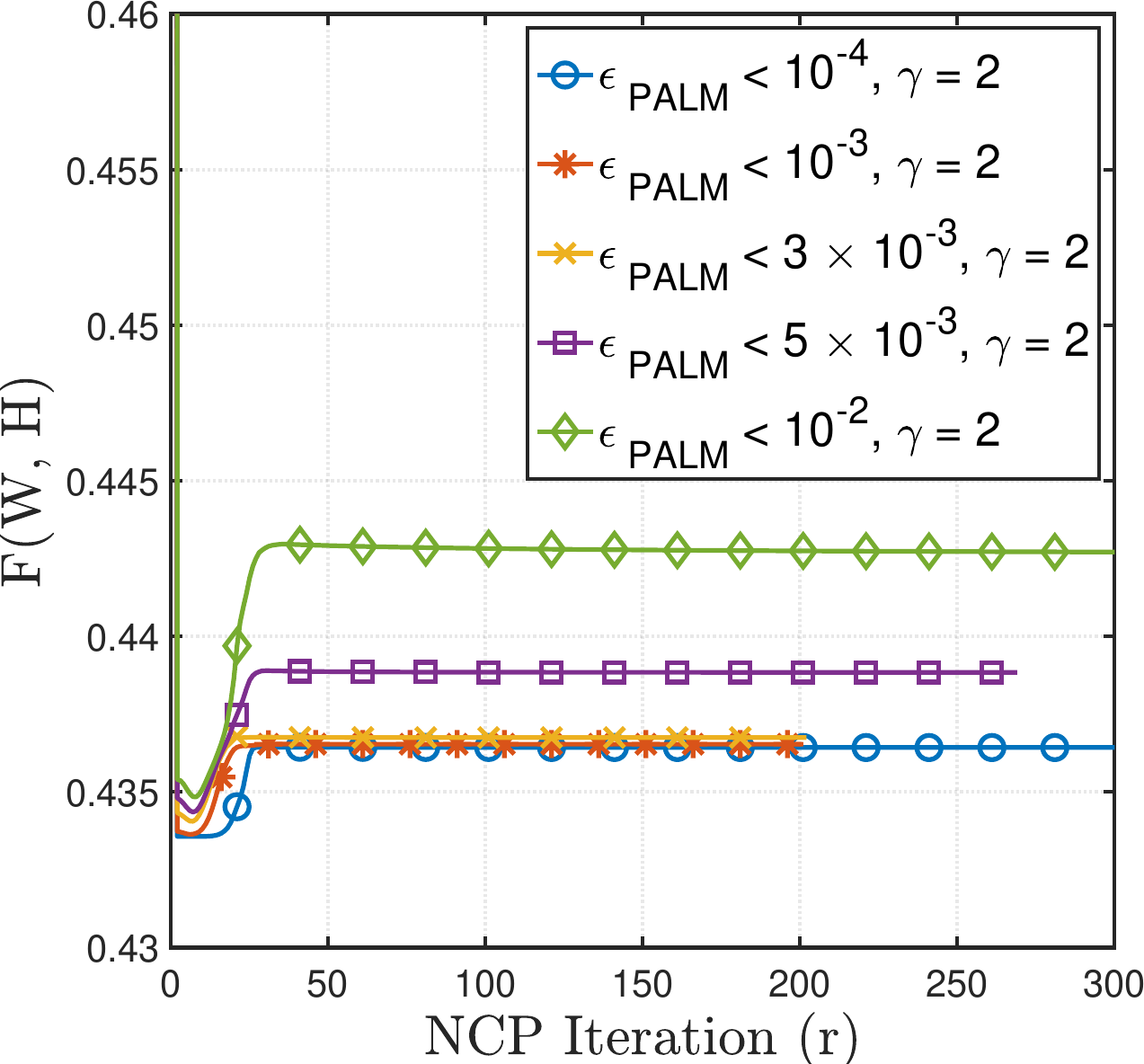}
		\label{fig:cost_gamma2}
	}
	\subfigure[\scriptsize ]{
		\includegraphics[width=4cm]{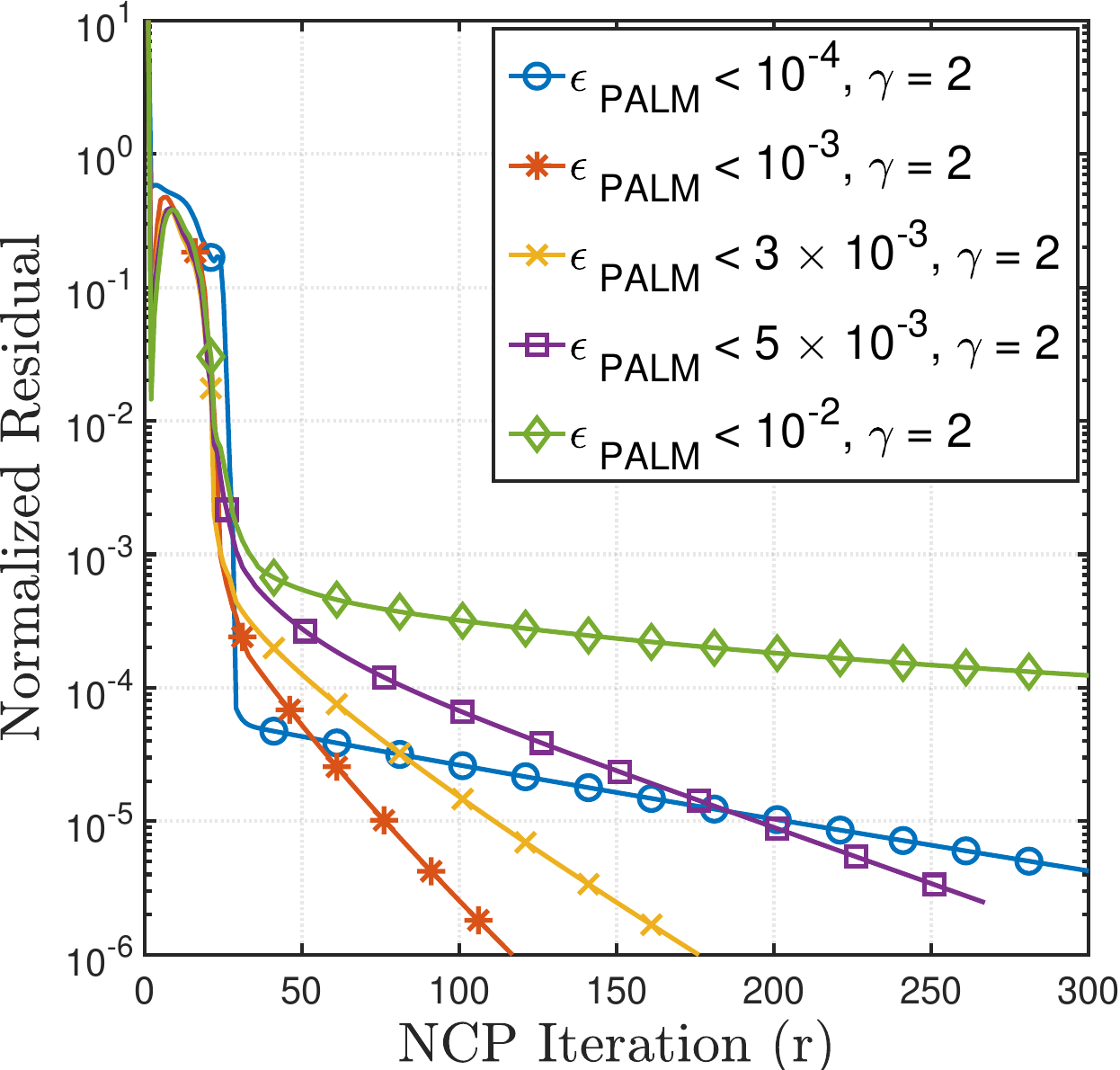}
		\label{fig:nr_gamma2}
	}
	\subfigure[\scriptsize ]{
		\includegraphics[width=4cm]{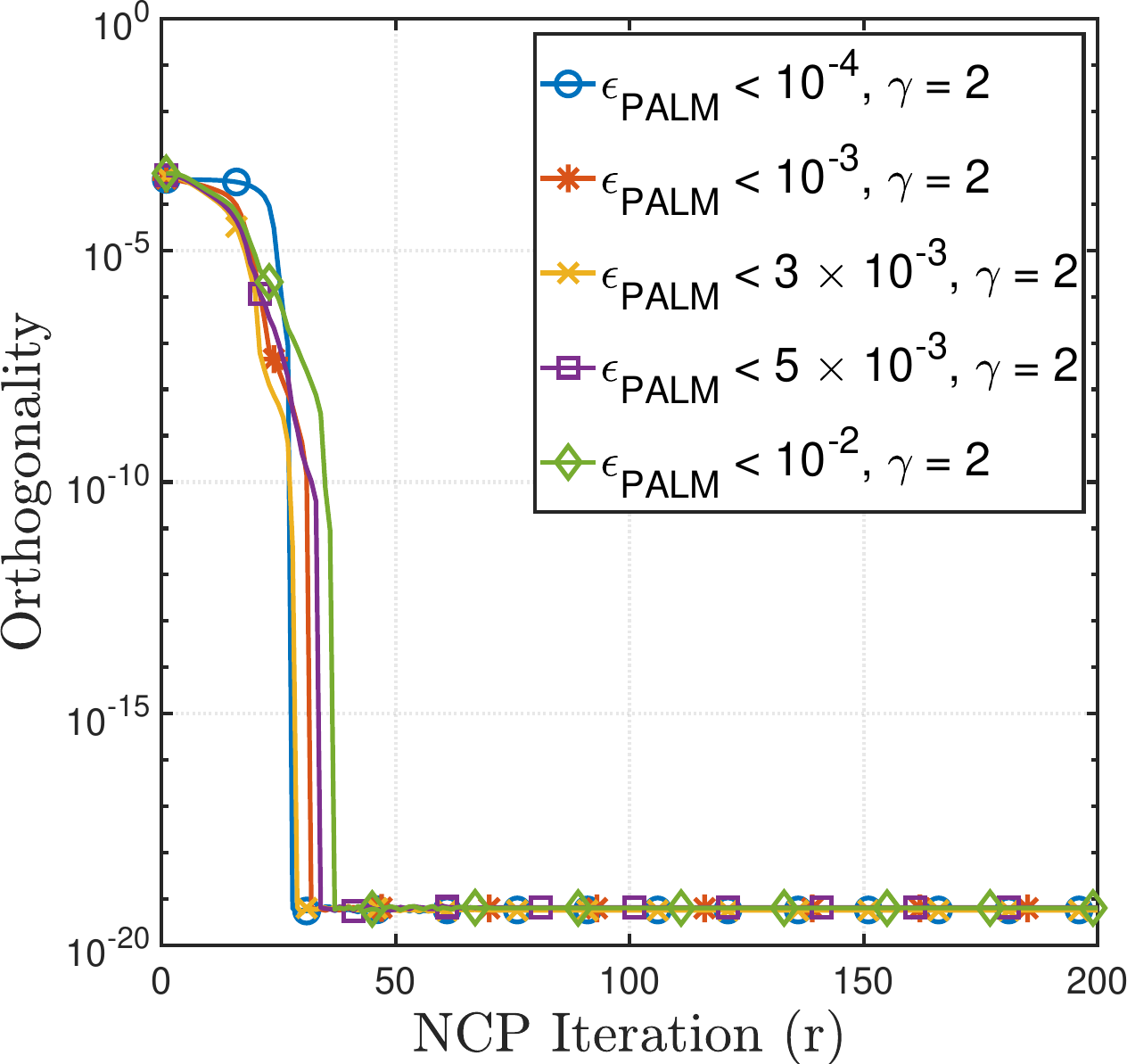}
		\label{fig:gamma2}
	}
	\caption{Convergence curves of SNCP on under different values of $\epsilon_{\rm PALM}$ and $\gamma$.}
	\vspace{-0.5cm}
	\label{fig:sncp_epsilon&gamma}
\end{figure}

\begin{figure} [t!]
	\centering
	\subfigure[\scriptsize ]{
		\includegraphics[width=4cm]{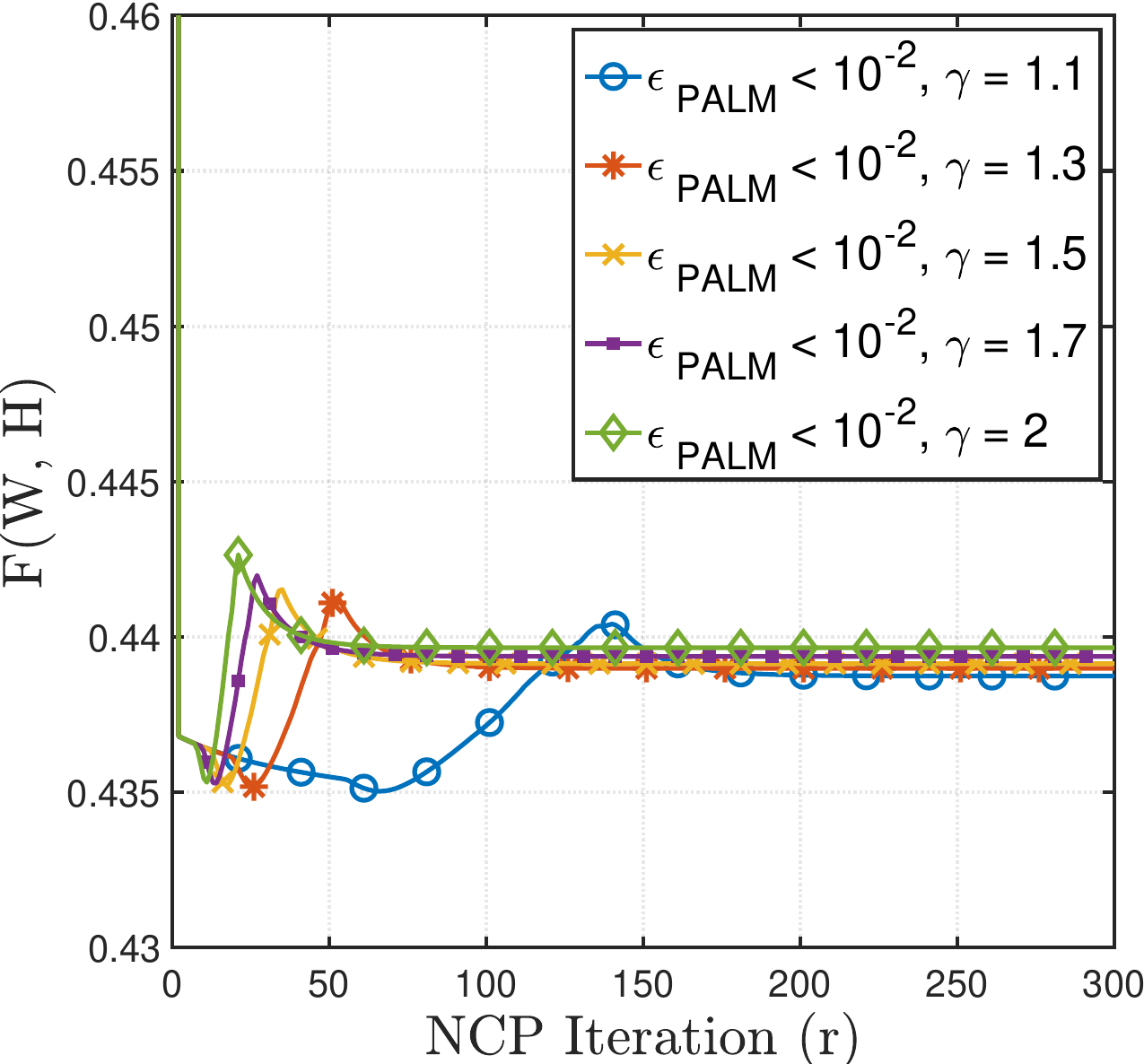}
		\label{fig:nsncp_cost_epsilon1e_2}
	}
	\subfigure[\scriptsize ]{
		\includegraphics[width=4cm]{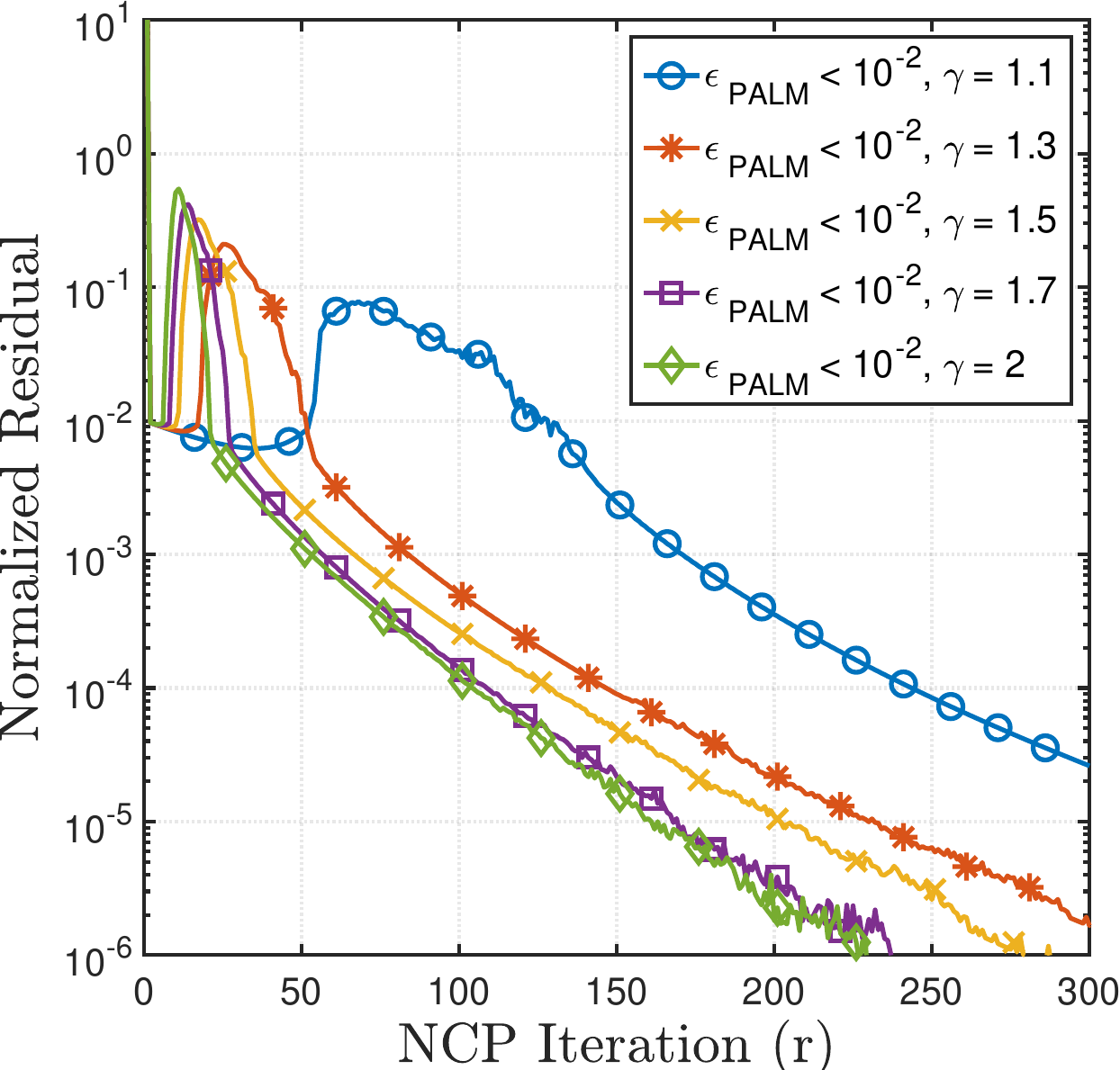}
		\label{fig:nsncp_nr_epsilon1e_2}
	}
	\subfigure[\scriptsize ]{
		\includegraphics[width=4cm]{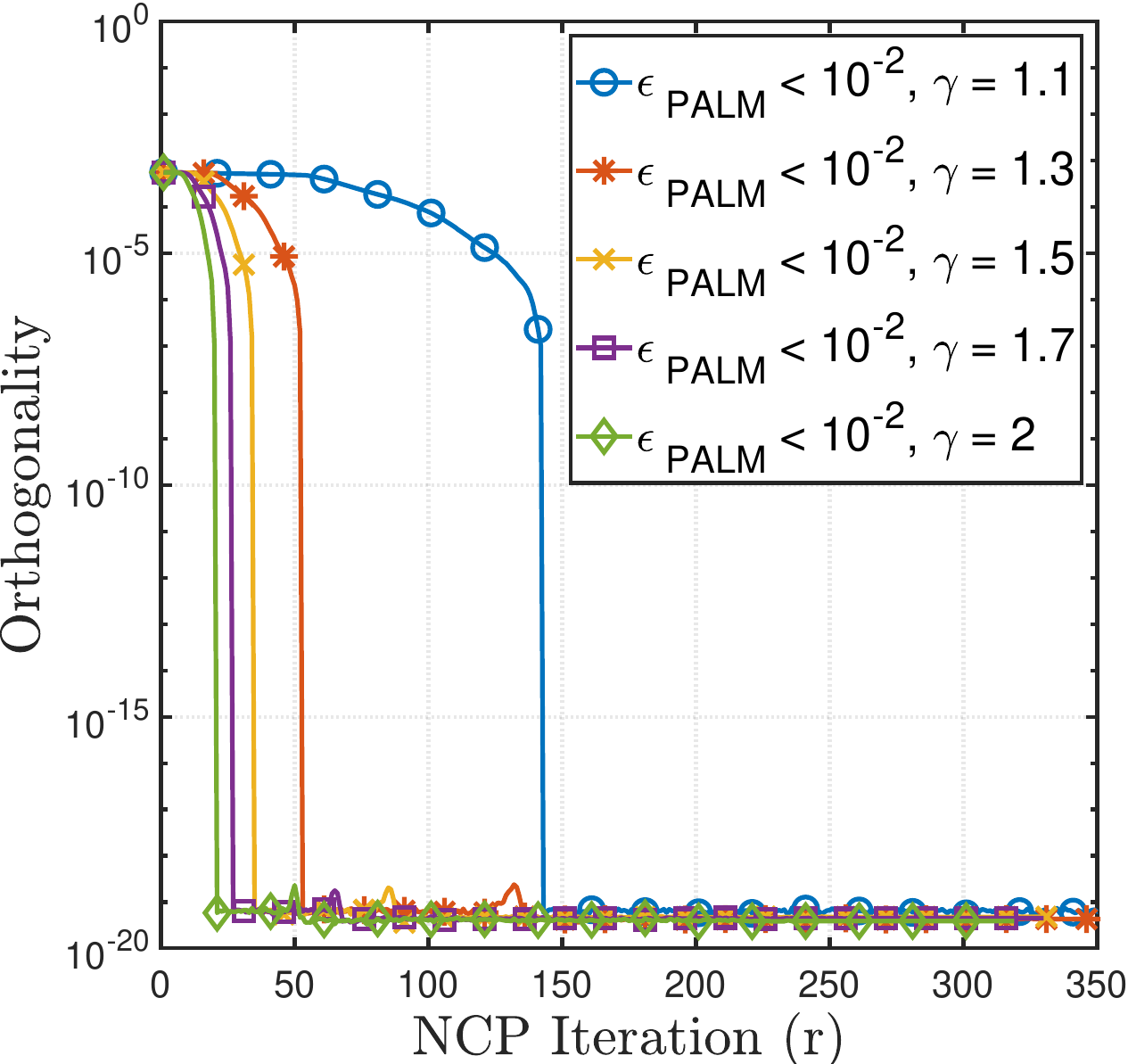}
		\label{fig:nsncp_ortho_epsilon1e_2}
	}
	\subfigure[\scriptsize ]{
		\includegraphics[width=4cm]{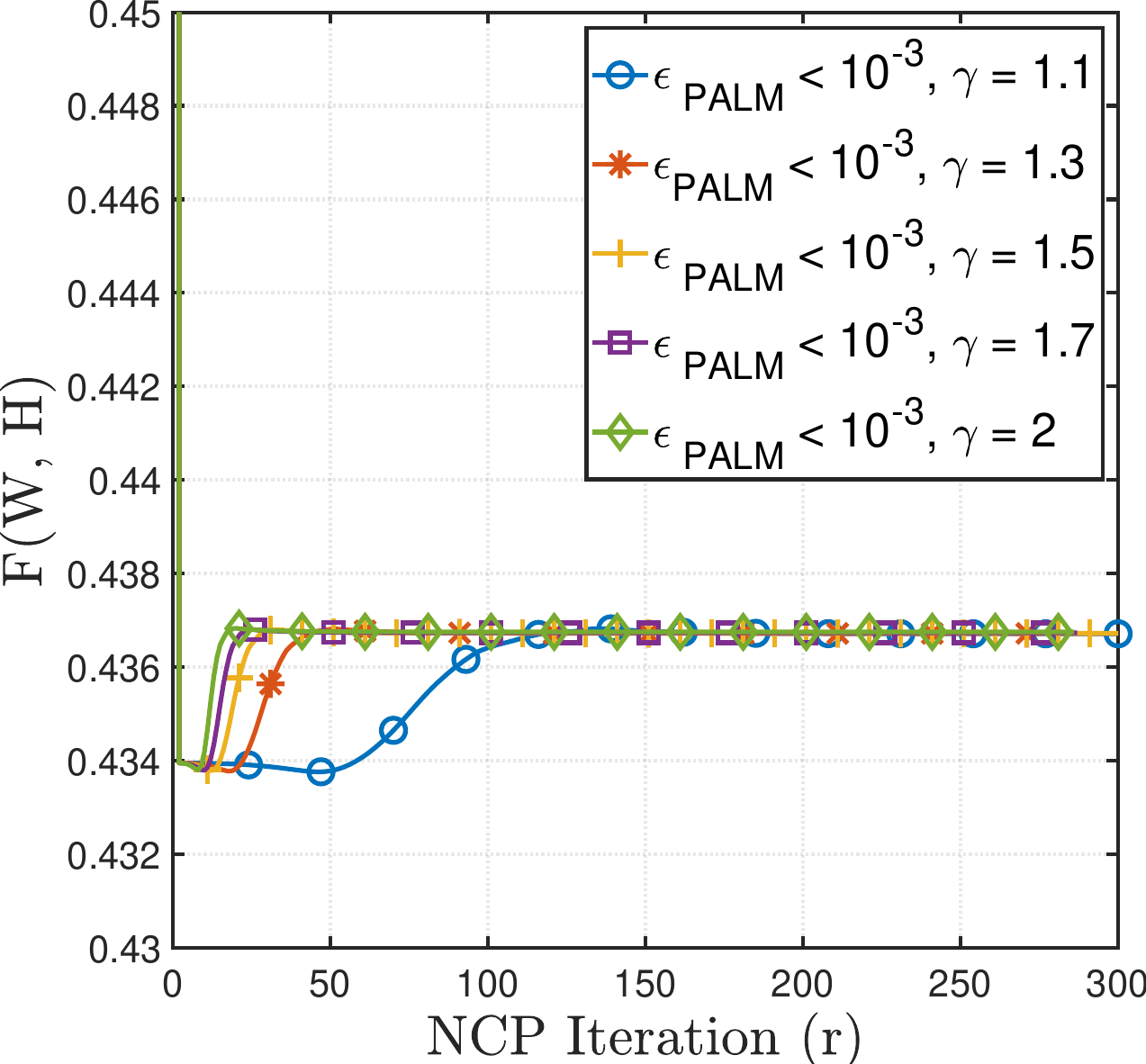}
		\label{fig:nsncp_cost_epsilon1e_3}
	}
	\subfigure[\scriptsize ]{
		\includegraphics[width=4cm]{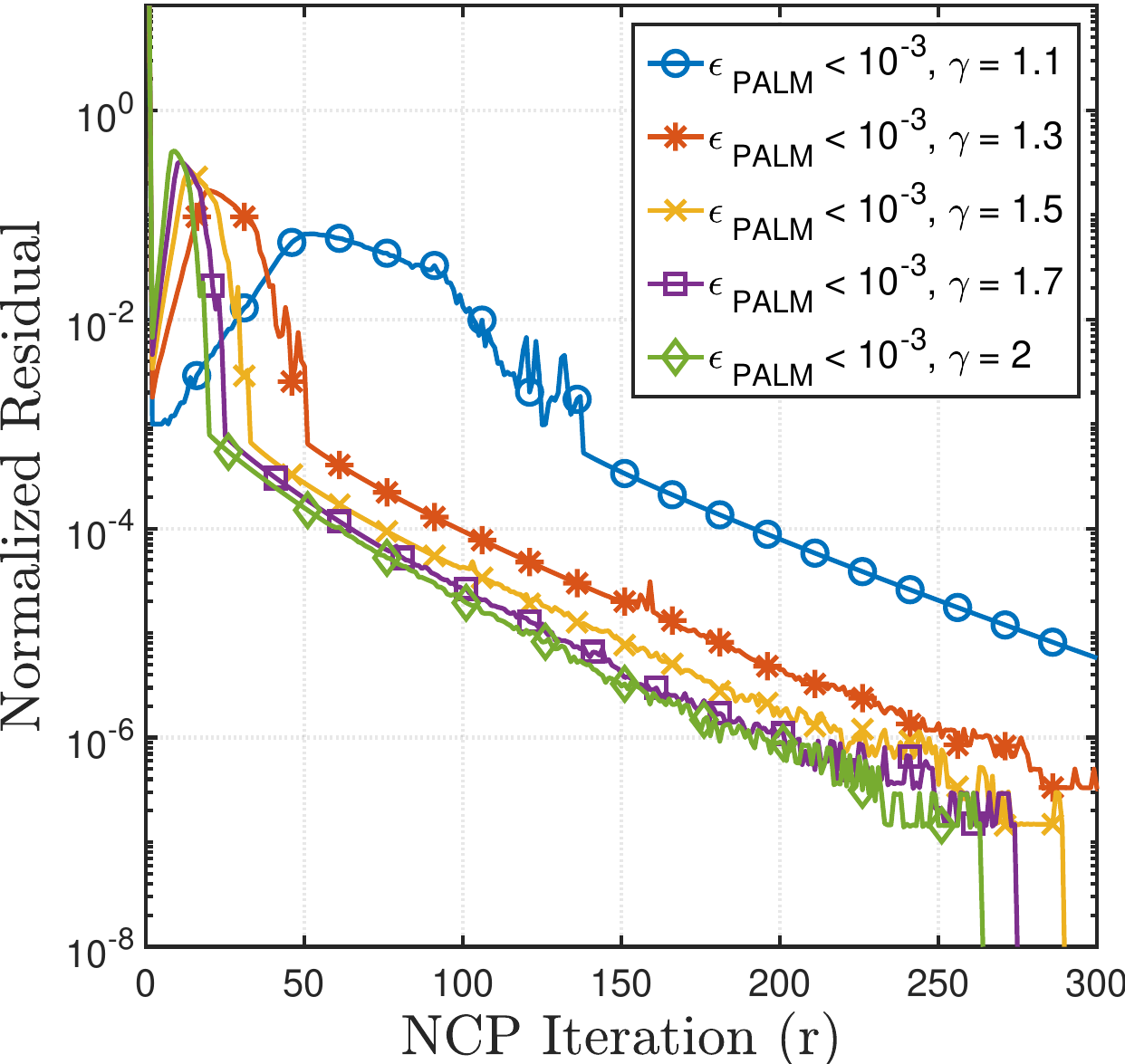}
		\label{fig:nsncp_nr_epsilon1e_3}
	}
	\subfigure[\scriptsize ]{
		\includegraphics[width=4cm]{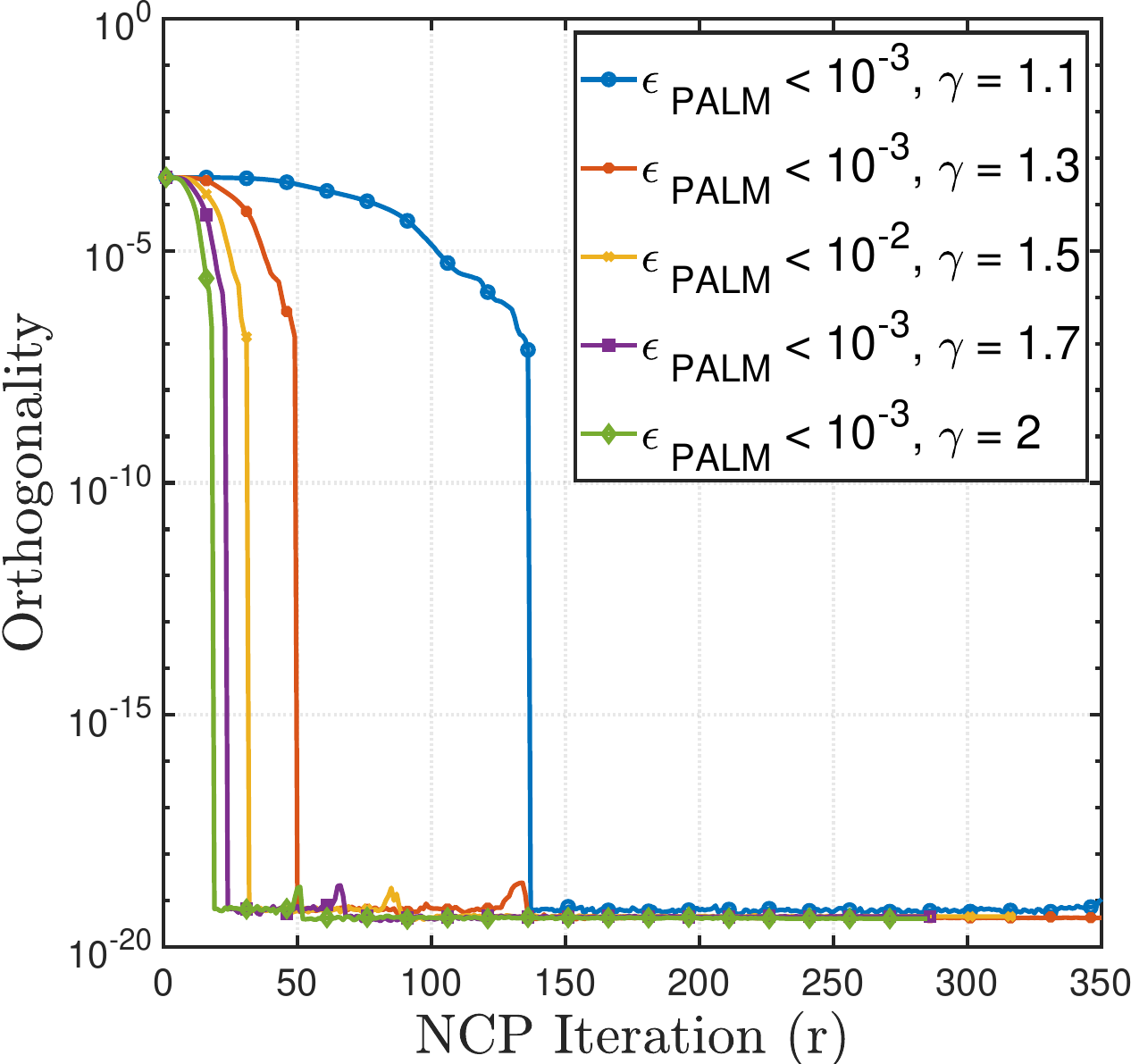}
		\label{fig:nsncp_ortho_epsilon1e_3}
	}
	\subfigure[\scriptsize ]{
		\includegraphics[width=4cm]{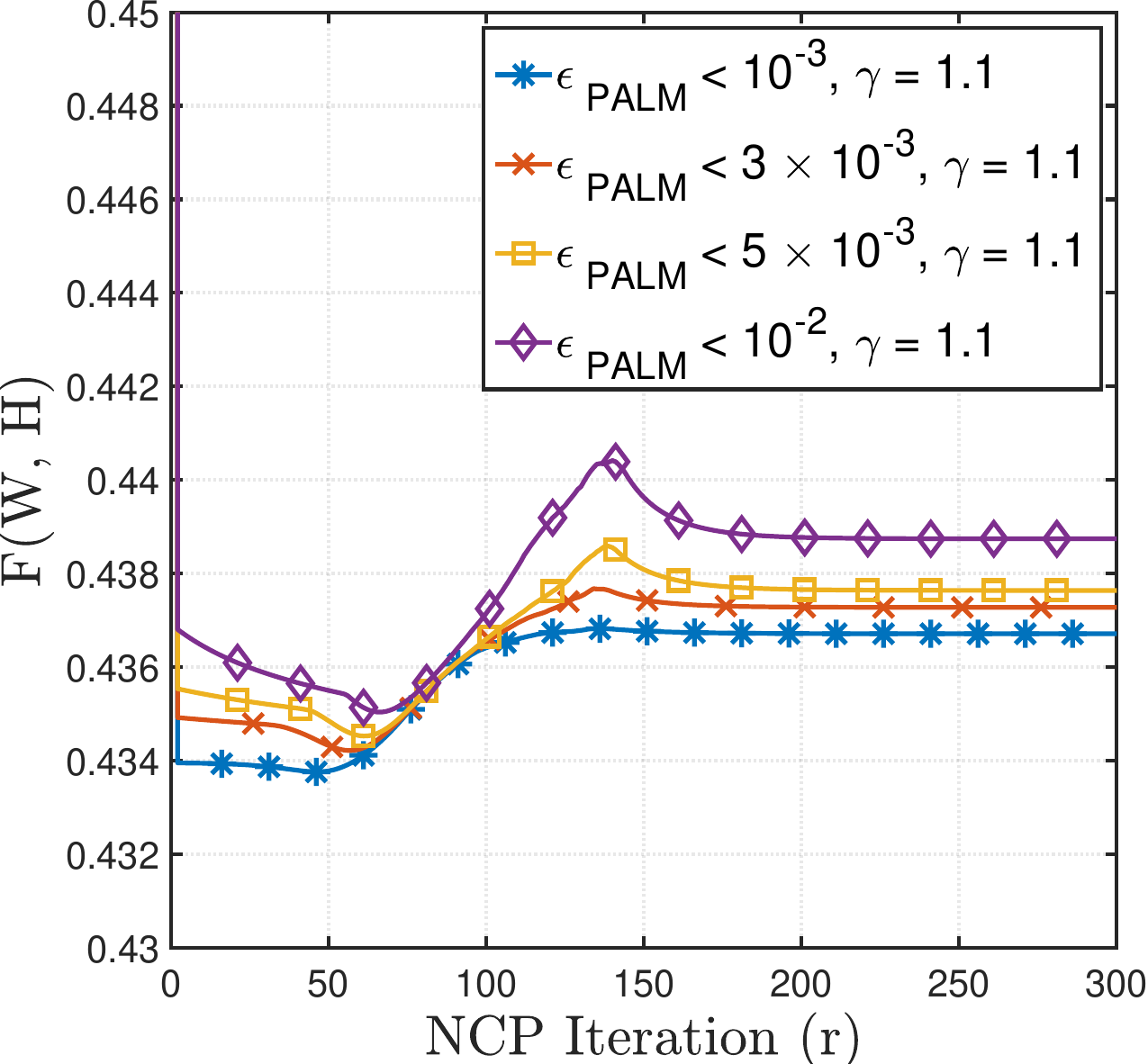}
		\label{fig:nsncp_cost_gamma1_1}
	}
	\subfigure[\scriptsize ]{
		\includegraphics[width=4cm]{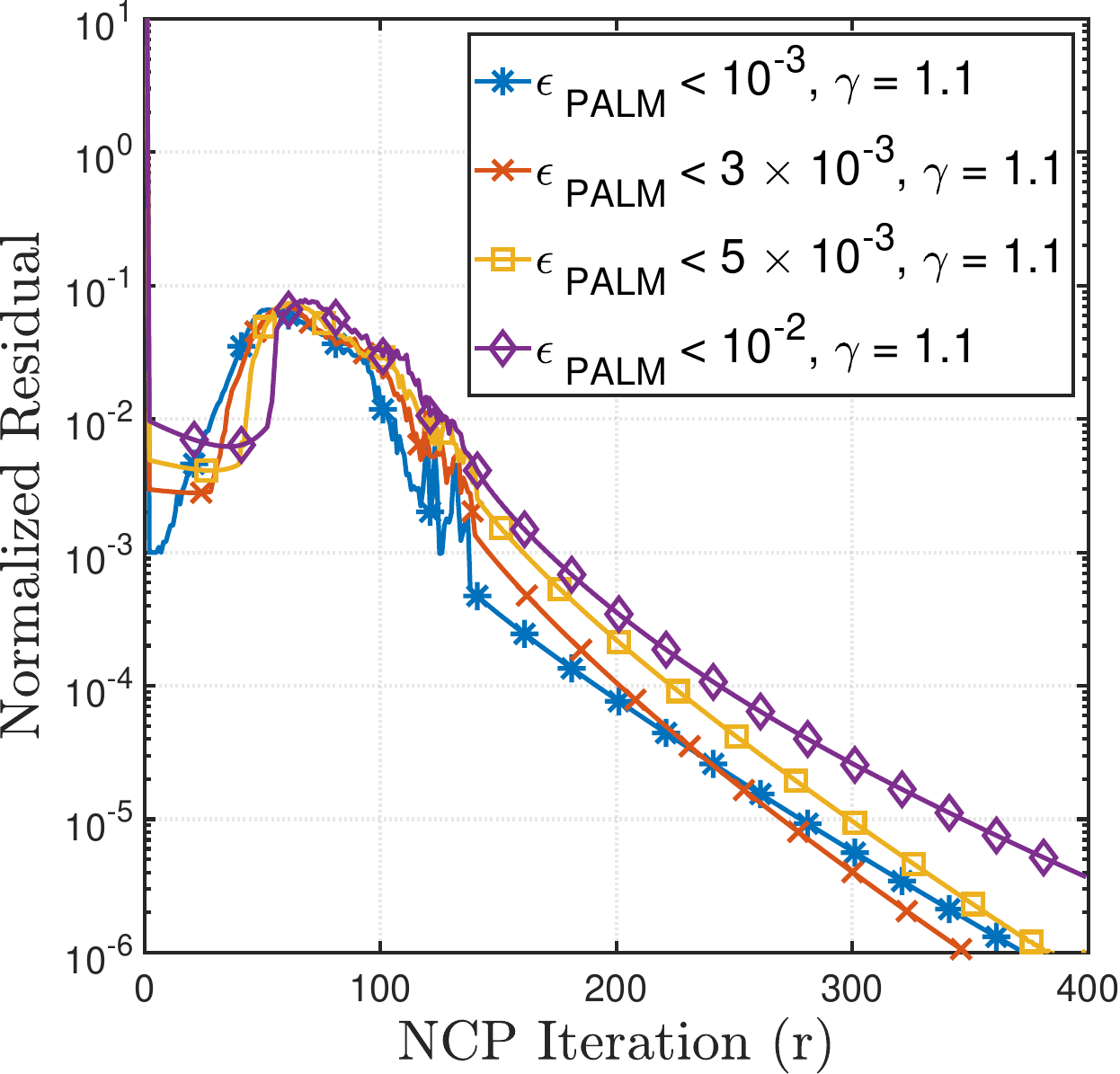}
		\label{fig:nsncp_nr_gamma1_1}
	}
	\subfigure[\scriptsize ]{
		\includegraphics[width=4cm]{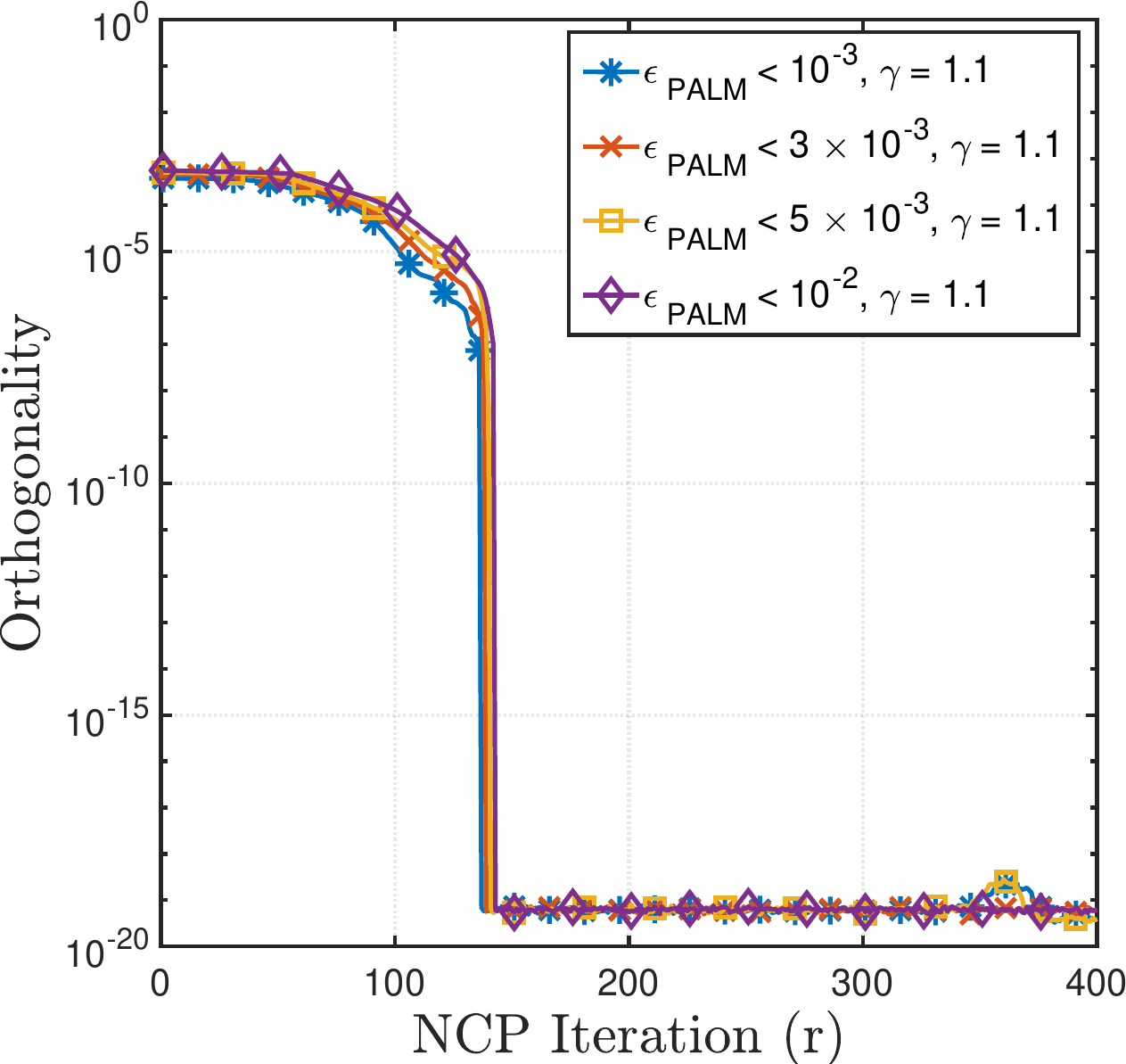}
		\label{fig:nsncp_ortho_gamma1_1}
	}
	\subfigure[\scriptsize ]{
		\includegraphics[width=4cm]{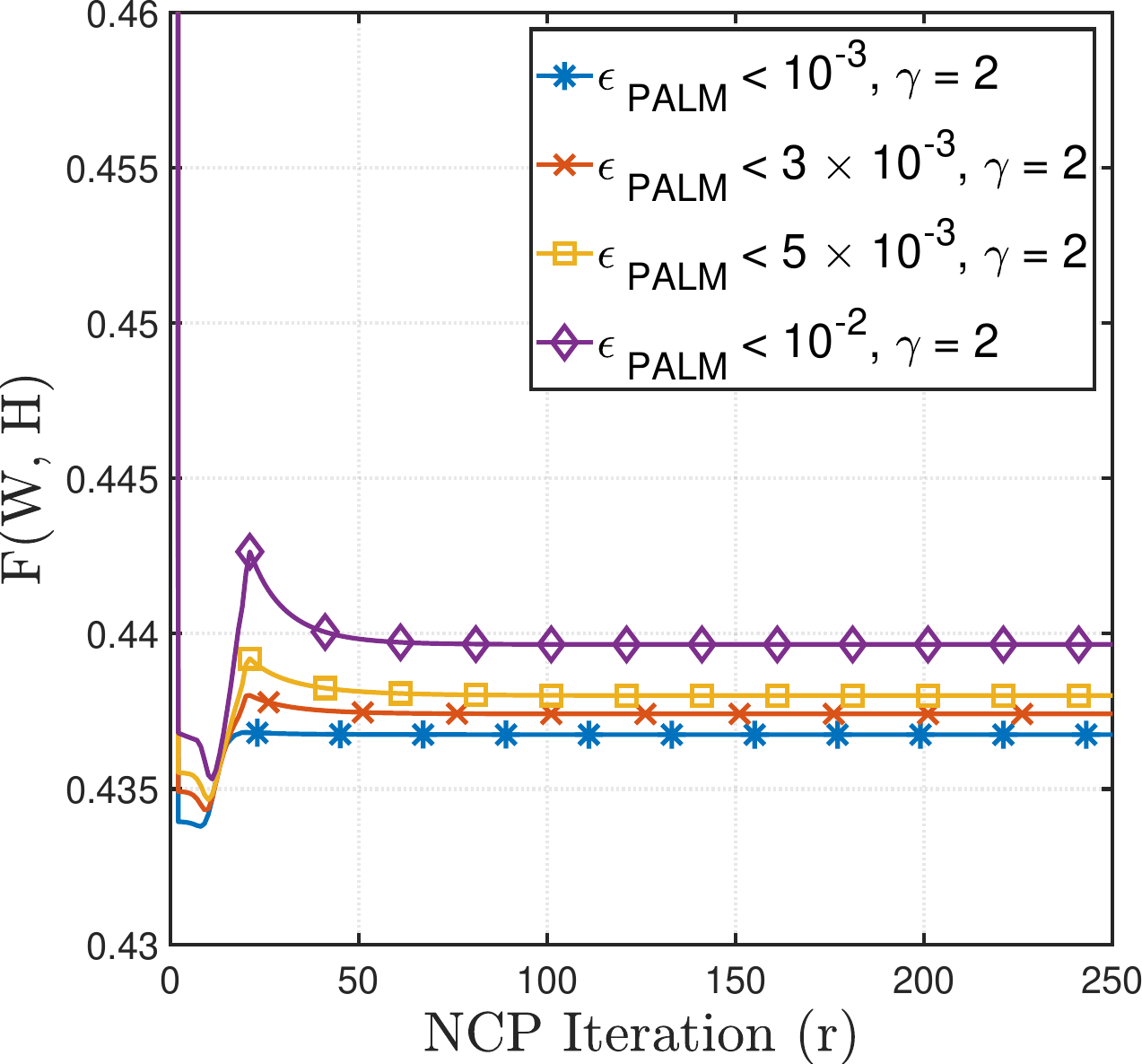}
		\label{fig:nsncp_cost_gamma2}
	}
	\subfigure[\scriptsize ]{
		\includegraphics[width=4cm]{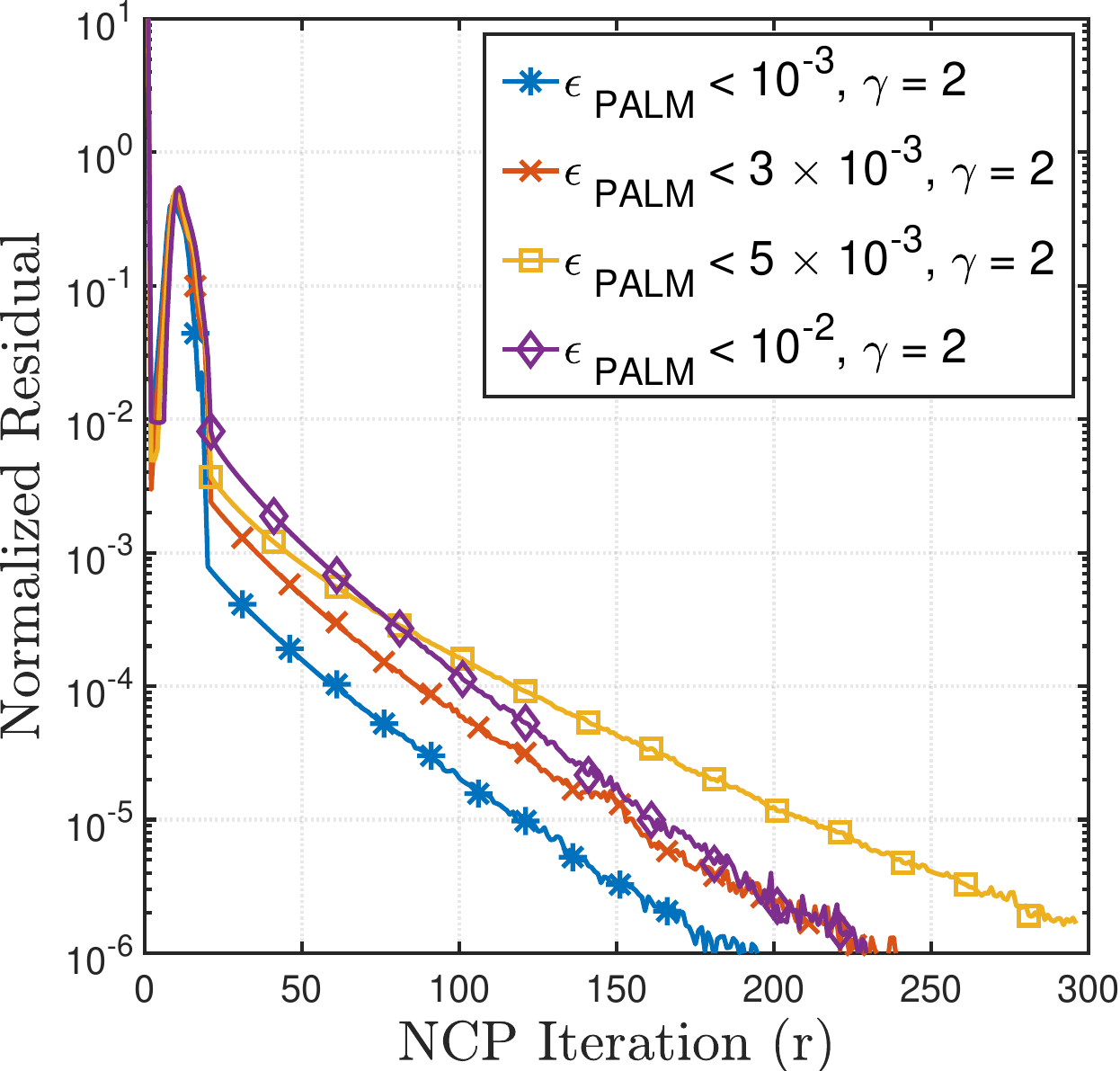}
		\label{fig:nsncp_nr_gamma2}
	}
	\subfigure[\scriptsize ]{
		\includegraphics[width=4cm]{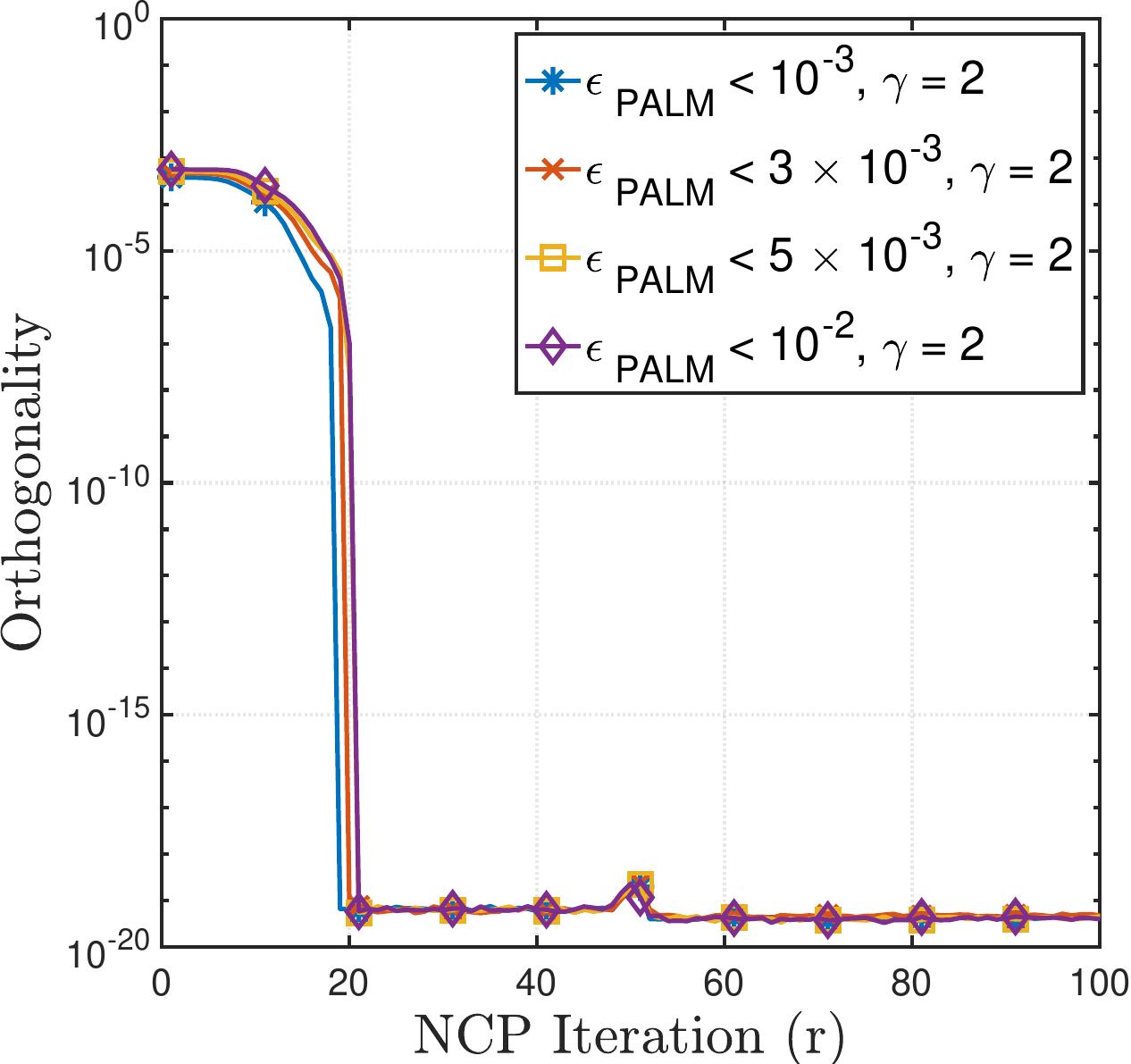}
		\label{fig:nsncp_gamma2}
	}\vspace{-0.3cm}
	\caption{Convergence curves of NSNCP on under different values of $\epsilon_{\rm PALM}$ and $\gamma$.}
	\vspace{-0.5cm}
	\label{fig:nsncp_epsilon&gamma}
\end{figure}

\section{Other choices of $(p, q, v)$ and their effects}

\subsection{The landscape of the penalized subproblem (15) with $(p, q, v)$}

We draw the contour of $\phi(\hb)$ with $\hb \in \Rbb^{2 \times 1}$, with different combinations of $(p, q, v)$ in Fig. \ref{fig:pqv_contour}. One can observe that if $(p, q, v)$ are bigger, there exits a larger area where the objective function (of problem (10) in the manuscript) will be of a small value. 

	\begin{figure} [H]
	\centering
	\subfigure[\scriptsize $p = 1, q = 2, v = 2$]{
		\includegraphics[width=4.5cm]{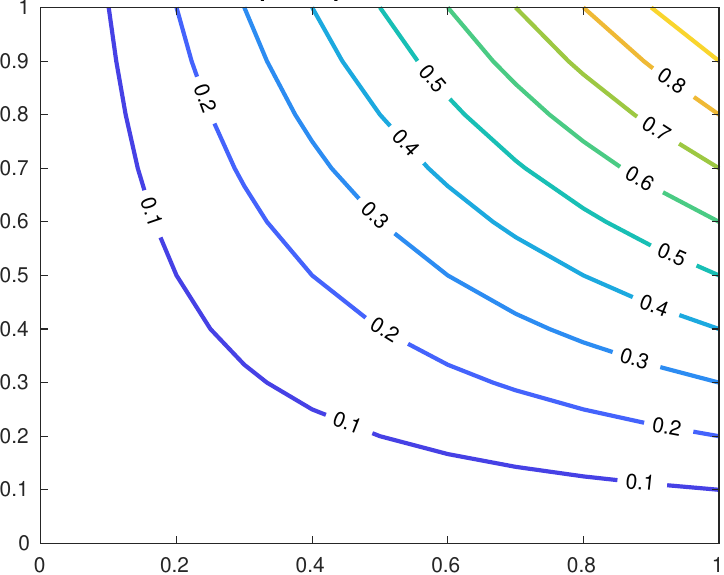}
		\label{fig:p1q2v2}
	}
	\subfigure[\scriptsize $p = 1, q = 2, v = 10$]{
		\includegraphics[width=4.5cm]{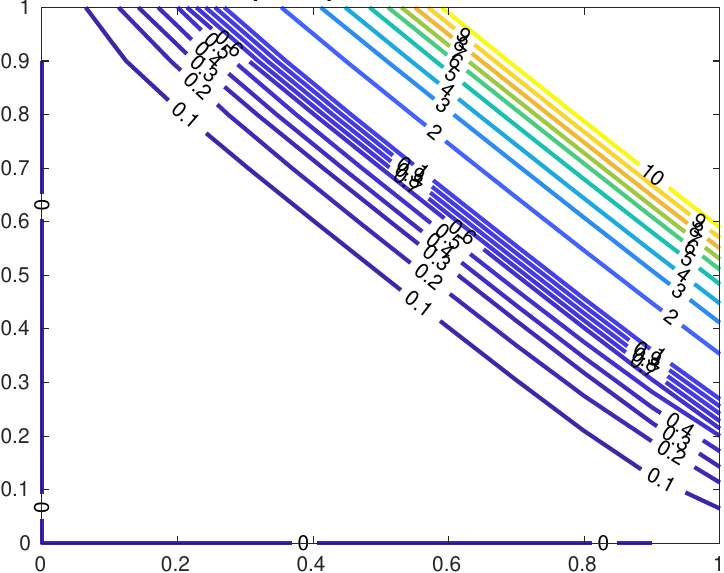}
		\label{fig:p1q2v10}
	}
	\subfigure[\scriptsize $p = 1, q = 3, v = 3$]{
		\includegraphics[width=4.5cm]{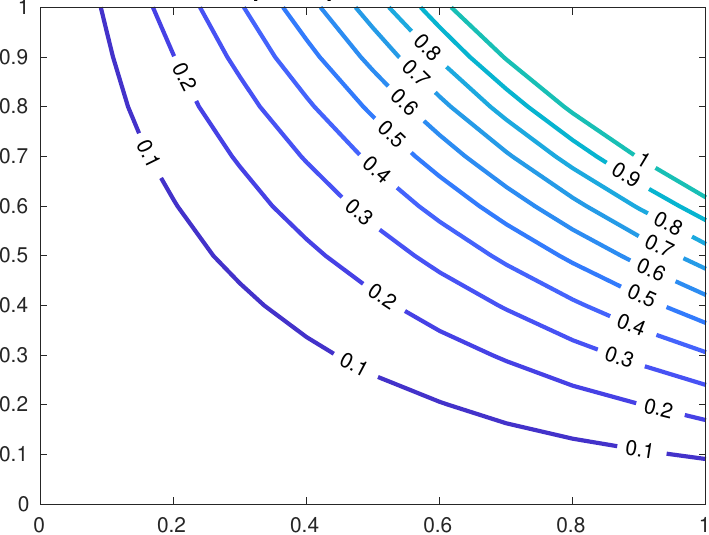}
		\label{fig:p1q3v3}
	}
	\subfigure[\scriptsize $p = 1, q = 3, v = 10$]{
		\includegraphics[width=4.5cm]{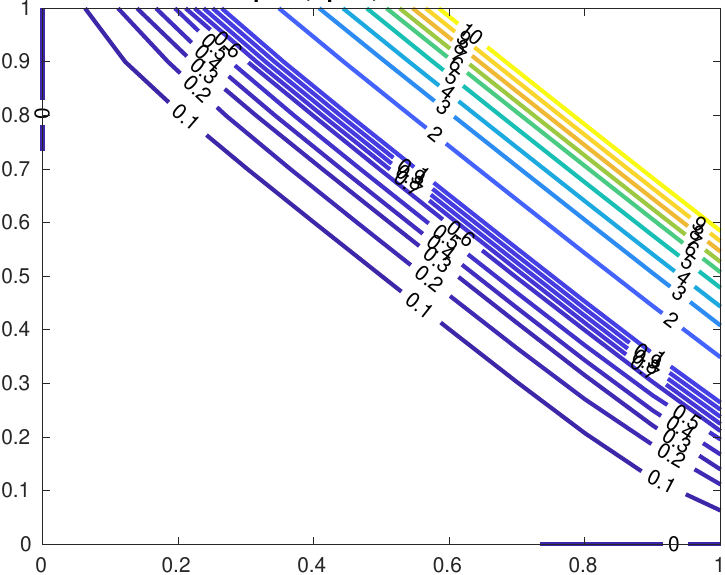}
		\label{fig:p1q3v10}
	}
	\subfigure[\scriptsize $p = 2, q = 3, v = 3$]{
		\includegraphics[width=4.5cm]{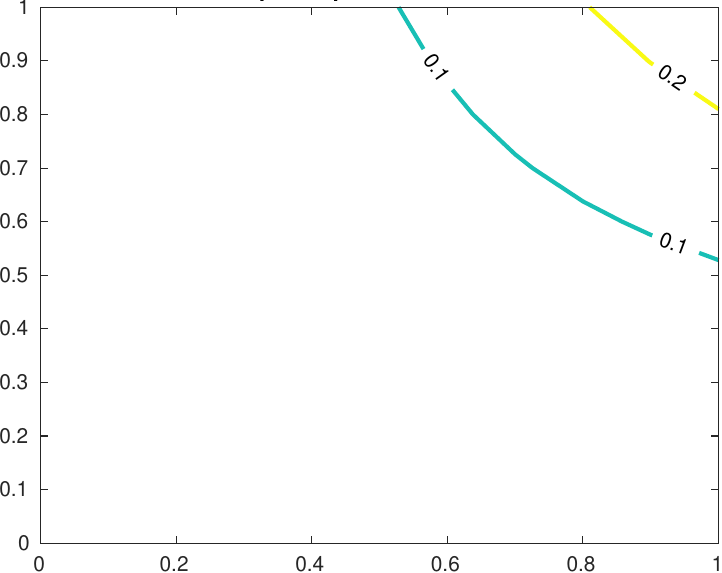}
		\label{fig:p2q3v3}
	}
	\subfigure[\scriptsize $p = 2, q = 3, v = 10$]{
		\includegraphics[width=4.5cm]{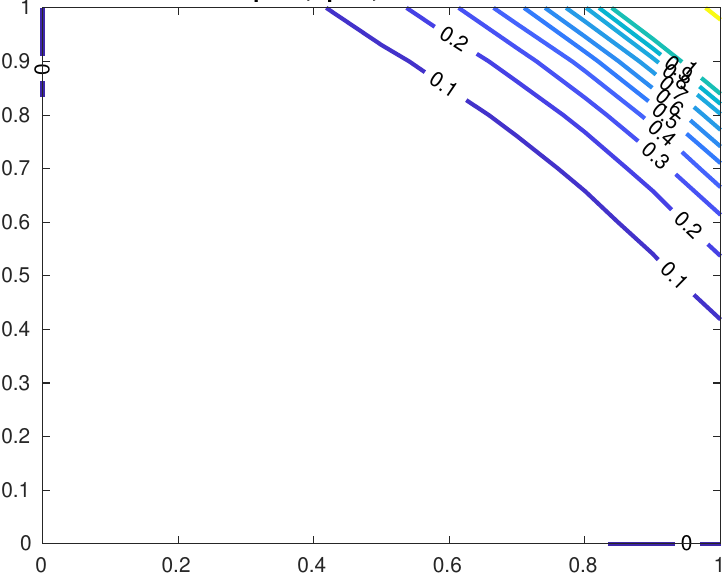}
		\label{fig:p2q3v10}
	}
	\caption{The contour of $\sum_{j = 1}^{N}\phi(\hb_j)$ with $\Hb \in \Rbb^{2 \times 1}$ with different combinations of $(p, q, v)$.}
	\vspace{-0.3cm}
	\label{fig:pqv_contour}
\end{figure}

\subsection{Algorithm performance with $(p, q, v)$}

We examine the effect of  $(p, q, v)$ on the performance of the proposed SNCP method. Fig. \ref{fig:pqv_fixpq} and Fig. \ref{fig:pqv_fixv} show the convergence curves of the SNCP method on synthetic data with SNR = -3dB. Note that we utilize the Armijo rule to compute the step sizes for Algorithm 1 since it is difficult to compute $\lambda_{\max}(\nabla_{\Hb}^2 G_{\rho}(\Wb^k, \Hb^k))$  and $\lambda_{\max}(\nabla_{\Wb}^2 G_{\rho}(\Wb^k, \Hb^{k+1}))$ when $v \geq q > 2$, and the results presented are generated by all algorithms with one same initialization. As seen in Fig. \ref{fig:pqv_fixpq}, with a bigger $v$, the SNCP method not only converges slower and cannot obtain an orthogonal $\Hb$ but also performs worse. We can also observe from Fig. \ref{fig:pqv_fixv} that when $q$ is much bigger than $p$, the performance of the SNCP method deteriorates in terms of both convergence speed and clustering accuracy.  

\begin{figure} [t!]
	\centering
	\subfigure[]{
		\includegraphics[width=4.5cm]{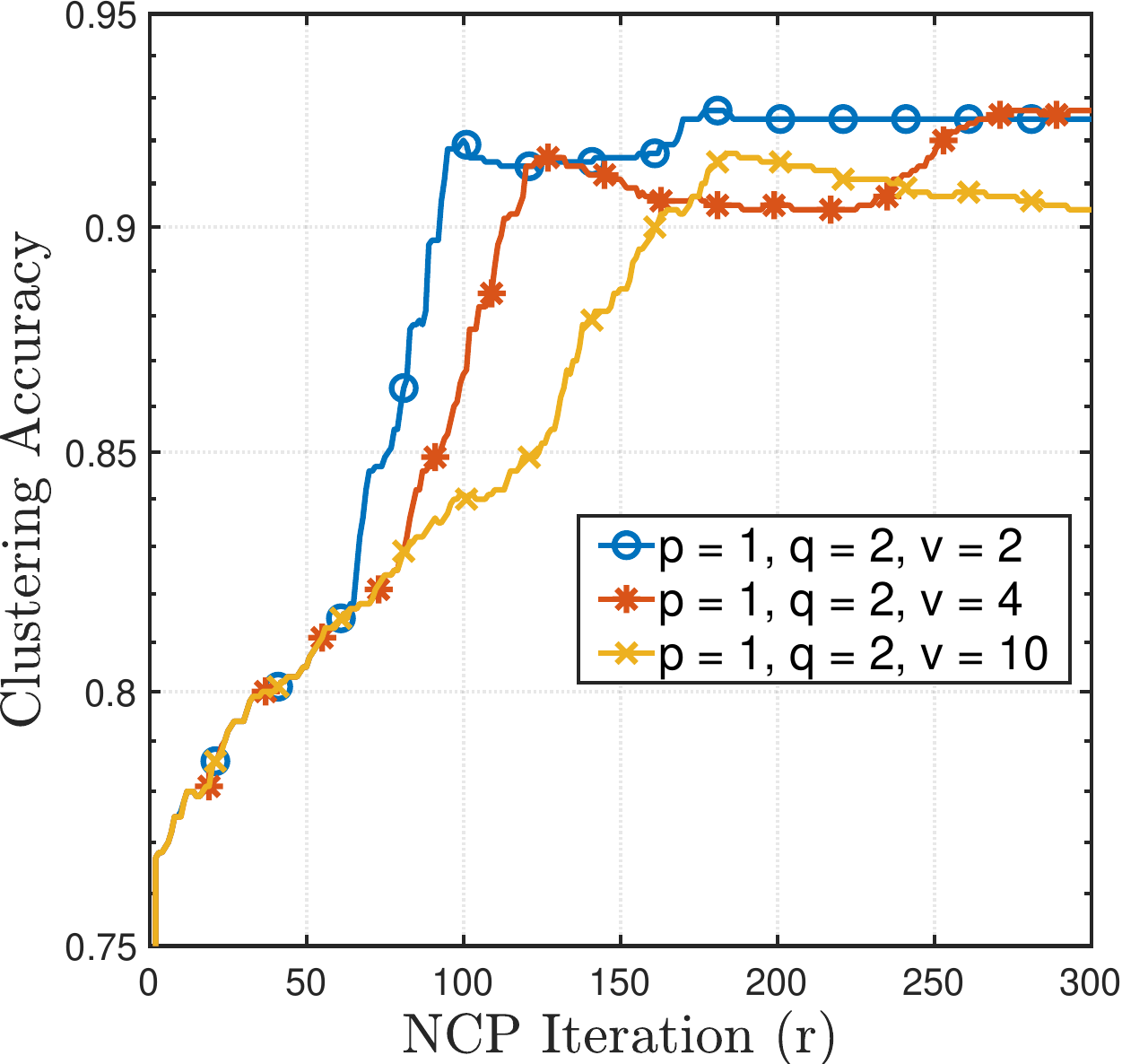}
	}
	\subfigure[]{
		\includegraphics[width=4.5cm]{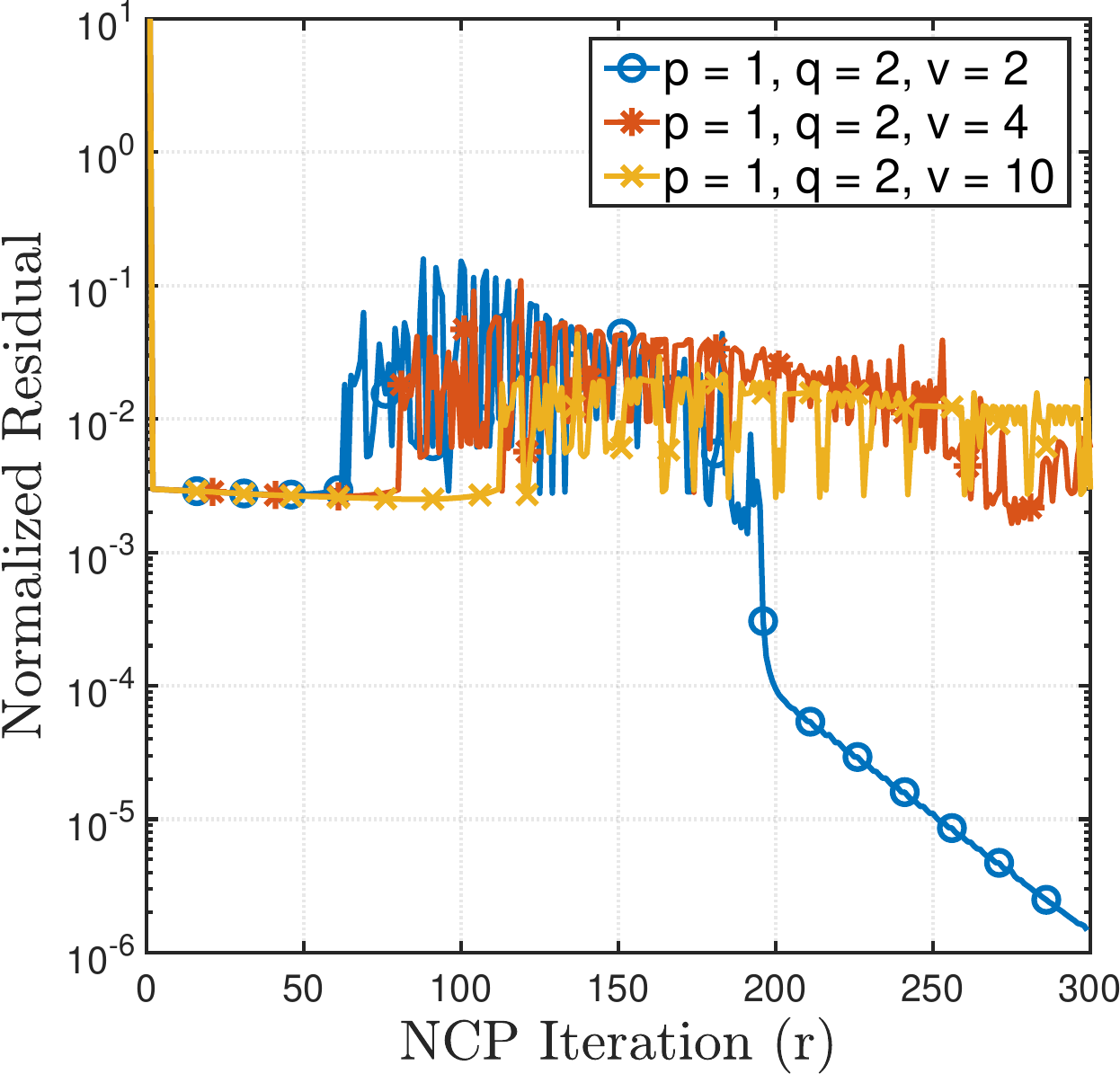}
	}
	\subfigure[]{
		\includegraphics[width=4.5cm]{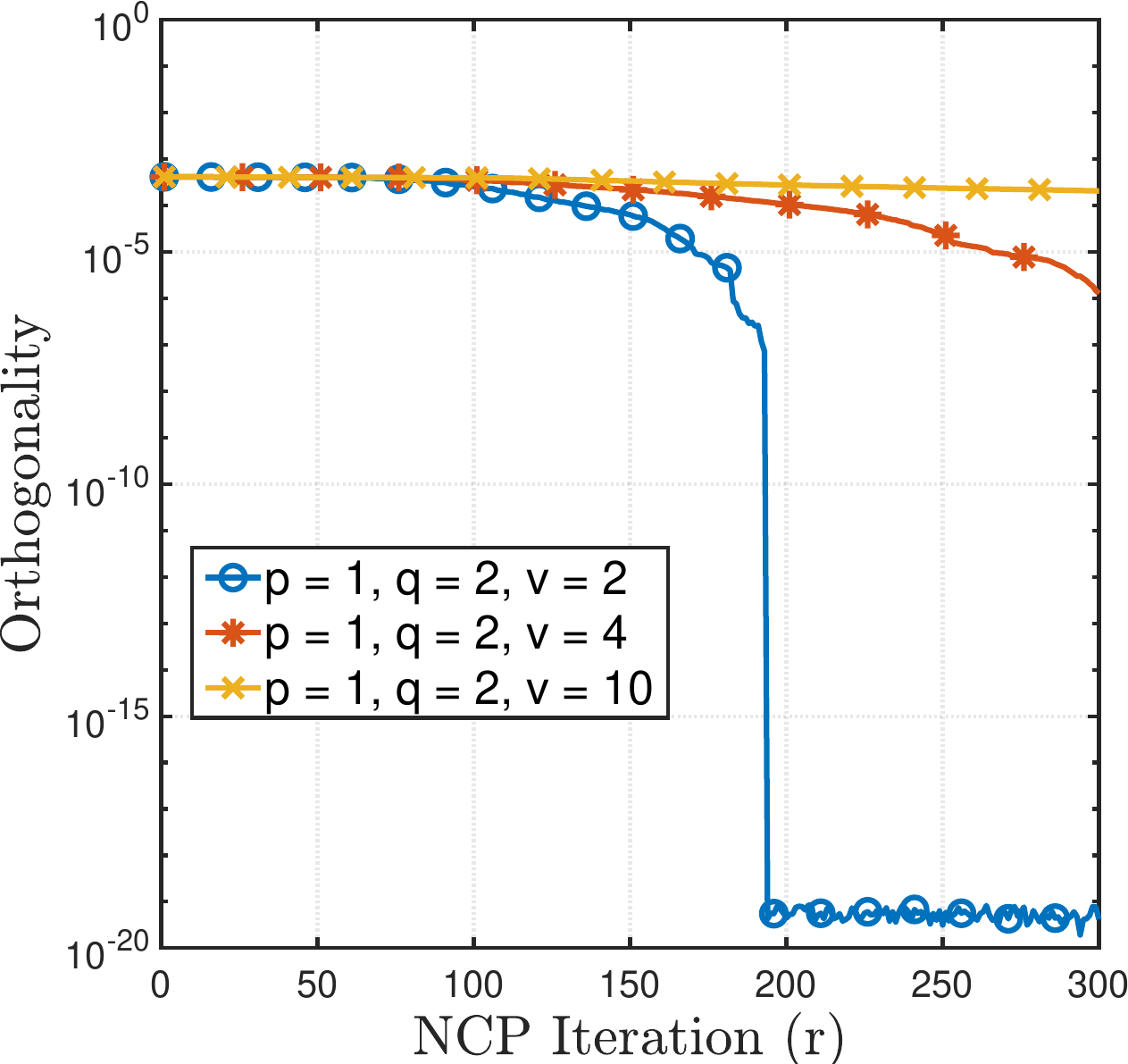}
	}
	\subfigure[]{
		\includegraphics[width=4.5cm]{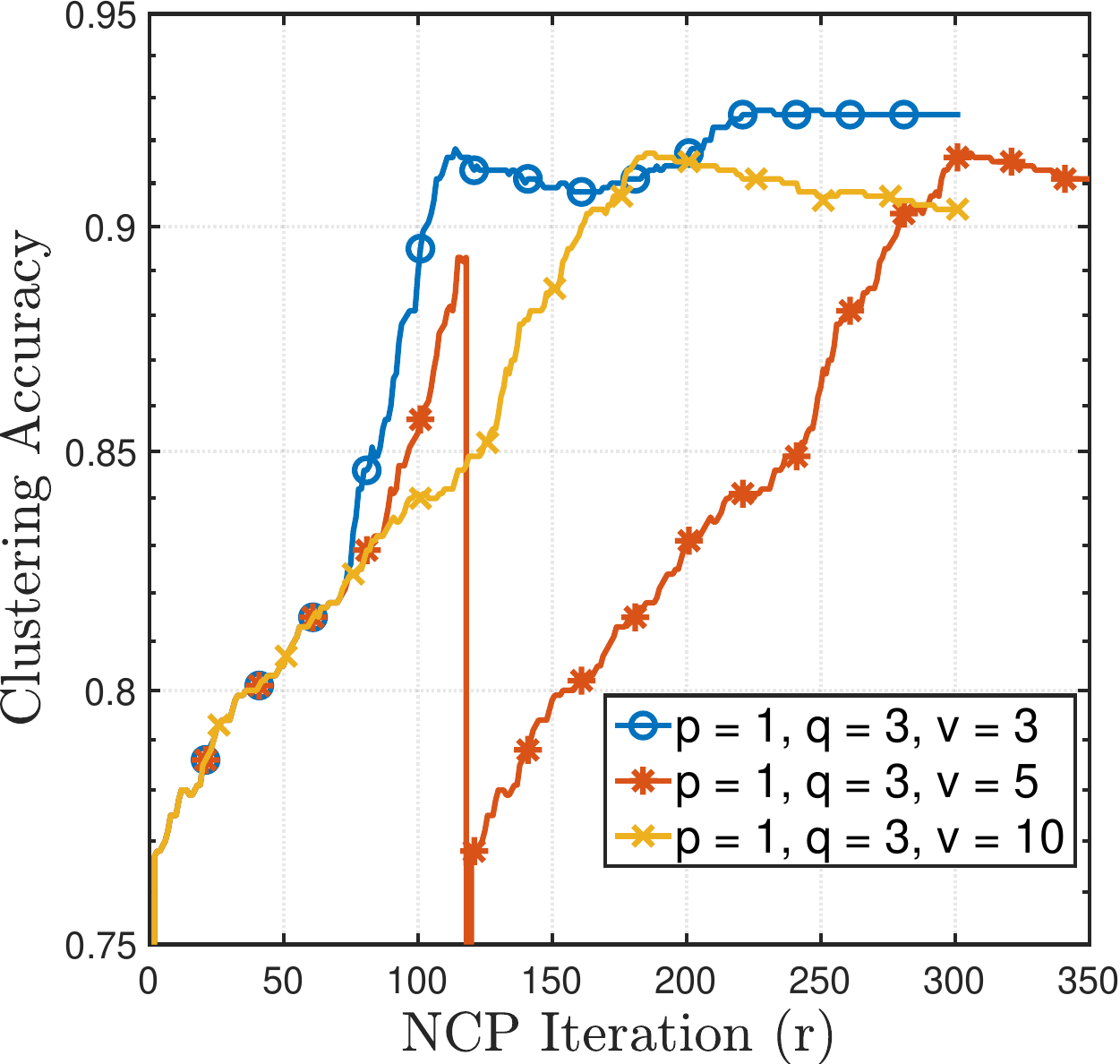}
	}
	\subfigure[]{
		\includegraphics[width=4.5cm]{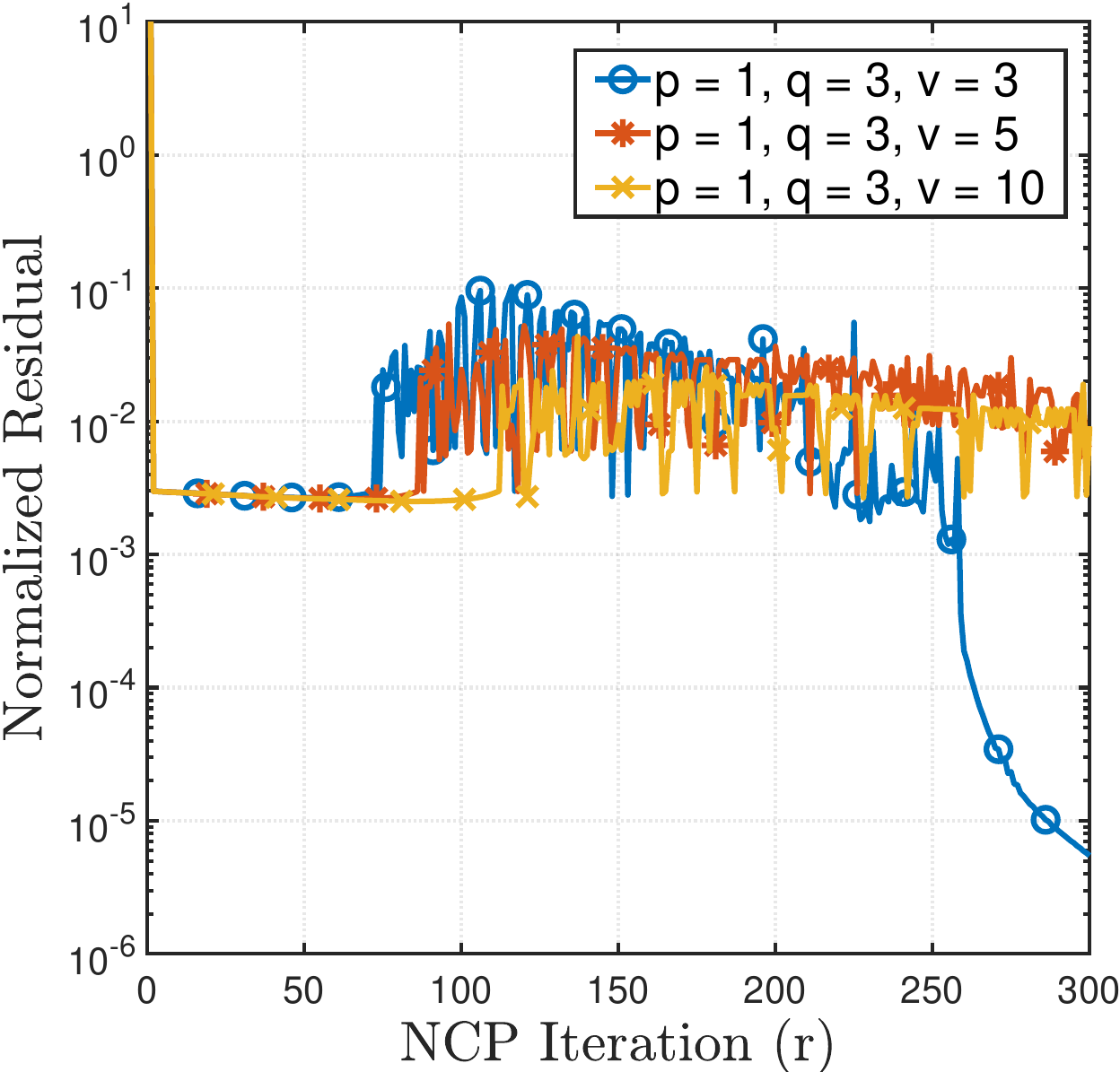}
	}
	\subfigure[]{
		\includegraphics[width=4.5cm]{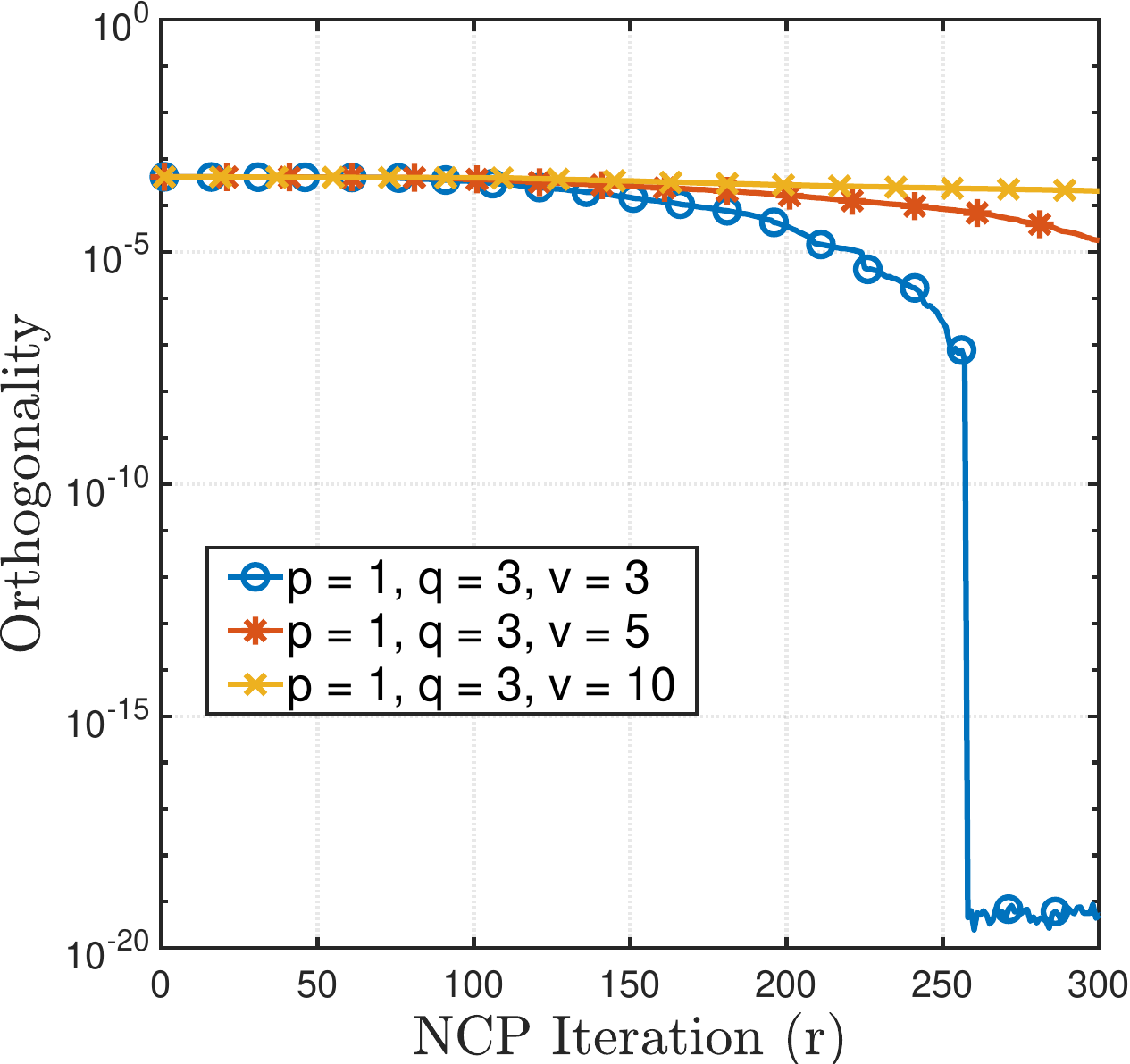}
	}
	\subfigure[]{
		\includegraphics[width=4.5cm]{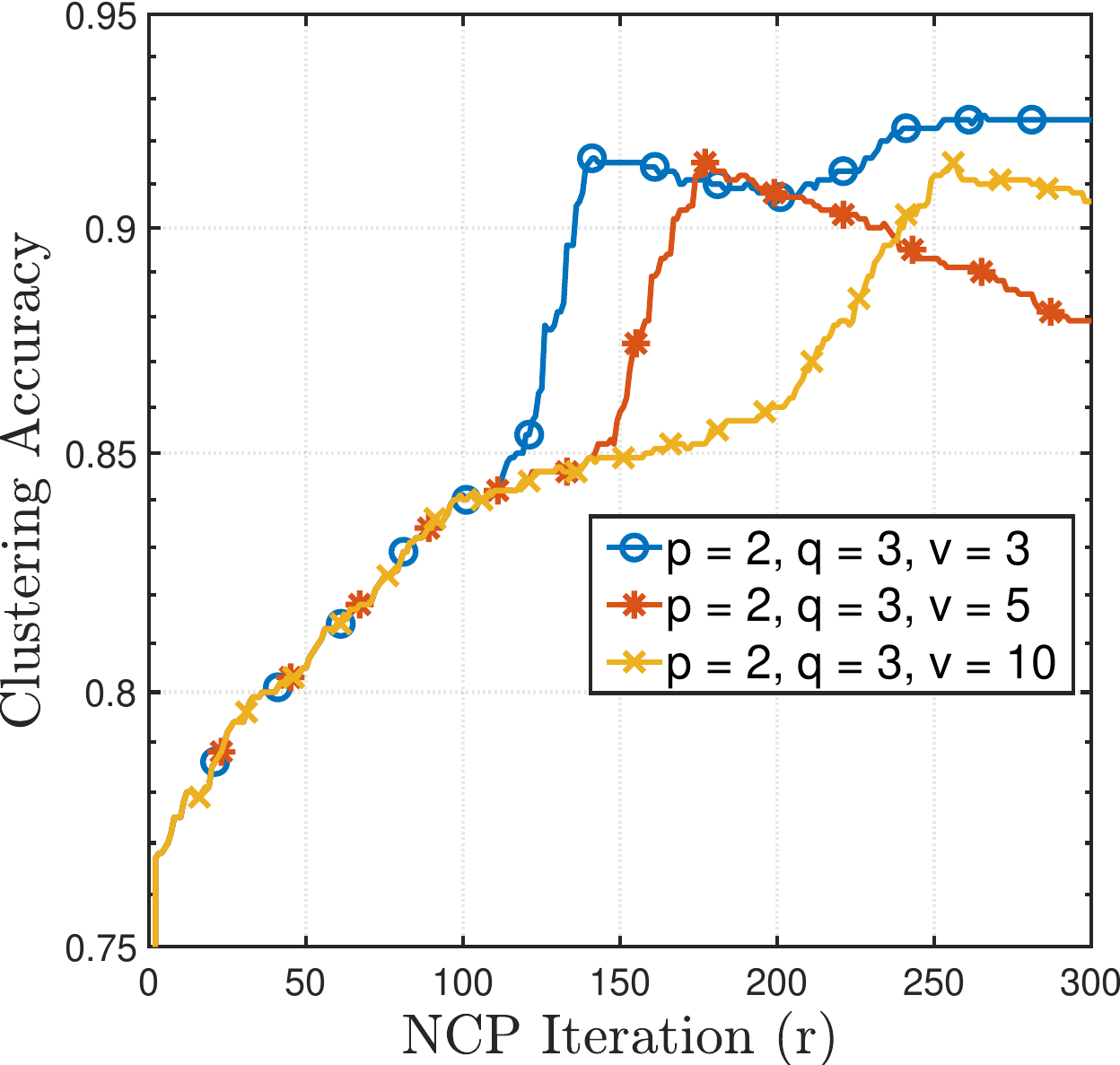}
	}
	\subfigure[]{
		\includegraphics[width=4.5cm]{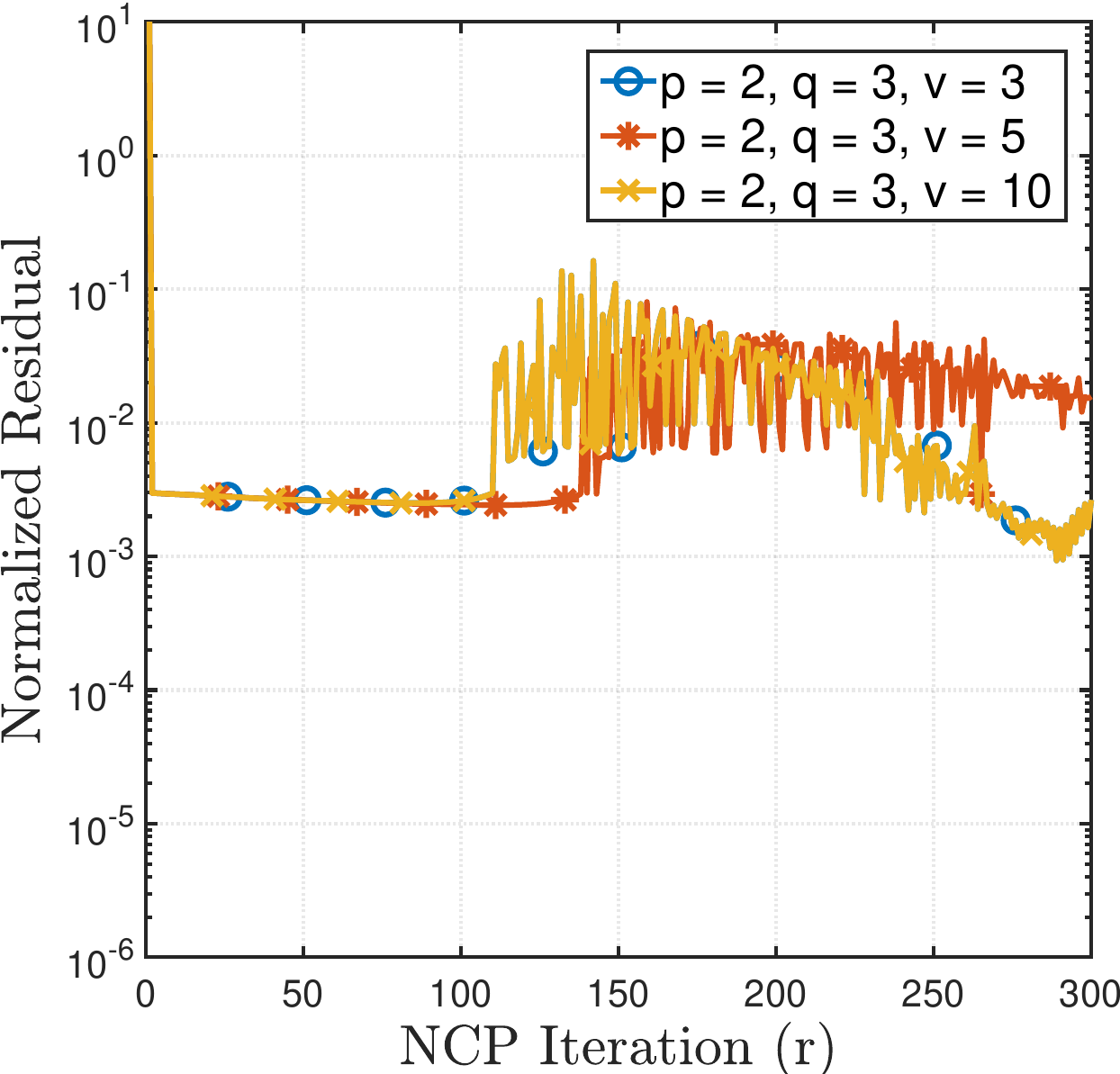}
	}
	\subfigure[]{
		\includegraphics[width=4.5cm]{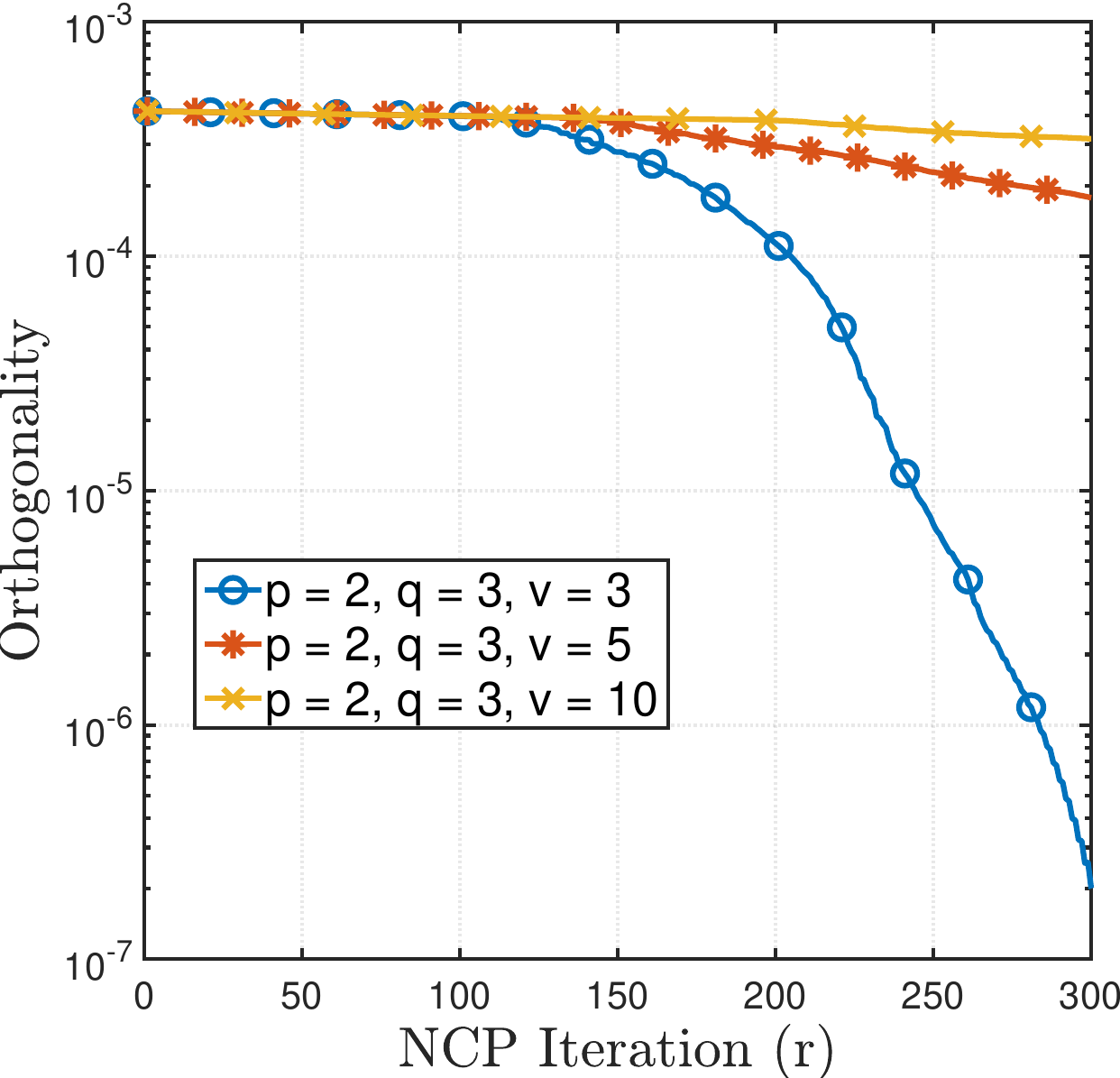}
	}
	\vspace{-0.25cm}
	\caption{Convergence curves versus NCP iteration of the proposed SNCP method with different values of $p, q, v$.}
	\label{fig:pqv_fixpq}
\end{figure}

\begin{figure}
	\centering
	\subfigure[]{
		\includegraphics[width=4.5cm]{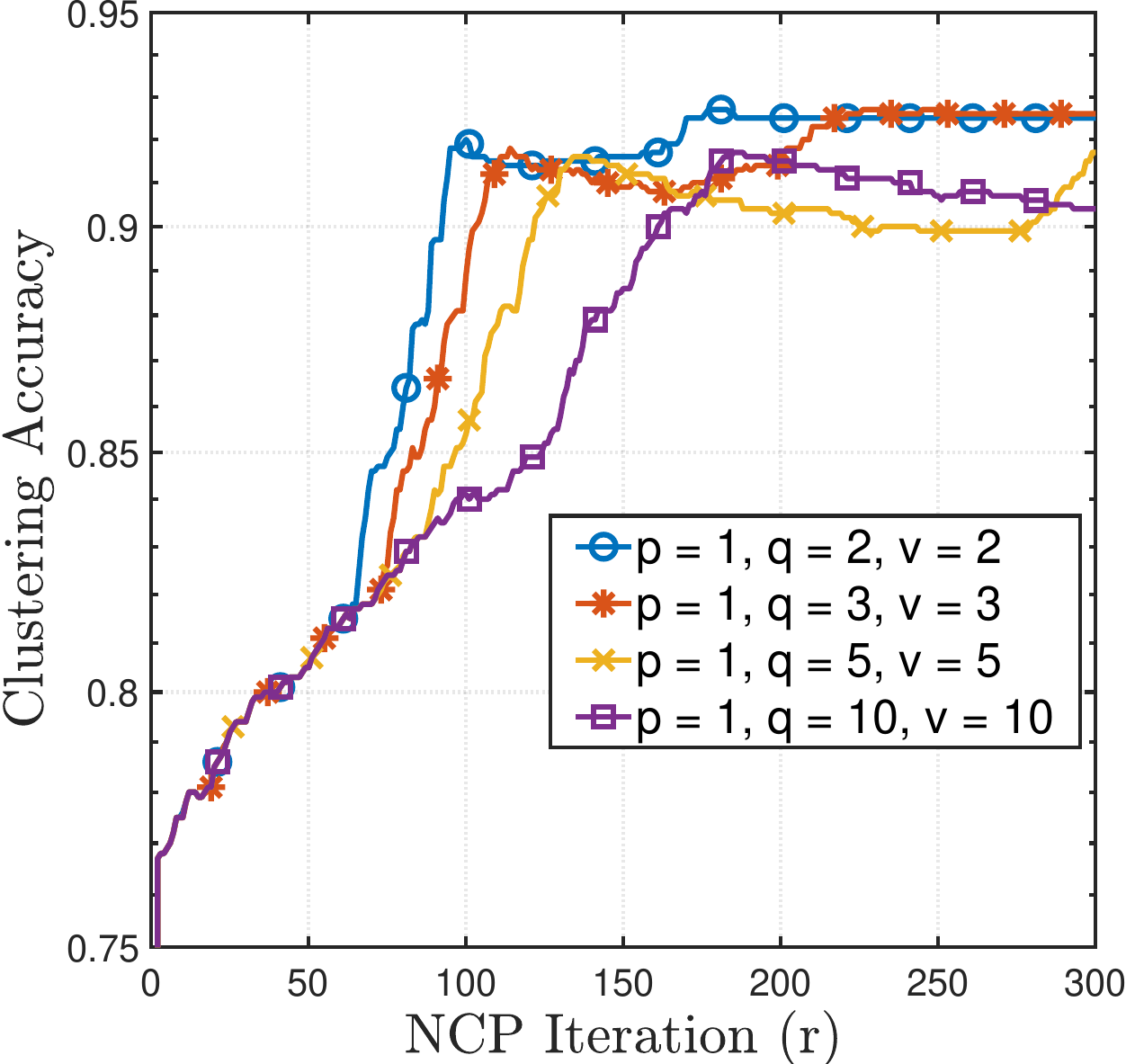}
	}
	\subfigure[]{
		\includegraphics[width=4.5cm]{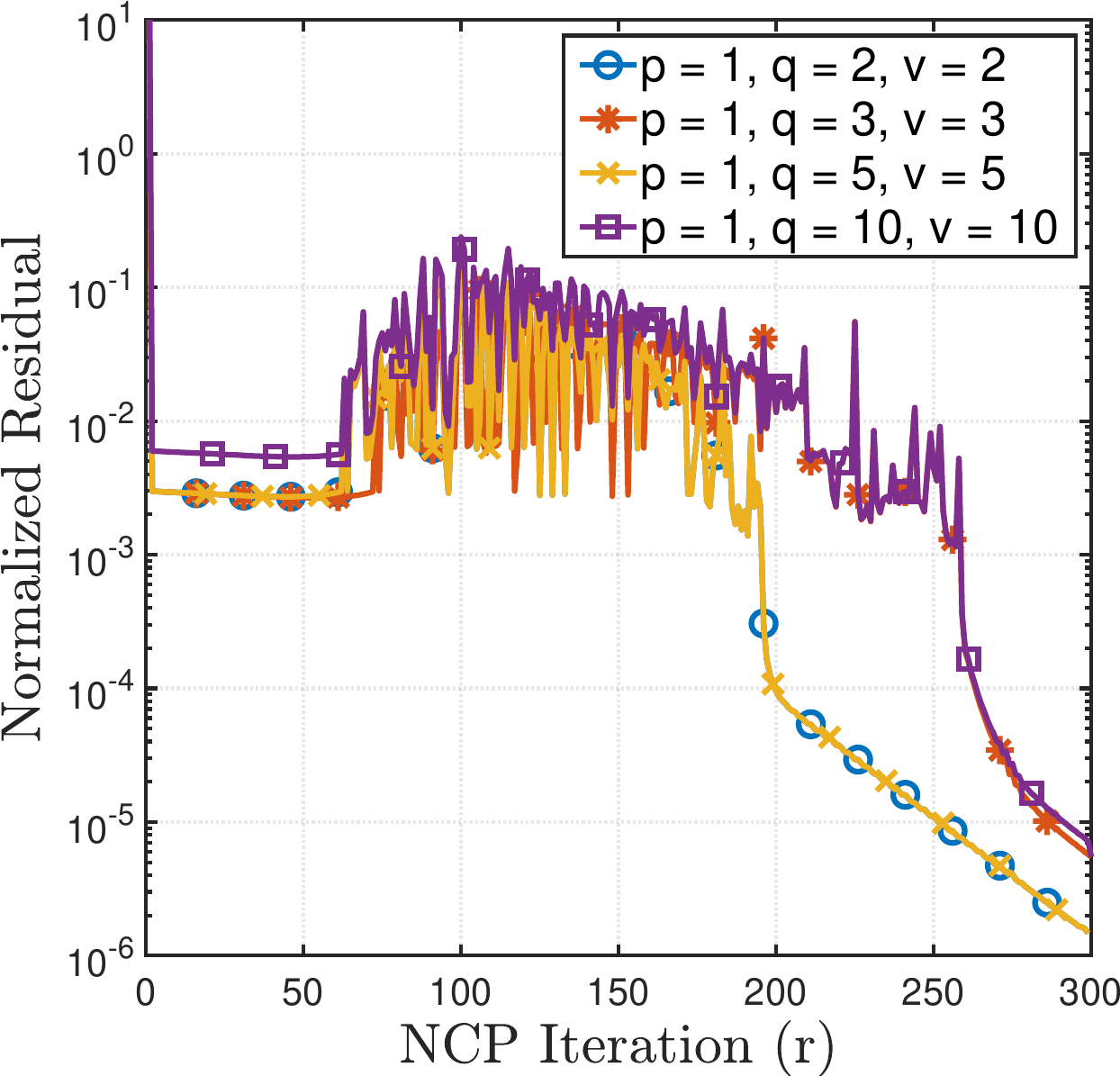}
	}
	\subfigure[]{
		\includegraphics[width=4.5cm]{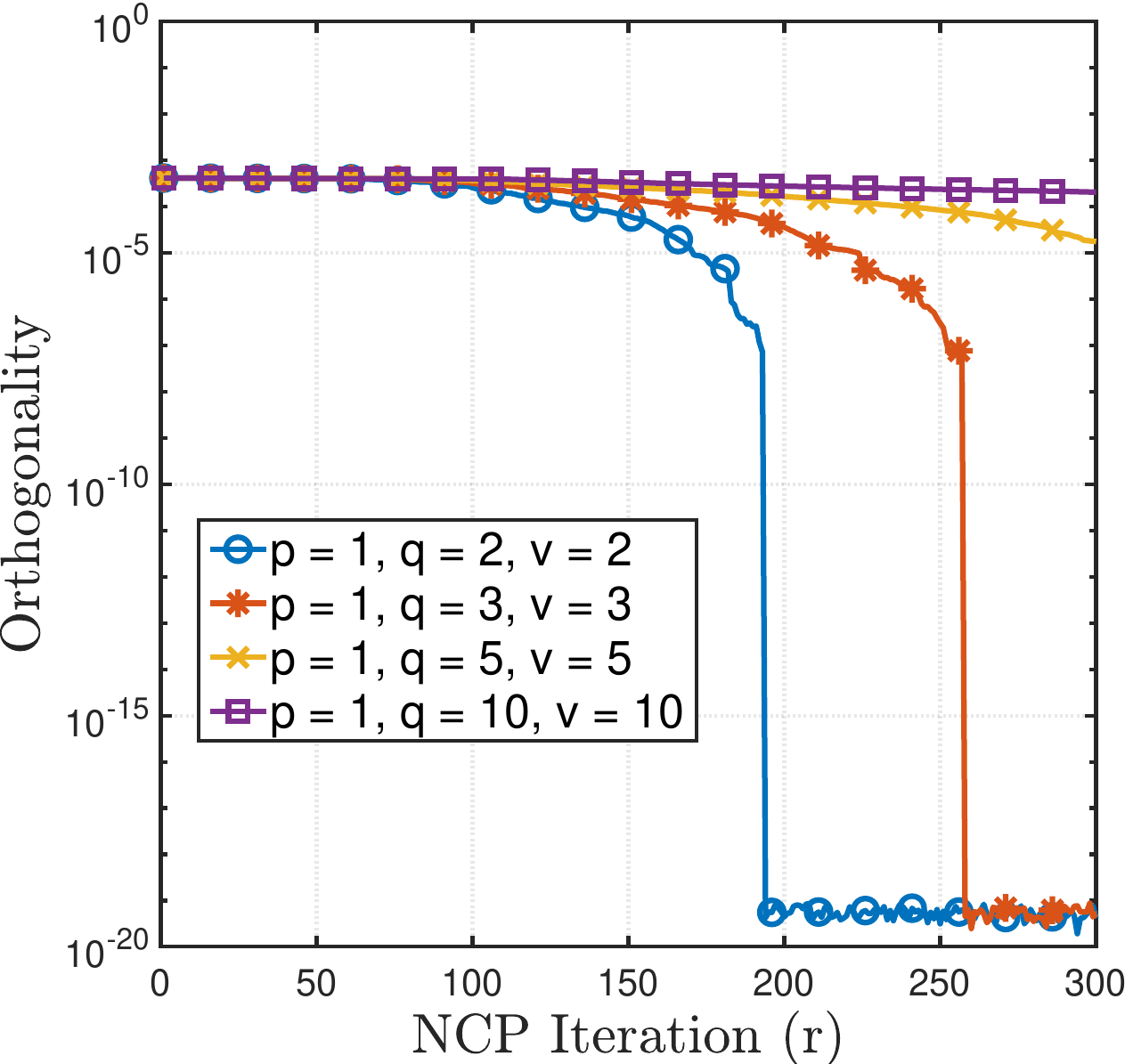}
	}
	\subfigure[]{
		\includegraphics[width=4.5cm]{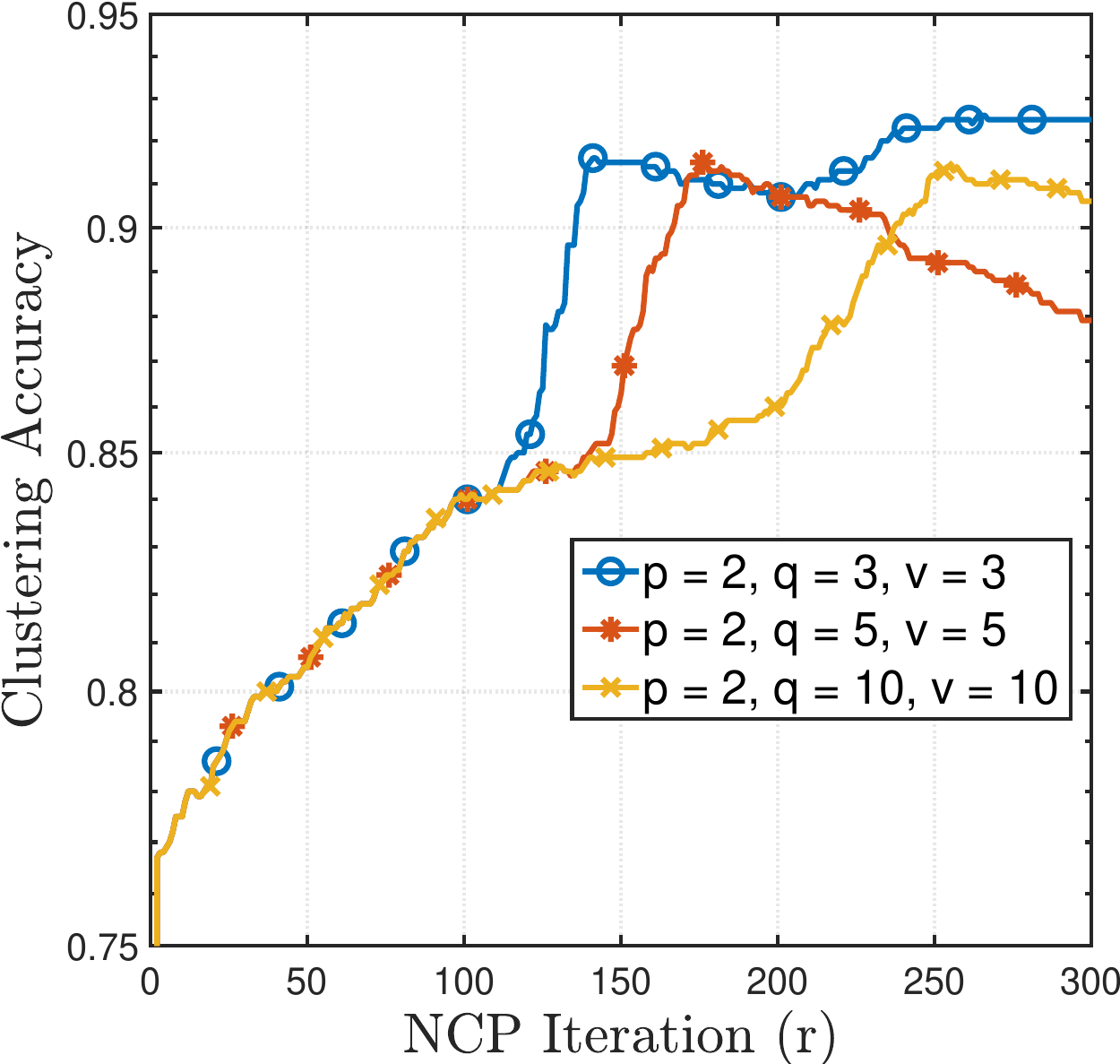}
	}
	\subfigure[]{
		\includegraphics[width=4.5cm]{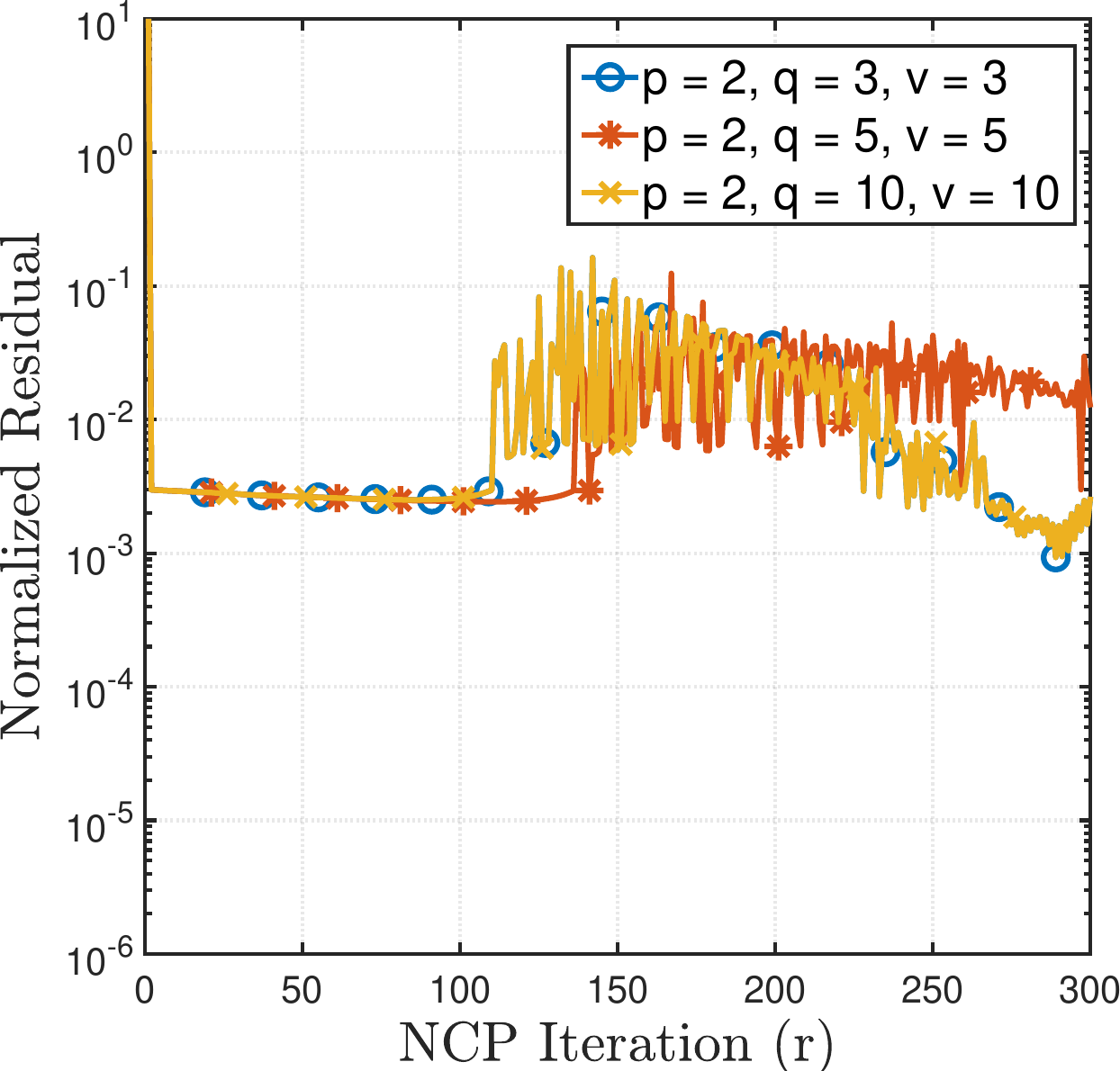}
	}
	\subfigure[]{
		\includegraphics[width=4.5cm]{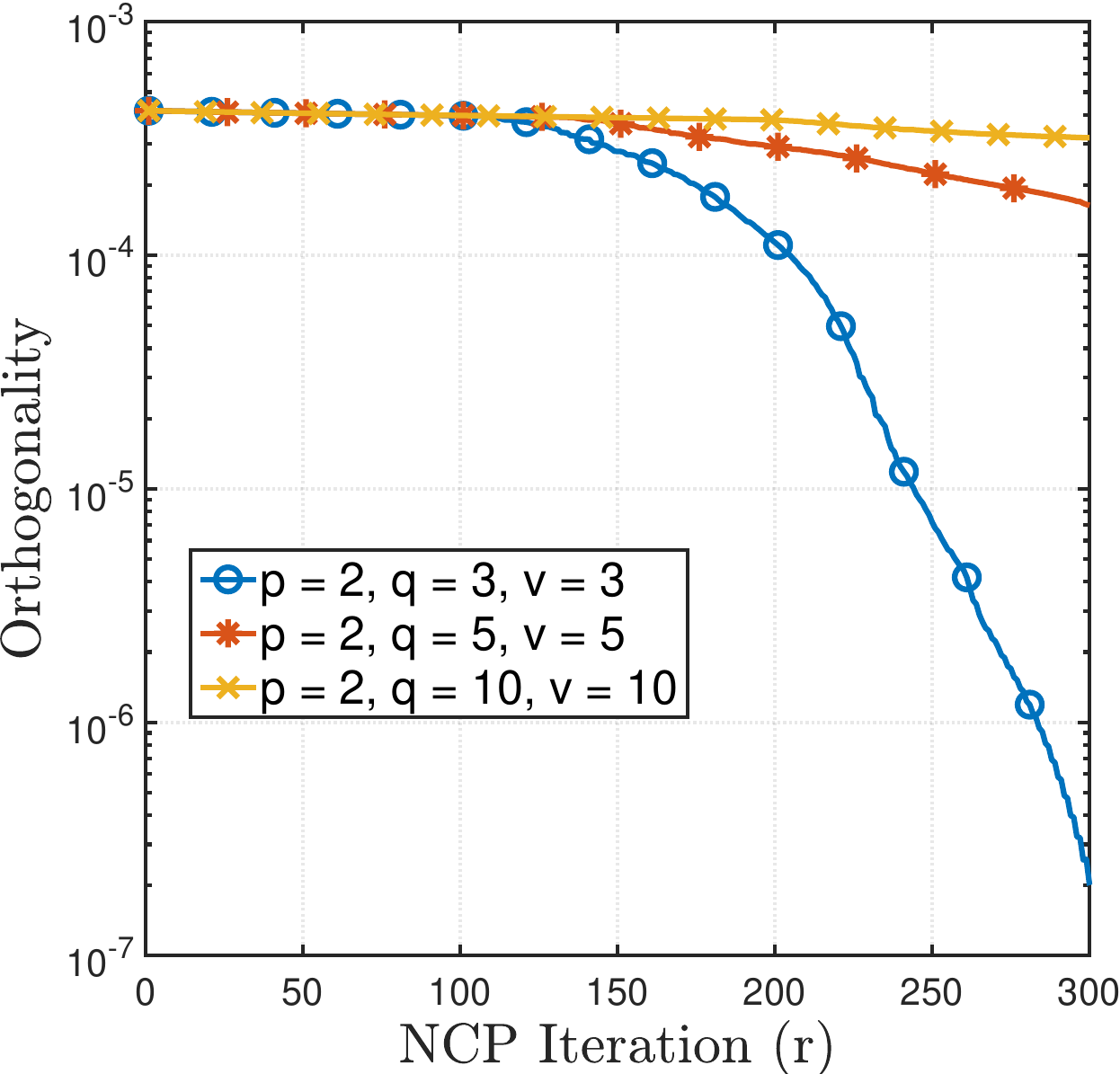}
	}
	\vspace{-0.25cm}
	\caption{Convergence curves versus of NCP iteration of the proposed SNCP method with different values of $p, q, v$.}
	\label{fig:pqv_fixv}
\end{figure}

\section{Complete results of clustering performance}

In this section, we present the complete results of Table I, II, III and IV in the manuscript including both the values of ACC and ARI. Specifically, Table \ref{table: clustering quality} corresponds to Table I, Table \ref{table: clustering TCGA data} corresponds to Table II, Table \ref{table: clustering doc data} corresponds to Table III, and Table \ref{table: doc data after DR} corresponds to Table IV.

 \begin{table}[H] 
		\centering \footnotesize
		\caption{Average clustering performance (\%) and CPU time (s) on the synthetic data for different values of SNR. 
		}\vspace{-0.2cm}
		\setlength{\tabcolsep}{3.0mm}
		\label{table: clustering quality}
		\begin{tabular}{|c|c|c|c|c|c|c|c|}
			\hline
			\multicolumn{2}{|c|}{SNR (dB)}              
			&       -5       &       -3       &       -1       &       1        &       3        &       5        \\ \hline
			
			\multirow{9}{0.7cm}{ARI}   
			&   KM    &     57.0     &     65.8      &     73.8      &     73.6      &     73.4      &     68.7      \\ \cline{2-8}
			&     KM++      &     57.1      &     66.9      &     63.7      &     66.8      &     65.3      &     72.9      \\ \cline{2-8}
			&     DTPP     &     78.5      &     82.8      &     83.0      &     84.4      &     83.5      &     86.9      \\ \cline{2-8}
			&     ONP-MF     &     56.6     &    87.0       &     80.2      &     89.6      &     89.6      &     89.2      \\ \cline{2-8}
			&     ONMF-S     &     71.2      &     73.4      &     75.7      &     76.3      &     76.8      &     77.1      \\ \cline{2-8}
			&     HALS     &     67.9      &     81.9      &     85.2      &     87.2      &     87.0      &     86.2      \\ \cline{2-8}
			&     SNCP     & \textbf{91.1} & \textbf{91.4} & \textbf{91.3} &  \textbf{91.5} & \textbf{91.6} &  \textbf{91.9} \\ \cline{2-8}
			&     NSNCP     & 89.7 &  90.3&  91.0 & 91.3 &  91.4& 91.5 \\ \hline
			\multirow{9}{0.7cm}{ACC}   
			&   KM    &     63.4      &     69.7     &     74.7      &     74.3      &     75.6      &     73.7      \\ \cline{2-8}
			&     KM++      &     64.1      &     70.1      &     70.3      &     70.8      &     69.2      &     75.5      \\ \cline{2-8}
			&     DTPP     &     81.6      &     85.1      &     85.9      &     87.0      &     86.6      &     89.4      \\ \cline{2-8}
			&     ONP-MF     &     66.8      &     88.3      &   83.1  &     89.9      &     90.0      &     90.4      \\ \cline{2-8}
			&     ONMF-S     &     77.4      &     79.0      &     80.2      &     81.3      &     81.7      &     81.9      \\ \cline{2-8}
			&     HALS     &     76.3      &     86.0      &     88.1      &     89.6      &     89.4      &     89.3      \\ \cline{2-8}
			&     SNCP     & \textbf{91.5} & \textbf{91.9} &  \textbf{92.0}& \textbf{92.5} & \textbf{92.8}  &  \textbf{93.4}\\
			\cline{2-8}
			& NSNCP     &  90.1 &  90.7 & 91.8 &  92.0 &  92.2 &  92.5 \\ \hline
			\multirow{9}{0.7cm}{Time}   
			&   KM    &     3.10      &     1.96     &     1.52      &     1.16      &     1.11      &     1.01      \\ \cline{2-8}
			&     KM++      &  3.09    &   2.44        &  2.05         &  1.70         &  1.59        &  1.59         \\ \cline{2-8}
			&     DTPP     &     30.4      &     34.0      &     38.6      &     31.1      &     35.1      &     30.7      \\ \cline{2-8}
			&     ONP-MF     &     1097      &     1123      &     1124    &     1153      &     1148      &     1129      \\ \cline{2-8}
			&     ONMF-S     &     68.4      &     97.1      &     116      &     116      &     90.5      &     104.6      \\ \cline{2-8}
			&     HALS     &     22.6      &     25.6      &     23.3      &     15.7      &     19.3      &     16.7      \\ \cline{2-8}
			&     SNCP     &  20.6 &  15.6 &  13.4 &  13.3 &  12.4 &  11.7\\ \cline{2-8}
			&     NSNCP    & 14.8 & 15.5 &  11.5 & 10.5 &  10.6 &  9.75\\ \hline		
		\end{tabular}
		\vspace{-0.35cm}
\end{table}

\begin{table}[H] 
	\centering \footnotesize
	\caption{Average clustering performance ($\%$) and CPU time (s) on the TCGA data. } 
	\setlength{\tabcolsep}{3mm}
	\label{table: clustering TCGA data}\vspace{-0.2cm}
	\begin{tabular}{|c|c|c|c|c|c|c|}
		\hline
		\multicolumn{2}{|c|}{Dataset}              
		&       1       &       2       &       3       &       4        &       5            \\ \hline
		\multicolumn{2}{|c|}{\#samples $N$}              
		&       1667       &       3086       &    3660    &  5314       &       11135          \\ \hline
		\multicolumn{2}{|c|}{\#cancers $K$}              
		&       5       &       10       &    15    &  20       &       33           \\ \hline
		\multirow{8}{0.7cm}{ARI} 
		&   KM    &     55.9     &     47.9      &     38.2     &     29.4     &     19.4        \\ \cline{2-7}
		&   KM++    &   59.3      &     33.5      &   30.1  &  21.1      &    17.2    \\ \cline{2-7}  
		&   DTPP    &  79.2      &     46.4      &     47.3      &   43.6     &   \textbf{35.9}       \\ \cline{2-7}
		&   ONMF-S      &  86.1  &  74.8   & 60.3  &   48.1     &   25.4    \\ \cline{2-7}
		&   ONP-MF      &   83.7     &  56.6       &   37.5    &   18.4   & 14.5 \\ \cline{2-7}
		&     HALS     &  89.6    &  63.0   &  48.1   & 43.6    &  30.9  \\\cline{2-7}
		&     SNCP  & 88.7 & 76.4 & 58.5 & 42.4 & 27.2  \\ \cline{2-7}
		&     NSNCP     & \textbf{90.8} & \textbf{77.2} &  \textbf{67.1} & \textbf{50.3} & 30.7   \\ \hline
		\multirow{8}{0.7cm}{ACC}
		&   KM    &     75.0      &   67.0      &   57.0      &  52.2      &    34.4    \\ \cline{2-7}
		&   KM++    &    75.5      &    55.3      &    53.7      &     48.4     &    34.5      \\\cline{2-7}   
		&   DTPP    &     79.2      &     58.5      &    58.0      &     57.0      &  \textbf{43.1}  \\ \cline{2-7}
		&     ONMF-S      &    85.8  &   71.8       &   61.9       & 58.2       &     38.4       \\ \cline{2-7}
		&     ONP-MF      &  84.0      &  68.5    &    48.1    &  32.9    & 14.8  \\ \cline{2-7}
		&     HALS     &  86.2  & 68.8   &  56.1   & 56.3   & 39.1 \\ \cline{2-7}
		&     SNCP     & 85.6 & 79.3 & 64.0 & 61.1 &41.1 \\ \cline{2-7}
		&     NSNCP     & \textbf{89.2} & \textbf{81.2} & \textbf{68.2} & \textbf{64.3} & 42.7 \\ \hline
		\multirow{8}{0.7cm}{Time}
		&   KM    &    3.02      &    7.54     &   12.3      &   34.7      &     115     \\ \cline{2-7}
		&   KM++    &     3.08      &     7.83      &     26.9      &  40.3  &     331         \\\cline{2-7}   
		&   DTPP    &   249      &  434    &    1357    &   2135      &     2454       \\ \cline{2-7}
		&     ONMF-S      & 89.9    &   772     &   1440     &  2492     &   19794  \\ \cline{2-7}
		&     ONP-MF      & 4886  &  10414       &   14232   &    16092  &       77222    \\ \cline{2-7}
		&     HALS     &     2.26     &   11.4      &    98.1     &  260     &   1605      \\ \cline{2-7}
		&     SNCP     & 41.6 &  249 & 307 & 486 & 1118 \\ \cline{2-7}
		&     NSNCP     & 22.1 & 153 & 240 & 526 &  1756 \\ \hline	
	\end{tabular}
\end{table}

\begin{table}[t!] 
	\centering \footnotesize
	\caption{Average clustering performance ($\%$) and CPU time (s) on TDT2 data. } 
	\setlength{\tabcolsep}{3mm}
	\label{table: clustering doc data}\vspace{-0.1cm}
	\begin{tabular}{|c|c|c|c|c|c|c|c|}
		\hline
		\multicolumn{2}{|c|}{Dataset}              
		&       1      &       2      &    3     &   4  &   5      &   6   \\ \hline	
		\multicolumn{2}{|c|}{\#terms $M$}              
		&       13133      &       24968       &      11079       &   20431    &   16067      &   29725       \\ \hline
		\multicolumn{2}{|c|}{\#docs $N$}              
		&       842       &      3292      &    631     &    1745       &       1079       &     4779        \\ \hline	
		\multirow{7}{0.7cm}{ARI} 
		&   KM    &     66.7      &     35.3      &     82.6      &     34.4      &     \textbf{55.9}      &  56.2  \\ \cline{2-8}
		&   KM++    &     71.0      &     35.8      &     82.9      &     32.2      &    55.1     & 53.2  \\
		\cline{2-8} 
		&   DTPP    &     72.0      &     33.0      &     81.0      &     45.8      &     36.7  & 54.7     \\ \cline{2-8}
		&   ONMF-S      &  78.8  &  33.8   &  84.7   &   50.3     &  41.1         &  56.0          \\ \cline{2-8}
		&   ONP-MF      &   78.8        &    28.0       &  86.1         &    46.1       &   48.4    & 52.0          \\ \cline{2-8}
		&     HALS     &  78.1    &  32.1   &  87.0    & 45.4    &    39.9  &  52.5\\ \cline{2-8}
		&     SNCP  & \textbf{88.3} & \textbf{36.4} & 87.1 & \textbf{51.5} & 48.9 & \textbf{60.7}  \\ \cline{2-8}
		&     NSNCP     & 87.9 & 35.8 & \textbf{88.2} & 50.5 & 44.8 & 52.7    \\ \hline
		\multirow{7}{0.7cm}{ACC}
		&   KM    &     77.9      &     51.7      &     84.4      &     49.3      &     61.3      & 70.2  \\ \cline{2-8}
		&   KM++    &     73.5      &     48.9      &     84.7      &     47.1      &     \textbf{61.7}      & 65.7  \\
		\cline{2-8} 
		&   DTPP    &     70.0      &     50.4      &     77.4      &     52.1      &     45.1     & 66.6  \\ \cline{2-8}
		&     ONMF-S      &    81.8  &   52.0       &    83.6       & 59.6          &     50.9      &     68.8        \\ \cline{2-8}
		&     ONP-MF      &   85.3        &   46.1        &  \textbf{89.5}         &    60.9       &  59.4    &  66.8          \\ \cline{2-8}
		&     HALS     &  77.9   & 47.8   &  85.1   & 54.3   &  48.7  &  64.4  \\ \cline{2-8}
		&     SNCP     & \textbf{86.1} & \textbf{56.8} & 88.1 & \textbf{62.0} & 58.2 & \textbf{70.7}\\ \cline{2-8}
		&     NSNCP     &  79.7 &  54.1 &  88.6 &  60.8 &  56.4 &  64.9  \\ \hline
		\multirow{7}{0.7cm}{Time}
		&   KM    &     13.6      &     195      &     6.82      &     65.0      &     27.4      &     318       \\ \cline{2-8}
		&   KM++    &     10.6      &     199      &     8.15      &    53.6  &     28.9      &     401      \\
		\cline{2-8}  
		&   DTPP    &     126      &     726     &     81      &    413      &     204      &   2282         \\ \cline{2-8}
		&     ONMF-S      & 405    &   8124     &   157     &   1915     &  633    &   17603    \\ \cline{2-8}
		&     ONP-MF      &    1200     &     7854      &     757      &     3660      &     1951      &  14302      \\ \cline{2-8}
		&     HALS     &     12.9      &    35.6      &     6.07      &   27.4      &    20.0      &     162      \\ \cline{2-8}
		&     SNCP     & 37.3 &  407 & 20.4 & 64.0 & 41.1  & 342  \\ \cline{2-8}
		&     NSNCP     & 24.7 &  335 &  16.3 &  50.0 &   37.6 &  424  \\ \hline	
	\end{tabular}
	\vspace{-0.3cm}
\end{table}

\begin{table}[H] 
	\centering \footnotesize
	\caption{Average clustering performance ($\%$) and CPU time (s) on dimension-reduced TDT2 data } 
	\label{table: doc data after DR}
	\setlength{\tabcolsep}{3mm}
	\label{table: clustering real data_DR}\vspace{-0.2cm}
	\begin{tabular}{|c|c|c|c|c|c|c|c|}
		\hline
		\multicolumn{2}{|c|}{Dataset}              
		&       1       &       2       &       3       &       4        &       5        &       6        \\ \hline
		\multicolumn{2}{|c|}{Time of SC }              
		&       13.8       &    402       &       7.11       &      90.7       &     30.7        &       1114       \\ \hline			
		\multirow{7}{0.7cm}{ARI} 
		&   SC+KM    &     77.7  &  58.8   &    78.8    &  81.7     &    88.2 & 72.1 \\ \cline{2-8}
		&   SC+KM++    &     96.7     &     77.8      &     97.9     &     \textbf{99.3}      &     98.0      & 82.7  \\ \cline{2-8}
		&   SC+HALS  & 93.3  & 78.3  & 96.6  & 97.7 & 99.1   & 80.2  \\ \cline{2-8}
		&   SC+ONP-MF    & 97.6 & 77.9 & \textbf{98.4}   &  98.9   &  99.3  & 79.9  \\ \cline{2-8}
		&   SC+SNCP    & \textbf{98.0} & \textbf{79.3} & 97.6   &  98.8   &  \textbf{99.4}  & \textbf{83.4}  \\ \hline
		\multirow{7}{0.7cm}{ACC}
		&   SC+KM    &     81.6      &    69.0   &   83.1      &   86.3   &     89.1   & 80.2  \\ \cline{2-8}
		&   SC+KM++    &     98.3      &    83.3      &     98.9     &     \textbf{99.6}      &     98.9      & 90.3 \\
		\cline{2-8}
		&   SC+HALS    & 95.8   & 82.8  &  97.8  & 98.8  & 99.4   & 90.8  \\ \cline{2-8}	
		&   SC+ONP-MF    &   98.6  &  82.3 &   \textbf{99.0}  &  99.3   &  \textbf{99.5}  & 89.7  \\ \cline{2-8}
		&   SC+SNCP    & \textbf{99.0} & \textbf{84.3} & 98.0   &  99.2 & 99.3  & \textbf{92.7}  \\ \hline
		\multirow{7}{0.7cm}{Time}
		&   SC+KM    &   0.24  & 1.34 &   0.17 &   0.45    &  0.33 & 2.33  \\ \cline{2-8}
		&   SC+KM++    &  0.25     & 1.96 & 0.18  &     0.46  &   0.34   &  2.18   \\ \cline{2-8}
		&   SC+HALS    &  0.78  &  2.45  & 0.81  & 1.65   & 0.75  & 1.52  \\ \cline{2-8}	
		&   SC+ONP-MF    &     13.6  &  385  &   7.29     &   90.0   &     26.0      & 790 \\ \cline{2-8}
		&   SC+SNCP    & 0.59 & 2.00 &0.55 & 1.31   & 1.21  & 3.17 \\ \hline  	
	\end{tabular}
\end{table}

\section{Sensitiveness to initial centroids}

We utilize a simple example to illustrate the advantage of our proposed NCP methods over K-means when faced with bad initial points. Specifically, we generate a dataset with 1000 samples belonging to 3 clusters, which is shown in Fig. \ref{fig: example}. We specifically choose three initial centroids for K-means and our proposed methods by randomly selecting three points from the biggest cluster, which is also shown in Fig. \ref{fig: example}.
\begin{figure} [H]
	\centering
	\includegraphics[width=5cm]{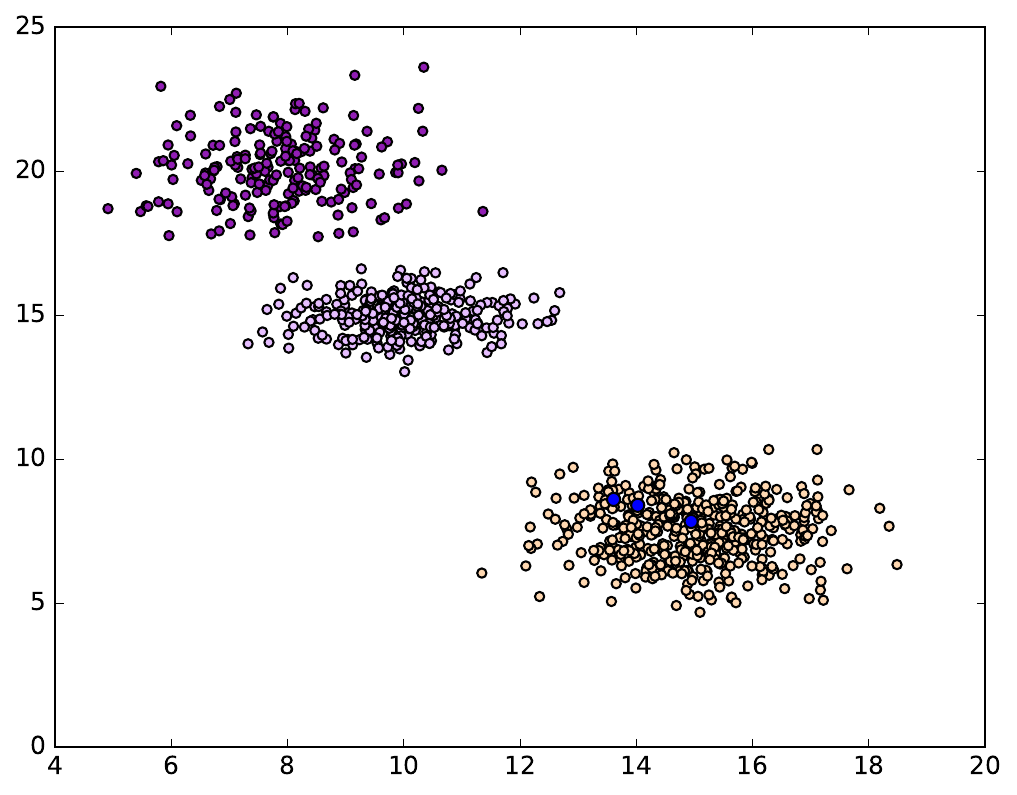}
	\caption{A dataset with 1000 samples in 3 clusters. Blue points indicate the three initial centroids.}
		\label{fig: example}
\end{figure}

\begin{figure} [t!]
	\centering\footnotesize
	\subfigure[\scriptsize K-means, Iter = 1]{
		\includegraphics[width=5cm]{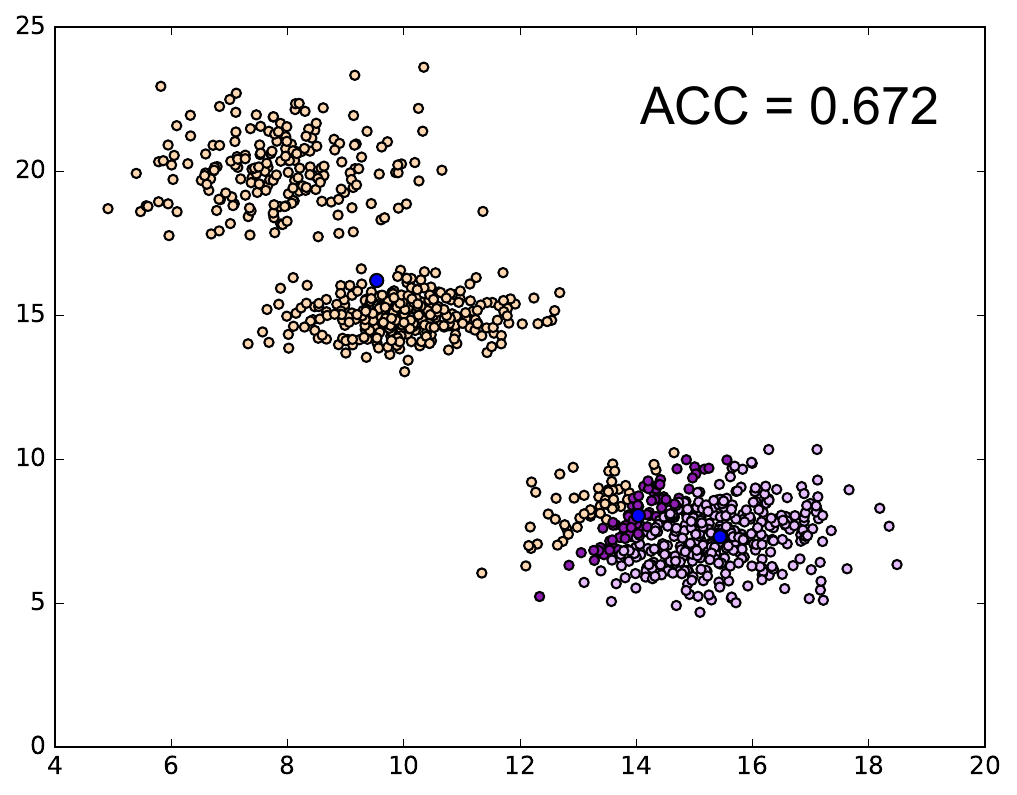}
	}\label{fig: km1}
	\subfigure[\scriptsize K-means, Iter = 2]{
		\includegraphics[width=5cm]{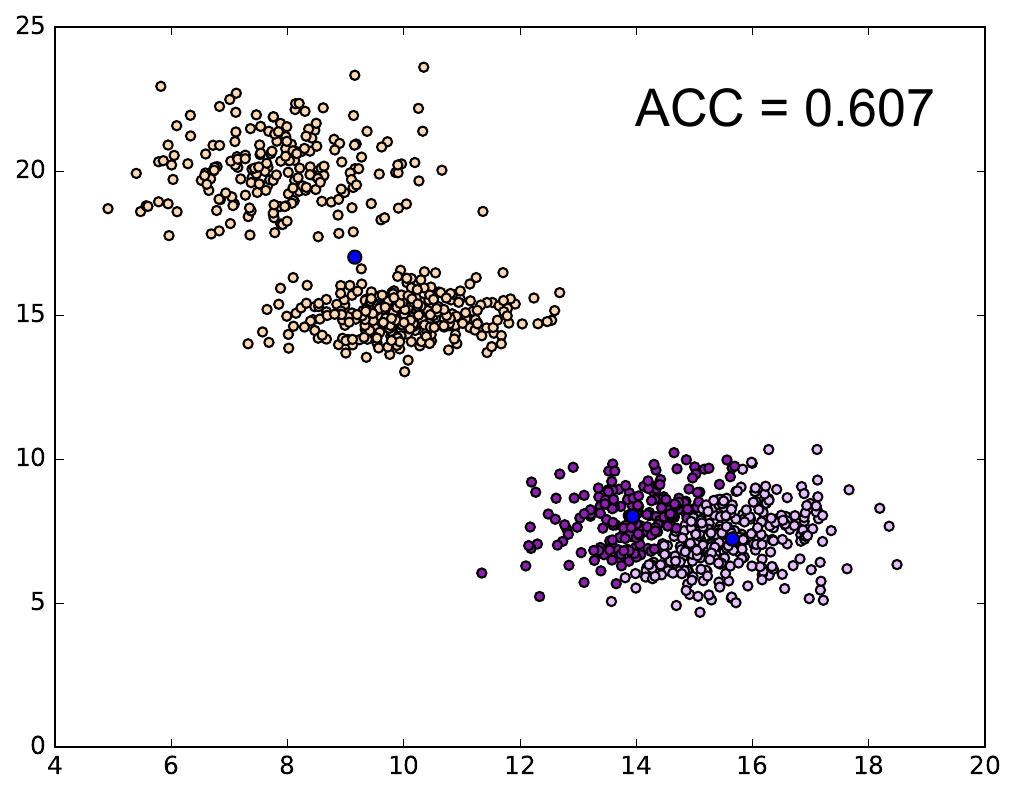}
	}\label{fig: km2}
	\subfigure[\scriptsize K-means, Iter = 6]{
		\includegraphics[width=5cm]{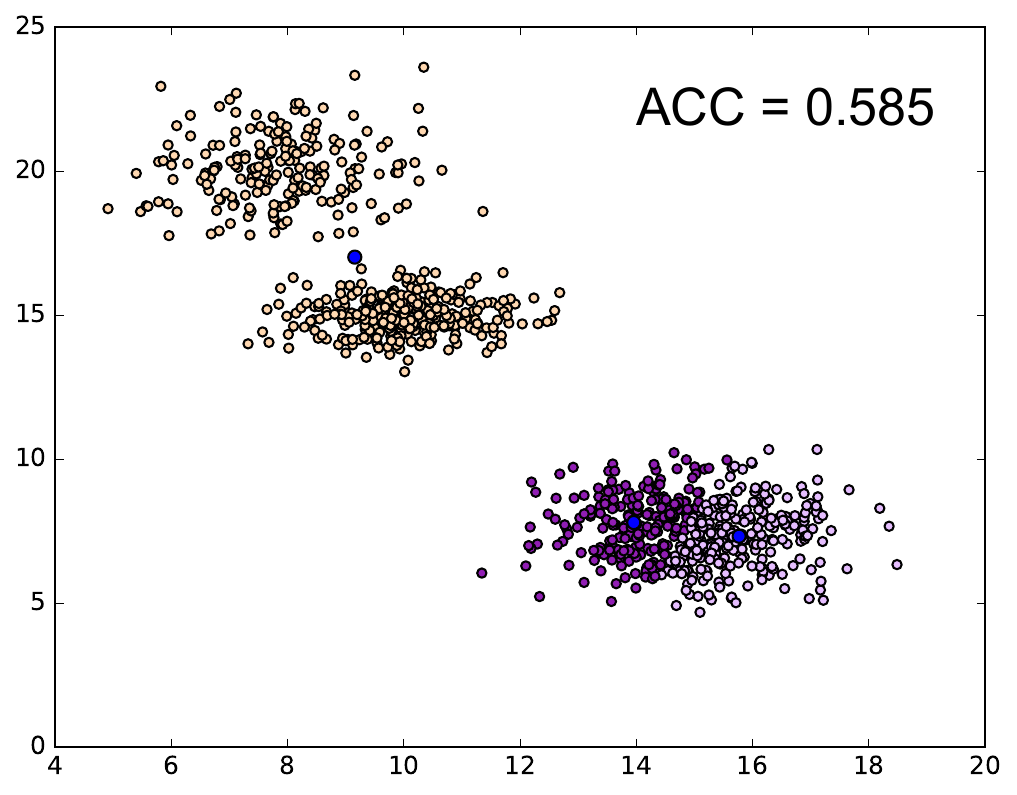}
	}\label{fig: km3}
	\subfigure[\scriptsize SNCP, NCP Iter = 1]{
		\includegraphics[width=5cm]{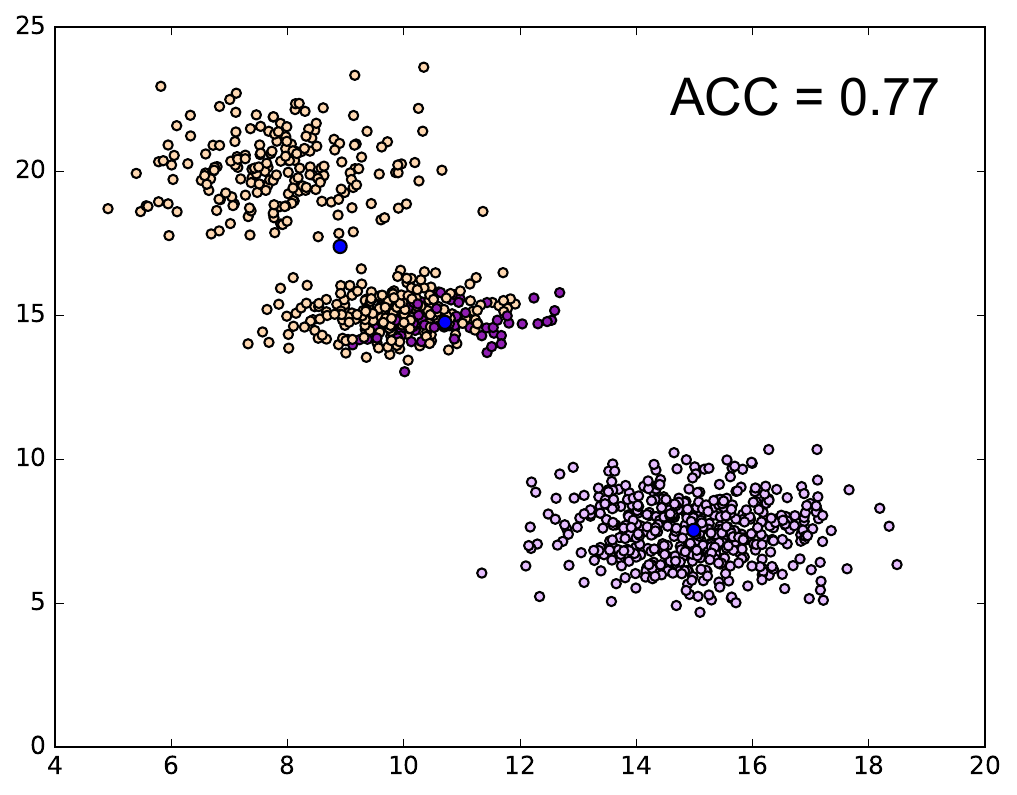}
	}\label{fig: sncp1}
	\subfigure[\scriptsize SNCP, NCP Iter = 3]{
		\includegraphics[width=5cm]{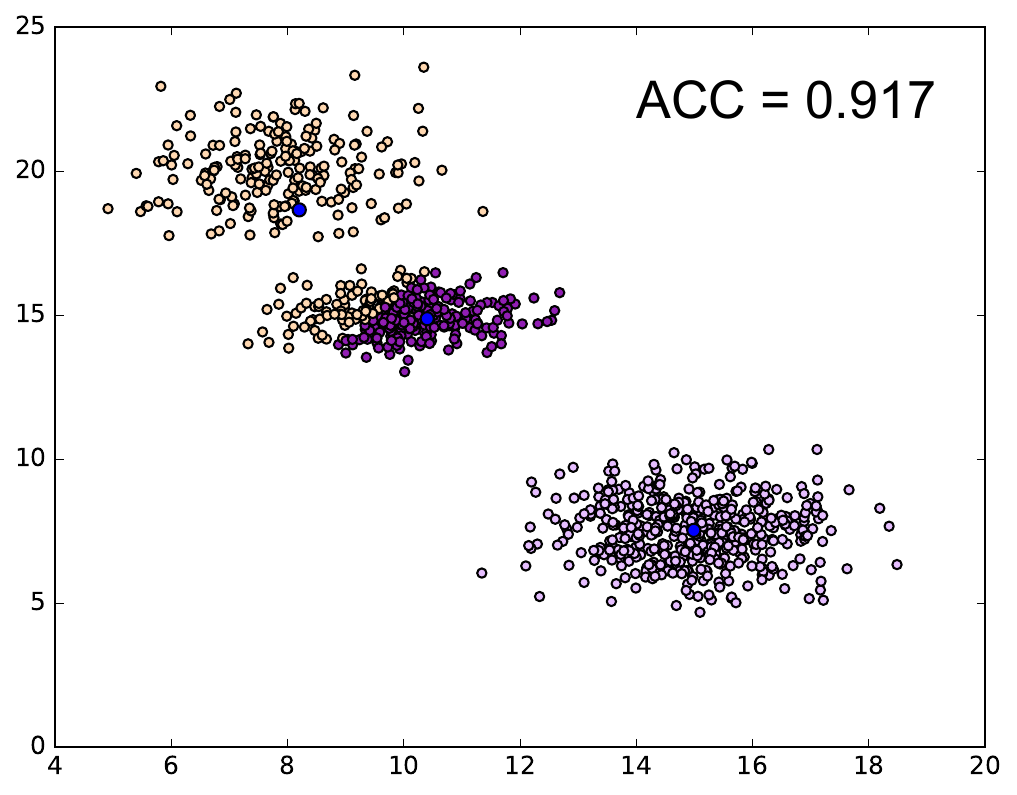}
	}\label{fig: sncp2}
	\subfigure[\scriptsize SNCP, NCP Iter = 41]{
		\includegraphics[width=5cm]{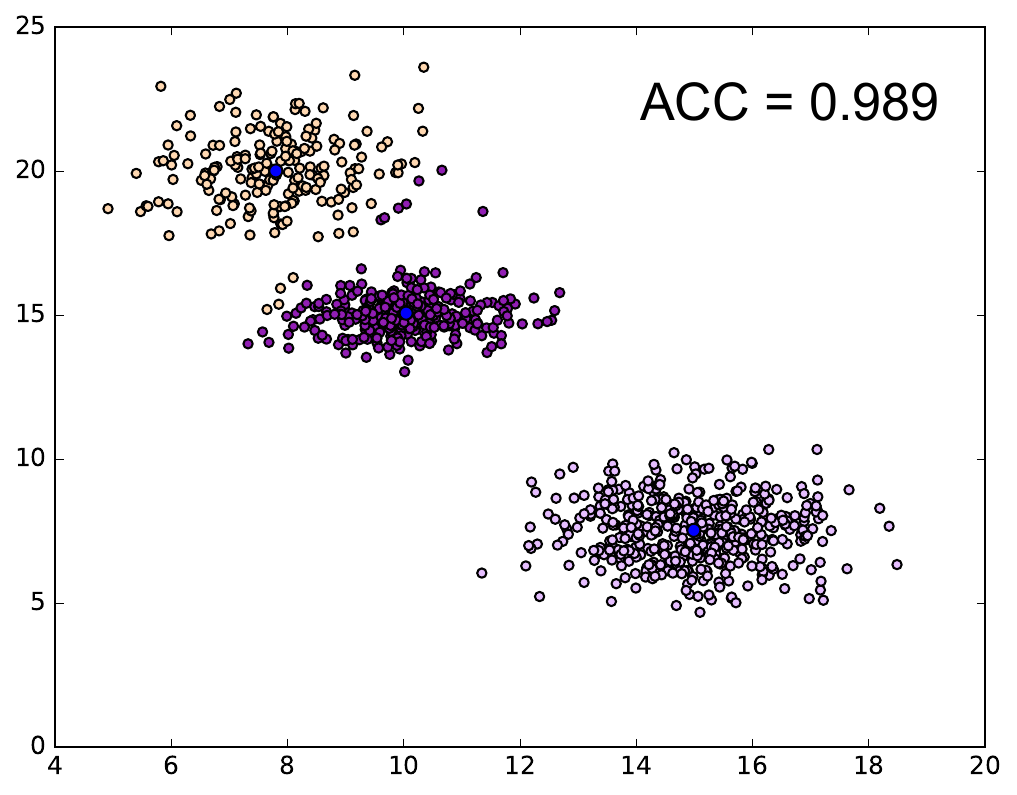}
	}\label{fig: sncp3}
	\subfigure[\scriptsize NSNCP, NCP Iter = 1]{
		\includegraphics[width=5cm]{./figures/cls_centers/sncp/res1}
	}\label{fig: nsncp1}
	\subfigure[\scriptsize SNCP, NCP Iter = 11]{
		\includegraphics[width=5cm]{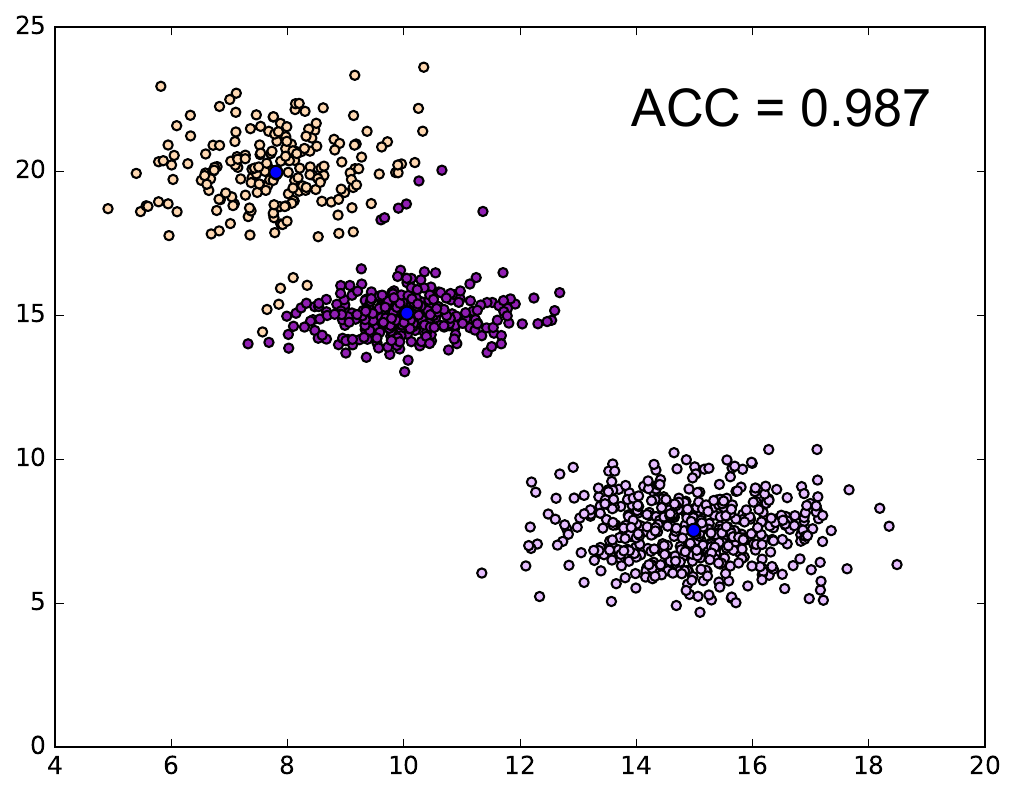}
	}\label{fig: nsncp2}
	\subfigure[\scriptsize SNCP, NCP Iter = 13]{
		\includegraphics[width=5cm]{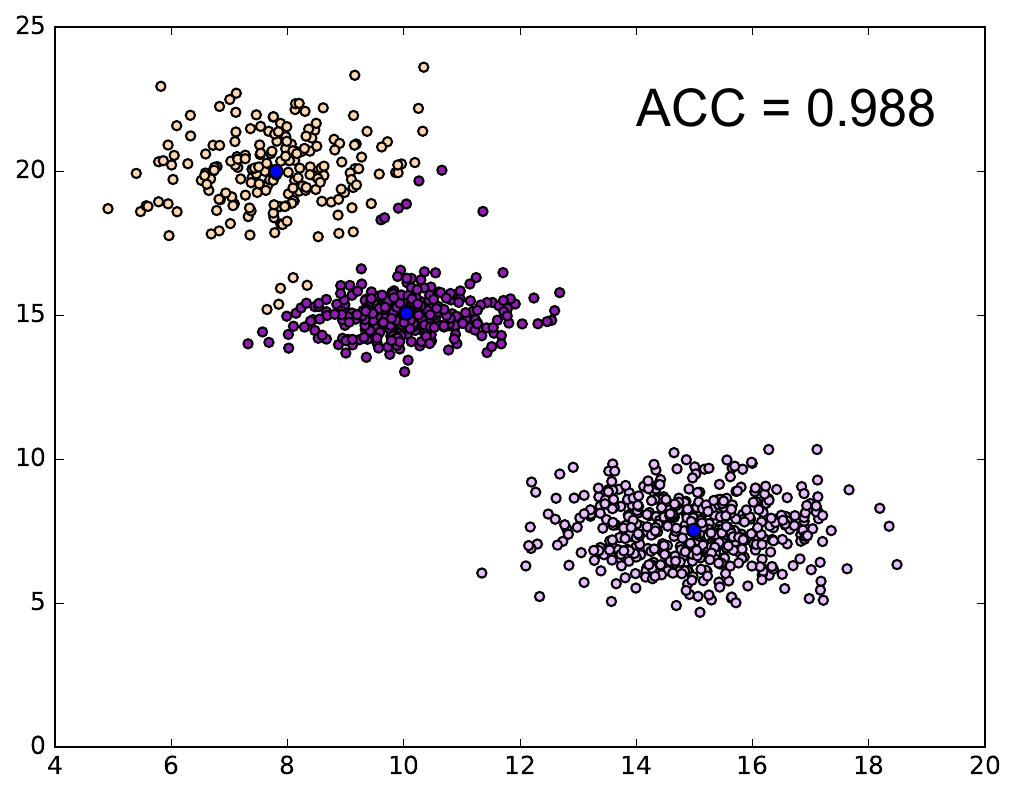}
	}\label{fig: nsncp3}
	\vspace{-0.25cm}
	\caption{The clustering results of on the TCGA dataset with one same initialization}
	\label{fig: cls_results}
\end{figure}

Fig. \ref{fig: cls_results} shows the clustering results of the three methods in different iterations. As seen in Fig.  \ref{fig: cls_results} (a)-(c), the K-means cannot identify the three clusters given bad initial centroids. Specifically, it mistakenly classify the two smaller clusters into one and utilize only one centroid to represent them. Compared with K-means, the proposed SNCP with the same initial centroids can gradually classify the dataset into three almost correct clusters as it proceeds from one NCP iteration to another.

\footnotesize